\documentclass[11pt]{article} 
\usepackage[letterpaper,margin=1in]{geometry}

\usepackage[T1]{fontenc}
\usepackage{xcolor}
\usepackage{microtype}
\usepackage{xspace}
\usepackage{verbatim}

\def\colorful{0}
\ifnum\colorful=1
\newcommand{\new}[1]{{\color{red} #1}}

\else
\newcommand{\new}[1]{{#1}}

\fi

\newif\ifred
\redtrue

\usepackage{titlesec}
\titleformat*{\paragraph}{\bfseries}

\usepackage{amsthm}
\usepackage{amsmath,amssymb,amsfonts,mathtools}
\usepackage{empheq}
\usepackage{dsfont}
\usepackage{mathrsfs}
\usepackage{bbm}

\usepackage{graphicx}
\usepackage{enumitem}
\usepackage{caption}
\usepackage{framed}

\usepackage{tikz}
\usetikzlibrary{patterns}
\usetikzlibrary{arrows,shapes,automata,backgrounds,petri,positioning}
\usetikzlibrary{shadows}
\usetikzlibrary{calc}
\usetikzlibrary{spy}
\usetikzlibrary{angles,quotes}
\usetikzlibrary{matrix}

\usepackage{pgf,pgfplots}
\usepackage{pgfmath,pgffor}
\pgfplotsset{compat=1.17}

 \usepackage{algorithm}
\usepackage[noend]{algpseudocode}

\definecolor[named]{ACMBlue}{cmyk}{1,0.1,0,0.1}
\definecolor[named]{ACMYellow}{cmyk}{0,0.16,1,0}
\definecolor[named]{ACMOrange}{cmyk}{0,0.42,1,0.01}
\definecolor[named]{ACMRed}{cmyk}{0,0.90,0.86,0}
\definecolor[named]{ACMLightBlue}{cmyk}{0.49,0.01,0,0}
\definecolor[named]{ACMGreen}{cmyk}{0.20,0,1,0.19}
\definecolor[named]{ACMPurple}{cmyk}{0.55,1,0,0.15}
\definecolor[named]{ACMDarkBlue}{cmyk}{1,0.58,0,0.21}

\usepackage{aliascnt}

\newtheorem{theorem}{Theorem}[section]

\newaliascnt{question}{theorem}

\aliascntresetthe{question}

\newaliascnt{cond}{theorem}

\aliascntresetthe{cond}

\newaliascnt{condition}{theorem}

\aliascntresetthe{condition}

\newaliascnt{lemma}{theorem}
\newtheorem{lemma}[lemma]{Lemma}
\aliascntresetthe{lemma}

\newaliascnt{corollary}{theorem}

\aliascntresetthe{corollary}

\newaliascnt{corrollary}{theorem}

\aliascntresetthe{corrollary}

\newaliascnt{informal theorem}{theorem}
\newtheorem{informal theorem}[informal theorem]{Theorem (informal statement)}
\aliascntresetthe{informal theorem}

\newaliascnt{proposition}{theorem}
\newtheorem{proposition}[proposition]{Proposition}
\aliascntresetthe{proposition}

\newaliascnt{claim}{theorem}
\newtheorem{claim}[claim]{Claim}
\aliascntresetthe{claim}

\newaliascnt{fact}{theorem}
\newtheorem{fact}[fact]{Fact}
\aliascntresetthe{fact}

\newaliascnt{obs}{theorem}

\aliascntresetthe{obs}

\newaliascnt{observation}{theorem}

\aliascntresetthe{observation}

\newaliascnt{remark}{theorem}
\newtheorem{remark}[remark]{Remark}
\aliascntresetthe{remark}

\newaliascnt{infproposition}{theorem}

\aliascntresetthe{infproposition}

\newaliascnt{assumption}{theorem}

\aliascntresetthe{assumption}

\newaliascnt{definition}{theorem}
\newtheorem{definition}[definition]{Definition}
\aliascntresetthe{definition}

\usepackage[colorlinks,citecolor=blue,linkcolor=magenta,bookmarks=true]{hyperref}
\usepackage[nameinlink,capitalise]{cleveref}

\crefname{theorem}{Theorem}{Theorems}
\crefname{question}{Question}{Questions}
\crefname{conj}{Conjecture}{Conjectures}
\crefname{cond}{Condition}{Conditions}
\crefname{condition}{Condition}{Conditions}
\crefname{lemma}{Lemma}{Lemmas}
\crefname{corollary}{Corollary}{Corollaries}
\crefname{corrollary}{Corollary}{Corollaries}
\crefname{informal theorem}{Theorem}{Theorems}
\crefname{proposition}{Proposition}{Propositions}
\crefname{claim}{Claim}{Claims}
\crefname{fact}{Fact}{Facts}
\crefname{obs}{Observation}{Observations}
\crefname{observation}{Observation}{Observations}
\crefname{remark}{Remark}{Remarks}
\crefname{infproposition}{Proposition}{Propositions}
\crefname{assumption}{Assumption}{Assumptions}
\crefname{definition}{Definition}{Definitions}
\crefname{example}{Example}{Examples}

\crefname{algocf}{Algorithm}{Algorithms}
\Crefname{algocf}{Algorithm}{Algorithms}
\crefname{algorithm}{Algorithm}{Algorithms}
\Crefname{algorithm}{Algorithm}{Algorithms}

\crefname{ineq}{Inequality}{Inequalities}
\creflabelformat{ineq}{#2{\upshape(#1)}#3}

\crefname{sub}{Subsection}{Subsections}
\creflabelformat{sub}{#2{\upshape(#1)}#3}

\crefname{subsection}{Subsection}{Subsections}
\creflabelformat{subsection}{#2{\upshape(#1)}#3}

\crefname{sdp}{SDP}{SDPs}
\creflabelformat{sdp}{#2{\upshape(#1)}#3}

\crefname{lp}{LP}{LPs}
\creflabelformat{lp}{#2{\upshape(#1)}#3}



\newcommand{\eqdef}{\coloneqq}

\newcommand{\lp}{\left}
\newcommand{\rp}{\right}
\newcommand\norm[1]{\left\| #1 \right\|}

\renewcommand\vec[1]{\mathbf{#1}}
\DeclareMathOperator*{\pr}{\mathbf{Pr}}
\DeclareMathOperator*{\E}{\mathbf{E}}
\newcommand{\proj}{\mathrm{proj}}

\newcommand{\normal}{\mathcal{N}}
\newcommand{\cN}{\mathcal{N}}
\newcommand{\rand}{\mathrm{Rand}}

\newcommand{\tr}{\mathrm{tr}}
\newcommand{\bx}{\mathbf{x}}
\newcommand{\by}{\mathbf{y}}
\newcommand{\bv}{\mathbf{v}}

\newcommand{\bz}{\mathbf{z}}
\newcommand{\bw}{\mathbf{w}}

\newcommand{\bc}{\mathbf{c}}
\newcommand{\e}{\mathbf{e}}

\newcommand{\bI}{\mathbf{I}}

\newcommand{\bA}{\mathbf{A}}
\newcommand{\bL}{\mathbf{L}}

\renewcommand{\v}{\vec v}

\newcommand{\err}{\mathrm{err}}

\newcommand{\B}{\mathbb{B}} 
\renewcommand{\H}{\mathcal H }

\newcommand{\p}{\mathbf{P}}

\newcommand{\R}{\mathbb{R}}
\newcommand{\s}{\mathbb{S}}
\newcommand{\Z}{\mathbb{Z}}
\newcommand{\N}{\mathbb{N}}
\newcommand{\eps}{\epsilon}

\newcommand{\poly}{\mathrm{poly}}

\newcommand{\var}{\mathbf{Var}}

\newcommand{\D}{\mathcal{D}}

\renewcommand{\B}{\mathbf{B}}

\newcommand{\sign}{\mathrm{sign}}

\newcommand{\Ind}{\mathds{1}}
\newcommand{\1}{\Ind}

\newcommand{\ba}{\mathbf{a}}
\newcommand{\bb}{\mathbf{b}}

\newcommand{\wt}{\widetilde}
\newcommand{\wh}{\widehat}

\newcommand{\z}{\mathbf{z}}
\newcommand{\op}{\mathrm{op}}

\newcommand{\x}{\vec x}
\newcommand{\w}{\vec w}
\renewcommand{\u}{\vec u}

\newcommand{\Hinge}{\mathrm{Hinge}}

\newcommand{\iid}{{i.i.d.}\ }
\newcommand{\abs}[1]{\lp| #1 \rp|}

\newcommand{\spa}{\mathrm{span}}

\newcommand{\card}[1]{|#1|}

\DeclareMathOperator{\Var}{Var}
\renewcommand\Pr{\pr}

\date{}

\begin{document}

\title{Beyond Multi-Index Models:\\
Learning Intersections of Degree-$2$ PTFs
}

\author{
Ilias Diakonikolas\thanks{Supported by NSF Medium Award CCF-2107079,
ONR Award number N00014-25-1-2268, and a Guggenheim Fellowship.}\\
University of Wisconsin-Madison\\
\texttt{ilias@cs.wisc.edu}
\and
Daniel M. Kane\thanks{Supported by NSF Medium Award CCF-2107547.}\\
University of California, San Diego\\
\texttt{dakane@ucsd.edu}
\and
Mingchen Ma\thanks{Supported by ONR Award number N00014-25-1-2268.}\\
University of Wisconsin-Madison\\
\texttt{mingchen@cs.wisc.edu}
\and
Lisheng Ren\thanks{Supported in part by NSF Medium Award CCF-2107079
and ONR Award number N00014-25-1-2268.}\\
University of Wisconsin-Madison\\
{\tt lren29@wisc.edu}
\and
Trung Tran\\
University of California, San Diego\\
\texttt{trt023@ucsd.edu}
}

\maketitle

\vspace{-0.5cm}

\begin{abstract}

We give an algorithm for PAC learning intersections of $k$
degree-2 polynomial threshold functions over $\R^d$ under the
standard Gaussian distribution. Our algorithm achieves classification error
$\eps$ using $d^{O(k)}2^{O_k(\log^3(1/\eps))}$ time and samples.
For every fixed $k$, this is polynomial in $d$ and
quasipolynomial in $1/\eps$, improving over the
$d^{O(k^2/\eps^2)}$ running time of the standard polynomial
regression approach. To our knowledge, our result is the first
to obtain a dimension exponent independent of the accuracy
parameter, even for $k=2$. We complement our algorithmic
result with a lower bound showing that
the complexity dependence of $d^{\Omega(k)}$ is unavoidable for any Statistical Query learning algorithm.

Lifting the input to the space of linear and quadratic features
turns the target into an intersection of $k$ halfspaces. This suggests
the classical approach of using the first two labeled moments to reduce
the learning problem to at most $k$ dimensions. In the lifted space,
however, these moments can reveal nothing about the relevant subspace.
Our starting point is
a Gaussian comparison theorem showing that, when the
defining quadratics are sufficiently regular, the labeled
moments needed for subspace recovery are approximately
preserved. To handle arbitrary defining quadratics,
we establish a new structural result showing that every
direction with significant resampling influence can be
detected through degree-$O(k)$ Hermite moments.
These results allow us to efficiently
identify influential directions and reduce the problem
to learning regular intersections through an adaptive
restriction procedure, yielding an initial algorithm
with running time $d^{O(k)}2^{\mathrm{poly}_k(1/\eps)}$.

To improve the exponential dependence on $1/\eps$,
we develop a two-stage dimension-reduction approach
that avoids the costly adaptive partitioning procedure.
After identifying a subspace containing all directions
with significant influence, we use a generalized
notion of regularity and Gaussian comparison to recover
the remaining relevant quadratic features through a new derivative-based moment operator. Together, the recovered features support
an intersection that closely approximates the target.
Combining these ingredients with a robust Hinge loss minimization method
gives our quasipolynomial-time learning algorithm.
\end{abstract}

\thispagestyle{empty}

\newpage

\tableofcontents

\thispagestyle{empty}

\newpage

\setcounter{page}{1}

\section{Introduction}
\label{sec:introduction}

We study the problem of PAC learning intersections of degree-$2$ polynomial threshold functions (PTFs)
under the Gaussian distribution. A degree-$2$ PTF is a Boolean function obtained by thresholding a
polynomial of degree at most two. An intersection of $k$ such functions labels a point positive if it
satisfies all $k$ quadratic inequalities.
When the defining polynomials are affine, this is the well-studied problem
of learning intersections of halfspaces~\cite{kwek1996pac,blum1997learning,KS04,Vempala10a,APS26,rev-PV26}.
Quadratic thresholds capture pairwise
interactions between coordinates and include basic geometric concepts such as balls and ellipsoids.
Intersections of quadratic PTFs also describe nonconvex regions, such as spherical shells.

Efficient algorithms are known for learning halfspaces and PTFs of fixed
degree, even in the distribution-free setting~\cite{rosenblatt1958perceptron,MT:94},
and for learning generalized linear and single-index models
under suitable assumptions~\cite{kakade2011efficient}.
For a single quadratic PTF, it suffices to solve a linear program after
lifting the examples to quadratic features. For an intersection, a negative
example tells us only that at least one of the defining inequalities failed.
Even for two halfspaces, polynomial-time distribution-free PAC learning
remains a tantalizing open problem.
In fact, \cite{APS26} recently gave the first
subexponential-time algorithm for this case; see also~\cite{rev-PV26}.

This open problem has motivated the study of intersections of halfspaces
under structured input distributions. Efficient algorithms are known
for intersections of a constant number of halfspaces under the uniform
distribution on a ball~\cite{blum1997learning}, Gaussian
distributions~\cite{Vempala10a,DKS18a}, and logconcave distributions
when the bounding hyperplanes pass through the mean~\cite{Vempala10}.
Finding a small set of relevant linear directions is also central to
recent work on learning multi-index models (MIMs) under Gaussian inputs;
see, e.g.,~\cite{ABM23,DiakonikolasIKZ25,DIKR25,DLB25}.

A MIM is a function of the form $f(\x)=g(\mathbf W\x)$,
where $\mathbf W\in\R^{r\times d}$ and the number of indices $r$
is small compared to $d$. The function $g$ is called the link function.
Beyond intersections of halfspaces, MIMs also include many
important function classes, such as shallow ReLU networks
of bounded width~\cite{diakonikolas2020algorithms,ChenKM21,chen2023learning,diakonikolas2024efficiently,
ChenN24}
and multiclass linear classifiers~\cite{DiakonikolasIKZ25, diakonikolas2025statistical}.
We refer the reader to~\cite{BH25} for a survey.

Despite their broad applicability, multi-index models (MIMs) impose a specific form
of low-dimensional structure: the target function must depend only on a small
number of {\em linear} projections of the input.
The restriction to \emph{linear} indices matters even for functions of
a single quadratic feature. For example, consider the spherical shell
\[
f(\x)=\Ind\{a\leq\norm{\x}_2^2\leq b\} \;,
\qquad 0<a<b<\infty \;.
\]
The label depends on a single quadratic feature, $\norm{\x}_2^2$, but
no projection onto a proper linear subspace of $\R^d$ determines it; existing guarantees for low-dimensional MIMs do not directly give useful learning guarantees for a function like this.

\new{The shell example points to a broader research direction beyond MIMs:
which functions of a few unknown {\em nonlinear} features can be learned
efficiently under Gaussian inputs? See~\cite{BH25,Nichani26} for
discussion. From the perspective of learning geometric concepts,
intersections of quadratic PTFs are arguably the most natural starting
point. Consider the well-studied problem of learning intersections
of halfspaces. If we keep the link function fixed and increase the degree
of the defining polynomials from one to two, we obtain precisely the class
studied here. The target then has the form}
\[
f(\x)=g(q_1(\x),\ldots,q_k(\x)),
\qquad
g(t_1,\ldots,t_k)=\Ind\{t_1\geq0,\ldots,t_k\geq0\},
\]
so the target depends on only $k$ quadratic features, even though it may require $d$ linear indices.
The difficulty is to find useful quadratic features from labels that
reveal only whether all $k$ inequalities hold. To see where the linear
case differs, consider an intersection of $k$ halfspaces under Gaussian
inputs. Its label depends only on the projection onto the span of its
normal vectors. Estimates of the first two labeled moments (aka degree at most $2$ Chow parameters),
$
\E[f(\x)\x]$ and
$\E[f(\x)(\x\x^\top-\bI_d)],
$
can be used to find a subspace $V$ of dimension $O(k)$ such that an intersection with all its normal vectors in $V$ approximates the target to error $\eps$. Searching over intersections in the recovered subspace gives, for example, a running time of $d^{O(1)}(k/\eps)^{O(k^2)}$, where $d$ is the ambient dimension and the exponent of $d$ is a universal constant independent of $k$ and $\eps$~\cite{Vempala10a,DKS18a}.

This reduction does not directly extend to quadratic PTFs: even a single
quadratic inequality need not be determined by the projection onto any
proper linear subspace.
One might instead lift the examples to the $\Theta(d^2)$-dimensional space of quadratic
features, where the target becomes an intersection of $k$ halfspaces. However, the
lifted distribution is not Gaussian. Its coordinates are dependent, and the
distributional properties used by the existing subspace-recovery algorithms no
longer hold. Choosing an orthonormal Hermite basis makes these coordinates
uncorrelated, but does not resolve this obstruction.

For arbitrary intersections of quadratic PTFs, the standard general-purpose (and the only known) approach is low-degree polynomial regression~\cite{KOS:08}.
Combining the Gaussian surface-area bound for PTFs~\cite{Kane11} with
the realizable learning guarantee of~\cite{KOS:08} gives a
polynomial-regression algorithm for this class with running time
$d^{O(k^2/\eps^2)}$.\footnote{Each quadratic PTF has Gaussian surface area $O(1)$,
so an intersection of $k$ such PTFs has Gaussian surface area $O(k)$.
The realizable learning bound of~\cite[Theorem~25]{KOS:08} is $d^{O(S^2/\eps^2)}$
for a class with Gaussian surface area at most $S$.}
Thus, even for $k=2$, the exponent of the dimension depends polynomially on $1/\eps$. This leaves the following basic question:
\begin{quote}
\emph{For every fixed $k$, can intersections of $k$ degree-$2$ PTFs be learned under the Gaussian distribution in time $d^{O_k(1)}F_k(1/\eps)$, for some function $F_k$ independent of $d$?}
\end{quote}
The above question allows arbitrary dependence on $1/\eps$; it only asks that the exponent of $d$
be independent of the accuracy parameter. Our main result answers it affirmatively, with
quasipolynomial dependence on $1/\eps$.

\subsection{Our Results and Technical Contributions}
\label{subsec:intro-results}

We first define the target concept class and the PAC learning problem.

\begin{definition}[Intersections of degree-$2$ PTFs]
\label{def:degree-2-ptf-intersections}
For $k\in\Z_+$, let $\H_k$ be the class of functions $f:\R^d\to\{0,1\}$ of the form $f(\x)=\Ind\{q_1(\x)\geq0,\ldots,q_k(\x)\geq0\}$, where each $q_i:\R^d\to\R$ is a polynomial of degree at most two.
\end{definition}

\begin{definition}[PAC learning under the Gaussian distribution]
\label{def:learning-degree-2-ptf-intersections}
Let $D$ be an unknown distribution over $\R^d\times\{0,1\}$ with marginal $D_\x=\cN(0,\bI_d)$, and suppose that $y=f^*(\x)$ almost surely for some $f^*\in\H_k$. Given \iid sample access to $D$, an accuracy parameter $\eps\in(0,1/2)$, and a confidence parameter $\delta\in(0,1)$, the goal is to output a hypothesis $\wh f:\R^d\to\{0,1\}$ such that, with probability at least $1-\delta$,
$
\err_D(\wh f)\eqdef\Pr_{(\x,y)\sim D}[\wh f(\x)\neq y]\leq\eps.
$
The hypothesis need not belong to $\H_k$.
\end{definition}

Our main algorithmic result is the following; see \Cref{thm:sec9-conclusion} for the formal statement and proof.
For simplicity, we state the bound for constant confidence.

\begin{theorem}[Main algorithmic result] \label{th main}
There is an algorithm that PAC learns $\H_k$ under the standard Gaussian distribution
to error $\eps\in(0,1/2)$ with probability at least $99\%$,
using $d^{O(k)}\,2^{O_k(\log^3(1/\eps))}$ samples and time.
\end{theorem}

Here the constant implicit in $O_k(\cdot)$ may depend on $k$. For every
fixed $k$, the running time is polynomial in $d$ and quasipolynomial
in $1/\eps$. Already for $k=2$, this gives $d^{O(1)}2^{O(\log^3(1/\eps))}$
time, whereas polynomial regression gives $d^{O(1/\eps^2)}$.
Our algorithm is improper: it learns quadratic features $\phi_1,\ldots,\phi_m$ and outputs a hypothesis of the form
\[
\wh f(\x)=\Ind\{P(\phi_1(\x),\ldots,\phi_m(\x))\geq0\},
\]
where $P$ is a polynomial and the number $m$ of learned quadratic
features is bounded by a function of $k$ and $\eps$.
Thus, the output is a single PTF (whose degree may be larger than two).

\begin{remark}
{\em In \Cref{th exponential time}, we also develop a different learning algorithm with time complexity $d^{O(k)}2^{\poly_k(1/\eps)}$, whose output hypothesis is a decision tree of intersections of quadratic PTFs. We refer the reader to \Cref{sec tree building} for details. }
\end{remark}

We also prove a lower bound in the Statistical Query (SQ)
model~\cite{Kearns:98}, that qualitatively matches the complexity of our algorithm for constant error $\eps$.
In the SQ model, the learner accesses the labeled distribution
through approximate expectations of bounded query functions.
The following theorem shows that a linear dependence
on $k$ in the dimension exponent is necessary for SQ algorithms; see \Cref{thm:SQ-LB-main} for the formal statement and proof.

\begin{theorem}[SQ lower bound]
\label{thm:intro-sq}
Fix a constant $\gamma\in(0,1/2)$. For $d$ sufficiently large as a function of $k$ and $\gamma$, any SQ algorithm that learns $\H_k$ under the standard Gaussian distribution to error at most $1/2-\gamma$, with success probability at least $2/3$, must either use a query of tolerance at most $d^{-\Omega(k)}$ or make at least $2^{d^{\Omega(1)}}$ queries.
\end{theorem}

Thus, the $d^{O(k)}$ dependence of our upper bound is essentially
optimal in the SQ framework, where complexity accounts for both
the number of queries and their inverse tolerance. This gives a
computational distinction from intersections of halfspaces.
In the realizable Gaussian setting, the moment-based algorithms
for halfspace intersections can be implemented using
$d^{O(1)}(k/\eps)^{O(k^2)}$ statistical queries and tolerance
$1/\poly(d,k,1/\eps)$, with a universal exponent of
$d$~\cite{Vempala10a,DKS18a}.\footnote{The first two labeled moments
can be estimated by bounded statistical queries after truncating
the Gaussian tails. The subsequent search over a finite cover
only requires estimates of classification error; see
\cite[Section~5.1]{DKS18a}.}
Our SQ lower bound already holds when all the defining quadratics depend
on the same unknown linear projection. It leaves open whether the
running time can be polynomial in both $d$ and $1/\eps$ for every fixed $k$.

\vspace{-0.2cm}

\new{
\paragraph{Technical contributions.}
Beyond our learning guarantees, we develop a collection of structural
and algorithmic tools that may be useful for other
learning problems involving nonlinear features under Gaussian inputs.
Our approach combines Gaussian analysis, algebraic arguments,
spectral methods, and polynomial approximation.
Specifically, we highlight the following:
\begin{enumerate}[leftmargin=*, itemsep=-1pt]
    \item \emph{Low-degree certificates for directional influence.}
    We develop a framework for analyzing the directional dependence of Boolean
    functions  and constructing the corresponding low-degree certificate
    through their Gaussian-weighted decision boundaries. By restricting a
    Boolean function to Gaussian lines, we show (\Cref{lm error-mass}) that
    nontrivial resampling influence along a direction implies a nontrivial
    amount of Gaussian-weighted boundary mass along that direction. We further
    introduce a Gaussian boundary operator (\Cref{def boundary operator}) that,
    through integration by parts, expresses the correlation between the Boolean
    function and a polynomial in terms of the change signed values of the
    polynomial at these decision boundaries (\Cref{lm M operator}). This
    provides a systematic way to translate information about the changes in a
    Boolean function into polynomial correlations. For intersections of $k$
    quadratic PTFs, we develop a quantitative construction of such
    certificates, showing that any direction with nontrivial resampling
    influence admits a polynomial witness of degree $O(k)$ with nontrivial
    correlation. (Sections~\ref{sec certificate polynomial}-\ref{sec direction finding}.)

    \item \emph{Robust algebraic tools for degenerate decision boundaries.}
   We develop quantitative structural techniques for analyzing degenerate
   polynomial decision boundaries, which allow the boundary-based certificate
   framework to work without separation or genericity assumptions. Our key
   approach combines algebraic tools, including discriminants and resultants,
   with Gaussian anti-concentration to translate approximate root degeneracies
   on a nontrivial fraction of Gaussian slices into global small-norm
   polynomial relations.
   In particular, we establish structural simplifications for nearly double
   roots (\Cref{sec double roots}) and a robust factorization theorem (\Cref{lm
small resultant to factorization}) for nearly shared roots.  These structural
alternatives allow us to simplify the defining intersection and construct
polynomial certificates without imposing separation or genericity assumptions
on its decision boundaries. (Section~\ref{sec double roots}-\ref{sec certificate completion}.)

    \item \emph{Generalized regularity and Gaussian comparison.}
    We develop a generalized notion of regularity (\Cref{def:subspace-regular})
    for quadratic polynomials relative to a subspace $V \subseteq \R^d$,
    allowing arbitrary dependence on $V$ while controlling the residual
    quadratic features in directions orthogonal to $V$.
    We show (\Cref{th error to regularity}) that an intersection with
    sufficiently small resampling influence outside $V$ can be approximated by
    one satisfying this regularity condition, even when its original defining
    polynomials are highly irregular. Under this generalized regularity
    condition, we establish a Gaussian comparison result
    (\Cref{lem:gaussian-hermite-moment-comparison}) that replaces the residual
    quadratic features by linear Gaussian features while preserving their
    dependence on $V$. Together, these results provide a framework for
    extending moment-based learning techniques from linear Gaussian models to
    nonlinear polynomial features.

    \item \emph{Derivative-based subspace recovery in polynomial feature spaces.}
We develop new structural
techniques for recovering relevant
directions in a space of quadratic polynomial features. Unlike standard
subspace recovery, the feature space we consider consists of residual linear, mixed, and quadratic terms, whose coordinates are not jointly Gaussian and do
not admit the usual notion of directional differentiation. To overcome this
difficulty, we first establish a structural result
(\Cref{lm structre idealized gaussian}) for intersections of PTFs in an
idealized Gaussian model, showing that influential directions in the residual
feature space can be certified by low-degree polynomial correlations. We then
introduce a new differential operator for quadratic features and its adjoint
under Gaussian measure (\Cref{lm derivative operator} and
\Cref{lm derivative operator construction}), allowing us to approximate the
corresponding directional derivative correlations using labeled moments of the
original input distribution. Under our generalized regularity condition, this
yields a bilinear form that approximates an operator of rank at most $k$.
Consequently, spectral methods recover a low-dimensional space of quadratic
features sufficient for learning the target, without requiring identification
of the original defining polynomials.

\item \emph{Robust polynomial approximation and hinge-loss minimization.}
Building on polynomial approximation techniques for learning intersections of
margin halfspaces~\cite{KS04}, we develop a robust version suitable for intersections
of quadratic PTFs under Gaussian inputs. Specifically, we construct polynomial
approximations with simultaneously small expected hinge loss and controlled
Gaussian $L_2$ norm, while preserving the quadratic feature space of the
target. The norm bound allows us to control the additional loss introduced by
errors in the recovered feature representation. Consequently, we can learn over
an approximately correct feature space through convex hinge-loss minimization.
\end{enumerate}
}

\vspace{-0.3cm}

\subsection{Technical Overview}

\vspace{-0.1cm}

\paragraph{Subspace Recovery for Intersections of Halfspaces.}
We first recall the approach for learning intersections of halfspaces under a Gaussian distribution.
Although learning intersections of halfspaces under general marginal distributions is a challenging
problem, whose computational complexity is far from being understood
(even for the case of $k=2$), the Gaussian distribution has nice structural properties
that enable efficient algorithms; see, e.g.,~\cite{blum1997learning,Vempala10a,DKS18a}.

Specifically, consider an instance of an intersection of $k$ halfspaces
$f(\x) = \Ind\{\w^{(1)}\cdot\x \ge t_1, \dots, \w^{(k)}\cdot\x \ge t_k\}$
and let $V$ be the (at most) $k$-dimensional subspace spanned by $\w^{(1)},\dots,\w^{(k)}$.
The intuition is that if we look at the correlation between $f(\x)$ and the first two moments
of the marginal distribution, $\E_{\x\sim \cN(0,\bI)}[f(\x)\x]$ and
$\E_{\x\sim \cN(0,\bI)}[f(\x)(\x\x^\top-\bI)]$, then for every $\v \perp V$,
$\v$ is orthogonal to $\E_{\x\sim \cN(0,\bI)}[f(\x)\x]$ and $\E_{\x\sim \cN(0,\bI)}[f(\x)(\x\x^\top-\bI)]$
has $0$ eigenvalue along $\v$, because $f(\x)$ is independent of $\v\cdot \x$.
On the other hand, if $f(\x)$ heavily depends on a unit direction $\v$,
then either $\abs{\v\cdot\E_{\x\sim\cN(0,\bI)}[f(\x)\x]}$
or $\abs{\v^\top\E_{\x\sim\cN(0,\bI)}[f(\x)(\x\x^\top-\bI)]\v}$ will be large.

As a result, if we take $W$ to be the subspace spanned by $\E_{\x\sim\cN(0,\bI)}[f(\x)\x]$ and the eigenvectors of $\E_{\x\sim\cN(0,\bI)}[f(\x)(\x\x^\top-\bI)]$ corresponding to its $k$ eigenvalues of largest absolute value, then there must be an instance whose defining vectors are in $W$, close to $f(\x)$.

As a result, if we brute-force search over $W$, we solve the learning problem in $d^{O(1)}(k/\eps)^{O(k^2)}$ time.
In fact, as shown in \cite{DKS18a}, even if the input distribution is corrupted by a small amount of nasty noise, we can still use the same method to learn a good hypothesis from the noisy distribution.

\vspace{-0.2cm}

\paragraph{Polynomial Regularity and Gaussian Comparison.}
Given the discussion above, when working with an intersection of $k$ PTFs, $f(\x) = \Ind\{q_1(\x) \ge 0, \dots, q_k(\x) \ge 0\}$, a natural attempt would be to use polynomial features. For a Gaussian distribution, we choose to use
$\Phi(\x) \in \R^{d(d+3)/2}$, where each coordinate of $\Phi(\x)$ is a normalized Hermite polynomial with degree at most 2, so that the coordinates are orthogonal under the inner product defined under the Gaussian distribution. Under this feature basis, any $q_i(\x) = \theta_i\cdot \Phi(\x) - t_i$.
The idea is that if $\Phi(\x)$ is Gaussian distributed,
then we can apply the same algorithm above to search for good parameters
$(\hat{\theta}_i,\hat{t}_i), i \in [k]$ and return the corresponding intersections of PTFs.
Unfortunately, this is not true, and the distribution of $\Phi(\x)$ is far from a Gaussian distribution.
However, inspired by techniques developed in pseudo-randomness literature \cite{diakonikolas2010bounded,diakonikolas2010bounding,de2014deterministic},
for a nice linear function $L: \R^{d(d+3)/2} \to \R$, $L(\Phi(\x))$, which corresponds to some
quadratic function $q(\x)$, is approximately Gaussian when $\x \sim \cN(0,\bI)$.
We call a degree-$2$ polynomial $p$ to be $\tau$-regular, if for every unit vector $\v$,
$\norm{\partial_\v p}^2 \le \tau^2 \var_{\x \sim \cN(0,\bI)}[p(\x)]$, where $\norm{\partial_\v p}$
is called the influence of $p$ along the $\v$ direction.
Equivalently, if we consider a quadratic function
$p(\x) = \bb\cdot\x + \x^\top \bA \x - \tr(\bA) - t$,
then the regularity condition is equivalent to saying that
$\norm{\bb}$ is small and $\norm{\bA}_{\op}$ is small.
In fact, if $p_i(\x) = \theta_i \cdot \Phi_i(\x), i \in [k]$,
are all regular for some sufficiently small $\tau$,
then the regularity condition guarantees that the
distribution of $(p_1(\x),\dots,p_k(\x))$, $\x \sim \cN(0,\bI)$
approximates a \emph{Gaussian replacement vector} $(\theta_1\cdot G,\dots,\theta_k\cdot G)$,
where $G$ is a standard Gaussian vector.
To see why this helps us, we prove a Gaussian comparison lemma
(\Cref{lem:gaussian-hermite-moment-comparison}) that for any smooth function
$\Psi: \R^k \to \R$ and every direction $\u \in \R^{d(d+3)/2}$, $j = 0,1,2$,
\begin{align*}
    \E_{\x\sim \cN(0,\bI)}[\Psi(p_1(\x),\dots,p_k(\x))h_j(\u \cdot \Phi(\x))] \approx\E_{G\sim \cN(0,\bI)}[\Psi(\theta_1\cdot G,\dots,\theta_k\cdot G)h_j(\u \cdot G)] \;.
\end{align*}
That is to say, if the defining polynomials of $f(\x)$, $q_1,\dots,q_k$ are sufficiently regular,
then the moment information and structure of the polynomial threshold function is sufficiently
close to an instance of intersection of halfspaces in $\R^{d(d+3)/2}$ under a Gaussian
distribution, with the same defining vectors.
Then we can simply treat an intersection of PTFs as its Gaussian replacement for the analysis.
As a result, we can directly run the robust learning algorithm from \cite{DKS18a}
over $\Phi(\x)$ to efficiently learn a hypothesis with error depending
on the regularity parameter $\tau$.

\vspace{-0.2cm}

\paragraph{Randomized Error and Extracting Influential Directions.}
Unfortunately, there is no guarantee that our defining polynomials $q_i$ will be regular,
and so our Gaussian comparison lemma may fail to hold.
There are two possible reasons why this might be the case.

On the one hand, it is possible that the polynomials that we chose
to represent $f$ are irregular, but that another choice would not have been.
For example, suppose $f(\x)$ contains a defining polynomial $q(\x) = \x_1^2+\x^2_2$, which is highly irregular.
On the other hand, since $q(\x) \ge 0$ always holds, we can remove it from $f(\x)$ and find an equivalent representation for the target function.
In fact, consider an irregular polynomial $p(\x)$. Since a bad direction must either correlate well with the linear part of $p$, or have large norm under the bilinear form defined by the quadratic part, we can find a low dimensional subspace (spanned by the linear part and the top few eigenvectors of the quadratic form) that is responsible for all of the bad directions.
If we are allowed to remove the dependence on this subspace
(for example, by replacing this component of the input by a random constant),
then we can reduce $p$ to the regular case.
\new{Our goal is to apply this procedure to each defining polynomial $q_i$
so that the resulting polynomials $\tilde q_i$ are regular and the
intersection they define is close to $f$.}

However, to achieve this without introducing much error,
we need to know that re-randomizing the input of $f$ along a low-dimensional subspace
is unlikely to change the output. This will hold so long as $f$ is not heavily influenced
by any particular direction (as we can change the input on a subspace
by changing it one coordinate at a time).
To quantify this influence, we introduce the notion of randomized error
along a direction $\v$ as
\begin{align*}
        \rand_f(\v) = \E_{\bw \sim \cN(0,\bI_{\v^\perp})}\left[ \Pr_{z,z' \sim \cN(0,1)}[f(\w+z \v) \neq f(\w+z' \v)] \right] \;.
\end{align*}
As we show in \Cref{th error to regularity}, if $\rand_f(\v)$ is small for every direction $\v$, then $f$ must be close to a $\tau$-regular instance up to a small $L_1$ error.

Unfortunately, sometimes $f$ will have directions of large influence, in which case the irregularity of its \new{representation} will be intrinsic.
In particular, there may be some directions $\v$ with $\rand_f(\v)$ reasonably large.
In such a case, no matter how we represent it, some of the defining polynomials of $f$
will have large influence in the $\v$-direction.
To deal with this issue, we will actually need to handle these directions separately.
The hope is that with sample access to $f$, we can directly identify these directions $\v$,
with $\rand_f(\v)$ being nontrivial large.
\new{
To do this, we show that a direction $\v$ will have large influence on $f$ only if
it has large influence on the low-degree part of $f$ (\Cref{th certificate}).
In particular, letting $F_\ell(\x) = \proj_{P_{\le \ell}}(f(\x))$ be the degree-$\ell$-part of $f$
for $\ell$ a large enough multiple of $k$, we show that for any direction $\v$
with $\rand_f(\v) \geq \alpha$ that it must hold that
$\E[(\partial_v F_\ell)^2] \geq \poly_k(\alpha)$. In particular,
this means that defining the influence matrix as
$G_\ell:= \E_{\x \sim \cN(0,\bI)}[\nabla F\nabla F ^\top]$,
that any influential direction for $f$ will have $\v^T G_\ell \v$ relatively large.
Hence, taking $V$ to be the subspace spanned by the most significant eigenvectors of $G_\ell$,
we can efficiently find a subspace $V$ so that for all $\v$ orthogonal to $V$,
$\rand_f(\v)$ is very small (see \Cref{th direction finding one shot}).
By \Cref{th certificate}, if $G_\ell$ has no large eigenvalues, then
$f$ has no direction of large resampling influence. If $\rand_f(\v)$
is large for some unit vector $\v$, then $G_\ell$ has a large
eigenvalue, and a top eigenvector gives a direction of large influence
on the low-degree part of $f$.
}

\vspace{-0.2cm}

\paragraph{Certificate Polynomial Construction.}

We now explain the main ideas behind our main structural result (\Cref{th certificate}).
Specifically, we show that for
an intersection of $k$ degree-$2$ PTFs $f$, if a direction $\v$ has $\rand_f(\v) \ge \alpha$,
then there must be some $r \in [2k]$ and a unit-norm polynomial $p(\w)$ over $\v^\perp$ such that
\begin{align*}
    \E_{\x \sim \cN(0,\bI)}[f(\x) h_r(\v\cdot\x)p(\w)] \ge \poly_k(\alpha),
\end{align*}
and $r + \deg(p) \le 2k$.
In other words, as long as $\rand_f(\v)$ is non-trivially large, then $f(\x)$
must have a non-trivial correlation with a normalized polynomial of degree at most $2k$,
with no-constant-in-$\v$ component.

To give the intuition behind \Cref{th certificate}, we decomposed the target function $f$
according to the value $\w \perp \v$.
Fix a direction $\v$ and write $\x=z\v+\w$, where $\w\perp\v$. For each fixed $\w$, the restriction $f_{\w}(z)=f(z\v+\w)$ is a one-dimensional union of intervals with at most $2k$ decision boundaries.
Furthermore, if $\rand_f(\v)\geq\alpha$, we show in \Cref{lm error-mass} that
these boundaries have total Gaussian-weighted mass at least $\Omega(\alpha)$.

\new{
As $f_{\w}(z)$ is a step function, directly computing correlations with $f$ will be difficult.
However, this means that $\partial_\v f_\w$ is a sum of $\delta$-functions
and much easier to work with. Furthermore, the operator $M_{\v}Q=(\v\cdot\x)Q-\partial_{\v}Q$ is
adjoint to the derivative operator over the Gaussian distribution.
Namely, for functions $g$ and $h$, we have $\E[(\partial_\v g)h] = \E[g M_\v h]$.
Applying this with $g=f_\w$, we get}
\[
\E_{\x}[f(\x)M_{\v}Q(\x)]
=\E_{\w}\left[\sum_{s\in S_{\w}}\Delta f_{\w}(s)Q(s\v+\w)\phi(s)\right] \;,
\]
where $S_{\w}$ denotes the set of decision boundaries of $f_{\w}$
and $\Delta f_\w(s) \in \{\pm 1\}$ is the sign change of $f_\w$ at boundary $s$.
Thus, our goal is to construct a low-degree polynomial $Q$ whose value at each boundary
has the same sign as the corresponding change in $f_{\w}$. Furthermore,
using the polynomial $P=M_\v Q$ has the advantage that the integral over $\v$ direction
of $P$ is $\E[M_\v Q \cdot 1] = \E[Q \cdot \partial_\v 1] = 0.$
Thus, this construction will naturally produce functions with no constant-in-$\v$-component.

For a generic intersection, such a polynomial has a simple construction.
Writing $f(\x)=\mathbf{1}\{q_1(\x)\geq0,\ldots,q_k(\x)\geq0\}$,
consider $Q_{\v}=\partial_{\v}(\prod_{i=1}^k q_i)$. At every decision boundary,
the sign of $Q_{\v}$ agrees with the direction of the jump,
implying that the inner product is always non-negative.
Furthermore, if the derivative of the active polynomial and the values
of all other defining polynomials are bounded away from zero,
then the boundary contributes a nontrivial positive amount to the correlation.
However, this argument can still fail to provide a good lower bound
if $\prod_{i=1}^k q_i$ has pairs of roots that are very close to each other---either due
to some of the $q_i$ having (approximate) double roots or some pairs approximately sharing roots.

\new{To handle these degeneracies, we develop several structural tools that combine algebraic
techniques with Gaussian concentration and anti-concentration to establish quantitative structural
dichotomies. Our starting point is that, after fixing $\w\perp\v$, each defining polynomial becomes
a univariate quadratic whose coefficients are low-degree polynomials in $\w$. This allows us to use
the discriminant to analyze nearly double roots and the resultant to analyze nearly shared roots.
Importantly, both quantities are themselves low-degree polynomials in $\w$,
so Gaussian anti-concentration allows us to convert their smallness
on a nontrivial fraction of slices into a global small-$L_2$-norm condition.
These conditions enable us to derive global structural results to simplify the defining polynomials.

For nearly double roots, we establish a dichotomy (\Cref{sec double roots}): either such roots
occur too infrequently to contribute significantly to the Gaussian-weighted decision boundary, or
the corresponding threshold function can be approximated by a function independent of the $\v$
direction. For nearly shared roots, we develop a robust factorization argument
(\Cref{sec factorization}) showing that either one polynomial is nearly independent of $\v$,
the two polynomials are nearly proportional, or they can be approximated by polynomials sharing a
common affine factor. In particular, if several defining polynomials both contain some affine
function $L$ as a linear factor, then this common affine factor $L$
allows us to construct a new certificate using the factors of the defining polynomials
(\Cref{lm sec55 affine certificate}).
In either case, if the roots are close,
these structural alternatives allow us to simplify the defining intersection,
up to a controlled approximation error, whenever the generic certificate
does not yield sufficient correlation.}

Finally, we combine these tools through an iterative construction to find the desired certificate
polynomial (\Cref{sec certificate completion}). Starting from the target function $f(\x)$, at each
step, we either obtain a certificate with nontrivial correlation or replace the current
intersection by a simpler one by introducing a small $L_1$ error. Importantly, after each
replacement we reconsider the actual decision boundaries of the new intersection, since the
previous boundaries need not be preserved. By the structural results for nearly double roots
and nearly shared roots that we derived above, each unsuccessful step removes or decomposes a
defining quadratic or identifies approximately proportional affine factors, so only $O(k)$ changes
are possible. By carefully controlling the accumulated approximation error and simplifying multiple
constraints together, we obtain the correlation guarantee of \Cref{th certificate}.

\vspace{-0.2cm}

\paragraph{A $d^{O(k)}2^{\poly_k(1/\eps)}$ Time Learning Algorithm.}
The technical ingredients developed thus far are already enough to give us a learning algorithm for
intersections of $k$ degree-$2$ PTFss \new{with $d^{O(k)}$ complexity dependence}.
As discussed above, our goal is to remove the dependence on the
directions that have high influence on $f(\x)$, by randomizing the input along those directions.
\new{Naively, one would simply compute the high influence subspace $V$ for $f$ and then condition
on the projection onto $V$ (or more precisely, divide $V$ into tiny bins and consider separately
the samples that project onto each bin, effectively removing the dependence on the $V$-direction).
Unfortunately, there is no guarantee that conditioning on the projection to $V$ will not cause
there to be new directions of large influence. So instead we must apply this procedure adaptively.
After restricting to a bin, we re-run our algorithm for finding one high influence direction, and
if we find one, we condition further by partitioning the bin into many small bins along the found
direction.}
Intuitively, we maintain a tree rooted at the target function $f$, whose leaves form a partition of
$\R^d$. Each node of the tree is associated with an intersection of $k$ PTFs, obtained by freezing the
input value of the instance at its parent node along a single direction through the above
branching process.

It remains to show that this procedure actually terminates in a reasonable number of steps.
To do this, we consider the $L_2$ norm of the low-degree Hermite projection of $f$.
Specifically, suppose that we branch $f$ along a direction $\v$ and get a series of functions $f^{(1)},\dots,f^{(m)}$, with $\x\cdot \v$ frozen to be different values.
\new{Since the $\v$ influence of the low-degree part of $f$ must be large
and since conditioning on the $\v$ direction effectively decreases the degree of this component,
we will have that $\sum_{i<\ell}\| f_{\le i}\|_2^2$ (where $f_{\le i}$ is the Hermite projection of
$f$ of degree at most $i$) is strictly less than its average over the $f^{(j)}$.
Since this sum is never more than $\ell$, for $\ell=O(k)$,}
the branching operation will terminate after at most $\poly_k(1/\eps)$ rounds.
Using Markov's inequality, we can show that the tree has $2^{\poly_k(1/\eps)}$ many leaves.
Furthermore, most of the leaf functions will have no directions of large influence, so as discussed
above, we may approximate it by an intersection of regular PTFs. Since these polynomials do behave
like Gaussians, we may employ the ideas of \cite{DKS18a} to find a low-dimensional space of
degree-$2$ polynomials that nearly contains the defining polynomials of our intersection. We can
then find these via a brute force search.
As a result, by running the learning algorithm for regular instances $2^{\poly_k(1/\eps)}$ times,
we learn a decision tree, with each leaf associated with an intersection of $k$ PTFs that is
$\poly(\eps)$ close to the target function $f$.

\vspace{-0.2cm}

\paragraph{Improved Algorithm via Hinge Loss Minimization.}
Unfortunately, the tree-based algorithm developed above has a
running time that depends exponentially on a polynomial in
$1/\eps$, since it repeatedly partitions the Gaussian space
along influential directions.
To improve this exponential dependence, we draw inspiration from
prior works (see, e.g.,~\cite{KS04}) that study learning intersections of halfspaces
with $\gamma$-margin.
Roughly speaking, if all examples $\x$ have unit norm and are $\gamma$-far from every decision
boundary of an intersection of $k$ halfspaces $f(\x)$, then there exists a polynomial $p$ of degree
$O_k(\log(1/\gamma))$ such that $\sign(p(\x)) = f(\x)$ for every example in the support of the
distribution. So, a linear programming approach that runs in $d^{O_k(\log(1/\gamma))}$ time can
output a single PTF that is close to $f$.
This approach has two shortcomings. First, such an algorithm has no tolerance to label noise.
Second, it does not work under any continuous distribution, where examples can be arbitrarily close
to the decision boundary. In fact, if we treat our target function $f(\x)$ as an intersection of
$k$ halfspaces in $\R^{d(d+3)/2}$, defined by $\theta_i\cdot \Phi(\x)$, then there is no similar
known way for us to use some polynomial approximation-based algorithm with degree $\poly\log(1/\eps)$.

\new{In \Cref{sec hinge loss}, we give a polynomial approximation that does
not require a margin assumption. For every intersection $f:\R^d\to\{0,1\}$
of $k$ quadratic PTFs, there is a polynomial $p$ of degree
$O_k(\log(1/\eps))$ such that
\[
 \E_{\x\sim\cN(0,\bI)}[\operatorname{Hinge}(2f(\x)-1,p(\x))]\le\eps,
 \qquad \|p\|_2\le 2^{O_k(\log^2(1/\eps))}.
\]
Here $\operatorname{Hinge}(y,t)=\max\{0,1-yt\}$ for $y\in\{\pm1\}$.}
Our construction is based on rational approximations of the
sign function. By combining these rational approximations
and clearing their positive denominators, we obtain a
polynomial that has a large margin away from the decision
boundaries. Gaussian anti-concentration and tail bounds
then allow us to control its expected hinge loss and
$L_2$ norm. Since hinge loss is convex in the polynomial
coefficients, we can learn such a polynomial efficiently
using stochastic convex optimization over the space of monomials.
Importantly, we show that the sign of the learned polynomial is $\eps$-close to the target function $f(\x)$.

However, directly performing this optimization over all
degree-$O_k(\log(1/\eps))$ polynomials in $\R^d$ would
require $d^{O_k(\log(1/\eps))}$ time, which is even less desirable than our previous $d^{O(k)}2^{\poly_k(1/\eps)}$ time algorithm.
The reason is that the search space has too large a dimension.
\new{Suppose instead that we are given a low-dimensional subspace
$\mathcal H\subseteq\mathcal P_{\le 2}(\R^d)$ such that the target
function $f$ can be represented as an intersection of $k$ quadratic
PTFs with defining polynomials in $\mathcal H$.}
Then by running the hinge loss minimization algorithm over $\mathcal{H}$, we get a $\dim(\mathcal{H})^{O_k(\log(1/\eps))}$ time algorithm.
Unfortunately, such a subspace is not available to us. And even if we do have one such a subspace, we can usually only promise that some intersection of PTFs from the subspace is close to $f$, which means the hinge loss minimization algorithm has to tolerate label noise.
In fact, due to the norm bound of the approximate polynomial, we need a much
stronger approximation guarantee than $\eps$ hinge loss error to make the algorithm noise tolerant.
If $\Pr_{\x\sim\cN(0,\bI)}[f(\x)\ne g(\x)]\le \rho$,
then the additional hinge loss incurred by a polynomial
of norm at most $W$ can be bounded by $2W\sqrt{\rho}$.
Thus, it suffices to find a space $\mathcal H$ that contains
an intersection with error $\rho=2^{-O_k(\log^2(1/\eps))}$.
Our goal is to find such a space with dimension at most
$2^{O_k(\log^2(1/\eps))}$, while maintaining a running
time of $d^{O(k)}$ times a quasipolynomial function of
$1/\eps$.

\paragraph{A Two-Stage Dimension Reduction Approach.}

To achieve this, we develop a two-stage dimension-reduction
procedure in \Cref{sec quasi}.
In the first stage, we apply \Cref{th certificate} and \Cref{th direction finding one shot} to find
a subspace $V\subseteq\R^d$ of dimension
$\operatorname{poly}_k(1/\rho)$, such that $f$ has
sufficiently small randomized error along every direction
orthogonal to $V$. Writing $\x=\z+\w$, where
$\z=\proj_V(\x)$ and $\w=\proj_{V^\perp}(\x)$.
For each defining polynomial $q_i$, we write
\begin{align*}
    & q_i(\x)  = a_i(\z) + (\bb_i + \bL_i^\top\z)^\top \w + \w^\top \mathbf{A}_i \w - \tr(\mathbf{A}_i) \\
    & N_{\u_i}(\z,\w)  = (\bb_i + \bL_i^\top\z)^\top \w + \w^\top \mathbf{A}_i \w - \tr(\mathbf{A}_i),
\end{align*}
where $a_i(\z)$ is a quadratic function that only depends on $\z$, and $\bA_i$ has no components in $V$,
while $\u_i = (\bb_i,\bL_i,\mathbf{A}_i)$ is the vector of coefficients of $N_{\u_i}$.
However, finding such a subspace $V$ does not mean
that the target is close to an intersection supported
only on $V$. Although every individual direction
$\v\in V^\perp$ has small randomized error,
the directions in $V^\perp$ may still collectively
have a significant influence on the target.
In terms of the above decomposition, this means
that the residual polynomials
$N_{\u_i}(\z,\w)$ cannot simply be discarded.
On the other hand, the space $\mathcal{S}$ spanned by polynomials of the form $N_\u(\z,\w)$ might
still have dimension $\Omega(d^2)$. As a result, we need a second stage of dimension reduction
to find a low-dimensional subspace of $\mathcal{S}$ that captures the influence
of these residual polynomials $N_{\u_i}, i \in [k]$.

The central idea that we use for this second stage of dimension reduction is based on a more
general notion of regularity and Gaussian comparison lemma
developed in \Cref{sec:gaussian-comparison}.
Specifically, let $V\subseteq \R^d$ be a subspace.
We say that a quadratic polynomial $p$ is $(\tau,V)$ regular,
if $\norm{\partial_\v p}^2 \le \tau^2 \var_{\x \sim \cN(0,\bI)}[p(\x)]$
for every unit vector $\v \perp V$.
In other words, the full regularity condition can be seen as a special case
of $(\tau,V)$ regularity, where we take $V = \{0\} $.
Unlike the tree-based algorithm, the $(\tau,V)$-regularity condition allows us
to flexibly keep the dependence on $\z$ instead of fixing its value.
In fact, the regularity condition guarantees that the distribution of
$(\z, N_{\u_1}(\z,\w),\dots,N_{\u_k}(\z,\w))$, $\z\sim \cN(0,\bI_V),\w \sim \cN(0,\bI_{V^\perp})$
approximates its Gaussian replacement $(\z,\u_1\cdot G,\dots, \u_k\cdot G)$.
This allows us to perform our analysis over a Gaussian model,
where we replace each $N_{\u_i}(\z,\w)$ by $\langle \u_i,G\rangle$,
for $G$ an independent standard Gaussian vector in the residual coefficient space.
The resulting function
\[
\tilde f(\z,G)
=\1\{\tilde q_i:= a_i(\z)+\langle \u_i,G\rangle\ge0,
\ \forall i\in[k]\}
\]
is an intersection of halfspaces in $G$ for every fixed
$\z$.
As we show in \Cref{lem:gaussian-comparision}, if $q_1,\dots,q_k$ are all $(\tau,V)$ regular, then for any smooth function $\Psi:\R^k \to \R$,
\begin{align*}
    \E_{\x\sim \cN(0,\bI)}[\Psi(q_1(\x),\dots,q_k(\x))] \approx\E_{\z,G}[\Psi(\tilde q_1(\z,G),\dots,\tilde q_k(\z,G))].
\end{align*}
Suppose that given sample access to $\tilde f$, we would like to identify
directions in $G$-space that are important for $\tilde f$.
In fact, by \Cref{th error to regularity}, the property of $V$ ensures that $f$ must be sufficiently close to an instance that is $(\tau,V)$-regular.
So, if this can be done, then the intuition is that we could potentially make use of the Gaussian
comparison techniques to identify the subspace spanned by the $N_{\u_i}$.
\new{For each fixed $\z$, the first two labeled moments in $G$ suffice to
find a subspace on which an intersection of halfspaces approximates
$\tilde f(\z,\cdot)$, as in~\cite{DKS18a}.}
\new{Unfortunately, applying this idea directly would require being able to take measurements conditional on $\z$, which can only be approximated using exponential in $\dim(V)$ samples.}
However, the nice structure of $\tilde f$ still makes it possible to achieve this goal
by using $O(k)$ many moments.
Specifically, in \Cref{lm structre idealized gaussian} we show that for every direction $\v$
in the $G$ space that has non-trivial influence on $\tilde f$,
there exists a polynomial $p$ of degree $2k$ such that
$\E_{\z,G}[\tilde f M_\v p]$ is non-trivially large.
This gives a bilinear form with input $\v$ and polynomial $p$.
In fact, using integration by parts,
\begin{align}\label{eq gaussian space}
    \E_{\z,G}[\tilde f(\z,G)M_\v p] = \E_{\z,G}[p \partial_\v \tilde f] = \sum_{i=1}^k (\v \cdot \u_i) \E_{\z,G} [p\frac{\partial \tilde f}{\partial \tilde q_i}],
\end{align}
we know that such a bilinear form has at most $k$ non-zero singular values. Thus, taking the subspace spanned by the top $k$ singular vectors of the bilinear form, we find all directions in $G$ space that are important to $\tilde f$.

However, some additional work is needed to apply these ideas.
First, we need to ensure that the Gaussian comparison techniques
developed in \Cref{sec:gaussian-comparison} suffice to imply
large singular values of the aforementioned bilinear form.
This is a particular issue as we only have a comparison for functions of the form $\Psi(q_1,\dots,q_k)$, not an arbitrary smooth function support over $\R^d$.
\new{Second, we work with a space $\mathcal{S}$ of polynomials of the form
$N_\u(\z,\w)$ and have access only to labeled examples $(\x,f(\x))$.
Our bilinear form involves a directional derivative along $\v$ in the
Gaussian replacement space, and it is not clear how to define the
corresponding operation on $\mathcal{S}$.}
We address these two issues separately.
First, the certificate polynomial $p$ that we developed in \Cref{lm structre idealized gaussian}
is a sum of products of the defining polynomials $\tilde q_i$, not an arbitrary polynomial.
Thus, each term in the summation of the bilinear form \eqref{eq gaussian space}
actually has the desired form of $\Psi(q_1,\dots,q_k)$.
\new{
Let $f_h$ be a smooth approximation of $f$. It suffices to construct a
bilinear form $B_f$ such that
\[
B_f(\u,p) \approx
\sum_{i=1}^k \langle \u,\u_i\rangle
\E\left[p\,\frac{\partial f_h}{\partial q_i}\right]
\]
uniformly over unit residual coefficient directions $\u$ and unit-norm
polynomials $p$ of degree at most $2k$. Let $\bar V$ be the span of
the top $k$ left singular vectors of $B_f$. Then $f$ is close to an
intersection whose defining polynomials lie in
\[
\mathcal H
=\mathcal P_{\le 2}(V)+\{N_\u:\u\in\bar V\} \;.
\]
}
It remains to address the second issue of how to build such a $B_f$.
The observation is that under the Gaussian distribution
$\E_{\x \sim \cN(0,\bI)}[\partial_\v f] = \E_{\x \sim \cN(0,\bI)}[(\v\cdot\x)f]$,
which means that the behavior of $\partial_\v f$ can be simulated through
standard polynomial multiplication under the Gaussian distribution.
The hope is that if we replace $\v$ by $N_\v(\z,\w)$ in the multiplication,
then this would satisfy our purpose. In fact, using standard integration by parts,
\begin{align*}
\E_{(\z,\w)\sim \cN(0,\bI)}[N_\v(\z,\w)f]=
\E_{(\z,\w)\sim \cN(0,\bI)}[(\bb + \bL^\top\z)^\top \nabla_\w f + \tr(\mathbf{A}\nabla^2_\w f)] \;.
\end{align*}
In \Cref{lm derivative operator}, if we denote by
$d_\v(f) = (\bb + \bL^\top\z)^\top \nabla_\w f + \tr(\mathbf{A}\nabla^2_\w f)$,
then under our regularity condition, we actually have that
$\E_{\z,\w}[p d_\u f]$ is sufficiently close to $\sum_{i=1}^k
\langle \u,\u_i\rangle
\E\left[
p\,\frac{\partial f_h}{\partial q_i}
\right]$ and can be used as $B_f$.
But since we do not have direct access to $f$ and cannot compute its gradients or Hessian,
we still need an equivalent representation of $\E_{\z,\w}[p d_\u f]$
that can be directly computed from samples. That is, we want to find some operator $T_\u$
such that $\E_{\z,\w}[fT_\u p] = \E_{\z,\w}[p d_\u f]$.
Such an operator can be reverse-engineered using integration by parts.
In \Cref{lm derivative operator construction}, we show that
$T_\v p =pN_\v(\z,\w)-(\bb+\bL^\top\z+2\mathbf A\w)^\top\nabla_\w p +\tr(\mathbf A\nabla_\w^2p),$
satisfies this property. Thus, we can find a subspace $\mathcal{H}$ of polynomials,
where an intersection of $k$ degree-$2$ PTFs supported on it is sufficiently close to $f$.

Finally, we apply the hinge-loss minimization algorithm
of \Cref{sec hinge loss} over the recovered space $\mathcal H$.
Since
\[
\dim(\mathcal H)
\le O(\dim(V)^2+k)
=2^{O_k(\log^2(1/\eps))} \;,
\]
and we only need polynomials of degree
$O_k(\log(1/\eps))$ in a basis of $\mathcal H$,
the resulting feature space has dimension at most $2^{O_k(\log^3(1/\eps))}.$
Combining the two dimension-reduction stages with
hinge-loss minimization gives our final learning
algorithm with running time $d^{O(k)}2^{O_k(\log^3(1/\eps))}.$

\vspace{-0.2cm}

\paragraph{Extensions and Open Problems.}
A natural open question is whether our quasipolynomial dependence on $1/\epsilon$ can be
improved to polynomial dependence for every fixed $k$, yielding a running time of
$d^{O(k)}(1/\epsilon)^{O_k(1)}$. It is a plausible conjecture that such an algorithm is
possible. A second direction concerns higher-degree PTFs.
Our hinge-loss approach extends to intersections of $k$ degree-$r$ PTFs
under the standard Gaussian distribution, yielding
a $d^{O_{k,r}(\log^2(1/\epsilon))}$-time learning algorithm for any fixed degree $r$.
It remains open whether our structural and dimension-reduction machinery can be
generalized beyond degree two to obtain a running time polynomial in $d$, with an exponent
depending only on $k$ and $r$ and not on $\epsilon$. More broadly, it would be interesting
to identify natural regularity or complexity conditions on the link function $g$ under
which functions of the form $f(x)=g(p_1(x),\ldots,p_k(x))$, with quadratic $p_i$, can be
learned efficiently, and to understand which of our techniques extend beyond the
intersection structure.

\vspace{-0.2cm}

\subsection{Related Work}

\vspace{-0.1cm}

\paragraph{Learning intersections of halfspaces.}
Understanding the computational complexity of learning intersections of
halfspaces is one of the central problems in computational learning theory. In
the distribution-free setting, various computational hardness results have
been established, such as hardness of proper learning
\cite{blum1988training,khot2008hardness}, representation-based
hardness~\cite{sherstov2010optimal,sherstov2013intersection},
cryptographic and Statistical Query hardness on learning
a super-constant number of halfspaces~\cite{klivans2007unconditional,klivans2009cryptographic,daniely2016complexity,tiegel2024improved}.
However, whether an intersection of two halfspaces can be learned
in polynomial time in the distribution-free setting remains open.
Motivated by this challenge, a line of work has focused on designing
learning algorithms under structured distributions, such as learning two
homogeneous halfspaces under symmetric distributions \cite{Baum:91},
low-weighted intersections \cite{KOS:04},
learning under margin assumptions \cite{KS04}
and under factorizable distributions \cite{DMRT25}. In particular, under the uniform distribution or Gaussian distributions,
efficient learning algorithms are known for learning an intersection of
a constant number of halfspaces \cite{blum1997learning,Vempala10a}.
These algorithms can also be made robust against adversarial noise~\cite{DKS18a}.
Under more general logconcave distributions, \cite{Vempala10} gives similar algorithms for learning intersections of homogeneous halfspaces,
and more recently \cite{diakonikolas2026robust} gave approximate agnostic learners
for functions of constantly many halfspaces under general logconcave marginals.

\vspace{-0.2cm}

\paragraph{Multi-index models and beyond.}
The moment-based subspace recovery method for learning intersections of halfspaces has also inspired a rich line of work on the complexity of
learning functions of a small number of linear projections (MIMs)
under structured distributions, such as shallow ReLU networks \cite{diakonikolas2020algorithms,chen2023learning,diakonikolas2024efficiently}, multiclass linear classifiers \cite{diakonikolas2025statistical,diakonikolas2026polynomial}
and general multi-index models \cite{ABM23,DLB25,DiakonikolasIKZ25,DIKR25}.
In particular, \cite{DiakonikolasIKZ25,DIKR25,diakonikolas2026robust} give
general frameworks for robustly learning multi-index models under Gaussian and
logconcave distributions. We also refer to~\cite{BH25} for a recent survey.

Beyond multi-index models, little is known
about efficient learning with multiple nonlinear features. Recent works have
made initial progress on learning structured hierarchical functions using
$3$-layer neural networks, including functions of a single polynomial
feature~\cite{NDL23,WNL24} and polynomial functions of multiple quadratic
features~\cite{FWNL25}. These guarantees rely on
structural and nondegeneracy assumptions on the underlying features and link
functions and do not extend to intersections of quadratic PTFs.
In contrast, our work develops structural and algorithmic tools for learning
intersections of arbitrary quadratic PTFs without additional
assumptions.

\vspace{-0.2cm}

\paragraph{Regularity and pseudorandomness for PTFs.}
A substantial line of work has developed regularity, decomposition, and
Gaussian approximation techniques for analyzing low-degree polynomial threshold
functions; see, e.g.,~\cite{diakonikolas2010bounding,diakonikolas2010regularity,MZstoc10,de2014deterministic,DeS14}.
A recurring theme is to separate the structured components of the defining
polynomials from sufficiently regular components. These techniques have played
an important role in deterministic approximate counting and pseudorandomness,
including for arbitrary Boolean functions of multiple
PTFs~\cite{diakonikolas2010bounded,de2014deterministic}.
In our setting, the defining polynomials are unknown. We must recover,
from labeled examples, a space of quadratic features that supports an
accurate approximation to the target. Our Gaussian comparison controls
the labeled moments used to recover this space.

\paragraph{AI Disclosure.}
The main framework and algorithmic results of this paper were
developed by the authors. In particular, the authors independently
established \Cref{th exponential time} and developed the hinge-loss
minimization framework of \Cref{sec hinge loss}.
The authors also observed that combining this framework with
a suitable dimension-reduction method could improve the running
time to $d^{O(k)}2^{\poly_k(\log(1/\eps))}$.

To achieve this improvement, the authors first investigated an
approach based on a stronger regularity lemma and
showed that it does not hold in the required generality.
The authors subsequently interacted with GPT-6 Astra to explore alternative
dimension-reduction strategies. These interactions led to an
initial version of the approach
using a slightly weaker notion of regularity, underlying \Cref{sec quasi}.
The authors then substantially simplified and refined this
approach, established its connection to the framework they had
previously developed, and obtained the final algorithm and
analysis presented in the paper.

During manuscript preparation, GPT-6.1 was also used to improve
the clarity of the exposition, check local proof arguments,
and identify errors in calculations in earlier drafts.

\paragraph{Organization of the Paper.}

In \Cref{sec notation}, we list basic notations used throughout the paper. In \Cref{sec:gaussian-comparison}, we introduce the notion of $(\tau,V)$-regularity for quadratic polynomials and develop a Gaussian comparison lemma based on the regularity condition. In \Cref{sec regular case reduction}, we show that any intersection of $k$ degree two PTFs who has no high influence directions in $V^\perp$ must be close to a regular instance. In \Cref{sec certificate polynomial}, we show our main structural result that a direction with high influence on the target function must have influence on its low degree part. In \Cref{sec direction finding}, we develop an efficient algorithm to find these high-influence directions using the structural result.
In \Cref{sec tree building}, we develop a $d^{O(k)}2^{\poly_k(1/\eps)}$ time learning algorithm based on the developed technical ingredients.
In \Cref{sec hinge loss}, we develop a robust hinge loss minimization algorithm that will be used to improve the dependence on $1/\eps
$. In \Cref{sec quasi}, we develop a two-stage dimension reduction method and complete the main learning algorithm. In \Cref{sec sq}, we provide a matching SQ lower bound.

\section{Preliminaries and Notations}\label{sec notation}

\paragraph{Notations}
Let $k \in \Z_+$, we use $[k]$ to denote $\{1,\dots,k\}$.
We use the convention $\sign(t)=1$ for $t\ge 0$ and
$\sign(t)=-1$ for $t<0$.
When passing between $\{0,1\}$-valued and $\{\pm1\}$-valued
functions, we use the correspondence $f\mapsto 2f-1$,
with inverse $g\mapsto (g+1)/2$.
Thus, in either convention, membership in an intersection is
represented by the label $1$.
The notation $O_k(\cdot)$ allows the implicit constant to depend
on $k$.
We use $\poly_k$ for an unspecified polynomial dependence whose
degree and constants may depend on $k$.
In a positive lower bound with argument $t\in(0,1)$, this denotes
a quantity of the form $c_k t^{C_k}$ for positive constants
$c_k,C_k$ depending only on $k$.
For $t\ge 1$, $\mathrm{quasipoly}_k(t)$ denotes a bound of the form
$2^{(O_k(\log(t))}$.

We use small boldface characters for vectors and capital bold characters for matrices and tensors.
For $\x \in  \R^d$ and $i \in [d]$, $\x_i$ denotes the $i$-th coordinate of $\x$.
Let $V$ be a subspace. We use $V^\perp$ to denote its orthogonal complement, and
use $\proj_V(\x)$ to denote the projection of $\x$ onto subspace $V$. We will also use $\x_V$ for short to denote $\proj_V(\x)$.
We use $\s^{d-1}$ to denote the unit sphere over $\R^d$

Unless otherwise specified, function norms are taken
under the standard Gaussian distribution on the relevant space.
For a function $f:\R^d\to\R$, we write
$\norm{f}_2=(\E_{\x\sim\cN(0,\bI)}[f(\x)^2])^{1/2}$.
For vectors, $\norm{\cdot}_2$ denotes the Euclidean norm; for matrices,
$\norm{\cdot}_{\op}$ and $\norm{\cdot}_F$ denote the operator and
Frobenius norms, respectively.
For convenience, we will also use $\norm{\cdot}$ to denote $\ell_2(L_2)$ norm.

\paragraph{Boolean Functions and Randomized Error}

\begin{definition}[Randomized Error]\label{def randmozied error}
    Let $f: \R^d \to \{0,1\}$ be a Boolean function and let $\v \in \s^{d-1}$ be a unit vector. We define the randomized error of $f$ along direction $\v$ to be
    \begin{align*}
        \rand_f(\v) = \E_{\bw \sim \cN(0,\bI_{\v^\perp})}\left[ \Pr_{z,z' \sim \cN(0,1)}[f(\w+z \v) \neq f(\w+z' \v)] \right],
    \end{align*}
    where $\w \sim \cN(0,\bI_{\v^\perp}) \perp \v$.
\end{definition}
Intuitively, $\rand_f(\v)$ quantifies the error of $f$ under Gaussian distribution when we rerandomize the direction along $\v$ but fix the other directions.
We give the following proposition of the randomized error.

\begin{proposition}\label{prop randomized error}
     Let $f: \R^d \to \{0,1\}$ be a Boolean function and let $\v \in \s^{d-1}$ be a unit vector. Then,
     \begin{align*}
         \rand_f(\v) = 2 \E_{\bw \sim \cN(0,\bI)}\left[ \Var_{z\sim \cN(0,1)}[f(\w+z \v)] \right],
     \end{align*}
      where $\w \sim \cN(0,\bI) \perp \v$.
\end{proposition}

\begin{proof}[Proof of \Cref{prop randomized error}]
    We first fix $\w \perp \v$. Since $z,z'$ are independent Gaussian, we have
    \begin{align*}
        & \Pr_{z,z' \sim \cN(0,1)}[f(\w+z \v) \neq f(\w+z' \v)] \\
        &= \Pr_{z,z' \sim \cN(0,1)}[f(\w+z \v) =1, f(\w+z' \v) = 0] + \Pr_{z,z' \sim \cN(0,1)}[f(\w+z \v) =0, f(\w+z' \v) = 1] \\
        & = 2 \Pr_{z\sim \cN(0,1)}[f(\w+z \v) =1]\Pr_{z\sim \cN(0,1)}[f(\w+z \v) =0] \\
        & = 2 \Var_{z\sim \cN(0,1)}[f(\w+z \v)].
    \end{align*}
    Since $\w$ is independent of $z$, by taking the expectation over $\w$, we have
    \begin{align*}
        \rand_f(\v) = 2 \E_{\bw \sim \cN(0,\bI)}\left[ \Var_{z\sim \cN(0,1)}[f(\w+z \v)] \right].
    \end{align*}
\end{proof}

\begin{definition}[Layer Decomposition]
    Let $f: \R^d \to \R$ be a function. Let $\v \in \s^{d-1}$ be a unit vector and let $\w \perp \v$. We define the one dimensional function $f_\w: \R \to \R$ as $f_\w(s) = f(s\v + \w)$. For a polynomial $q: \R^d \to \R$, we also use $q^{(\w)}$ to denote $q_\w$.
\end{definition}

\paragraph{Hermite polynomials.}
For each integer $r\ge 0$, let $h_r$ denote the normalized
probabilists' Hermite polynomial of degree $r$, with positive
leading coefficient.
In particular,
$h_0(t)=1$, $h_1(t)=t$, and $h_2(t)=(t^2-1)/\sqrt{2}$.
These polynomials satisfy
\[
    \E_{Z\sim\cN(0,1)}[h_r(Z)h_s(Z)]
    = \Ind\{r=s\},
    \qquad
    h_r'(t)=\sqrt{r}\,h_{r-1}(t)
    \quad (r\ge 1).
\]
For a multi-index $\alpha\in\Z_{\ge 0}^d$, write
$|\alpha|=\sum_{j=1}^d\alpha_j$ and define
$H_\alpha(\x)=\prod_{j=1}^d h_{\alpha_j}(x_j)$.
The polynomials $H_\alpha$ form an orthonormal basis for the
space of square-integrable functions under $\cN(0,\bI_d)$, and
$H_\alpha$ has total degree $|\alpha|$.

\begin{definition}[Hermite Projection]
Let $k \in \N$, and $V \subseteq \R^d$ be a subspace. We denote by $P_{\le k}(V)$ the space of polynomials supported on $V$ with degree no more than $k$.
    Let $f: \R^d \to \R$ be a function. We denote by $f_{\le k}$ to be the projection of $f$ onto the space spanned by Hermite polynomials with degree no larger than $k$.
\end{definition}

\paragraph{Gaussian polynomial inequalities.}
For $q\ge 1$, we write
$\norm{P}_q=(\E_{\x\sim\cN(0,\bI)}[\abs{P(\x)}^q])^{1/q}$.
We will use the following Gaussian polynomial inequalities throughout.

\begin{fact}[Gaussian hypercontractivity, see, e.g.,~{\cite[Chapter 11]{Don14}}]
\label{fct:gaussian-hypercontractivity}
Let $P:\R^d\to\R$ be a polynomial of degree at most $r$.
For every real $q\ge 2$,
$\norm{P}_q \le (q-1)^{r/2}\norm{P}_2$.
In particular, there are universal constants $c,C>0$ such that
if $\deg(P)\le 2$, then, for every $t\ge 1$,
$\Pr_{\x\sim\cN(0,\bI)}
    [\abs{P(\x)}>t\norm{P}_2]
    \le C\exp(-ct)$.
\end{fact}

\begin{fact}[Gaussian anti-concentration {\cite[Theorem 8]{CW:01}}]
\label{fct:gaussian-anticoncentration}
There is a universal constant $C>0$ such that the following holds.
Let $r\ge 1$ be an integer and let $P:\R^d\to\R$ be a nonzero
polynomial of degree at most $r$.
For every $t>0$ and $a\in\{1,2\}$,
    $\Pr_{\x\sim\cN(0,\bI)}
    [\abs{P(\x)}\le t\norm{P}_a]
    \le Crt^{1/r}$.
\end{fact}

\begin{fact}[Gaussian product-norm bounds]
\label{fct:gaussian-product-norm}
For every pair of nonnegative integers $r,s$, there are constants
$c_{r,s},C_{r,s}>0$, depending only on $r,s$, such that
    $c_{r,s}\norm{P}_2\norm{Q}_2
    \le \norm{PQ}_2
    \le C_{r,s}\norm{P}_2\norm{Q}_2$
for all polynomials $P,Q:\R^d\to\R$ of degrees at most $r,s$,
respectively.
\end{fact}

\begin{proof}[Proof of \Cref{fct:gaussian-product-norm}]
By Cauchy--Schwarz and \Cref{fct:gaussian-hypercontractivity},
$\norm{PQ}_2\le\norm{P}_4\norm{Q}_4
\le 3^{(r+s)/2}\norm{P}_2\norm{Q}_2$.

For the lower bound, the claim is immediate if either polynomial
is constant. Otherwise, \Cref{fct:gaussian-anticoncentration}
provides constants $a_r,a_s>0$, depending only on $r,s$, such that
\[
    \Pr[\abs{P}<a_r\norm{P}_2]\le 1/4,
    \qquad
    \Pr[\abs{Q}<a_s\norm{Q}_2]\le 1/4.
\]
With probability at least $1/2$, both complementary inequalities
hold. Hence
$\norm{PQ}_2^2\ge
a_r^2a_s^2\norm{P}_2^2\norm{Q}_2^2/2$,
which proves the lower bound.
\end{proof}

\section{Gaussian Approximation via Regularity}\label{sec:gaussian-comparison}

In this section, we develop Gaussian comparison tools that allow us to reduce the learning of regular intersections of degree-$2$ PTFs to learning intersections of halfspaces under Gaussian distributions.
The starting observation is that every degree-$2$ polynomial can be represented as a linear function of the degree-$1$ and degree-$2$ Hermite features.
Thus, replacing the lifted feature vector by a standard Gaussian vector transforms an intersection of degree-$2$ PTFs into an intersection of halfspaces, for which powerful dimension-reduction techniques are available.
The main challenge is to justify this replacement in a way that preserves the information needed for learning.
In particular, we need to show that the relevant low-degree moments are preserved sufficiently well to recover a useful subspace, and that a subspace capturing the Gaussian surrogate also yields a good approximation to the original target.

Our comparison results are developed in a more general setting that will also be useful in later sections.
Specifically, we consider $(\tau,V)$-regular polynomials, which may have unrestricted dependence on a fixed subspace $V$ but satisfy regularity conditions in the orthogonal directions.
The Gaussian replacement then preserves the original coordinates in $V$ and acts only on the remaining polynomial contributions.
The fully regular case corresponds to $V=\{0\}$.
We first establish the moment-comparison and subspace-transfer guarantees in this general setting.
We then specialize to the fully regular case and combine these results with the Gaussian halfspace recovery algorithm of \cite{DKS18a} to obtain an efficient learner for regular intersections of degree-$2$ PTFs.

\begin{definition}[$(\tau,V)$-regularity]
\label{def:subspace-regular}
Let $\tau\in(0,1)$, let $V\subseteq\R^d$ be a subspace, and let
$p:\R^d\to\R$ be a polynomial of degree at most two. For a unit vector
$\bv\in\R^d$, define its directional influence by
$\inf_p(\bv)
    \eqdef
    \E_{\bx\sim\normal(0,\bI_d)}
    \left[(\bv\cdot\nabla p(\bx))^2\right]$.
We say that $p$ is $(\tau,V)$-regular if
$\inf_p(\bv)\le
\tau^2\var_{\bx\sim\normal(0,\bI_d)}[p(\bx)]$
for every unit vector $\bv\in V^\perp$.
We call $p$ $\tau$-regular when it is $(\tau,\{0\})$-regular.
\end{definition}
\new{Regularity ensures that no input direction has a large influence relative to the total variance. In the fully regular case, an orthogonal change of coordinates expresses the centered polynomial as a sum of independent terms, each contributing only a small fraction of the polynomial's variance. This gives a central-limit intuition for approximating the polynomial by a Gaussian with the same mean and variance.}

\paragraph{Notation and Gaussian replacement.}
Throughout this section, for $\bx\sim\normal(0,\bI_d)$, we use
$\bz=\proj_V\bx$ and $\bw=\proj_{V^\perp}\bx$ to denote its components
in $V$ and $V^\perp$, respectively, so that $\bx=\bz+\bw$.
Polynomial $L_2$ norms are taken under the corresponding standard
Gaussian distribution. Let $\Phi_2:\R^d\to\R^N$, where
$N=d(d+3)/2$, be the vector of normalized Gaussian Hermite
polynomials of degrees one and two. These features have mean zero
and satisfy
$\E[\Phi_2(\bx)\Phi_2(\bx)^\top]=\bI_N$.

Every polynomial $p_i$ of degree at most two has a decomposition
\[
    p_i(\bx)
    =
    q_i(\bz)
    +\bb_i^\top\bw
    +\bz^\top\bL_i\bw
    +\bw^\top\bA_i\bw-\tr(\bA_i),
\]
where $\bb_i\in V^\perp$, $\bL_i:V^\perp\to V$ is linear, and
$\bA_i:V^\perp\to V^\perp$ is symmetric. Here
$q_i(\bz)=\E[p_i(\bx)\mid\bz]$, so the remaining terms have
conditional mean zero given $\bz$.
The following fact translates regularity into bounds on these
coefficients.

\begin{fact}[Coefficient bounds from regularity]
\label{fct:regular-to-coefficient}
Suppose $p_i$ is $(\tau,V)$-regular, and let
$\sigma_i^2\eqdef\var_{\bx}[p_i(\bx)]$.
Then the coefficients in the decomposition above satisfy
$\norm{\bb_i}_2\le\tau\sigma_i$,
$\norm{\bL_i}_\op\le\tau\sigma_i$, and
$\norm{\bA_i}_\op\le\tau\sigma_i/2$.
Moreover,
$\norm{\bb_i}_2^2+\norm{\bL_i}_F^2
+2\norm{\bA_i}_F^2\le\sigma_i^2$.
\end{fact}

\begin{proof}[Proof of \Cref{fct:regular-to-coefficient}]
For every unit vector $\bv\in V^\perp$, differentiating the
decomposition and using the independence of $\bz$ and $\bw$ gives
$
    \E_{\bx}\left[(\bv\cdot\nabla p_i(\bx))^2\right]
    =(\bb_i^\top\bv)^2+\norm{\bL_i\bv}_2^2
     +4\norm{\bA_i\bv}_2^2
    \le\tau^2\sigma_i^2$,
where the inequality follows from regularity.
Taking suprema over unit vectors $\bv\in V^\perp$ yields the
coefficient bounds. Finally, Gaussian orthogonality gives
$\sigma_i^2=\var_{\bz}[q_i(\bz)]
+\norm{\bb_i}_2^2+\norm{\bL_i}_F^2
+2\norm{\bA_i}_F^2$, proving the last bound.
\end{proof}
Let
$    S_V
    \eqdef
    \left\{
        \boldsymbol{\theta}\in\R^N:
        \E\left[
            \boldsymbol{\theta}\cdot\Phi_2(\bx)
            \mid\bz
        \right]=0
    \right\}$
be the subspace of residual coefficient vectors. We can therefore
write
$p_i(\bx)=q_i(\bz)+\boldsymbol{\theta}_i\cdot\Phi_2(\bx)$
for a unique $\boldsymbol{\theta}_i\in S_V$.
Throughout this section, we use $\vec{G}\sim\normal(0,\bI_N)$,
independent of $\bx$, for the Gaussian replacement vector. Define
the Gaussian replacement of $p_i$ by
$\wt p_i(\bz,\vec{G})
    \eqdef
    q_i(\bz)+\boldsymbol{\theta}_i\cdot\vec{G}$.
The same vector $\vec{G}$ is used for all $i$, preserving the pairwise
covariances of the residual contributions. For an intersection
$f(\bx)=\Ind\{p_i(\bx)\ge0\text{ for all }i\in[k]\}$, write
$\wt f(\bz,\vec{G})
=\Ind\{\wt p_i(\bz,\vec{G})\ge0\text{ for all }i\in[k]\}$
for its Gaussian surrogate. When $V=\{0\}$, each $q_i$ is constant,
and $\wt f$ is an intersection of $k$ halfspaces under a standard
Gaussian distribution.

\subsection{Gaussian Moment Comparison}

We now derive the Hermite moment comparison, treating $m=0,1,2$ in turn.

\begin{lemma}[Gaussian Hermite moment comparison]
\label{lem:gaussian-hermite-moment-comparison}
Let $\tau\in(0,1)$, and let $p_1,\ldots,p_k$ have the
decompositions and Gaussian replacements defined above.
For every $i\in[k]$, assume that $\norm{p_i}_2=1$ and that
the coefficient norms $\norm{\bb_i}_2$, $\norm{\bL_i}_\op$,
and $\norm{\bA_i}_\op$ are all at most $\tau$.
Let $\psi:\R^k\to\R$ be a $C^4$ function
$D_r(\psi)\eqdef
\max_{i_1,\ldots,i_r\in[k]}
\norm{\partial_{i_1}\cdots\partial_{i_r}\psi}_\infty
<\infty$ for $r\in[4]$.

For an integrable function $F$ of $\bx$, define
$\mathcal{P}_V F\eqdef F-\E[F\mid\bz]+\E[F]$.
Then, for every unit vector $\bv\in S_V$ and every
$m\in\{0,1,2\}$,
\[
\begin{aligned}
\Bigl|
\E_{\bx}\!\left[
    \psi(p_1(\bx),\ldots,p_k(\bx))\,
    \mathcal{P}_V\!\left[h_m(\bv\cdot\Phi_2(\bx))\right]
\right]
-
\E_{\bz,\vec{G}}\!\left[
    \psi(\wt p_1(\bz,\vec{G}),\ldots,\wt p_k(\bz,\vec{G}))\,
    h_m(\bv\cdot\vec{G})
\right]
\Bigr|
\le\eps_m,
\end{aligned}
\]
where $h_m$ is the normalized univariate Hermite polynomial
of degree $m$, and
$\eps_0
=O\!\left(k^2\tau D_2(\psi)\right)$,
$\eps_1
=O\!\left(k^3\tau\max_{1\le r\le3}D_r(\psi)\right)$,
$\eps_2
=O\!\left(k^4\tau\max_{1\le r\le4}D_r(\psi)\right)$.
\end{lemma}

\new{The role of the comparison is easiest to see in the fully regular case $V=\{0\}$. The Gaussian surrogate is then an intersection of halfspaces, whose first two labeled moments provide the information needed for subspace recovery. To compare these moments, we work with a smooth function $\psi$ of the defining polynomials and consider its correlations with the Hermite weights $h_m(\bv\cdot\Phi_2(\bx))$, for $m\in\{0,1,2\}$. These capture the expectation and the directional first and centered second moments, respectively. Our goal is to show that regularity preserves these quantities under Gaussian replacement, with error controlled by $\tau$ and the derivatives of $\psi$.

We first establish the smooth comparison below, corresponding to the case without a Hermite weight. We gradually replace the residual polynomial contributions by a Gaussian vector with the same covariance, keeping the variables in $V$ unchanged. After Gaussian integration by parts, the matching covariance terms cancel, and regularity controls the remaining error. To compare the first and second moments, we use integration by parts again to express the weighted expectations in terms of derivatives of $\psi$, up to similarly small errors, and apply the smooth comparison to these derivatives. When $V$ is nontrivial, the conditional centering in the statement removes the contribution of the squared probe that depends only on the retained variables. The probe itself need not be regular, since the small factors controlling the errors come from the defining polynomials.}

\begin{lemma}[Gaussian comparison] \label{lem:gaussian-comparision}
Let $\tau,V,p_1,\ldots,p_k$ satisfy the  assumptions of
\Cref{lem:gaussian-hermite-moment-comparison}, and use the same notation. Let $F:\R^k\to\R$ be a $C^2$
function such that
$D_2(F)\eqdef\max_{i,j\in[k]}\norm{\partial_i\partial_j F}_\infty<\infty$.
Then
\[
\Big|
    \E_{\x}\big[F(p_1(\x),\ldots,p_k(\x))\big]\\
    -\E_{\x,\z}\big[
        F(\widetilde p_1(\x_V,\z),\ldots,
          \widetilde p_k(\x_V,\z))
    \big]
    \Big|
    \le O(k^2\tau D_2(F)).
\]
\end{lemma}

\begin{proof}[Proof of \Cref{lem:gaussian-comparision}]
To prove \Cref{lem:gaussian-comparision}, we keep the variables in $V$ unchanged and
gradually replace the residual terms by a Gaussian vector with
the same covariance. Throughout, we use the assumptions and
notation of \Cref{lem:gaussian-hermite-moment-comparison}.
For $t\in[0,1]$, define
$    p_{t,i}
    =
    q_i(\x_V)
    +\sqrt{t}\,\big(p_i(\x)-q_i(\x_V)\big)
    +\sqrt{1-t}\,\boldsymbol{\theta}_i\cdot\z$,
where $\z$ is an independent standard Gaussian vector, and set
    $\p_t=(p_{t,1},\ldots,p_{t,k})$ and
    $H(t)=\E_{\x,\z}[F(\p_t)]$.
The two expectations in the lemma are $H(1)$ and $H(0)$. Therefore, to bound $|H(0)-H(1)|$, it suffices for us to bound $H'(t)$ for all $t\in (0,1)$. We will apply Gaussian integral by parts on $H'(t)$ and then show that the expression we get can be bounded by regularity.

For convenience, write $\u=\x_V$, $\by=\x_\perp$, and $\ba_i(\u)=\bb_i+\bL_i^\top \u$,
    $R_i=p_i(\x)-q_i(\u)$
    and
    $G_i=\boldsymbol{\theta}_i\cdot\z$.
Then
$
    R_i=\ba_i(\u)^\top \by+\by^\top \bA_i \by-\tr(\bA_i).
$
Define
\[
\begin{aligned}
    \Sigma_{ij}
    &:=\E[R_iR_j]
    =\boldsymbol{\theta}_i\cdot\boldsymbol{\theta}_j
    =\E_\u[\ba_i(\u)^\top \ba_j(\u)]+2\tr(\bA_i\bA_j),\\
    \Gamma_{ij}(\u,\by)
    &:=
    \big(\ba_i(\u)+\bA_i \by\big)^\top
    \big(\ba_j(\u)+2\bA_j\by\big).
\end{aligned}
\]
Thus $\Sigma$ is the common covariance matrix of the centered
vectors $(R_i)_{i=1}^k$ and $(G_i)_{i=1}^k$.
Unless otherwise indicated, expectations below are over the
independent Gaussian vectors $\u,\by,\z$.

\begin{fact}[Integration-by-parts identity]
\label{fact:gaussian-interpolation-identity}
For every $t\in(0,1)$,
    $H'(t)
    =
    \frac12\sum_{i,j=1}^k
    \E\left[
        (\Gamma_{ij}-\Sigma_{ij})\,
        \partial_i\partial_j F(\p_t)
    \right]$.
\end{fact}

\begin{proof}[Proof of \Cref{fact:gaussian-interpolation-identity}]
Differentiating with respect to $t$ gives
\[
    H'(t)
    =
    \frac{1}{2\sqrt{t}}\sum_i
    \E[R_i\,\partial_i F(\p_t)]
    -
    \frac{1}{2\sqrt{1-t}}\sum_i
    \E[G_i\,\partial_i F(\p_t)].
\]
For fixed $\u$, Gaussian integration by parts gives
    $\E_\by[R_i g(\by)]
    =
    \E_\by\left[
        (\ba_i(\u)+\bA_i \by)^\top\nabla_\by g(\by)
    \right]$.
For the quadratic terms, this follows by applying the identity
$\E_\by[y_r v(\by)]=\E_\by[\partial_r v(\by)]$ with
$v(\by)=y_sg(\by)$; the resulting constant term is canceled by
$\tr(\bA_i)$.

Holding $\u,\z$ fixed, apply this identity with
$g(\by)=\partial_i F(\p_t)$. Since
    $\nabla_\by p_{t,j}
    =\sqrt{t}\,\big(\ba_j(\u)+2\bA_j\by\big)$,
the chain rule gives
    $\E[R_i\,\partial_i F(\p_t)]
    =
    \sqrt{t}\sum_j
    \E[\Gamma_{ij}\,\partial_i\partial_j F(\p_t)]$.
Similarly, holding $\u,\by$ fixed and integrating by parts in $\z$,
we obtain $\E[G_i\,\partial_i F(\p_t)]
    =
    \sqrt{1-t}\sum_j
    \Sigma_{ij}\,\E[\partial_i\partial_j F(\p_t)]$.
Substituting these identities into $H'(t)$ proves the claim.

These operations are justified by Gaussian moment bounds,
since bounded second derivatives imply that $F$ and its first
derivatives have at most quadratic and linear growth,
respectively.
\end{proof}

\begin{fact}[Regularity bound]
\label{fact:gaussian-interpolation-error}
For every $i,j\in[k]$,
    $\E[|\Gamma_{ij}-\Sigma_{ij}|]
    \le
    \norm{\Gamma_{ij}-\Sigma_{ij}}_{L_2}
    \le C\tau$,
where $C>0$ is a universal constant.
\end{fact}

\begin{proof}
Since $\E[R_i\mid \u]=0$ and $\E[p_i^2]=1$, we have
$\norm{\bb_i}_2^2+\norm{\bL_i}_F^2+2\norm{\bA_i}_F^2
    =\E[R_i^2]\le 1$.
In particular,
    $\E_\u[\norm{\ba_i(\u)}_2^2]
    =\norm{\bb_i}_2^2+\norm{\bL_i}_F^2\le1$.
Expanding the discrepancy gives
\[
\begin{aligned}
    \Gamma_{ij}-\Sigma_{ij}
    =
    \ba_i(\u)^\top \ba_j(\u)
    -\E_\u[\ba_i(\u)^\top \ba_j(\u)]
    +\big(2\bA_j \ba_i(\u)+\bA_i \ba_j(\u)\big)^\top \by
    +2\big(\by^\top \bA_i\bA_j\by-\tr(\bA_i\bA_j)\big).
\end{aligned}
\]
For a standard Gaussian vector $\mathbf{Z}$ and any square matrix $M$,
    $\norm{\mathbf{Z}^\top M\mathbf{Z}-\tr(M)}_{L_2}
    \le\sqrt{2}\,\norm{M}_F$.
Using this identity and independence of $\u$ and $\by$, we obtain
\[
\begin{aligned}
    \norm{\ba_i^\top \ba_j-\E_\u[\ba_i^\top \ba_j]}_{L_2}
    &\le
    \norm{\bL_i \bb_j+\bL_j \bb_i}_2
    +\sqrt{2}\,\norm{\bL_i\bL_j^\top}_F
    \le C\tau,\\
    \norm{(2\bA_j \ba_i+\bA_i \ba_j)^\top \by}_{L_2}
    &\le
    2\norm{\bA_j}_{\op}+\norm{\bA_i}_{\op}
    \le3\tau,\\
    2\norm{\by^\top \bA_i\bA_j\by-\tr(\bA_i\bA_j)}_{L_2}
    &\le
    2\sqrt{2}\,\norm{\bA_i\bA_j}_F
    \le C\tau.
\end{aligned}
\]
The final bounds use
$\norm{\bL_i}_{\op},\norm{\bA_i}_{\op}\le\tau$,
the coefficient bounds above, and
$\norm{MN}_F\le\norm{M}_{\op}\norm{N}_F$.
The triangle inequality and Cauchy--Schwarz prove the claim.
\end{proof}

By \Cref{fact:gaussian-interpolation-identity,fact:gaussian-interpolation-error},
    $|H'(t)|
    \le
    \frac12D_2(F)\sum_{i,j=1}^k
    \E[|\Gamma_{ij}-\Sigma_{ij}|]
    \le Ck^2\tau D_2(F)$.
Integrating over $t\in[0,1]$, using continuity at the endpoints,
gives
    $|H(1)-H(0)|\le Ck^2\tau D_2(F)$,
as desired.
\end{proof}

We now prove \Cref{lem:gaussian-hermite-moment-comparison}
using the unweighted comparison established above.
The case $m=0$ follows directly from
\Cref{lem:gaussian-comparision}.
For $m=1,2$, Gaussian integration by parts removes the Hermite
weights, leaving derivatives of $\psi$ and remainder terms
controlled by the coefficient bounds.
We then apply \Cref{lem:gaussian-comparision} to these derivatives
to obtain the desired comparison.
\begin{proof}[Proof of \Cref{lem:gaussian-hermite-moment-comparison}]
For convenience, write
    $\p=(p_1(\x),\ldots,p_k(\x))$,
    $\widetilde \p=
    (\widetilde p_1(\x_V,\z),\ldots,\widetilde p_k(\x_V,\z))$
and let $\u=\x_V$ and $\by=\x_\perp$.

\paragraph{Case $m=0$.}
There is no Hermite weight in this case, so we apply \Cref{lem:gaussian-comparision}
directly to $\psi$.

Since $h_0=1$ and $\mathcal{P}_V[1]=1$, \Cref{lem:gaussian-comparision} gives
    $\left|\E[\psi(\p)]-\E[\psi(\widetilde \p)]\right|
    \le Ck^2\tau D_2(\psi)$.

\paragraph{Case $m=1$.}
We first use integration by parts to remove the Hermite factor,
obtaining first derivatives of $\psi$ and a remainder controlled
by regularity. We then apply \Cref{lem:gaussian-comparision} to these first derivatives.

Write
    $r=\v\cdot\Phi_2(\x)$,
    $\widetilde r=\v\cdot\z$ and
    $c_i=\v\cdot\boldsymbol{\theta}_i$.
Since $\v\in S_V$, we can write
\[
    r=\ba_\v(\u)^\top \by+\by^\top \B_{\v{}}\by-\tr(\B_\v),
    \qquad
    \ba_\v(\u)=\bb_\v+\bL_\v^\top \u,
\]
where $\B_\v$ is symmetric. In particular, $\E[r\mid \u]=0$, so
$
    \mathcal{P}_V[h_1(r)]=r.
$
The normalization $\norm{\v}_2=1$ gives $\norm{\bb_\v}_2^2+\norm{\bL_\v}_F^2+2\norm{\B_\v}_F^2
    =\E[r^2]=1$.
Consequently,
    $\E_\u[\norm{\ba_\v(\u)}_2^2]\le1$,
    $|c_i|\le\norm{\boldsymbol{\theta}_i}_2\le1$.
Also write $\ba_i(\u)=\bb_i+\bL_i^\top \u$, so that
$
    \nabla_\by p_i=\ba_i(\u)+2\bA_i \by.
$

We establish the first-moment expansion for a general smooth
function $F:\R^k\to\R$, so that we can also use it in the case
$m=2$. For fixed $\u$, Gaussian integration by parts gives
    $\E_\by[r\,g(\by)]
    =
    \E_\by[(\ba_\v(\u)+\B_{\v{}}\by)^\top\nabla_\by g(\by)]$.
Applying this identity to $g(\by)=F(\p)$, define
\[
    \Gamma_{\v{}i}
    =
    (\ba_\v(\u)+\B_{\v{}}\by)^\top(\ba_i(\u)+2\bA_i \by),
    \qquad
    \Delta_i=\Gamma_{\v{}i}-c_i.
\]
Since $\E[\Gamma_{\v{}i}]=c_i$, we obtain
\[
    \E[rF(\p)]
    =
    \sum_{i=1}^k c_i\,\E[\partial_i F(\p)]
    +\mathcal{R}_1(F),
\]
where
$
    \mathcal{R}_1(F)
    =
    \sum_{i=1}^k
    \E[\Delta_i\,\partial_i F(\p)].
$

To bound the remainder, expand
\[
\begin{aligned}
    \Delta_i
    =
    \ba_\v(\u)^\top \ba_i(\u)
    -\E_\u[\ba_\v(\u)^\top \ba_i(\u)]
    +\big(2\bA_i \ba_\v(\u)+\B_\v \ba_i(\u)\big)^\top \by
    +2\big(\by^\top \B_{\v{}}\bA_i \by-\tr(\B_{\v{}}\bA_i)\big).
\end{aligned}
\]
Using the same Gaussian second-moment estimates as in the proof
of \Cref{lem:gaussian-comparision},
\[
\begin{aligned}
    \norm{\Delta_i}_{L_2}
    \le
    \norm{\bL_i \bb_\v+\bL_\v \bb_i}_2
    +\sqrt{2}\,\norm{\bL_{\v{}}\bL_i^\top}_F
    +2\norm{\bA_i}_{\op}
    +\norm{\B_{\v{}}\bb_i}_2+\norm{\B_{\v{}}\bL_i^\top}_F
    +2\sqrt{2}\,\norm{\B_{\v{}}\bA_i}_F
    \le C\tau.
\end{aligned}
\]
The last inequality follows from the normalization of $\v$ and
the assumptions
    $\norm{\bb_i}_2, \norm{\bL_i}_{\op},
    \norm{\bA_i}_{\op}\le\tau$.
Therefore,
    $|\mathcal{R}_1(F)|\le Ck\tau D_1(F)$.

On the Gaussian side, integration by parts in $\z$ gives
    $\E[\widetilde r\,\psi(\widetilde \p)]
    =
    \sum_{i=1}^k c_i\,\E[\partial_i\psi(\widetilde \p)]$.
Combining this identity with the first-moment expansion,
\[
\begin{aligned}
    \left|
    \E[r\,\psi(\p)]
    -\E[\widetilde r\,\psi(\widetilde \p)]
    \right|
    \le
    \sum_{i=1}^k |c_i|
    \left|
    \E[\partial_i\psi(\p)]
    -\E[\partial_i\psi(\widetilde \p)]
    \right|
    +Ck\tau D_1(\psi).
\end{aligned}
\]
Applying \Cref{lem:gaussian-comparision} to each $\partial_i\psi$, and using
$|c_i|\le1$, bounds this by
\[
    Ck^3\tau D_3(\psi)+Ck\tau D_1(\psi)
    \le Ck^3\tau\max_{1\le j\le3}D_j(\psi).
\]

\paragraph{Case $m=2$.}
Conditional centering removes the $\u$-dependent variance of the
probe. We then apply integration by parts and reuse the
first-moment expansion to reduce the comparison to second
derivatives of $\psi$, which are handled by \Cref{lem:gaussian-comparision}.

Let
    $s(\u)=\E_\by[r^2]
    =\norm{\ba_\v(\u)}_2^2+2\norm{\B_\v}_F^2$.
Since $\E[r^2]=1$,
    $\mathcal{P}_V[h_2(r)]
    =
    \frac{r^2-s(\u)}{\sqrt{2}}$.
Apply the integration-by-parts identity from the previous case
to $g(\by)=r\psi(\p)$. Since
\[
    (\ba_\v+\B_{\v{}}\by)^\top\nabla_\by r
    =
    s(\u)+3\ba_\v^\top \B_{\v{}}\by
    +2\big(\by^\top \B_\v^2\by-\tr(\B_\v^2)\big),
\]
we obtain
\begin{equation}
    \label{eq:hermite-comparison-second-ibp}
    \E[(r^2-s(\u))\psi(\p)]
    =
    \sum_i\E[r\Gamma_{\v{}i}\,\partial_i\psi(\p)]
    +\E[J_\v\psi(\p)],
\end{equation}
where
    $J_\v
    =
    3\ba_\v(\u)^\top \B_{\v{}}\by
    +2\big(\by^\top \B_\v^2\by-\tr(\B_\v^2)\big)$.

For the first term in
\eqref{eq:hermite-comparison-second-ibp}, write
$\Gamma_{\v{}i}=c_i+\Delta_i$ and apply the first-moment expansion
to $F=\partial_i\psi$. This gives
\[
\begin{aligned}
    \sum_i\E[r\Gamma_{\v{}i}\,\partial_i\psi(\p)]
    =
    \sum_{i,j}c_ic_j
    \E[\partial_i\partial_j\psi(\p)]
    +\sum_i c_i\,\mathcal{R}_1(\partial_i\psi)
    +\sum_i\E[r\Delta_i\,\partial_i\psi(\p)].
\end{aligned}
\]
By the bound on $\mathcal{R}_1$ above and $|c_i|\le1$,
    $\left|\sum_i c_i\,\mathcal{R}_1(\partial_i\psi)\right|
    \le Ck^2\tau D_2(\psi)$.
Also, Cauchy--Schwarz gives
\[
\begin{aligned}
    \left|\sum_i\E[r\Delta_i\,\partial_i\psi(\p)]\right|
    \le
    D_1(\psi)\sum_i\norm{r}_{L_2}\norm{\Delta_i}_{L_2}
    \le Ck\tau D_1(\psi).
\end{aligned}
\]

It remains to bound the second term in
\eqref{eq:hermite-comparison-second-ibp}.
Another application of Gaussian integration by parts gives
\[
    \E[J_\v\psi(\p)]
    =
    \sum_i
    \E\left[
        (3\ba_\v+2\B_{\v{}}\by)^\top \B_\v\nabla_\by p_i\,
        \partial_i\psi(\p)
    \right].
\]
The normalization of $\v$ implies
    $\E[\norm{3\ba_\v+2\B_{\v{}}\by}_2^2]\le C$,
while the coefficient bounds give
\[
\begin{aligned}
    \E[\norm{\B_\v\nabla_\by p_i}_2^2]
    =
    \norm{\B_{\v{}}\bb_i}_2^2
    +\norm{\B_{\v{}}\bL_i^\top}_F^2
    +4\norm{\B_{\v{}}\bA_i}_F^2
    \le C\tau^2.
\end{aligned}
\]
Therefore, Cauchy--Schwarz yields
    $|\E[J_\v\psi(\p)]|
    \le Ck\tau D_1(\psi)$.

Combining these estimates establishes the second-moment expansion
\[
    \E[(r^2-s(\u))\psi(\p)]
    =
    \sum_{i,j}c_ic_j\,
    \E[\partial_i\partial_j\psi(\p)]
    +\mathcal{R}_2(\psi),
\]
with
    $|\mathcal{R}_2(\psi)|
    \le C\big(k\tau D_1(\psi)+k^2\tau D_2(\psi)\big)$.

On the Gaussian side, applying integration by parts twice gives
\[
    \E[\psi(\widetilde \p)h_2(\widetilde r)]
    =
    \frac{1}{\sqrt{2}}
    \sum_{i,j}c_ic_j\,
    \E[\partial_i\partial_j\psi(\widetilde \p)].
\]
Applying \Cref{lem:gaussian-comparision} to each $\partial_i\partial_j\psi$, and using
$|c_i|\le1$, bounds the difference between the two second Hermite
moments by
\[
    C\big(
        k\tau D_1(\psi)
        +k^2\tau D_2(\psi)
        +k^4\tau D_4(\psi)
    \big)
    \le
    Ck^4\tau\max_{1\le j\le4}D_j(\psi).
\]
This proves the claimed bound for $m=2$.
\end{proof}

\subsection{Gaussian Subspace Transfer}

Having compared the moments of the two models, we next show how
a subspace recovered for the Gaussian surrogate can be used to
approximate the original target.
The following lemma shows that, if $\wt f(\bz,\vec{G})$ has small
randomized error in every unit direction of $S_V$ orthogonal to
a subspace $W\subseteq S_V$, then $f(\bx)$ is close to an
intersection whose residual coefficient vectors lie in $W$,
while allowing unrestricted degree-$2$ dependence on $\bz$.
The approximation error is controlled by $\tau$ and the
directional randomized error.
This provides the approximation guarantee needed to restrict
the hypothesis search after recovering $W$ in the Gaussian model.

\begin{lemma}[Gaussian subspace transfer]\label{lm subspace preserve}
Let $\tau,V,p_1,\ldots,p_k$ satisfy the assumptions of \Cref{lem:gaussian-hermite-moment-comparison},
and use the same notation. Define
$
f(\x)=\mathbf{1}\{p_1(\x)\ge0,\ldots,p_k(\x)\ge0\}$, and
$
\widetilde f(\x_V,\z)
=
\mathbf{1}\{
q_i(\x_V)+\boldsymbol{\theta}_i\cdot\z\ge0
\text{ for all }i\in[k]
\}$.
Let $\delta\in[0,1]$, and
let $W\subseteq S_V$ be a subspace such that
    $\rand_{\widetilde f}(\w)\leq\delta$ for every unit vector $\w\in W^\perp$
where $W^\perp$ is taken within $S_V$.

Then there exists an intersection
$g(\x)=\mathbf{1}\{g_1(\x)\ge0,\ldots,g_k(\x)\ge0\}$
such that
$g_i\in
P_{\le2}(V)\oplus
\{\u\cdot\Phi_2(\x):\u\in W\}$
for every $i\in[k]$, where $P_{\le2}(V)$ denotes
the space of polynomials of degree at most two depending only
on $\x_V$, such that
$\Pr_{\x\sim\cN(0,\bI_d)}[f(\x)\ne g(\x)]
\le Ck^C(\delta+\tau)^{1/20}$,
for a universal constant $C>1$.
\end{lemma}
\begin{proof}[Proof of \Cref{lm subspace preserve}]
Throughout this proof, $C_k$ denotes a positive quantity bounded
by $Ck^C$ for a universal constant $C>1$; its value may increase
from line to line. We prove
$\Pr[f(\x)\ne g(\x)]
\le C_k(\delta+\tau)^{1/20}$.
The hypothesis says that, in the Gaussian model, $W$ captures almost all
the label information beyond $\x_V$. Our goal is to obtain an approximation
whose residual coefficients lie in $W$ and then use the same coefficients
in the original model. Direct projection into $W$ need not preserve the
coefficient bounds required for Gaussian comparison. We therefore first
find a controlled coefficient span, retain the part of that span almost
contained in $W$, and perform Gaussian comparison before making the final
projection.

We use two standard estimates. First, if $P,Q$ are degree-at-most-two
polynomials of a standard Gaussian input, $\norm{P}_2\ge 1/2$, and
$\norm{P-Q}_2\le s$, then
\begin{equation}
\Pr[\mathbf{1}\{P\ge0\}\ne\mathbf{1}\{Q\ge0\}]
\le C s^{2/5}.
\label{eq:subspace-transfer-sign-stability}
\end{equation}
Indeed, degree-two Gaussian anticoncentration and Markov's inequality give,
for $t>0$,
\[
\begin{aligned}
\Pr[\mathbf{1}\{P\ge0\}\ne\mathbf{1}\{Q\ge0\}]
\le \Pr[|P|\le t]+\Pr[|P-Q|>t]\le C\sqrt{t}+\frac{s^2}{t^2}.
\end{aligned}
\]
Taking $t=s^{4/5}$ proves the claim for $s>0$; the case $s=0$ is immediate.

Second, \Cref{lem:gaussian-comparision} implies the following comparison for disagreement
probabilities. For a joint tuple of $2k$ defining polynomials whose $L_2$
norms lie in $[1/2,1]$ and whose residual coefficient bounds are at most
$\rho$, the disagreement probability of the two corresponding intersections
differs between the original and Gaussian models by at most $C_k\rho^{1/5}.$
The claim is immediate when $\rho=0$, and follows from the trivial
bound $1$ when $\rho\ge1/2$, after increasing $C_k$.
Suppose $0<\rho<1/2$. First normalize each polynomial by its positive $L_2$ norm,
which increases its coefficient bounds by at most a factor of two.
Smooth each threshold on a scale $t$. If $A,B\in[0,1]$ are the resulting
smooth approximations to the two intersections, then $A+B-2AB$ smoothly
approximates their disagreement indicator and has second partial
derivatives bounded by $Ct^{-2}$ for a universal constant $C$,
since the undifferentiated factors lie in $[0,1]$.
Anticoncentration and a union bound give smoothing error in each model
at most $Ck\sqrt{t}$, since the Gaussian replacement preserves each
polynomial's $L_2$ norm and is itself a degree-at-most-two polynomial
of Gaussian variables. Applying \Cref{lem:gaussian-comparision} to the
$2k$ normalized polynomials bounds the smooth comparison error by
$Ck^2\rho t^{-2}$.
Taking $t=\rho^{2/5}$ gives a total bound of $Ck^2\rho^{1/5}$.

Write $\proj_U$ for orthogonal projection onto a coefficient subspace $U$.
Let $\eps=\delta+\tau$, and fix a sufficiently small universal
constant $c\in(0,2^{-8})$. We may assume that $\eps\le ck^{-2}$.
Otherwise,
$
1\le c^{-1/20}k^{1/10}\eps^{1/20},
$
so the trivial error bound $1$ gives the required conclusion
with a polynomial prefactor in $k$.
Choose parameters $a,b\in(0,1/2)$, to be specified at the end.

\paragraph{Step 1: Spectrally truncate the residual coefficient vectors.}
Although each $\boldsymbol{\theta}_i$ has small residual coefficients, a unit vector
in their span can require a large linear combination of the $\boldsymbol{\theta}_i$'s
when these vectors nearly cancel. We remove the directions responsible
for this poor conditioning. These directions have little total weight in
the defining polynomials, so removing them changes the target only slightly.

\begin{fact}[Spectral truncation]
\label{fact:subspace-transfer-truncation}
Let
$\Theta:\R^k\to S_V$,
$\Theta\boldsymbol{\alpha}=\sum_{i=1}^k\alpha_i\boldsymbol{\theta}_i$,
$M=\Theta\Theta^\top$,
and let $T$ be the eigenspace of $M$ with eigenvalues at least $a^2$.
Then
$\norm{\boldsymbol{\theta}_i-\proj_T\boldsymbol{\theta}_i}_2\le a$
for every $i\in[k]$.
Moreover, for every unit vector $\v\in T$, writing
\[
\v\cdot\Phi_2(\x)
=
\bb_{\v}^\top\x_\perp
+\x_V^\top \bL_{\v}\x_\perp
+\x_\perp^\top \bA_{\v}\x_\perp
-\tr(\bA_{\v}),
\]
we have
\[
\max\left\{
\norm{\bb_{\v}}_2,
\norm{\bL_{\v}}_{\op},
\norm{\bA_{\v}}_{\op}
\right\}
\le \frac{\sqrt{k}\tau}{a}.
\]
\end{fact}

\begin{proof}[Proof of \Cref{fact:subspace-transfer-truncation}]
By the definition of $T$,
$\norm{(\bI-\proj_T)\Theta}_\op\le a$.
Applying this operator to each coordinate vector of $\R^k$ proves
the first claim.

Every unit $\v\in T$ has a representation $\v=\Theta\boldsymbol{\alpha}$ with
$\norm{\boldsymbol{\alpha}}_2\le a^{-1}$, obtained from the singular value decomposition
of $\Theta$ restricted to the retained singular directions.
The maps taking a residual coefficient vector to its $\bb,\bL,\bA$ coefficients
are linear. The triangle inequality and Cauchy--Schwarz therefore give
\[
\begin{aligned}
\max\left\{
\norm{\bb_{\v}}_2,
\norm{\bL_{\v}}_{\op},
\norm{\bA_{\v}}_{\op}
\right\}
\le \tau\sum_i|\alpha_i|\le \sqrt{k}\tau\norm{\boldsymbol{\alpha}}_2\le \frac{\sqrt{k}\tau}{a}.
\end{aligned}
\]
\end{proof}

Set
\[
\boldsymbol{\theta}_i^0=\proj_T\boldsymbol{\theta}_i,
\qquad
p_i^0(\x)=q_i(\x_V)+\boldsymbol{\theta}_i^0\cdot\Phi_2(\x).
\]
Let
$
f^0(\x)
=
\mathbf{1}\{p_i^0(\x)\ge0\text{ for all }i\in[k]\}.
$
Let $\widetilde p_i^0(\u,\z)=q_i(\u)+\boldsymbol{\theta}_i^0\cdot\z$ and let
$\widetilde f^0$ be the corresponding Gaussian intersection.

Orthogonality of the Hermite features gives
$\norm{p_i-p_i^0}_2\le a$, and the same bound holds in the Gaussian model.
Consequently, \eqref{eq:subspace-transfer-sign-stability} and a union bound imply
\begin{equation}
\Pr[f\ne f^0]\le C_k a^{2/5},
\qquad
\Pr[\widetilde f\ne\widetilde f^0]\le C_k a^{2/5}.
\label{eq:subspace-transfer-truncation-error}
\end{equation}
Also, since the residual features have conditional mean zero given $\x_V$,
\begin{equation}
\begin{aligned}
n_i^2:=\norm{p_i^0}_2^2
=\norm{q_i}_2^2+\norm{\boldsymbol{\theta}_i^0}_2^2
=1-\norm{\boldsymbol{\theta}_i-\boldsymbol{\theta}_i^0}_2^2
\ge1-a^2.
\end{aligned}
\label{eq:subspace-transfer-truncated-norm}
\end{equation}
Write $\u=\x_V$, and let
$\z,\z'\sim\cN(0,\bI_{S_V})$ be independent of each other and of $\u$.
Let
$
U=\operatorname{span}
\{\proj_{W^\perp}\boldsymbol{\theta}_i:i\in[k]\}
\subseteq W^\perp,
$
and let $\v_1,\ldots,\v_r$ be an orthonormal basis of $U$,
where $r=\dim U\le k$.
The function $\widetilde f(\u,\z)$ depends on the $W^\perp$
component of $\z$ only through its projection onto $U$.
Resampling the coordinates along $\v_1,\ldots,\v_r$ one at a time
and applying a union bound therefore gives
\[
\Pr\left[
\widetilde f(\u,\z)
\ne
\widetilde f(\u,\proj_W\z+\proj_{W^\perp}\z')
\right]
\le
\sum_{j=1}^r\rand_{\widetilde f}(\v_j)
\le r\delta
\le k\delta.
\]
Both $\z$ and $\proj_W\z+\proj_{W^\perp}\z'$ are standard Gaussian.
Applying \eqref{eq:subspace-transfer-truncation-error} at both
endpoints and using another union bound gives
\begin{equation}
\Pr\left[
\widetilde f^0(\u,\z)
\ne
\widetilde f^0(\u,\proj_W\z+\proj_{W^\perp}\z')
\right]
\le\eta,
\qquad
\eta:=k\delta+C_k a^{2/5}.
\label{eq:subspace-transfer-truncated-resampling}
\end{equation}
Here and below, constants denoted by $C_k$ may increase from line to line.

\paragraph{Step 2: Retain the directions of $T$ almost contained in $W$.}
A direction of $T$ that lies substantially outside $W$ is substantially
disturbed by rerandomizing $W^\perp$. Since this rerandomization rarely
changes the label, the target has little dependence on such directions.
The following fact makes this statement quantitative.

\begin{fact}[Gaussian resampling onto a nearby subspace]
\label{fact:subspace-transfer-resampling}
Let $T,W$ be subspaces of a finite-dimensional Euclidean space $S$.
Let $\z,\z'$ be independent standard Gaussians on $S$, and let $\u$ be
independent of them. Suppose $H(\u,\z)\in\{0,1\}$ depends on $\z$ only through
$\proj_T\z$, and
\[
\Pr[
H(\u,\z)\ne H(\u,\proj_W\z+\proj_{W^\perp}\z')
]
\le\eta.
\]
Let $E\subseteq T$ be the eigenspace of
$
K=\proj_T\proj_W\proj_T:T\to T
$
with eigenvalues at least $1-b^2$.
Then every $\e\in E$ satisfies
$
\norm{\e-\proj_W \e}_2\le b\norm{\e}_2.
$
Moreover, if $\bc$ is an independent standard Gaussian on $T\cap E^\perp$,
then
\[
\Pr[H(\u,\z)\ne H(\u,\proj_E\z+\bc)]
\le\frac{\eta}{b^2}.
\]
\end{fact}

\begin{proof}[Proof of \Cref{fact:subspace-transfer-resampling}]
Choose an orthonormal eigenbasis $\e_1,\ldots,\e_r$ of $K$, with eigenvalues
$\lambda_j\in[0,1]$. In this basis,
\[
\proj_T\z
\qquad\text{and}\qquad
\proj_T(\proj_W\z+\proj_{W^\perp}\z')
\]
are standard Gaussian vectors with cross-covariance
$\operatorname{diag}(\lambda_1,\ldots,\lambda_r)$.

For fixed $\u$, expand $H$ in the normalized multivariate Hermite basis:
\[
H\left(\u,\sum_{j=1}^r s_j\e_j\right)
=
\sum_{\alpha\in\N^r}
\widehat H_\u(\alpha)\,H_\alpha(\mathbf{s}).
\]
Because $H$ is Boolean, disagreement probability equals expected squared
difference. Orthogonality and the Gaussian Hermite correlation identity give
\begin{align}
\Pr[
H(\u,\z)\ne H(\u,\proj_W\z+\proj_{W^\perp}\z')
]
=
2\sum_\alpha
\left(1-\prod_{j=1}^r\lambda_j^{\alpha_j}\right)
\E_\u[\widehat H_\u(\alpha)^2].
\label{eq:subspace-transfer-hermite-resampling}
\end{align}

Let $J=\{j:\lambda_j<1-b^2\}$, so that $T\cap E^\perp$ is spanned by the
$\e_j$ with $j\in J$. Independently replacing precisely these coordinates
gives disagreement probability
$
2\sum_{\alpha:\,\exists j\in J,\ \alpha_j>0}
\E_\u[\widehat H_\u(\alpha)^2].
$
For each term in this sum,
$
1-\prod_j\lambda_j^{\alpha_j}\ge b^2.
$
Comparison with \eqref{eq:subspace-transfer-hermite-resampling}
proves the resampling bound. Finally, for $\e\in E$,
\[
\norm{\e-\proj_W \e}_2^2
=
\norm{\e}_2^2-\langle \e,K\e\rangle
\le b^2\norm{\e}_2^2,
\]
which proves the first claim.
If $T=\{0\}$, both conclusions are immediate.
\end{proof}

Apply \Cref{fact:subspace-transfer-resampling} to
$H=\widetilde f^0$, using
\eqref{eq:subspace-transfer-truncated-resampling}.
Write $B=T\cap E^\perp$ and
\[
\boldsymbol{\theta}_i^E=\proj_E\boldsymbol{\theta}_i^0,
\qquad
\boldsymbol{\theta}_i^B=\proj_B\boldsymbol{\theta}_i^0.
\]
For an independent standard Gaussian $\bc$ on $B$, define
$
\widetilde r_i(\u,\z,\bc)
=
q_i(\u)+\boldsymbol{\theta}_i^B\cdot \bc+\boldsymbol{\theta}_i^E\cdot\z,
$
and let
$
\widetilde h(\u,\z,\bc)
=
\mathbf{1}\{
\widetilde r_i(\u,\z,\bc)\ge0\text{ for all }i\in[k]
\}.
$
The fact gives
\begin{equation}
\Pr[
\widetilde f^0(\u,\z)\ne\widetilde h(\u,\z,\bc)
]
\le\frac{\eta}{b^2}.
\label{eq:subspace-transfer-gaussian-approximation}
\end{equation}
Thus we have removed the directions of $T$ outside $E$ at a controlled
cost. The remaining residual directions are both controlled by
\Cref{fact:subspace-transfer-truncation} and close to $W$.

\paragraph{Step 3: Transfer while the coefficients still lie in the controlled span.}
We now return to the original Hermite features, defining
$
r_i(\x,\bc)
=
q_i(\x_V)+\boldsymbol{\theta}_i^B\cdot \bc+\boldsymbol{\theta}_i^E\cdot\Phi_2(\x),
$
and
$
h(\x,\bc)
=
\mathbf{1}\{r_i(\x,\bc)\ge0\text{ for all }i\in[k]\}.
$
It is important to keep $\bc$ random during this step. Fixing $\bc$ prematurely
could make some defining polynomials nearly zero in $L_2$, preventing a
uniform anticoncentration bound. Keeping $\bc$ random preserves their norms:
\begin{equation}
\begin{aligned}
\norm{r_i}_{L_2(\x,\bc)}^2
&=\norm{q_i}_2^2+\norm{\boldsymbol{\theta}_i^B}_2^2+\norm{\boldsymbol{\theta}_i^E}_2^2\\
&=n_i^2\in[1-a^2,1].
\end{aligned}
\label{eq:subspace-transfer-augmented-norm}
\end{equation}

Treat the coordinates of $\bc$ as additional retained Gaussian input
coordinates, enlarging $V$ to $V'=V\oplus B_\bc$, where $B_\bc$ denotes
this added coordinate space. Then
$q_i(\x_V)+\boldsymbol{\theta}_i^B\cdot \bc$ is a degree-at-most-two polynomial
on that enlarged subspace.

The residual coefficient vectors of the joint tuple
$
(p_1^0,\ldots,p_k^0,r_1,\ldots,r_k)
$
are $\boldsymbol{\theta}_i^0$ and $\boldsymbol{\theta}_i^E$, respectively. They lie in $T$ and have
Euclidean norm at most one. By \Cref{fact:subspace-transfer-truncation},
all their coefficient bounds are at most $\sqrt{k}\tau/a$.
Embedding these residuals in the enlarged Hermite feature space introduces
only zero coefficients on the additional features involving $\bc$.

By \eqref{eq:subspace-transfer-truncated-norm} and
\eqref{eq:subspace-transfer-augmented-norm}, the $L_2$ norms of all
$2k$ polynomials lie in $[1/2,1]$.
Applying the error comparison bound to their joint
disagreement event, and then using
\eqref{eq:subspace-transfer-gaussian-approximation}, gives
\begin{equation}
\Pr_{\x,\bc}[f^0(\x)\ne h(\x,\bc)]
\le
\frac{\eta}{b^2}
+
C_k\left(\frac{\tau}{a}\right)^{1/5}.
\label{eq:subspace-transfer-original-approximation}
\end{equation}
Indeed, with the choices $a=\eps^{3/4}$ and $b=\eps^{1/8}$
made below, our smallness assumption ensures $a,b\in(0,1/2)$.
Normalization increases the residual coefficient bounds by at
most a factor of two, and
\[
\frac{2\sqrt{k}\tau}{a}
\le 2\sqrt{k}\eps^{1/4}
\le 2c^{1/4}<1.
\]
Thus the normalized tuple satisfies the assumptions of
\Cref{lem:gaussian-comparision}.

\paragraph{Step 4: Move the coefficients into $W$ and fix the extra Gaussian.}
We have completed Gaussian comparison while the residual coefficients
remain in $E$. It remains to move these coefficients into $W$.
This last move is small in coefficient norm, so it can be handled directly
by sign stability.

Set
\[
\w_i=\proj_W\boldsymbol{\theta}_i^E,
\qquad
g_i(\x,\bc)
=
q_i(\x_V)+\boldsymbol{\theta}_i^B\cdot \bc+\w_i\cdot\Phi_2(\x),
\]
and let
\[
g_\bc(\x)
=
\mathbf{1}\{g_i(\x,\bc)\ge0\text{ for all }i\in[k]\}.
\]
\Cref{fact:subspace-transfer-resampling} gives
$
\norm{\boldsymbol{\theta}_i^E-\w_i}_2
\le b\norm{\boldsymbol{\theta}_i^E}_2
\le b.
$
Therefore, by orthonormality of the Hermite features,
$
\norm{r_i-g_i}_{L_2(\x,\bc)}\le b.
$
Using \eqref{eq:subspace-transfer-augmented-norm},
\eqref{eq:subspace-transfer-sign-stability}, and a union bound, we obtain
$
\Pr_{\x,\bc}[h(\x,\bc)\ne g_\bc(\x)]
\le C_k b^{2/5}.
$
Combining this with \eqref{eq:subspace-transfer-truncation-error} and
\eqref{eq:subspace-transfer-original-approximation} yields
\[
\Pr_{\x,\bc}[f(\x)\ne g_\bc(\x)]
\le
C_k\left[
a^{2/5}
+\frac{\delta+a^{2/5}}{b^2}
+\left(\frac{\tau}{a}\right)^{1/5}
+b^{2/5}
\right].
\]
Choose
\[
a=\eps^{3/4},
\qquad
b=\eps^{1/8},
\qquad
\eps=\delta+\tau.
\]
The four terms are bounded, respectively, by
$\eps^{3/10}$,
$\eps^{3/4}+\eps^{1/20}$,
$\eps^{1/20}$, and $\eps^{1/20}$.
Hence
$
\Pr_{\x,\bc}[f(\x)\ne g_\bc(\x)]
\le C_k\eps^{1/20}.
$

Finally, averaging over $\bc$ gives a deterministic $\bc_*$ with the same
error bound over $\x$. Let $g=g_{\bc_*}$. Since $\boldsymbol{\theta}_i^B\cdot \bc_*$ is a
constant and $\w_i\in W$, every defining polynomial satisfies
\[
g_i(\,\cdot\,,\bc_*)
\in
P_{\le2}(V)\oplus
\{\w\cdot\Phi_2(\,\cdot\,):\w\in W\}.
\]
This proves the required conclusion.
\end{proof}

\subsection{Learning Regular Intersections}
We now specialize to the fully regular case $V=\{0\}$ and combine
the preceding results with the moment-based dimension reduction
of \cite{DKS18a} to obtain a learning algorithm for regular
intersections.
The algorithm estimates the first two labeled moments in the
lifted feature space.
The moment comparison in
\Cref{lem:gaussian-hermite-moment-comparison} shows that these
estimates approximate the corresponding moments of the Gaussian
surrogate.
The Gaussian recovery guarantee then yields a coefficient subspace
$W$ of dimension at most $k$, outside which the surrogate has small
directional randomized error.
By \Cref{lm subspace preserve}, the original target admits a good
approximation by an intersection supported on $W$.
We complete the algorithm by searching over a finite net of such
intersections and selecting a hypothesis using independent
validation samples.

\begin{theorem}[Efficiently Learning a Regular Instance]\label{th regular instance}
There is an algorithm with the following guarantee.
Let $d,k\geq 1$ be integers and let $\eps\in(0,1/2)$.
Let $D$ be a distribution over $\R^d\times\{0,1\}$ with marginal
$D_{\x}=\cN(0,\bI_d)$.
Suppose there exists an intersection
$f(\x)
    =
    \mathbf{1}\{p_1(\x)\geq 0,\ldots,p_k(\x)\geq 0\}$,
where each $p_i$, $i\in[k]$, is a polynomial of degree at most two
and is individually $(\tau,\{0\})$-regular, where
$\tau \leq \left(\frac{\eps}{k}\right)^{c}$,
$\Pr_{(\x,y)\sim D}[f(\x)\neq y]
    \leq \left(\frac{\eps}{k}\right)^{c}$
and $c$ is a sufficiently large universal constant.

Then given $k,\eps$ and i.i.d.\ sample access to $D$, the algorithm uses $d^{O(1)}\left(k/\eps\right)^{O(k^2)}$
time and sample complexity and with probability at least $2/3$ outputs
a hypothesis $h:\R^d\to\{0,1\}$ with error
$\Pr_{(\x,y)\sim D}[h(\x)\neq y]\leq\eps$.
\end{theorem}
\begin{algorithm}[htbp]
\caption{RegularLearning}
\label{alg:regular-learning}
\begin{algorithmic}[1]
\State \textbf{Input:} $d,k,\eps$, and i.i.d.\ sample access to $D$.
\State \textbf{Output:} A hypothesis $h:\R^d\to\{0,1\}$.

\medskip

\State Set $N=d(d+3)/2$, $\lambda=(\eps/k)^A$, and
$r=(\eps/(Ck))^{5/2}$, for sufficiently large universal constants $A,C$.
\State Using fresh samples, estimate
\[
    \E_{(\x,y)\sim D}[y\Phi_2(\x)]
    \quad\text{and}\quad
    \E_{(\x,y)\sim D}
    [y(\Phi_2(\x)\Phi_2(\x)^\top-\bI_N)]
\]
by $\widehat{\boldsymbol{\mu}}$ and $\widehat M$, with Euclidean and
operator norm errors at most $\lambda/6$, respectively, and joint
failure probability at most $1/6$.
\State Form $\widehat T=[\,\widehat{\boldsymbol{\mu}}\;\;\widehat M\,]$.
Let $W$ be the span of its top $\min\{k,N\}$ left singular vectors,
and compute an orthonormal basis $\v_1,\ldots,\v_m$ of $W$.
\State Construct an $r$-net $\mathcal N$ of the unit Gaussian $L_2$
sphere in
\[
    \mathcal Q_W
    =\spa\{1,\v_1\cdot\Phi_2,\ldots,\v_m\cdot\Phi_2\},
\]
using a Euclidean net of the coefficient unit sphere. Include the
constant polynomials $1$ and $-1$.
\State Form the candidate family
\[
    \mathcal H
    =
    \left\{
        \x\mapsto\mathbf{1}\{q_i(\x)\geq0
        \text{ for all }i\in[k]\}
        :q_1,\ldots,q_k\in\mathcal N
    \right\}.
\]
\State Draw an independent validation sample
$\{(\x^{(j)},y^{(j)})\}_{j=1}^{n_{\mathrm{val}}}$, where
$n_{\mathrm{val}}=\lceil C\eps^{-2}\log(12|\mathcal H|)\rceil$.
\State \textbf{Return:} A hypothesis
\[
    \widehat h\in\arg\min_{h\in\mathcal H}
    \frac{1}{n_{\mathrm{val}}}
    \sum_{j=1}^{n_{\mathrm{val}}}
    \mathbf{1}\{h(\x^{(j)})\neq y^{(j)}\}.
\]
\end{algorithmic}
\end{algorithm}
\begin{proof}
We analyze \Cref{alg:regular-learning}. The proof uses
\Cref{lem:gaussian-hermite-moment-comparison} to compare the estimated
moments with their Gaussian counterparts and \Cref{lm subspace preserve}
to transfer the resulting subspace approximation to the original
distribution.

Let
\[
    \nu=\Pr_{(\x,y)\sim D}[f(\x)\neq y],
    \qquad N=d(d+3)/2.
\]
Replacing identically zero defining polynomials by $1$ and rescaling
the others by positive constants, assume that $\norm{p_i}_2=1$ for every
$i\in[k]$. Write
\[
    p_i(\x)=c_i+\boldsymbol{\theta}_i\cdot\Phi_2(\x),
    \qquad
    \widetilde f(\z)
    =
    \mathbf{1}\{c_i+\boldsymbol{\theta}_i\cdot\z\geq 0
    \text{ for all }i\in[k]\}.
\]
Here $\z\sim\cN(0,\bI_N)$.
\Cref{fct:regular-to-coefficient} supplies the coefficient bounds
required by
\Cref{lem:gaussian-hermite-moment-comparison,lm subspace preserve}
with $V=\{0\}$.

Define
\[
    \boldsymbol{\mu}
    =\E_{\z}[\widetilde f(\z)\z],
    \qquad
    M=\E_{\z}[
        \widetilde f(\z)(\z\z^\top-\bI_N)
    ],
    \qquad
    T=[\,\boldsymbol{\mu}\;\;M\,].
\]
Let $\widehat{\boldsymbol{\mu}}$, $\widehat M$, and $\widehat T$
be the estimates computed by the algorithm. We compare $\widehat T$
with $T$ by bounding the contributions of Gaussian replacement,
label disagreement, and sampling error.

Recall that $\lambda=(\eps/k)^A$. Smoothing the thresholds in \Cref{lem:gaussian-hermite-moment-comparison}
and applying Gaussian anticoncentration bounds the Gaussian-comparison
error by $k^{O(1)}\tau^{\Omega(1)}$, with universal exponents.
The label disagreement contributes $O(\sqrt{\nu})$: for every unit
vector $\v$, Gaussian hypercontractivity gives
\[
    \norm{\v\cdot\Phi_2}_2=1,
    \qquad
    \norm{(\v\cdot\Phi_2)^2-1}_2=O(1),
\]
so Cauchy--Schwarz bounds the perturbations of the two moment blocks
in Euclidean and operator norm, respectively.

The entries of the observed moment blocks have universally bounded
variance. Entrywise median-of-means estimates with accuracy
$\lambda/(6N)$ therefore give the accuracies required by the algorithm
using $\poly(d,1/\lambda)$ time and samples. With
probability at least $5/6$, we consequently have
\[
    \norm{\widehat T-T}_{\op}
    \leq
    k^{O(1)}\tau^{\Omega(1)}
    +O(\sqrt{\nu})+\lambda/3
    \leq\lambda,
\]
where the last inequality holds when the constant $c$ in the theorem
is sufficiently large.

Let $W$ be the subspace computed by the algorithm. Since $\widetilde f$ depends only on the projection onto
$\spa\{\boldsymbol{\theta}_1,\ldots,
\boldsymbol{\theta}_k\}$, the range of $T$ lies in this subspace.
Hence $\operatorname{rank}(T)\leq k$, and
    $\norm{\proj_{W^\perp}T}_{\op}\leq 2\lambda$.
It follows that, for every unit $\v\in W^\perp$, the correlation of
$\widetilde f$ with any centered, normalized univariate polynomial
of degree at most two in $\v\cdot\z$ is $O(\lambda)$.
By \cite[Proposition~5.2]{DKS18a},
\[
    \rand_{\widetilde f}(\v)
    \leq
    O\left(
        k^{4/11}\lambda^{1/11}
        \log^{2/11}(2k/\lambda)
    \right)
    =:\delta_0.
\]

\Cref{lm subspace preserve} therefore gives an
intersection $g$ whose defining polynomials belong to
\[
    \mathcal Q_W
    =
    \R\oplus
    \{\w\cdot\Phi_2:\w\in W\},
\]
and such that
\[
    \Pr_{\x\sim\cN(0,\bI_d)}[f(\x)\neq g(\x)]
    \leq
    k^{O(1)}(\delta_0+\tau)^{1/20}
    \leq\eps/4.
\]
Choosing $A$ sufficiently large and then $c$ sufficiently large
ensures this inequality.

It remains to show that the candidate family $\mathcal H$ contains
a good hypothesis. Since $\v_1,\ldots,\v_m$ are orthonormal
coefficient vectors, the polynomials
$1,\v_1\cdot\Phi_2,\ldots,\v_m\cdot\Phi_2$ are orthonormal under
$\cN(0,\bI_d)$. Thus the Euclidean coefficient net used by the
algorithm is an $r$-net in Gaussian $L_2$ norm. Since $m\leq k$,
    $|\mathcal H|\leq(C/r)^{k(k+1)}$.
Normalize the nonzero defining polynomials of $g$, replacing zero
constraints by $1$. The sign-stability estimate from the proof of
\Cref{lm subspace preserve} shows that rounding these
polynomials to the net changes the intersection on a set of
probability at most $Ckr^{2/5}$. The algorithm's choice
$r=(\eps/(Ck))^{5/2}$ therefore ensures that $\mathcal H$ contains
a hypothesis $h_0$ with
\[
    \Pr_{(\x,y)\sim D}[h_0(\x)\neq y]
    \leq
    \Pr_{\x}[f(\x)\neq g(\x)]
    +\Pr_{\x}[g(\x)\neq h_0(\x)]
    +\nu
    \leq 3\eps/4,
\]
where the probabilities over $\x$ are under $\cN(0,\bI_d)$.

Conditioned on the preceding samples, the family $\mathcal H$ is
fixed and the validation samples are independent. Hoeffding's
inequality and a union bound imply that, with probability at least
$5/6$, every candidate's empirical error differs from its true error
by at most $\eps/8$. The empirical minimizer returned by the algorithm
therefore has true error at most $\eps/4$ above that of $h_0$, and
hence at most $\eps$.

The overall success probability is at least $2/3$.
Moment estimation and subspace recovery take
$\poly(d,k/\eps)$ time, while
    $|\mathcal H|=(k/\eps)^{O(k^2)}$.
The claimed time and sample bounds follow.
\end{proof}

\section{Reduction to the Regular Case using Randomness}\label{sec regular case reduction}

In this section, we show that if the randomized error is sufficiently
small along every direction orthogonal to $V$, then the target is
close to an intersection whose defining polynomials are
$(\tau,V)$-regular. The intuition for this is simple.
We write each original defining polynomial as $q_i(\x) =  q_i(\z,\w) = a_i(\z) + (b_i + L_i^\top\z)^\top \w + \w^\top \mathbf{A}_i \w - \tr(\mathbf{A}_i) $, where $\z = \proj_V(\x)$ and $\w = \proj_{V^\perp}(\x)$. We call $(b_i + L_i^\top\z)^\top \w + \w^\top \mathbf{A}_i \w - \tr(\mathbf{A}_i)$ the remainder of $q_i$.
Notice that a remainder consists of a linear term $b_i\cdot \w$, a mixed term $(L_i^\top\z)\cdot \w$ and a pure quadratic term defined by $\mathbf{A}_i$.
We select the subspace $U \subseteq V^\perp$ to be the subspace spanned by all $b_i$, right singular vectors of $L_i$ corresponding to large singular values and large eigenvectors of $\mathbf{A}_i$.
The intuition here is that if we fix the component of $\x$ in $U$ at some fixed point. Now the mixed term of the resulting polynomial has no large singular values and the quadratic term of it has no large eigenvalues. In fact, combining the fact that every direction in $V^\perp$ has a tiny randomized error and Gaussian anti-concentration, we are able to show that there exists a choice of the coordinates in $U$ such that the resulting intersection is close to the original one and each defining polynomial has a not too small norm.

However, this does not imply regularity yet, because the resulting polynomial could have new large linear terms.
To deal with this, we apply the same idea above but only find a fix along the subspace spanned by these $k$ linear terms.

\new{As a result, the residual linear coefficient vectors and the operator norms of the residual mixed and quadratic coefficient matrices are small, while each resulting polynomial has a nonnegligible Gaussian $L_2$ norm. With our parameter choices, we distinguish two cases. If the variance is sufficiently large relative to the residual coefficient bounds, the directional-influence identity in the proof of \Cref{fct:regular-to-coefficient} gives $(\tau,V)$-regularity. Otherwise, the variance is small compared with the squared $L_2$ norm, so the polynomial has a nonnegligible mean. Chebyshev's inequality then shows that its threshold function is close to a constant.}

\begin{theorem}[Regular approximation from small randomized error]
\label{th error to regularity}
There is a universal constant $C>1$ such that the following holds.
Let $k\ge1$, $\eps,\tau\in(0,1)$, and let $V\subseteq\R^d$ be a
subspace. Suppose that $f:\R^d\to\{0,1\}$ is an intersection of at
most $k$ degree-$2$ PTFs and
\begin{align*}
    \rand_f(\v)\le\alpha_k(\eps,\tau)
    :=\frac{\tau^2\eps^2}{Ck^{12}},
\end{align*}
for every unit vector $\v \in  V^\perp$.
Then there are $k'\le k$ polynomials $q_1,\ldots,q_{k'}$ of degree
at most two such that $g(\x)=\mathbf{1}\{q_1(\x)\ge0,\ldots,q_{k'}(\x)\ge0\}$
satisfies $\Pr_{\x\sim\cN(0,\bI_d)}[f(\x)\ne g(\x)]\le\eps,$
and each $q_i$ is $(\tau,V)$-regular.
\end{theorem}

We first give two facts about random restrictions. The first allows
us to choose the second set of directions after fixing the first set.

\begin{fact}\label{fact randomized error restriction}
Let $f,h:\R^d\to\{0,1\}$ be Boolean functions. For every unit
vector $\v\in\R^d$, we have
\[
    \rand_h(\v)\le\rand_f(\v)
    +2\Pr_{\x\sim\cN(0,\bI_d)}[f(\x)\ne h(\x)].
\]
Moreover, let $U$ be a subspace with orthonormal basis
$\v_1,\ldots,\v_m$. For $\z\in U$, define
$f_{\z}(\x)=f(\z+\proj_{U^\perp}\x)$. Then
\[
    \E_{\z\sim\cN(0,\bI_U)}
    \left[\Pr_{\x\sim\cN(0,\bI_d)}
    [f(\x)\ne f_{\z}(\x)]\right]
    \le\sum_{j=1}^m\rand_f(\v_j).
\]
\end{fact}

\begin{proof}[Proof of \Cref{fact randomized error restriction}]
Let $\x\sim\cN(0,\bI_d)$ and $s\sim\cN(0,1)$ be independent,
and set $\x'=\proj_{\v^\perp}\x+s\v$. By a union bound,
\begin{align*}
    \rand_h(\v)
    &=\Pr_{\x,s}[h(\x)\ne h(\x')]\\
    &\le\Pr_{\x,s}[h(\x)\ne f(\x)]
       +\Pr_{\x,s}[f(\x)\ne f(\x')]
       +\Pr_{\x,s}[f(\x')\ne h(\x')]\\
    &=\rand_f(\v)
       +2\Pr_{\x\sim\cN(0,\bI_d)}[f(\x)\ne h(\x)],
\end{align*}
where the last equality uses that both $\x$ and $\x'$ are standard
Gaussians.

For the second claim, draw $\x\sim\cN(0,\bI_d)$ and
$\z\sim\cN(0,\bI_U)$ independently, and replace the coordinates
of $\x$ along $\v_1,\ldots,\v_m$ one at a time by those of $\z$.
At each step, the input is standard Gaussian, and the probability
that the label changes at step $j$ is $\rand_f(\v_j)$.
Taking a union bound over the $m$ steps proves the claim.
\end{proof}

The next lemma ensures that the defining polynomials do not become
too small in $L_2$ after fixing the coordinates.

\begin{lemma}\label{lm norm under random restriction}
There is a universal constant $K\ge1$ such that the following holds.
Let $p:\R^d\to\R$ be a nonzero polynomial of degree at most two,
and let $U\subseteq\R^d$ be a subspace. For $\z\in U$, define
$p_{\z}(\x)=p(\z+\proj_{U^\perp}\x)$. Then, for every
$\kappa\in(0,1)$,
\[
    \Pr_{\z\sim\cN(0,\bI_U)}
    [\norm{p_{\z}}_2<\kappa\norm{p}_2]
    \le K\sqrt{\kappa}.
\]
\end{lemma}

\begin{proof}[Proof of \Cref{lm norm under random restriction}]
Define $Q(\z)=\E_{\w\sim\cN(0,\bI_{U^\perp})}[p(\z+\w)^2].$
Then $Q$ is a nonnegative polynomial of degree at most four,
$Q(\z)=\norm{p_{\z}}_2^2$, and $\E_{\z\sim\cN(0,\bI_U)}[Q(\z)]=\norm{p}_2^2.$

By the Carbery--Wright inequality for degree-$4$ polynomials,
\[
    \Pr_{\z\sim\cN(0,\bI_U)}
    [Q(\z)<\kappa^2\norm{p}_2^2]
    \le K(\kappa^2)^{1/4}=K\sqrt{\kappa}.
\]
\end{proof}

\begin{proof}[Proof of \Cref{th error to regularity}]
We may write
\[
    f(\x)=\mathbf{1}\{p_1(\x)\ge0,\ldots,p_k(\x)\ge0\},
    \qquad \norm{p_i}_2=1 \quad\text{for every }i\in[k].
\]
Indeed, we can add constant-true constraints, replace an identically
zero polynomial by $1$, and positively rescale each defining
polynomial. If $V^\perp=\{0\}$, the conclusion holds with $g=f$.
We therefore assume that $V^\perp\ne\{0\}$.

Let $K$ be the constant from \Cref{lm norm under random restriction},
and set
\[
    \alpha=\frac{\tau^2\eps^2}{Ck^{12}},\qquad
    \kappa=(4Kk)^{-2},\qquad
    \delta^2=\frac{\tau^2\eps\kappa^4}{200k^2}.
\]
In particular, $\delta\in(0,1)$. Write $W=V^\perp$, and, for
$\z\in V$ and $\w\in W$, decompose
\[
    p_i(\z+\w)
    =a_i(\z)+\mathbf{b}_i^\top\w
     +\z^\top\mathbf{L}_i\w
     +\w^\top\mathbf{A}_i\w-\operatorname{tr}(\mathbf{A}_i),
\]
where $\mathbf{L}_i:W\to V$ is linear and
$\mathbf{A}_i:W\to W$ is symmetric. By Gaussian orthogonality,
\begin{equation}
    \norm{\mathbf{b}_i}_2^2
    +\norm{\mathbf{L}_i}_F^2
    +2\norm{\mathbf{A}_i}_F^2\le1.
    \label{eq regularity coefficient norm}
\end{equation}

\emph{First restriction.}
Let $U\subseteq W$ be the span, over all $i\in[k]$, of
$\mathbf{b}_i$, the right singular vectors of $\mathbf{L}_i$ with
singular value greater than $\delta$, and the eigenvectors of
$\mathbf{A}_i$ with eigenvalue of absolute value greater than
$\delta$. By \eqref{eq regularity coefficient norm},
    $m:=\dim U\le3k\delta^{-2}$.
Let $R=W\cap U^\perp$. By the choice of $U$, for every $i\in[k]$,
\begin{equation}
    \proj_R\mathbf{b}_i=0,\qquad
    \norm{\mathbf{L}_i\proj_R}_{\mathrm{op}}\le\delta,
    \qquad
    \norm{\mathbf{A}_i\proj_R}_{\mathrm{op}}
    =\norm{\proj_R\mathbf{A}_i}_{\mathrm{op}}\le\delta.
    \label{eq regularity remaining coefficients}
\end{equation}
For $\mathbf{t}\in U$, define
\[
    p_{i,\mathbf{t}}(\x)
    =p_i(\proj_V\x+\mathbf{t}+\proj_R\x),\qquad
    f_{\mathbf{t}}(\x)
    =f(\proj_V\x+\mathbf{t}+\proj_R\x).
\]
By \Cref{fact randomized error restriction},
\[
    \E_{\mathbf{t}\sim\cN(0,\bI_U)}
    \left[\Pr_{\x\sim\cN(0,\bI_d)}
    [f(\x)\ne f_{\mathbf{t}}(\x)]\right]
    \le m\alpha.
\]
By Markov's inequality, with probability at least $3/4$ over
$\mathbf{t}\sim\cN(0,\bI_U)$, the inner probability is at most
$4m\alpha$. Moreover, \Cref{lm norm under random restriction}
and a union bound imply that
\[
    \Pr_{\mathbf{t}\sim\cN(0,\bI_U)}
    [\exists i\in[k]:\norm{p_{i,\mathbf{t}}}_2<\kappa]
    \le kK\sqrt{\kappa}=1/4.
\]
We can therefore fix $\mathbf{t}\in U$ such that
\begin{equation*}
    e_1:=\Pr_{\x\sim\cN(0,\bI_d)}
    [f(\x)\ne f_{\mathbf{t}}(\x)]\le4m\alpha,
    \qquad
    \norm{p_{i,\mathbf{t}}}_2\ge\kappa
    \quad\text{for every }i\in[k].
    \label{eq regularity first restriction}
\end{equation*}
When $m=0$, the error is zero and the same conclusion holds with
$\mathbf{t}=0$.

For $\z\in V$ and $\w\in R$, the restricted polynomial has the form
\[
    p_{i,\mathbf{t}}(\z+\w)
    =a_{i,\mathbf{t}}(\z)+\mathbf{b}_{i,\mathbf{t}}^\top\w
     +\z^\top\mathbf{L}_{i,R}\w
     +\w^\top\mathbf{A}_{i,R}\w
     -\operatorname{tr}(\mathbf{A}_{i,R}),
\]
where
\[
    \mathbf{b}_{i,\mathbf{t}}=2\proj_R\mathbf{A}_i\mathbf{t},
    \qquad
    \mathbf{L}_{i,R}=\mathbf{L}_i|_R,
    \qquad
    \mathbf{A}_{i,R}=\proj_R\mathbf{A}_i|_R.
\]
The operators $\mathbf{L}_{i,R}$ and $\mathbf{A}_{i,R}$ have norm
at most $\delta$, but $\mathbf{b}_{i,\mathbf{t}}$ need not be small.
We remove these linear terms by a second restriction.

\emph{Second restriction.}
Let
\[
    H=\operatorname{span}\{\mathbf{b}_{i,\mathbf{t}}:i\in[k]\}
    \subseteq R,\qquad S=R\cap H^\perp.
\]
We have $\dim H\le k$. Although $H$ depends on $\mathbf{t}$,
we have already fixed $\mathbf{t}$. By
\Cref{fact randomized error restriction}, for every unit
$\v\in W$,
\begin{equation*}
    \rand_{f_{\mathbf{t}}}(\v)
    \le\rand_f(\v)+2e_1\le\alpha+2e_1.
    \label{eq regularity restricted randomized error}
\end{equation*}
For $\mathbf{s}\in H$, define
\[
    r_{i,\mathbf{s}}(\x)
    =p_i(\proj_V\x+\mathbf{t}+\mathbf{s}+\proj_S\x),\qquad
    h_{\mathbf{s}}(\x)
    =f(\proj_V\x+\mathbf{t}+\mathbf{s}+\proj_S\x).
\]
Applying \Cref{fact randomized error restriction} to
$f_{\mathbf{t}}$ and $H$, we obtain
\[
    \E_{\mathbf{s}\sim\cN(0,\bI_H)}
    \left[\Pr_{\x\sim\cN(0,\bI_d)}
    [f_{\mathbf{t}}(\x)\ne h_{\mathbf{s}}(\x)]\right]
    \le k(\alpha+2e_1).
\]
Also, $\E_{\mathbf{s}\sim\cN(0,\bI_H)}[\norm{\mathbf{s}}_2^2]
=\dim H\le k$. Applying Markov's inequality to these two quantities,
and applying \Cref{lm norm under random restriction} to
$p_{i,\mathbf{t}}$, we can fix $\mathbf{s}\in H$ such that
\begin{equation}
    \begin{aligned}
        e_2&:=\Pr_{\x\sim\cN(0,\bI_d)}
        [f_{\mathbf{t}}(\x)\ne h_{\mathbf{s}}(\x)]
        \le4k(\alpha+2e_1),\\
        \norm{\mathbf{s}}_2^2&\le4k,\\
        \norm{r_{i,\mathbf{s}}}_2
        &\ge\kappa\norm{p_{i,\mathbf{t}}}_2\ge\kappa^2
        \qquad\text{for every }i\in[k].
    \end{aligned}
    \label{eq regularity second restriction}
\end{equation}
Indeed, each of the first two conditions fails with probability at
most $1/4$, and the probability that any of the norm conditions fails
is at most $kK\sqrt{\kappa}=1/4$.

Fix such an $\mathbf{s}$, and write $r_i=r_{i,\mathbf{s}}$ and
$h=h_{\mathbf{s}}$. The error after the two restrictions is bounded by
\begin{align}
    \Pr_{\x\sim\cN(0,\bI_d)}[f(\x)\ne h(\x)]
    \le e_1+e_2\nonumber
    \le(1+8k)e_1+4k\alpha\nonumber
    \le4(1+8k)m\alpha+4k\alpha\nonumber
    \le112k^2\delta^{-2}\alpha.
    \label{eq regularity restriction error}
\end{align}

\emph{Regularity of the remaining polynomials.}
Since $\proj_S\mathbf{b}_{i,\mathbf{t}}=0$, the residual
coefficients of $r_i$ are
\[
    \widetilde{\mathbf{b}}_i=2\proj_S\mathbf{A}_{i,R}\mathbf{s},
    \qquad
    \widetilde{\mathbf{L}}_i=\mathbf{L}_i|_S,
    \qquad
    \widetilde{\mathbf{A}}_i=\proj_S\mathbf{A}_i|_S.
\]
By \eqref{eq regularity remaining coefficients} and
\eqref{eq regularity second restriction},
\[
    \norm{\widetilde{\mathbf{b}}_i}_2\le4\delta\sqrt{k},
    \qquad
    \norm{\widetilde{\mathbf{L}}_i}_{\mathrm{op}}\le\delta,
    \qquad
    \norm{\widetilde{\mathbf{A}}_i}_{\mathrm{op}}\le\delta.
\]
Every $r_i$ ignores the coordinates in $U\oplus H$.
For a unit vector $\v\in V^\perp$, let $\w=\proj_S\v$.
Using the coefficient identity from \Cref{fct:regular-to-coefficient}, we obtain
\begin{align*}
    \operatorname{inf}_{r_i}(\v)
    &=\E_{\x\sim\cN(0,\bI_d)}
      [(\v\cdot\nabla r_i(\x))^2]\nonumber\\
    &=(\widetilde{\mathbf{b}}_i^\top\w)^2
      +\norm{\widetilde{\mathbf{L}}_i\w}_2^2
      +4\norm{\widetilde{\mathbf{A}}_i\w}_2^2\nonumber\\
    &\le(16k+5)\delta^2\le21k\delta^2.
    \label{eq regularity influence bound}
\end{align*}
Thus the second restriction introduces only small linear coefficients,
and no further restriction is needed.

It remains to compare this bound with the variance of $r_i$. Set
    $\lambda=\frac{\eps\kappa^4}{8k}$.
If $\operatorname{Var}_{\x\sim\cN(0,\bI_d)}[r_i(\x)]\ge\lambda$,
let $q_i=r_i$. By the choice of $\delta$, for every unit
$\v\in V^\perp$,
\[
    \operatorname{inf}_{q_i}(\v)
    \le21k\delta^2\le\tau^2\lambda
    \le\tau^2\operatorname{Var}_{\x\sim\cN(0,\bI_d)}[q_i(\x)].
\]
Therefore, $q_i$ is $(\tau,V)$-regular.

Otherwise, let $\mu_i=\E_{\x\sim\cN(0,\bI_d)}[r_i(\x)]$.
By \eqref{eq regularity second restriction},
\[
    \mu_i^2=\norm{r_i}_2^2
       -\operatorname{Var}_{\x\sim\cN(0,\bI_d)}[r_i(\x)]
    \ge\kappa^4-\lambda\ge\kappa^4/2>0.
\]
Set $q_i=1$ if $\mu_i>0$ and $q_i=-1$ if $\mu_i<0$.
This constant polynomial is $(\tau,V)$-regular. By Chebyshev's
inequality,
\begin{align*}
    &\Pr_{\x\sim\cN(0,\bI_d)}
    [\mathbf{1}\{r_i(\x)\ge0\}\ne\mathbf{1}\{q_i\ge0\}]\\
    &\qquad\le\Pr_{\x\sim\cN(0,\bI_d)}
       [\abs{r_i(\x)-\mu_i}\ge\abs{\mu_i}]\\
    &\qquad\le
       \frac{\operatorname{Var}_{\x\sim\cN(0,\bI_d)}[r_i(\x)]}
            {\mu_i^2}
       \le\frac{2\lambda}{\kappa^4}=\frac{\eps}{4k}.
\end{align*}

Let $g(\x)=\mathbf{1}\{q_i(\x)\ge0\text{ for every }i\in[k]\}$.
Combining the preceding bound on
$\Pr_{\x\sim\cN(0,\bI_d)}[f(\x)\ne h(\x)]$
with a union bound gives
\[
    \Pr_{\x\sim\cN(0,\bI_d)}[f(\x)\ne g(\x)]
    \le112k^2\delta^{-2}\alpha+\eps/4.
\]

Taking $C$ sufficiently large makes the first term in the preceding
bound at most $\epsilon/2$, so the error is at most
$3\epsilon/4\le\epsilon$.

Finally, positive rescaling preserves both the threshold and
$(\tau,V)$-regularity, so we may normalize every $q_i$ to have
$L_2$ norm one. Constant-true constraints can be omitted, and an
intersection containing a constant-false constraint can be represented
by the single polynomial $-1$. If all constraints are omitted, use
the single polynomial $1$. Thus $g$ is an intersection of at most
$k$ degree-$2$ PTFs, regular with respect to the original subspace $V$.
\end{proof}

\section{Certificate Polynomial for Intersections of $k$ Degree-$2$ PTFs}\label{sec certificate polynomial}
In this section, we develop the main structural result, \Cref{th certificate}, for intersections of degree-$2$ polynomial threshold functions. We show that given a Boolean function $f(\x)$ that is a product of an intersection of $k$ degree-$2$ PTFs and any Boolean function $g(\w)$ that is independent of a direction $\v \in \R^d$. Suppose the randomized error of $f$ along $\v$ is at least $\alpha$, then there is a certificate polynomial of degree at most $2k$ correlates with $f$. In particular, such a polynomial can be written as a product as one polynomial that only depends on $\v$ direction and one independent of $\v$ direction.
As we show in \Cref{sec direction finding}, such a structure enables us to efficiently find such a direction $\v$.

\begin{theorem}\label{th certificate}
     Let $f(\x) = g(\w)\Ind\{q_1(\x)\ge 0,\dots,q_k(\x) \ge 0\}: \R^d \to \{0,1\}$, where for $i \in [k]$, $q_i$ is a degree-$2$ polynomial, $\w = \proj_{\v^\perp}(\x)$ and $g(\w) \in \v^{\perp} \to \{0,1\}$, $\v \in \s^{d-1}$.
     Suppose that $\rand_f(\v) \ge \alpha$, for some $\alpha \in (0,1)$, then there exists some $r \in [2k]$ and a polynomial $p(\w): \v^\perp \to \R$, with $\norm{p} = 1$ such that
    \begin{align*}
        \E_{\x\sim \cN(0,\bI)}[f(\x)h_r(\v\cdot\x)p(\w)] \ge \Omega(\alpha^{2^{O(k)}}),
    \end{align*}
    and $r+\deg(p) \le 2k$.
\end{theorem}

To gain an intuition for why some degree $2k$ polynomial can correlate with such a function $f$, we fix some $\w \perp \v$ and treat $f(\x)$ as a one-dimensional function along $\v$ direction.
If $g(\w) = 1$, then $f(\x)$ is the indicator of a union of at most $k+1$ intervals along the $\v$ direction, with at most $2k$ jump points.
This implies that for such a fixed layer $\w$, a degree $2k$ polynomial along direction $\v$ that crosses the boundaries of these intervals has a correlation with $f$.
However, such a construction does not imply a certificate polynomial required in \Cref{th certificate}, because if we average the correlation for such a polynomial over $\w$ spaces, the correlations can cancel with each other. Thus, some multiplies $p(\w)$ is needed to avoid such a cancellation.

Though, such an intuition does not directly give a way of constructing the required certificate, it motivates breaking down $f(\x)$ into layers by fixing $\w \in \v^\perp$ and looking at the decision boundary of $f$ as a one-dimensional function along $\v$. We introduce the following operator that will be heavily used throughout this work.

\subsection{Boundary Operator and Weighted Decision Boundary}\label{sec M operator}

\begin{definition}[Boundary Operator]\label{def boundary operator}
    Let $P$ be the class of polynomials over $\R^d$ and $\v \in \s^{d-1}$, we define the boundary operator $M_\v : P \to P$ as $M_\v(p) = (\x\cdot\v) p - \partial_\v p$.
\end{definition}

Before presenting the properties of the boundary operator, we first give the following facts about one-dimensional Hermite polynomials as well as the derivative of one-dimensional Boolean functions.
\begin{fact}\label{fact hermite 1D}
    Let $k \in \Z_+$ and let $h_k(z):\R \to \R$ be the 1D degree-$k$ Hermite polynomial. Then the following identities hold.
    \[
    d h_k(z)/dz=\sqrt{k}h_{k-1}(z),
    \qquad
    zh_k(z)=\sqrt{k+1}h_{k+1}(z)+\sqrt{k}h_{k-1}(z).
\]
\end{fact}

\begin{fact}\label{fact dirac}
     Let $f: \R \to \{0,1\}$ be a step function with a set of finite jump points $S$. For every $s \in S$, define $\Delta_f(s) = \lim_{\tau \to 0^+}f(s+\tau)-\lim_{\tau \to 0^+}f(s-\tau)$.
     Then, in the sense of distributions, $df/dx = \sum_{s \in S}\Delta_f(s)\delta_s$, where $\delta_s$ is the Dirac delta distribution at $s$.

\end{fact}

The boundary operator has two important properties following \Cref{fact hermite 1D} that will be used for our proof. The first one is \Cref{fact representation}.

\begin{fact}\label{fact representation}
    Let $p(\x)$: $\R^d \to \R$ be a polynomial with degree $r$. Let $\v \in \s^{d-1}$. Write $p(\x) = \sum_{i=0}^rh_i(\v\cdot\x)M_i(\w)$, where $\w = \proj_{\v^\perp}(\x)$ and $M_i(\w)$ is a polynomial over $\w$ with degree at most $r-i$.
    Then, $M_\v p(\x) = \sum_{i=0}^r \sqrt{i+1}h_{i+1}(\v \cdot \x)M_i(\w).$
\end{fact}
Roughly speaking, for every polynomial $p(\x)$, the Hermite decomposition of $M_\v p$ has no term that is totally independent of the $\v$ direction. In other words, if $\E_{\x\sim \cN(0,\bI)}[f(\x)M_\v p(\x)] \ge \beta$, then \Cref{fact representation} implies that there must be some $i \ge 1$ such that $\E_{\x\sim \cN(0,\bI)}[f(\x)h_i(\v\cdot\x)M_{i-1}(\w)] \ge \beta/(r+1)^{3/2}$.
Thus, to prove \Cref{th certificate}, it suffices to construct some certificate polynomial $p(\x)$ such that $M_\v p$ has non-trivial correlation with $f$.

The second property, \Cref{lm M operator} is an application of integration by parts. Roughly speaking, consider the correlation between a one-dimensional Boolean function $f(\x)$ along direction $\v$, and a polynomial $M_\v p$ under a Gaussian distribution. The correlation can be written as the sum of the value of $p$ (up to the sign change of $f$) at every boundary point weighted by its Gaussian density.

\begin{lemma}\label{lm M operator}

    Let $f: \R^d \to \{0,1\}$ be a Boolean function. For $\v \in \s^{d-1}$, write $\x = z\v+\w$, where
$z = \v\cdot\x$ and $\w = \proj_{\v^\perp}(\x)$. For every $\w \perp \v$, denote by $f_\w(z) = f(z\v+\w)$. Then for every polynomial $p: \R^d \to \R$, it holds
    \begin{align*}
        \E_{z \sim \cN(0,1)}[f_\w(z) M_\v p(z\v+\w)] = \E_{z \sim \cN(0,1)}[f'_\w(z)p(z\v+\w)]
    \end{align*}
    In particular, if $f_\w(z)$ is a step function with the set of finite jump points $S_\w$, then
    \begin{align*}
        \E_{z \sim \cN(0,1)}[f_\w(z) M_\v p(z\v+\w)] = \sum_{s \in S_\w}\Delta_{f_\w}(s)p(s\v+\w)\phi(s).
    \end{align*}
\end{lemma}

\begin{proof}[Proof of \Cref{lm M operator}]

We prove \Cref{lm M operator} using integration by parts.
We have
\begin{align*}
    \E_{z \sim \cN(0,1)}[f'_\w(z) p(z\v+\w)] & = \int_{\R} f'_\w(z) p_\w(z)\phi(z) dz
    = -\int_\R f_\w(z) (p_\w(z)\phi(z))' \\
    & = -\int_\R f_\w(z) (p'_\w(z)\phi(z)-zp_\w(z)\phi(z)) = \int_\R f_\w(z) (zp_\w(z)-p‘_\w(z))\phi(z) \\
    & = \E_{z \sim \cN(0,1)}[f_\w(z) M_\v p(z\v+\w)].
\end{align*}
If $f_\w(z)$ is a step function with finite jump points, then we have $f'_\w(z) = \sum_{s \in S_\w} \Delta_{f_\w}(s)\delta_s(z)$, which gives
\begin{align*}
        \E_{z \sim \cN(0,1)}[f_\w(z) M_\v p(z\v+\w)] = \sum_{s \in S_\w}\Delta_{f_\w}(s)p(s\v+\w)\phi(s).
\end{align*}

\end{proof}

\Cref{lm M operator} suggests if at every boundary point, the sign of $p$ matches with the sign of the derivative of $f$, then the correlation between $f$ and $M_\v p$ must be strictly positive.
However, even if $\Delta_{f_\w}(s)p(s\v+\w)>0$ is very large for every boundary point $s$, it does not implies $f(\x)$ has a large correlation with $M_\v p$, because after weighted by $\phi(s)$, the whole sum could still be small.
To give a quantitative lower bound for the correlation, we have to connect these weights with $\rand_f(\v)$. Specifically, we give the following lemma using the Gaussian isoperimetric inequality.

\begin{lemma}\label{lm error-mass}
    Let $f(\x): \R^d \to \{0,1\}$ be an intersection of $k$ degree-$2$ PTFs.
    Let $\v \in \s^{d-1}$ be a unit vector. For $\w \perp \v$, write $f_\w(s) = f(s\v+\w): \R \to \{0,1\}$ and let $S_\w \subseteq \R$ be the set of jump points of $f_\w$.
    There is a constant $C>1$, such that for every $\alpha \in (0,1)$ and for every $\v \in \s^{d-1}$, if $\rand_f(\v)>\alpha$, then
    \begin{align*}
        \E_{\w \sim \cN(0,\bI)}\left[\sum_{s \in S_\w} \phi(s)\right]> \alpha/C.
    \end{align*}
\end{lemma}

To prove \Cref{lm error-mass}, we will make use of the following 1-dimensional Gaussian isoperimetric inequality.

\begin{fact}\label{fact gaussian isopermetry}
    Let $S \subseteq \R$ be a set such that $\card{\partial S}<\infty$. Denote by $p_S = \Pr_{z \sim \cN(0,\bI)}[z \in S]$. Then
    \begin{align*}
        \sum_{s \in \partial S} \phi(s) \ge \phi(\Phi^{-1}(p_S)),
    \end{align*}
    where $\phi$ is the density function of a 1-dimensional Gaussian distribution and $\Phi(t) = \int_{-\infty}^t \phi(z)dz$.
\end{fact}

\begin{proof}[Proof of \Cref{lm error-mass}]
    We first show that there exists a constant $c>0$ such that for every $p \in [0,1]$, $\phi(\Phi^{-1}(p)) \ge cp(1-p)$.
    Without loss of generality, we assume that $p\le 1/2$. Let $s \le 0$ such that $\Phi(s) = p$. We show that $\phi(s)\ge cp(1-p)$.
    We consider two cases. If $\abs{s} \le 1$, then we have
    \begin{align*}
        \phi(s) \ge \phi(1) \ge 2\phi(1)/2 \ge 2\phi(1)p.
    \end{align*}
    If $\abs{s}>1$, then it holds $\Phi(s) \le \phi(s)/s$, which gives $\phi(s) \ge s \Phi(s) \ge p$.
    This implies that for every $p \in (0,1)$, it holds $\phi(\Phi^{-1}(p)) \ge 2\phi(1)p(1-p)$.

    By \Cref{fact gaussian isopermetry}, we know that
    \begin{align*}
        \sum_{s \in S_\w}\phi(s) &\ge \phi(\Phi^{-1}(\Pr_{\z\sim \cN(0,\bI)}[f_\w(\z)=1])) \\
        &\ge 2\phi(1)(\Pr_{\z\sim \cN(0,\bI)}[f_\w(\z)=1])(1-\Pr_{\z\sim \cN(0,\bI)}
        [f_\w(\z)=1]) \\
        & = 2\phi(1)\var_{\z\sim \cN(0,\bI)}[f_\w(\z)]
    \end{align*}
By taking expectation over $\w$ and \Cref{prop randomized error}, we get
    \begin{align*}
        \E_{\w \sim \cN(0,\bI)}[\sum_{s \in S_\w} \phi(s)] \ge  \phi(1)\rand_f(\v)>\phi(1)\alpha.
    \end{align*}
\end{proof}

\Cref{lm error-mass} implies that if $\rand_f(\v)$ is large for some direction $\v$, then in expectation of weighted sum of $\phi(s)$ over the boundary points of $f_\w$ is not too small.
This implies any $p$ such that $\Delta_{f_\w}(s)p(s\v+\w)>0$ is non-trivially large for most of the boundary points; then $M_\v p$ can be used as the certificate polynomial.

With this motivation, before giving the construction of our certificate polynomial, we first characterize the structure of boundary points of a one-dimensional intersection of PTF.
\begin{fact}\label{fact 1D boundary}
    Let $f(z) = \Ind \{p_1(z) \ge t_1,\dots, p_k(z) \ge t_k\}$, where for $i \in [k]$, $p_i(z)$ is a 1 dimensional degree-$2$ polynomial and $t_i \in \R$, be an intersection of $k$ degree-$2$ PTFs over $\R$. Let $S$ be the set of jump points of $f$. Then $S$ satisfies the following properties.
    \begin{enumerate}
        \item For each $s \in S$, let $C_s=\{p_i \mid p_i(s) - t_i = 0, i \in [k]\}$, then $C_s \neq \emptyset$ and $p_i(s)-t_i>0$ for every $p_i \not \in C_s$.
        \item For each $s \in S$, and for each $p_i \in C_s$ with $p'_i(s) \neq 0$, $\Delta_f(s) = \sign(p'_i(s))$.
    \end{enumerate}
\end{fact}

\begin{proof}[Proof of \Cref{fact 1D boundary}]
    To prove the first item, we first suppose that $C_s = \emptyset$, which implies that $q_i(s) - t_i \neq 0$ for each $i \in [k]$. By continuity, there exists a neighborhood of $s$, such that for every $z$ in the neighborhood and for every $i \in [k]$, $\sign(q_i(z)-t_i) = \sign(q_i(s)-t_i)$. This means $f$ does not change its sign at $s$, which gives a contradiction. Similarly, assume there exists some $q_i \not \in C_s$ such that $q_i(s) - t_i<0$. Then there exists a neighborhood of $s$ such that for every $z$ in the neighborhood, $q_i(z)-t_i<0$, which implies that $f(z) = 0$ and gives a contradiction.

    We next prove the second item. We first show that for every $q_i \in C_s$, either $q'_i(s) = 0$ or $\sign(q'_i(s)) = \sign(q'_j(s))$ for some other $q_j \in C_s$. Suppose not, then there exists $q_i, q_j \in C_s$ such that $q'_i(s)q'_j(s) < 0$. Since $q_i(s) - t_i = q_j(s) - t_j = 0$. By continuity, this implies that there exists a small neighborhood of $s$ such that $\sign(q_i(z)-t_i)\sign(q_j(z)-t_j)<0$ for every $z$ in the neighborhood. This implies that $f$ has no sign change at $s$. Now since $q_i(s)-t_i > 0$, for every $q_i \not \in C_s$ and $q'_i(s)$ has the same sign for $q_i \in C_s$, we know that $\Delta_f(s) = \sign(q'_i(s))$, for a $q'_i(s) \neq 0$.
\end{proof}

\subsection{Certificate Polynomials for Generic Instance and Root Decompositions}\label{sec certificate generic}

According to \Cref{fact 1D boundary}, we know that once we fix some $\w \perp \v$, the decision boundaries of $f_\w$ is a collection of roots of the defining polynomials $q_i = p_i - t_i$. Furthermore, for each boundary point $s$, we can cluster these $k$ polynomials into two groups. The first group of polynomials has $s$ as their roots, and furthermore their derivatives at $s$ have the same sign of $\Delta_{f_\w}(s)$. The other group of polynomials are strictly positive at $s$.

Based on this observation, we design a certificate polynomial, whose sign of value at every decision boundary of $f_\w$, for every $\w$, matches $\Delta_{f_\w}(s)$. Specifically, if we consider polynomial $Q_\v(\x):=\partial_\v(\prod_{i=1}^k q_i(\x))=\sum_{i=1}^k \prod_{j \neq i} q_j(\x) \partial_\v q_i(\x)$, then at each $\x =s\v + \w$, where $s$ is a boundary of $f_\w$, each of the $k$ terms in the summation is either $0$ or has the same sign as $\Delta_{f_\w}(s)$.
In particular, if a boundary point $s$ is a root of a single $q_i$ and $q_i$ does not have a double root at $s$, then $Q_\v$ gives a strictly positive contribution to the correlation between $f$ and $M_\v Q_\v$.

However, we need to derive a quantitative lower bound for the correlation in order to use it to design an algorithm. To do this,
we introduce parameters $\eta_d,\eta_c \in (0,1)$. $\eta_d$ is used to quantify the derivative of $q_i$ at its root $s$.
Intuitively, if $\abs{\partial_\v q_i}$ is small at $s$ (and the coefficient of $z^2$ in $q_i(z\v+\w)$ is bounded away from zero), then there should be another root of $q_i$ around $s$ and $s$ is approximately a double root of $q_i$.
$\eta_c$ is used to quantify the value of $q_i$ at $s$, where $s$ is not a root of $q_i$ but a root of a different defining polynomial $q_j$. The intuition here is that if $\abs{q_i}$ is tiny at $s$, then there must be a root of $q_i$ around $s$ and $s$ a an approximately root shared by both $q_i$ and $q_j$.
Given this intuition, we can partition the boundary points $S_\w$ of $f_\w$ into three sets.

\begin{definition}\label{def boundary partition}
    Let $f(\x) = g(\w)\Ind\{p_1(\x)\ge t_1,\dots,p_k(\x) \ge t_k\}: \R^d \to \{0,1\}$, where for $i \in [k]$, $p_i$ is a degree-$2$ polynomial with $\E_{\x\sim \cN(0,\bI)}[p_i(\x)] = 0, \var_{\x\sim \cN(0,\bI)}[p_i(\x)] = 1, t_i \in \R$ and $g(\w): \v^\perp \to \{0,1\}$, $\v \in \s^{d-1}$. For every $\w \perp \v$, denote by $q^{(\w)}(z) = q(z\v+\w)$.
    For $i \in [k]$, denote by $q_i(\x) = p_i - t_i$ and $Q_\v(\x) = \partial_\v (\prod_{i=1}^k q_i)$, where $\v \in \s^{d-1}$.
    For $\eta_d,\eta_c \in (0,1)$, $\w \perp \v$, we define
    \begin{enumerate}
        \item $G_\w(\eta_d,\eta_c)\subseteq S_\w$ to be the set of jump points $s$ of $f_\w$ such that there exists some $i \in [k]$ such that $\abs{(q^{(\w)}_i)'(s)}\ge \eta_d$ and for every $j \neq i$, $q^{(\w)}_j(s)\ge \eta_c$.
        \item $B_\w(\eta_d,\eta_c) \subseteq S_\w$ to be the set of jump points $s$ of $f_\w$, such that there exists some $i \in [k]$, such that $q^{(\w)}_i(s) = 0$ but $\abs{(q^{(\w)}_i)'(s)} \le \eta_d$.
        \item $C_\w(\eta_d,\eta_c) = S_\w \setminus( G_\w(\eta_d,\eta_c) \cup B_\w(\eta_d,\eta_c))$
    \end{enumerate}
\end{definition}

By \Cref{lm error-mass}, we know that if $\rand_f(\v)$ is non-trivially large, then $\E_{\w\sim \cN(0,\bI)}[\sum_{s \in G_\w(\eta_d,\eta_c)}\phi(s) + \sum_{s \in B_\w(\eta_d,\eta_c)}\phi(s) + \sum_{s \in C_\w(\eta_d,\eta_c)}\phi(s)]$ must be non-trivially large. By our above discussion, if the contribution from roots from $G_\w(\eta_d,\eta_c)$ is not too small, $M_\v Q_\v$ gives a nontrivially large correlation with respect to $f$. Formally, we give the following \Cref{lm generic certificate}.

\begin{lemma}[Certificate Polynomials for Generic Intersections of PTFs]\label{lm generic certificate}
    Let $f(\x) = g(\w)\Ind\{p_1(\x)\ge t_1,\dots,p_k(\x) \ge t_k\}: \R^d \to \{0,1\}$, where for $i \in [k]$, $p_i$ is a degree-$2$ polynomial with $\E_{\x\sim \cN(0,\bI)}[p_i(\x)] = 0, \var_{\x\sim \cN(0,\bI)}[p_i(\x)] = 1, t_i \in \R$ and $g(\w): \v^\perp \to \{0,1\}$, $\v \in \s^{d-1}$
    For $i \in [k]$, denote by $q_i(\x) = p_i - t_i$ and $Q_\v(\x) = \partial_\v (\prod_{i=1}^k q_i)$, where $\v \in \s^{d-1}$.
    For $\eta_d,\eta_c \in (0,1)$, $\w \perp \v$, \new{define $G_\w(\eta_d,\eta_c) \subseteq S_\w$} such that for each $s \in G_\w(\eta_d,\eta_c)$, there exists some $i \in [k]$ such that $\abs{(q^{(\w)}_i)'(s)}\ge \eta_d$ and for every $j \neq i$, $q^{(\w)}_j(s)\ge \eta_c$. Suppose that $\E_{\w \sim \cN(0,\bI)}[\sum_{s \in G_\w(\eta_d,\eta_c)} \phi(s)] \ge \alpha$, then
    \begin{align*}
        \E_{\x\sim \cN(0,\bI)}[f(\x)M_\v Q_\v(\x)] \ge \eta_d\eta_c^{k-1}\alpha.
    \end{align*}
\end{lemma}

\begin{proof}[Proof of \Cref{lm generic certificate}]
    For every $\w \perp \v$, denote by $f_\w(z) = f(z\v+\w)$ and define $S_\w\subseteq\R$, the set of jump points of $f_\w$. Notice that $\E_{\x\sim \cN(0,\bI)}[f(\x)M_\v Q_\v(\x)] = \E_{\w\sim \cN(0,\bI)}[\E_{z\sim \cN(0,1)}[f_\w(z)M_\v Q_\v(z\v+\w)]]$.
    By \Cref{lm M operator}, we have
    \begin{align*}
        \E_{\x\sim \cN(0,\bI)}[f(\x)M_\v Q_\v(\x)] &= \E_{\w\sim \cN(0,\bI)}[\E_{z\sim \cN(0,1)}[f_\w(z)M_\v Q_\v(z\v+\w)]] \\
        & = \E_{\w\sim \cN(0,\bI)}[\sum_{s \in S_\w}\Delta_{f_\w}(s)Q_\v(s\v+\w)\phi(s)] \\
        & = \E_{\w\sim \cN(0,\bI)}[\sum_{s \in S_\w}\sum_{i=1}^k (q^{(\w)}_i)'(s)\prod_{j\neq i}q^{(\w)}_j(s)  \Delta_{f_\w}(s)\phi(s)].
    \end{align*}
   By \Cref{fact 1D boundary}, for every $s \in S_\w$ and for every $i \in [k]$, $(q^{(\w)}_i)'(s)\prod_{j\neq i}q^{(\w)}_j(s)  \Delta_{f_\w}(s) \ge 0$. Furthermore, the strict inequality holds if and only if $(q^{(\w)}_i)'(s) \neq 0$ and $q^{(\w)}_j(s) > 0, j \neq i$, in which case, $(q^{(\w)}_i)'(s) \Delta_{f_\w}(s) = \abs{(q^{(\w)}_i)'(s)}$.

   This implies that
   \begin{align*}
       \E_{\x\sim \cN(0,\bI)}[f(\x)M_\v Q_\v(\x)] \ge \E_{\w\sim \cN(0,\bI)}[\sum_{s \in G_\w(\eta_d,\eta_c)}\sum_{i=1}^k \abs{(q^{(\w)}_i)'(s)}\prod_{j\neq i}q^{(\w)}_j(s)  \phi(s)] \ge \eta_d\eta_c^{k-1}\alpha.
   \end{align*}
\end{proof}

Although for a generic good instance $f$, the simple construction of $Q_\v$ can be used as a certificate polynomial, $Q_\v$ can easily fail to give enough correlation for some very natural instance. For example, even if $f(\x)$ is a simple intersection of two PTFs defined by $q_1(\x),q_2(\x)$, the coefficients of $q_1,q_2$ could be approximately parallel, which implies almost no roots will be in $G_\w(\eta_d,\eta_c)$ and $M_\v Q_\v$ gives nearly zero correlation with respect to $f$. For larger $k$, the configuration of these $k$ polynomials can be even more complicated.
Thus, the most challenging part of proving \Cref{th certificate} will be to construct certificate polynomials that work even if no roots are generic good.
Our proof will provide analysis for $B_\w(\eta_d,\eta_c)$(double roots) and $C_\w(\eta_d,\eta_c)$(commonly sharing roots) separately.
In \Cref{sec double roots}, we use the discriminant of a quadratic as a tool to analyze the case where double roots have enough Gaussian mass. In \Cref{sec factorization}, we develop a robust factorization lemma for quadratic functions under Gaussian distribution and use it to analyze the case where commonly sharing roots have enough Gaussian mass. Finally, In \Cref{sec certificate completion}, we combine all the techniques developed in \Cref{sec certificate polynomial} to prove \Cref{th certificate}.

\subsection{Analysis for Double Roots}\label{sec double roots}

Recall our discussion in the previous subsection, the generic polynomial that we construct will only give a tiny correlation, if a defining polynomial has a small derivative at its root.
In this section, we provide analysis for this case. For a fixed polynomial $q(\x)$, $\v \in \s^{d-1}$, and $\w \perp \v$, we consider the one dimensional $q^{(\w)}(z) = q(z\v+\w)$.
The observation is that we can view $q^{(\w)}(z)$ as a one-dimensional quadratic function along $\v$ direction, whose coefficients are polynomials in $\w$.
Specifically, we can always write $q^{(\w)}(z) = az^2 + b(\w)z + c(\w)$, where $\deg(b) \le 1$ and $\deg(c) \le 2$.
Following elementary algebra, for $s$, a root of $q^{(\w)}$, the square of the derivative of
$q^{(\w)}(s)$ is exactly equal to the discriminant of $q^{(\w)}$. i.e. $\Delta(\w):=b(\w)^2-4ac(\w)$.
This implies that if $\abs{(q^{(\w)})'(s)}$ is large, then $\Delta(\w)$ must be large.
On the other hand, if $\Delta(\w)$ is small, then after normalization, the two roots of $q^{(\w)}$ must be very close to each other, which implies that by introducing a small error, $\Ind\{q^{(\w)} \ge 0\}$ is nearly 0 or 1.

Although, such an analysis is for a single fixed $\w$, since $\Delta(\w)$ is a polynomial of degree at most 2, Gaussian concentration and anti-concentration properties enable us to extend the analysis to all layers $\w$.
Specifically, in \Cref{lm no double roots}, we show that is $\norm{\Delta(\w)}$ is not very small, then $\E_{\w \sim \cN(0,\bI)}[\sum_{s \in B_\w(\eta_d,\eta_c)} \phi(s)]$ must be very small and can be ignored. On the other hand, in \Cref{lm double roots}, we show that if $\norm{\Delta(\w)}$ is small, then $\Ind\{q(\x) \ge 0\}$ is close to some function $g(\w)$ that only depends on $\v^\perp$ and thus can be removed from the intersection.

\begin{lemma}\label{lm no double roots}
  Let $p(\x)$ be a degree-$2$ polynomial such that  $\E_{\x \sim \cN(0,\bI)}[p(\x)]= 0$ and $\var_{\x \sim \cN(0,\bI)}[p(\x)]=1$. Let $t \in \R$ and $q(\x): = p(\x) - t$.
       For $\v \in \s^{d-1}$, write $\x = z\v+\w$, where
$z = \v\cdot\x$ and $\w = \proj_{\v^\perp}(\x)$. For every $\w \perp \v$, denote by $q^{(\w)}(z) = q(z\v+\w)$.
    Write $q^{(\w)}(z) = az^2 + b(\w)z + c(\w)$, where $a \in \R$, $b(\w),c(\w)$ are polynomial over $\w$, with $\deg(b)\le 1$ and $\deg(c) \le 2$. Let $\Delta(\w) = b^2(\w)-4ac(\w)$.
    For every $\alpha \in (0,1/4)$ and $\eta_d \in (0,1)$, suppose that $\E_{\w \sim \cN(0,\bI)}[\Delta(\w)^2] > \alpha$, then $ \Pr_{\w\sim\cN(0,\bI)}[\exists s, q^{(\w)}(s) = 0, \abs{(q^{(\w)})'(s)} \le \eta_d] \le O(\eta_d/\alpha^{1/4})$.
\end{lemma}

\begin{proof}[Proof of \Cref{lm no double roots}]
Denote by $D(\w,z) = (q^{(\w)})'(z)$.
    For every $\w \perp \v$ and for every $z \in \R$, we have $D(\w,z) = 2az + b$ and thus $D^2(\w,z) = 4a^2z^2 + b^2(\w) +4ab(\w)z$.
    If $q^{(\w)}(z) = 0$, then we have
    \begin{align*}
        D^2(\w,z) = b^2-4ac(\w) = \Delta(\w).
    \end{align*}
This implies that if $s_1,s_2$ are two roots of $q^{(\w)}(z)$, then $\abs{(q^{(\w)})'(s_1)}=\abs{(q^{(\w)})'(s_2)} = \sqrt{\Delta(\w)}$.
This implies that
\begin{align*}
    \Pr_{\w\sim\cN(0,\bI)}[\exists s, q^{(\w)}(s) = 0, \abs{(q^{(\w)})'(s)} \le \eta_d] \le \Pr_{\w \sim \cN(0,\bI)}[\abs{\Delta(\w)} \le \eta_d^2] \le O(\sqrt{\eta_d^2/\norm{\Delta(\w)}}) \le O(\eta_d/\alpha^{1/4}).
\end{align*}
\end{proof}

\begin{lemma}\label{lm double roots}
    Let $p(\x)$ be a degree-$2$ polynomial such that
    $\E_{\x \sim \cN(0,\bI)}[p(\x)] = 0$ and
    $\var_{\x \sim \cN(0,\bI)}[p(\x)] = 1$.
    Let $t \in \R$ and $q(\x) := p(\x)-t$.
    For $\v \in \s^{d-1}$, write $\x = z\v+\w$, where
    $z = \v \cdot \x$ and $\w = \proj_{\v^\perp}(\x)$.
    For every $\w \perp \v$, denote by
    $q^{(\w)}(z) = q(z\v+\w)$.
    Write  $q^{(\w)}(z) = az^2+b(\w)z+c(\w),$
    where $a \in \R$ and $b(\w),c(\w)$ are polynomials over
    $\w$ with $\deg(b) \le 1$ and $\deg(c) \le 2$.
    Let $\Delta(\w) = b(\w)^2-4ac(\w)$.
    For every $\alpha \in (0,1/4)$, suppose that $\E_{\w \sim \cN(0,\bI_{\v^\perp})}
        [\Delta(\w)^2] \le \alpha.$
    Then there exists a Boolean function
    $g : \v^\perp \to \{0,1\}$ such that
    \[
        \Pr_{\x \sim \cN(0,\bI)}
        [g(\w) \neq \Ind\{q(\x) \ge 0\}]
        \le \Tilde{O}(\alpha^{1/16}).
    \]
\end{lemma}

\begin{proof}[Proof of \Cref{lm double roots}]
    We first normalize $q$ rather than truncate the threshold $t$.
    Since $p$ has mean zero and variance one,
        $s := \norm{q} = \sqrt{1+t^2} \ge 1$.
    Replacing $q$ by $q/s$ replaces $a,b,c$ by
    $a/s,b/s,c/s$, respectively, and its discriminant by
    $\Delta/s^2$. This positive rescaling does not change
    $\Ind\{q \ge 0\}$, and $\E_{\w \sim \cN(0,\bI_{\v^\perp})}
        \left[\left(\frac{\Delta(\w)}{s^2}\right)^2\right]
        \le \alpha.$
    Thus, reusing the same notation for the rescaled quantities,
    we may assume that $        \norm{q} = 1,
        \norm{\Delta} \le \alpha^{1/2}.$

    Set $\lambda = \alpha^{1/3}$.
    We consider two cases. In each case, we construct a polynomial
    $Q$ of degree at most two and a Boolean function $g(\w)$ such
    that $\Ind\{Q(\x) \ge 0\} = g(\w)$ almost surely and $        \norm{q-Q} \le C\alpha^{1/6},$
    where $C>0$ is a universal constant.

    \paragraph{Case 1: $\abs{a} \ge \lambda$.}
    Completing the square gives
    \[
        q^{(\w)}(z)
        =
        a\left(z+\frac{b(\w)}{2a}\right)^2
        -\frac{\Delta(\w)}{4a}.
    \]
    Define
    \[
        Q(z\v+\w)
        :=
        a\left(z+\frac{b(\w)}{2a}\right)^2,
        \qquad
        g(\w) := \Ind\{a>0\}.
    \]
    Since $b$ is affine, $Q$ has degree at most two.
    Moreover, $a \neq 0$, so the square vanishes only on a
    Gaussian-null set. Hence
    $\Ind\{Q(\x) \ge 0\} = g(\w)$ almost surely.
    Finally,
    \[
        \norm{q-Q}
        =
        \frac{\norm{\Delta}}{4\abs{a}}
        \le
        \frac{\alpha^{1/2}}{4\lambda}
        =
        \frac14\alpha^{1/6}.
    \]

    \paragraph{Case 2: $\abs{a} < \lambda$.}
    Define  $d(\w) := c(\w)+a
        =
        \E_{z \sim \cN(0,1)}[q(z\v+\w)].$
    Then
    \[
        q(z\v+\w)
        =
        a(z^2-1)+b(\w)z+d(\w).
    \]
    Gaussian orthogonality gives
        $1 = \norm{q}^2
        = 2a^2+\norm{b}^2+\norm{d}^2$.
    In particular, $\norm{d} \le 1$, and therefore
    \[
        \abs{
            \E_{\w \sim \cN(0,\bI_{\v^\perp})}[c(\w)]
        }
        \le
        \norm{d}+\abs{a}
        \le 1+\abs{a}.
    \]
    Using the discriminant identity
    $\Delta(\w)=b(\w)^2-4ac(\w)$, we obtain
    \begin{align*}
        \norm{b}^2
        &=
        \E_{\w \sim \cN(0,\bI_{\v^\perp})}[\Delta(\w)]
        +
        4a\E_{\w \sim \cN(0,\bI_{\v^\perp})}[c(\w)]\\
        &\le
        \norm{\Delta}
        +
        4\abs{a}
        \abs{
            \E_{\w \sim \cN(0,\bI_{\v^\perp})}[c(\w)]
        }\\
        &\le
        \alpha^{1/2}+4\abs{a}(1+\abs{a})\\
        &\le
        \alpha^{1/2}+8\lambda,
    \end{align*}
    where the last inequality uses $\abs{a}<\lambda<1$.

    Set
    \[
        Q(z\v+\w) := d(\w),
        \qquad
        g(\w) := \Ind\{d(\w) \ge 0\}.
    \]
    Again by Gaussian orthogonality,
    \begin{align*}
        \norm{q-Q}^2
        =
        2a^2+\norm{b}^2
        \le
        2\lambda^2+\alpha^{1/2}+8\lambda
        \le
        11\alpha^{1/3}.
    \end{align*}
    Thus $\norm{q-Q} \le \sqrt{11}\alpha^{1/6}$, as required.

    \paragraph{Converting the perturbation into disagreement.}
    In both cases, $\norm{q}=1$ and
    $\norm{q-Q}^2 \le C\alpha^{1/3}$.
    For every $u>0$, Gaussian anticoncentration for degree-two
    polynomials and Markov's inequality give
    \begin{align*}
        &\Pr_{\x \sim \cN(0,\bI)}
        [\Ind\{q(\x) \ge 0\} \neq g(\w)]\\
        &\qquad=
        \Pr_{\x \sim \cN(0,\bI)}
        [\Ind\{q(\x) \ge 0\} \neq \Ind\{Q(\x) \ge 0\}]\\
        &\qquad\le
        \Pr_{\x \sim \cN(0,\bI)}[\abs{q(\x)} \le u]
        +
        \Pr_{\x \sim \cN(0,\bI)}
        [\abs{q(\x)-Q(\x)} \ge u]\\
        &\qquad\le
        C\sqrt{u}+\frac{\norm{q-Q}^2}{u^2}\\
        &\qquad\le
        C\sqrt{u}+\frac{C\alpha^{1/3}}{u^2}.
    \end{align*}
    Taking $u=\alpha^{2/15}$ yields
    \[
        \Pr_{\x \sim \cN(0,\bI)}
        [\Ind\{q(\x) \ge 0\} \neq g(\w)]
        \le C\alpha^{1/15}.
    \]
\end{proof}

Together, \Cref{lm no double roots} and \Cref{lm double roots} imply that by selecting a proper threshold for $\norm{\Delta}$, any defining quadratic polynomial of the target function either has nearly no roots in $B_\w(\eta_d,\eta_c)$ in expectation, or up to a small error, we can remove it from the intersection and absorb it into some function $g(\w)$ that only depends on $\w$.
This implies that if $\E_{\w \sim \cN(0,\bI)}[\sum_{s \in G_\w(\eta_d,\eta_c)} \phi(s)]$ is small, then most of the roots can be thought to be in $C_\w(\eta_d,\eta_c)$.
In \Cref{sec factorization}, we provide a systematic way to analyze and construct certificate polynomials in this case.

\subsection{Robust Factorization of Quadratic Polynomials under Gaussian Distribution}\label{sec factorization}

We now turn to the case where a significant portion of the decision boundary arises from roots that are approximately shared by multiple defining polynomials. As discussed in \Cref{sec double roots}, we may assume that a defining polynomial has a not too small derivative at its roots for most of $\w \perp \v$.
However, the generic certificate given by \Cref{lm generic certificate} will still give only arbitrarily small correlation, if at a root of one defining polynomial, some other polynomial has nearly zero value, even though the corresponding boundary contributes significantly to the randomized error. In \Cref{lm small root distance}, we show that in this case, the two polynomials must have roots that are sufficiently close to each other.
However, this does not mean we can merge these commonly sharing roots naively into one root, because there dependence on $\w$ can be very complicated and even for a fixed $\w$, the arrangement of the $2k$ roots of the $k$ polynomials can be very complicated.
Thus, to handle this case, we need a certificate that takes advantage of the structure of these nearly shared roots rather than treating them as separate boundaries.

Our key observation is that if two polynomials have nearly sharing roots at a non-trivial fraction of layers, $\w \perp \v$, then the two polynomials must approximately share some global algebraic structures.
Such an observation starts from Euclidean algorithms for finding a greatest common divisor for one-dimensional polynomials.
Similar to the previous section, for each fixed $\w \perp \v$, and for each $i \in [k]$, we write $q_i^{(\w)}(z)=a_iz^2 + b_i(\w)z + c_i(\w)$ as a univariate polynomial in $z$.
Recall that two univariate polynomials have a common root if and only if their resultant (\Cref{def resultant}) vanishes. This implies that if the resultant of $q_1^{(\w)}(z),q_2^{(\w)}(z)$ is sufficiently small, we are able to find univariate affine polynomials $L,\ell_1,\ell_2$ such that $\norm{q_i^{(\w)}(z)-L\ell_i}$ is small.
In fact, in \Cref{fact resultant}, we point out the resultant of $q^{(\w)}_1,q^{(\w)}_2$ as a function of $\w$ is a polynomial of degree at most 4. Thus, using Gaussian anti-concentration, in \Cref{lm close root to small resultant}, we show that as long as $q_1,q_2$ have nearly shared roots on a non-trivial fraction of Gaussian layers, the $L_2$ norm of their resultant must be small.
The hope is that we would like to show in this case we have multivariate affine functions $L,\ell_1,\ell_2$ such that $q_i \approx L\ell_i$.
However, a direct application of Euclidean algorithm over these two polynomials does not imply this.
Although the defining polynomials are univariate quadratics after fixing $\w$, their coefficients are themselves polynomials in $\w$. The Euclidean algorithm therefore naturally produces rational functions of $\w$, which need not correspond to low-degree polynomial factors in the original variables. Indeed, as we will mention later, even if two polynomials share a root on every slice, they need not have a common affine factor.

This motivates our main technical result, \Cref{lm small resultant to factorization}, which gives a robust factorization lemma for two multivariate quadratics whose resultant is small.
In \Cref{lm small resultant to factorization}, we show that when the resultant is sufficiently small, there are only three cases: either one quadratic is nearly independent of $\v$ direction, or the two quadratics are nearly parallel or they can be approximately factorized into the form of $L\ell_i$. If we are in the first two cases, then we can simply remove one defining polynomial from the intersection and work on the rest of the ones. In \Cref{sec certificate completion}, we will show that once we find some $L$ using the factorization lemma that contributes enough mass for the boundary points, then we can use $L$ as a pivot to cluster the $k$ defining polynomials into two groups to construct a new certificate polynomials that bypass the difficulty of roots sharing.

\paragraph{Root Distance and Resultants}

To begin with, we start by showing that for a given two quadratics, under suitable normalization, if one polynomial evaluates very small at a root of another polynomial, then it has a nearby root.

\begin{lemma}\label{lm small root distance}
     For $i \in [2]$, let $p_i:\R^d \to \R$ be a degree-$2$ polynomial with  $\E_{\x \sim \cN(0,\bI)}[p_i(\x)]= 0$ and $\var_{\x \sim \cN(0,\bI)}[p_i(\x)]=1$.
    For $i \in [2]$, denote by $q_i(\x) = p_i(\x)-t_i$, $t_i \in \R$. For $\v \in \s^{d-1}$, write $\x = z\v+\w$, where
$z = \v\cdot\x$ and $\w = \proj_{\v^\perp}(\x)$. For every $\w \perp \v$, denote by $q_i^{(\w)}(z) = q_i(z\v+\w)$. Let $s_1,s_2$ be two roots of $q^{(\w)}_i(z)$, if exist and let $t \in \R$. Suppose that $\abs{s_1-t} \le \abs{s_2-t}$. Let $\eta_d,\eta_c \in (0,1)$. Suppose that $\abs{(q_i^{(\w)})'(s_1)}\ge \eta_d$ and $\abs{q^{(\w)}_i(t)} \le \eta_c$, then $\abs{s_1-t} \le 2\eta_c/\eta_d$.
\end{lemma}

\begin{proof}[Proof of \Cref{lm small root distance}]
Write $q=q_i^{(\w)}$. If $q$ is affine, then
$\abs{q(t)}=\abs{q'(s_1)}\abs{t-s_1}
\ge \eta_d\abs{t-s_1}$, which gives the desired bound.

Otherwise, write $q(z)=a_i(z-s_1)(z-s_2)$.
By the triangle inequality and the assumption
$\abs{s_1-t}\le\abs{s_2-t}$,
\[
    \abs{s_1-s_2}
    \le \abs{s_1-t}+\abs{s_2-t}
    \le 2\abs{s_2-t}.
\]
Since $\abs{q'(s_1)}=\abs{a_i}\abs{s_1-s_2}$, we obtain
\begin{align*}
    \eta_c
    &\ge \abs{q(t)}
     = \abs{a_i}\abs{s_1-t}\abs{s_2-t} \\
    &\ge \frac12\abs{q'(s_1)}\abs{s_1-t}
     \ge \frac{\eta_d}{2}\abs{s_1-t}.
\end{align*}
Thus $\abs{s_1-t}\le 2\eta_c/\eta_d$.
\end{proof}

Next, we introduce the key concept, resultant that will be considered in the rest of the section.

\begin{definition}[Resultant]\label{def resultant}
    For $i \in [2]$, let $q_i(z) = \sum_{r=0}^{d_i} a_{ir}z^{r}: \R \to \R$. The resultant for $(q_i,q_j)$ is defined as $R(q_i,q_j) = a_{id_i}^{d_j}\prod_{t=1}^{d_i}q_j(r_t)$, where for $t \in [d_i]$, $r_t$ is the $t$th root (multiple roots and complex roots are also considered) of $q_i$.
\end{definition}
It is not hard to see that two polynomials have an exact same root if and only if their resultant is zero. Besides, resultant also has the following useful property that will be used in our analysis later.
\begin{fact}\label{fact resultant order}
    Let $q_i: \R \to \R$ be a polynomial, $i \in [2]$. Then $\abs{R(q_1,q_2)} = \abs{ R(q_2,q_1)}$.
\end{fact}
The central property that will be used in our proof is that a resultant for 2 quadratics $q^{(\w)}_1,q^{(\w)}_2$ is always a polynomial in $\w$.

\begin{fact}\label{fact resultant}
    Let $q_1,q_2:\R^d \to \R$ be two polynomials with degree at most $2$, then $R(q^{(\w)}_1,q^{\w}_2)$ is a polynomial over $\R^d$ with degree at most 4. Furthermore,
    for $\w \perp \v$ and $i \in [2]$, write $q_i^{(\w)} = a_iz^2+b_i(\w)z + c(\w)$, where $a_i \in \R$, $b_i(\w),c_i(\w)$ are polynomials over $\w$, with $\deg(b_i)\le 1$ and $\deg (c_i) \le 2$.
  Define polynomials over $\w$, $U(\w) = a_2b_1(\w)-a_1b_2(\w)$ and $V(\w) =a_2c_1(\w) - a_1c_2(\w)$.
  If $a_1 \neq 0$, then
  \begin{align*}
      R(q^{(\w)}_1,q^{(\w)}_2) = a_1^{\deg(q_2) - 2} (U(\w)^2c_1(\w)/a_1 -U(\w)V(\w)b_1(\w)/a_1 + V^2(\w)).
  \end{align*}
\end{fact}

\begin{proof}[Proof of \Cref{fact resultant}]
  For $\w \perp \v$ and $i \in [2]$, write $q_i^{(\w)} = a_iz^2+b_i(\w)z + c(\w)$, where $a_i \in \R$, $b_i(\w),c_i(\w)$ are polynomials over $\w$, with $\deg(b)\le 1$ and $\deg (c) \le 2$.
  Define polynomials over $\w$, $U(\w) = a_2b_1(\w)-a_1b_2(\w)$ and $V(\w) =a_2c_1(\w) - a_1c_2(\w)$.
  Notice that
  \begin{align}\label{eq euclidean}
      a_2q_1^{(\w)}(z) - a_1q_2^{(\w)}(z) = U(\w)z +V(\w).
  \end{align}

\paragraph{Case 1: $a_1 \neq 0$}
In the first case, we consider the case where $a_1 \neq 0$, which implies that $q_1^{(\w)}$ always has degree 2.
 Let $r_1,r_2$ be two roots for $q^{(\w)}_1$
  and recall that $R(q_1^{(\w)},q_2^{(\w)})=a_1^{\deg(q_2)}q_2^{(\w)}(r_1)q_2^{(\w)}(r_2)$. We consider three different cases.
  By \eqref{eq euclidean}, we know that $q_2^{(\w)}(r_i) = -a_1^{-1}(U(\w)r_i+V(\w))$. This implies that
  \begin{align*}
      (U(\w)r_1+V(\w))(U(\w)r_2+V(\w)) & = U(\w)^2(r_1r_2) + U(\w)V(\w)(r_1+r_2) + V^2(\w) \\
      & = U(\w)^2c_1(\w)/a_1 -U(\w)V(\w)b_1(\w)/a_1 + V^2(\w),
  \end{align*}
  which is a polynomial over $\w$ with degree at most $4$.
  In particular,
  \begin{align} \label{eq resultant representation}
      R(q^{(\w)}_1,q^{(\w)}_2) = a_1^{\deg(q_2) - 2} (U(\w)^2c_1(\w)/a_1 -U(\w)V(\w)b_1(\w)/a_1 + V^2(\w)).
  \end{align}
\paragraph{Case 2: $a_1=0$, $b_1 \neq 0$}
In this case, we know that the only root of $q^{(\w)}_1$ is $r_1 = - c_1(\w)/b_1(\w)$, which gives $R(q^{(\w)}_1,q^{(\w)}_2) = b_1(\w)^{\deg(q_2^{(\w)})}q_2^{(\w)}(- c_1(\w)/b_1(\w))$.
By expanding $q^{(\w)}_2$, we have three cases. If $a_2 \neq 0$, then
\begin{align*}
    R(q^{(\w)}_1,q^{(\w)}_2) = a_2c_1^2(\w) - b_1(\w)b_2(\w)c_1(\w) + b_1^2(\w)c_2(\w),
\end{align*}
which has a degree at most 4. If $a_2 = 0$ but $b_2 \neq 0$, then
\begin{align*}
    R(q^{(\w)}_1,q^{(\w)}_2) = b_1(\w)c_2(\w) - b_2(\w)c_1(\w),
\end{align*}
which has degree at most 3. If $a_2 = 0, b_2 = 0$, then $R(q^{(\w)}_1,q^{(\w)}_2) = c_2(\w)$, which has a degree at most 2.
\paragraph{Case 3: $a_1=0,b_1=0$}
In this case, we always have $ R(q^{(\w)}_1,q^{(\w)}_2) = c_1(\w)^{\deg (q_2)}$, which has degree at most $4$.
\end{proof}

Using \Cref{fact resultant}, we next show that if there is a descent probability over $\w$ such that $\abs{q^{(\w)}_j}$ is small at a root of $q_i^{(\w)}$, then it must be the case that $R(q^{(\w)}_i,q^{(\w)}_j)$ has a small norm.

\begin{lemma}\label{lm close root to small resultant}
    For $i \in [k]$, let $p_i(\x)$ be a degree-$2$ polynomial with $\E_{\x \sim \cN(0,\bI)}[p_i(\x)]= 0$ and $\var_{\x \sim \cN(0,\bI)}[p_i(\x)]=1$. Let $t_i \in \R$ and $q_i(\x): = p_i(\x) - t_i$.
    Let $f(\x) = \Ind\{q_1(\x)\ge 0,\dots,q_k(\x) \ge 0\}$. Let $\eta_d,\eta_c \in (0,1)$. For $\w \perp \v$, let $C^{(\w)}_{12}$ be the set of jump points of $f_\w$ such that $q^{(\w)}_1(s) = 0$, and $\abs{q^{(\w)}_2(s)} \le \eta_c$. Suppose that $\E_{\w\sim \cN(0,\bI)}[\sum_{s \in C^{(\w)}_{12}}\phi(s)] \ge \alpha$, then  $\norm{R(q^{(\w)}_1,q^{(\w)}_2)} \le O(\eta_c\log(1/\alpha)/\alpha^4)$.
\end{lemma}

\begin{proof}[Proof of \Cref{lm close root to small resultant}]
We write $q^{(\w)}_i(z) = a_i z^2 + b_i(\w)z + c_i(\w)$, for $i \in [2]$, where $\deg(b_i) \le 1$ and $\deg(c_i) \le 2$. Notice that The degree of $q^{(\w)}_i(z)$ is at most 2. Our proof will thus consider two cases on the degree of $q_i^{(\w)}$.

\paragraph{Case 1: $\deg(q_1^{(\w)}) = 2$}
We start with the case where $\deg(q_1^{(\w)}) = 2$.
Let $r_1,r_2$ be the two roots of $q^{(\w)}_1$. Then, we have
\begin{align*}
    q_2^{(\w)}(r_1) - q_2^{(\w)}(r_2) & = a_2(r_1-r_2)(r_1+r_2) +b_2(\w)(r_1-r_2) \\
    & = (r_1-r_2)(a_2(r_1+r_2) +b_2(\w)) \\
    & = (r_1-r_2)(b_2(\w)-a_2b_1(\w)/a_1) \\
    & = -(r_1-r_2)U(\w)/a_1.
\end{align*}
On the other hand,
\begin{align*}
    a_1(r_1-r_2) = -(2a_1r_2+b_1(\w)) = -(q^{(\w)}_1)'(r_2).
\end{align*}
Together, we obtain that
\begin{align*}
    a_1^2(q_2^{(\w)}(r_1)-q_2^{(\w)}(r_2)) = -a_1(r_1-r_2)U(\w) = U(\w)(q^{(\w)}_1)'(r_2).
\end{align*}
This implies that
\begin{align*}
    R(q^{(\w)}_1,q^{(\w)}_2) & = a_1^{\deg(q_2)}q_2^{(\w)}(r_1)q_2^{(\w)}(r_2)\\
    & = q_2^{(\w)}(r_1)(a_1^{\deg(q_2)-2}(a_1^2q^{(\w)}_2(r_1) + a_1(r_1-r_2)U(\w))) \\
    & = q_2^{(\w)}(r_1)(a_1^{\deg(q_2)-2}(a_1^2q^{(\w)}_2(r_1) - U(\w)(q^{(\w)}_1)'(r_2)))
\end{align*}
We consider the degree of $q^{(\w)}_2$. If $\deg(q^{(\w)}_2) = 2$, then
\begin{align}\label{eq resultant rewrite}
    R(q^{(\w)}_1,q^{(\w)}_2) = q_2^{(\w)}(r_1)(a_1^2q^{(\w)}_2(r_1) - U(\w)(q^{(\w)}_1)'(r_2))
\end{align}
Similarly, if $\deg(q^{(\w)}_2) = 1$, then we have
\begin{align}\label{eq resultant rewrite deg 1}
        R(q^{(\w)}_1,q^{(\w)}_2) = q_2^{(\w)}(r_1)(a_1q^{(\w)}_2(r_1) + b_2(\w)(q^{(\w)}_1)'(r_2)).
\end{align}
And if $\deg(q^{(\w)}_2) =0$, then we have $R(q^{(\w)}_1,q^{(\w)}_2) = (q_2^{(\w)}(r_1))^2$.

Recall that $q_i^{(\w)}(z) = a_i(z^2-1) + b_i(\w)z + c_i(\w)+a_i = p_i(z\v+\w) - t_i$. Since $\E_{\x\sim \cN(0,\bI)}[p_i(\x)] = 0$ and $\var_{\x\sim\cN(0,\bI)}[p_i(\x)] = 1$.
This implies that $\abs{a_i} \le 1/\sqrt{2}$ and $\norm{b_i(\w)} \le \sqrt{2}$.
By the tail probability of Gaussian distribution, we know that for $T:=O(\sqrt{\log(1/\alpha)})$, $\Pr_{\w\sim\cN(0,\bI)}[\abs{b_i}>T] \le \alpha/32$.

For each $\w \perp \v$, we denote by $Z^{(\w)}$ the set of roots $s$ of $q^{(\w)}_1$ such that $\abs{q^{(\w)}_2(s)} \le \eta_c$, with $\phi(s)\ge\alpha/16$. Then, for each $s \in Z^{(\w)}$, we have $\abs{s} \le T$.
We will next show that for every $\w$, such that $\abs{b_1(\w)} \le T$, $\abs{b_2(\w)} \le T$ and $Z^{(\w)} \neq \emptyset$, $R(q_1^{(\w)},q_2^{(\w)})$ is small.
Let $r_1 = s$. If $\deg(q^{(\w)}_2) = 2$, then
by \eqref{eq resultant rewrite}, we have
\begin{align*}
   \abs{R(q^{(\w)}_1,q^{(\w)}_2) }& = \abs{q_2^{(\w)}(s)(a_1^{2}q^{(\w)}_2(s)+U(\w)(q_1^{(\w)})'(s))}\\
    &\le \eta_c^2 + \eta_c\abs{U(\w)}\abs{2a_1s+b_1(\w)} \\
    & \le \eta_c^2 + \eta_c(\abs{a_1b_2(\w)}+\abs{a_2b_1(\w)})\abs{2a_1s+b_1(\w)} \\
    & \le 10\eta_c T^2 = O(\eta_c\log(1/\alpha)).
\end{align*}
If $\deg(q^{(\w)}_2) = 1$, then by \eqref{eq resultant rewrite deg 1}, we have
\begin{align*}
   \abs{R(q^{(\w)}_1,q^{(\w)}_2) }& = \abs{q_2^{(\w)}(s)(a_1q^{(\w)}_2(s) - b_2(\w)(q^{(\w)}_1)'(s))}\\
    &\le \eta_c^2 + \eta_c\abs{b_2(\w)}\abs{2a_1s+b_1(\w)} \\
    & \le 10\eta_c T^2 = O(\eta_c\log(1/\alpha)).
\end{align*}
If $\deg(q^{(\w)}_2) = 0$, then trivially, we have $\abs{R(q^{(\w)}_1,q^{(\w)}_2) } \le \eta_c^2$.

Now since $\E_{\w \sim \cN(0,\bI)}[\sum_{s \in C_{12}(\w)}\phi(s)] \ge \alpha$, we know that $\E_{\w \sim \cN(0,\bI)}[\sum_{s \in Z(\w)}\phi(s)] \ge 7\alpha/8$. Since $\phi(s) \le 1/\sqrt{2\pi}$, this implies that
\begin{align*}
  \Pr_{\w\sim \cN(0,\bI)}[\abs{R(q^{(\w)}_1,q^{(\w)}_2) } \le O(\eta_c\log(1/\alpha))] \ge   \Pr_{\w\sim \cN(0,\bI)}[\abs{b_1(\w)}\le T,\abs{b_2(\w)}\le T,\card{Z}^{(\w)}\ge 1] \ge \alpha/2.
\end{align*}
Since $R(q^{(\w)}_i,q^{(\w)}_j)$ is a polynomial with degree at most $4$ by \Cref{fact resultant}, using Carbery–Wright anti-concentration inequality, we have for every $\tau>0$, we have
\begin{align*}
\Pr_{\w\sim \cN(0,\bI)}[\abs{R(q^{(\w)}_1,q^{(\w)}_2)}\le\tau] \le O((\tau/\norm{R(q^{(\w)}_1,q^{(\w)}_2)})^{1/4}).
\end{align*}
By choosing $\tau = O(\eta_c\log(1/\alpha))$, we obtain that $\norm{R(q^{(\w)}_1,q^{(\w)}_2)} \le O(\eta_c\log(1/\alpha)/\alpha^4)$.

\paragraph{Case 2: $\deg(q^{(\w)}_1) = 1$}

In the rest of the proof, we work on the case where $\deg(q_1^{(\w)}) = 1$. In this case, we have $q^{(\w)}_1(z) = b_1(\w)z + c_1(\w)$ and
\begin{align*}
    R(q^{(\w)}_1,q^{(\w)}_2) = b_1(\w)^{\deg (q^{(\w)}_2)}q_2^{(\w)}(r),
\end{align*}
where $r$ is the root of $q^{(\w)}_1(z)$. This gives for each $r \in Z^{(\w)}$ and for every $\w$ such that $\abs{b_1} \le T$,
\begin{align*}
    \abs{R(q^{(\w)}_1,q^{(\w)}_2)} \le \eta_c \max\{1,\abs{b_1(\w)}^2\} \le O(\eta_c\log (1/\alpha))
\end{align*}

Now since $\E_{\w \sim \cN(0,\bI)}[\sum_{s \in C_{12}(\w)}\phi(s)] \ge \alpha$, we know that $\E_{\w \sim \cN(0,\bI)}[\sum_{s \in Z(\w)}\phi(s)] \ge 7\alpha/8$. Since $\phi(s) \le 1/\sqrt{2\pi}$, this implies that
\begin{align*}
  \Pr_{\w\sim \cN(0,\bI)}[\abs{R(q^{(\w)}_1,q^{(\w)}_2) } \le O(\eta_c\log(1/\alpha))] \ge   \Pr_{\w\sim \cN(0,\bI)}[\abs{b_1(\w)}\le T,\abs{b_2(\w)}\le T,\card{Z}^{(\w)}\ge 1] \ge \alpha/2.
\end{align*}
Since $R(q^{(\w)}_i,q^{(\w)}_j)$ is a polynomial with degree at most $4$ by \Cref{fact resultant}, using Carbery–Wright anti-concentration inequality, we have for every $\tau>0$, we have
\begin{align*}
\Pr_{\w\sim \cN(0,\bI)}[\abs{R(q^{(\w)}_1,q^{(\w)}_2)}\le\tau] \le O((\tau/\norm{R(q^{(\w)}_1,q^{(\w)}_2)})^{1/4}).
\end{align*}
By choosing $\tau = O(\eta_c\log(1/\alpha))$, we obtain that $\norm{R(q^{(\w)}_1,q^{(\w)}_2)} \le O(\eta_c\log(1/\alpha)/\alpha^4)$.
\end{proof}

\paragraph{Robust Factorization Lemma}
With \Cref{lm close root to small resultant}, we next present the main technical lemma, \Cref{lm small resultant to factorization}. We show that under the condition of \Cref{lm close root to small resultant}, there are only three cases for a pair of degree 2 polynomials. Before giving \Cref{lm small resultant to factorization}, we first provide some intuition about what algebraic structure we can obtain from small resultants.

Our intuition starts from the Euclidean algorithm for finding the greatest common divisor for a pair of quadratics.
Consider two univariate quadratic polynomials $q_i = a_i z^2 + b_i z + c_i$, $i \in [2]$. The first step of Euclidean algorithm is to eliminate the leading quadratic term, by considering
\begin{align*}
    a_2q_1 - a_1q_2 = Uz + V,
\end{align*}
where in \Cref{fact resultant}, we define $U = a_2b_1 - a_1b_2$ and $V = a_2c_1 - a_1c_2$.
If $U \neq 0$ and $q_1,q_2$ exactly share a root, then the root $r$ must be $-V/U$, which provides a way to factorizing the two polynomials into the form of $q_i = L \ell_i$.
The resultant of the two quadratics provides a quantitative version of this observation: it satisfies $R(q_1,q_2)=U^2q_1(r)/a_1$. Consequently, whenever the resultant is small and $U$ is bounded away from zero, both polynomials take small values at $r$. By polynomial division, each $q_i$ can then be approximated by a product $(z-r)\ell_i(z)$, with an approximation error controlled by the resultant. On the other hand, if $U$ is nearly zero, then small resultant implies that $V$ is also nearly zero, which implies that $q_1$ and $q_2$ are nearly in parallel.
Thus, a small resultant suggests either an approximate common affine factor or approximate proportionality of the two polynomials.

However, in our case, the coefficients of $q^{(\w)}_i$ are polynomials on $\w$. If we directly carry the analysis of Euclidean algorithm, then the formal candidate of a common root is $r = - V(\w)/U(\w)$, which is a rational function in $\x$, instead of a low degree polynomial. Furthermore, since $U(\w)$ could be very small for some $\w \perp \v$, the instability from these slices would amplify the error in the approximate setting.
Furthermore, there is also a genuine algebraic obstruction to demanding a common affine factor. For example, consider a simple example in $\R^2$, where $q_1^{(w)}(z) = q_2^{(w)}(z) = zw - 1$. Although the two function shares an exact root $1/w$, for every $w \neq 0$, they cannot be written as a product of two affine polynomials in $(z,w)$. In fact, even if we consider a case of $q_1^{(w)}(z) = q_1^{(w)}(z) = z^2 - w^2 -1 = (z - \sqrt{w^2+1})(z + \sqrt{w^2+1})$, where an exact factorization is possible, the factorization might be an irrational function in $w$.

We therefore seek a structural conclusion with several alternatives rather than an unconditional common-affine-factor theorem. Under the normalization and boundary-mass assumptions of \Cref{lm close root to small resultant}, the goal is to show that either one defining polynomial is approximately independent of $\v$ direction, the two defining polynomials are approximately proportional, or both can be approximated by products of affine polynomials sharing a common affine factor.

\begin{lemma}\label{lm small resultant to factorization}
     For $i \in [2]$, let $p_i(\x)$ be a degree-$2$ polynomial with $\E_{\x \sim \cN(0,\bI)}[p_i(\x)]= 0$ and $\var_{\x \sim \cN(0,\bI)}[p_i(\x)]=1$. Let $t_i \in \R$ and $q_i(\x): = p_i(\x) - t_i$.
     Let $f(\x) = \Ind\{q_1(\x) \ge 0, q_2(\x) \ge 0 \}$. Let $\eta_c \in (0,1)$ and $\alpha \in (0,1)$.
     For $\w \perp \v$, denote by $F(\w) \subseteq \R$, the set of jump points $s$ of $f_\w$, such that $q^{(\w)}_1(s) = 0$,  and $0\le q^{(\w)}_2(s) \le \eta_c$.
     Let $\alpha \in (0,1)$. Suppose that $\E_{\w \sim \cN(0,\bI)}[\sum_{s \in F(\w)}\phi(s)] \ge \alpha$ and $\abs{t_i} \le O(\log(1/\alpha))$. Denote by $\eps_{fac} = \Tilde{O}( (\eta_c/\alpha^4)^{1/2000}\alpha^{-1})$. Then the one of the follows holds.
     \begin{enumerate}
         \item for some $i \in [2]$, there exists a degree-2 polynomial $\Tilde{q}_i(\w)$ such that $\norm{q_i -\Tilde{q}_i(\w)} \le \eps_{fac}$.

         \item there exists some $\lambda \in \R$ such that $\norm{q_i-\lambda q_{3-i}} \le \eps_{fac}$, for some $i \in [2]$,
         \item  there exist polynomials $L,\ell_1,\ell_2$ with degree at most 1 such that $\norm{q_i(\x)-L(\x)\ell_i(\x)} \le \eps_{fac}$ for both $i \in [2]$.
     \end{enumerate}
\end{lemma}

Before giving the proof, we first provide the intuition of the idea before \Cref{lm small resultant to factorization}.
Recall that \Cref{lm close root to small resultant} shows that the resultant of the two defining polynomials has a small Gaussian $L_2$ norm. Our goal is to use this information to obtain an approximate common affine factor, unless the polynomials fall into one of the degenerate configurations described in the lemma. Following the intuition from the Euclidean algorithm, we consider two cases, depending on whether at least one polynomial has a sufficiently large quadratic coefficient along the direction $\v$.

First, suppose that $\abs{a_1}$ is bounded below by some parameter \(\kappa>0\). We eliminate the quadratic terms and write
\begin{align*}
a_2q_1-a_1q_2=U(\w)z+V(\w),
\end{align*}
where $U = a_2b_1(\w) - a_1b_2(\w)$ is affine and $V = a_2c_1(\w) - a_1c_2(\w)$ is quadratic in $\w$. If this remainder is sufficiently small, then the two defining polynomials are approximately proportional, giving the second alternative of the lemma.
Otherwise, we show that $U(\w)$ must have a nontrivial $L_2$ norm. Indeed, the resultant can be expressed in terms of $U$ and $V$, and if $\norm{U}$ is very small, the smallness of the resultant would force $V$ to be small as well, contradicting our assumption that the remainder is nontrivial. This lower bound on $\norm{U}$ is important because it allows us to make the polynomial division suggested by the Euclidean algorithm quantitatively stable.

The main technical step is then to construct an affine polynomial $h(\w)$ such that $V(\w)\approx U(\w)h(\w)$, without directly dividing by $U(\w)$. We distinguish two cases according to the size of the linear part of $U$. If it is sufficiently large, we perform polynomial division along the corresponding direction and obtain a remainder that is independent of that direction. By restricting the resultant to the hyperplane $U(\w) = 0$, we can control this remainder using the small $L_2$ norm of the resultant and the bounded degree of the polynomials. On the other hand, if the linear part of $U$ is small, then its constant part must be bounded away from zero. In this case, we use the degree-four Hermite component of the resultant to show that the quadratic part of $V$ must be small, allowing us to approximate $V/U$ by an affine polynomial. In either case, we obtain an affine $h$ satisfying $\norm{V-Uh}$ being small.
 We then define $L = z + h(\w)$ and use the resultant identity, together with a lower bound on the norm of products of low-degree Gaussian polynomials, to show that both $q_1$ and $q_2$ are close to products of $L$ with affine polynomials. This gives the desired approximate common factor. The case where $\abs{a_2}$ is bounded below follows symmetrically.

It remains to consider the case where both $\abs{a_1}$ and $\abs{a_2}$ are small. In this regime, the two defining polynomials are approximately linear in $\v$ direction, so we instead work with $\bar q_i(z,\w)=b_i(\w)z+c_i(\w)$. If both $b_i$ have small norm, the polynomials are approximately independent of $\v$, giving the first alternative. Otherwise, we introduce the corresponding elimination polynomial $T(\w)=b_2(\w)c_1(\w)-b_1(\w)c_2(\w)$. Using the original assumption that the two polynomials have nearly shared roots on a nontrivial fraction of Gaussian slices, we show that $T$ has a small $L_2$ norm. We then distinguish whether $b_1$ and $b_2$ are approximately proportional. If they are, the smallness of $T$ implies that the defining polynomials are approximately proportional as well. Otherwise, an argument analogous to the previous case shows that $c_1$ and $c_2$ are approximately divisible by $b_1$ and $b_2$, respectively, with a common affine quotient. This again yields an approximate common affine factor. Combining these cases establishes the desired structural alternatives.

Now, we are able to give the proof of \Cref{lm small resultant to factorization}. We first give the following polynomial division lemma that will be used several times in our proof.

\begin{lemma}[Polynomial division under a Gaussian distribution]
\label{lm polynomial division}
Let $\x \sim \cN(0,\bI)$ and $\v \in \s^{d-1}$. Write $z = \x \cdot \v$ and $\w = \proj_{\v^\perp}(\x)$.
Consider an affine function $L(z\v+\w)=\lambda z+B(\w),\lambda\ne0,$ where $B$ is an affine function on $\w$. Let $r_L(\w)=-B(\w)/\lambda$ be its root in $z$. Let $q$ be a polynomial of degree at most two, written as $q(z\v+\w)=az^2+b(\w)z+c(\w),$ $\deg b\le1,\quad\deg c\le2$.
The following hold.
\begin{enumerate}
    \item There are unique polynomials $\ell(\x)$ and $m(\w)$, with
$\deg\ell\le1$ and $\deg m\le2$, such that $q(\x)=L(\x)\ell(\x)+m(\w),$ given by
\begin{align*}
    &\ell(z\v+\w)
       =\frac a\lambda z+\frac{b(\w)}\lambda
                           -\frac{aB(\w)}{\lambda^2},\\
    &m(\w)
       =c(\w)-\frac{B(\w)b(\w)}\lambda
                    +\frac{aB(\w)^2}{\lambda^2}
         =q(r_L(\w)\v+\w).
\end{align*}
If $q$ is affine, $\ell$ is constant.
\item For a universal constant $C$,
\begin{equation}
    \|\ell\|_2\le
      \frac C{|\lambda|}
       \left(1+\frac{\|B\|_2}{|\lambda|}\right)\|q\|_2.
    \label{eq affine division quotient norm}
\end{equation}
For every polynomial $P$ of degree at most $l\le4$,
\begin{equation}
    \bigl\|P(r_L(\w)\v+\w)\bigr\|_{L^2(\w)}
      \le C_l
        \left(1+\frac{\|B\|_2}{|\lambda|}\right)^l
        \|P\|_{L^2(\x)},
    \label{eq affine hyperplane restriction}
\end{equation}
where $C_l$ depends only on $l$.
\item Let $E\subseteq\v^\perp$ have Gaussian probability at least
$\beta\in(0,1/2]$. Suppose that $s:E\to\mathbb R$ is measurable,
$|s(\w)|\le T$, and, for every $\w\in E$, $|q(s(\w)\v+\w)|\le\varepsilon$ and $|L(s(\w)\v+\w)|\le\eta$, then
\begin{equation}
    \|m\|_2\le C\beta^{-2}
       \left(\varepsilon+
        \eta\bigl(T+\sqrt{\log(2/\beta)}\bigr)\|\ell\|_2\right).
    \label{eq affine division small values}
\end{equation}
In particular, when $L(s(\w)\v+\w)=0$ on $E$,
\begin{equation}
    \|q-L\ell\|_2=\|m\|_2\le C\beta^{-2}\varepsilon.
    \label{eq affine division exact roots}
\end{equation}

\end{enumerate}

\end{lemma}

\begin{proof}[Proof of \Cref{lm polynomial division}]
The first item can be proved by directly checking the definition of $\ell$ and $m$.
For uniqueness, regard the
polynomials as polynomials in $z$ with coefficients depending on $\w$.
A nonzero multiple of $L$, whose leading coefficient in $z$ is the
nonzero constant $\lambda$, cannot be independent of $z$.

To prove \eqref{eq affine division quotient norm}, write $    q=a(z^2-1)+b(\w)z+(c(\w)+a).$
Gaussian orthogonality gives
    $2a^2+\|b\|_2^2+\|c+a\|_2^2=\|q\|_2^2$.
The bound follows from the division step and the
triangle inequality.

For \eqref{eq affine hyperplane restriction}, expand in normalized
Hermite polynomials along $\v$:
\[
    P(z\v+\w)=\sum_{j=0}^l h_j(z)P_j(\w),\qquad
    \deg P_j\le l-j,\qquad
    \sum_{j=0}^l\|P_j\|_2^2=\|P\|_2^2.
\]
The function $r_L$ is affine. Gaussian moment bounds imply
\[
    \|h_j(r_L(\w))\|_4
       \le C_l(1+\|r_L\|_2)^j,
\]
and hypercontractivity gives $\|P_j\|_4\le C_l\|P_j\|_2$.
H\"older's inequality and the triangle inequality therefore give
\begin{align*}
    \bigl\|P(r_L(\w)\v+\w)\bigr\|_2
\le\sum_{j=0}^l\|h_j(r_L)\|_4\|P_j\|_4
    \le C_l(1+\|r_L\|_2)^l\|P\|_2.
\end{align*}
Substitute $\|r_L\|_2=\|B\|_2/|\lambda|$.

Finally, write $\ell(z\v+\w)=\gamma z+D(\w)$, where $D$ is
affine. Then $|\gamma|\le\|\ell\|_2$ and
$\|D\|_2\le\|\ell\|_2$. Gaussian tails imply that, except on a
set of probability at most $\beta/2$,
$
    |D(\w)|\le C\|\ell\|_2\sqrt{\log(2/\beta)}.
$
On a subset of $E$ of probability at least $\beta/2$ we consequently
have
$
    |\ell(s(\w)\v+\w)|
       \le C\bigl(T+\sqrt{\log(2/\beta)}\bigr)\|\ell\|_2.
$
Evaluate the polynomial division at the original selected
point $s(\w)\v+\w$. It gives
\[
    |m(\w)|
      \le\varepsilon+
         C\eta\bigl(T+\sqrt{\log(2/\beta)}\bigr)\|\ell\|_2
\]
on that set. Degree-two Gaussian anti-concentration says that
$\Pr_{\w \sim \cN(0,\bI)}[|m(\w)|\le h]\le C(h/\|m\|_2)^{1/2}$ for
$m\ne0$. The probability lower bound $\beta/2$ proves
\eqref{eq affine division small values}. The case $m=0$ is immediate,
and setting $\eta=0$ gives \eqref{eq affine division exact roots}.
\end{proof}

Now, we are able to prove \Cref{lm small resultant to factorization}

\begin{proof}[Proof of \Cref{lm small resultant to factorization}]
Recall that, for $i\in[2]$,
$
    q_i^{(\w)}(z)=a_i z^2+b_i(\w)z+c_i(\w).
$
Write $B=1+C\log(1/\alpha)$, with $C$ sufficiently large. The
normalization in the statement gives
$
    |a_i|\le1, \|b_i\|_2\le1, \|c_i\|_2\le B.
$
Let $\kappa,\xi,\rho,\tau'\in(0,1)$ be cutoffs,
to be specified at the end, with $\tau'\le\sqrt\rho/4$.

\paragraph{Case 1: $|a_1|\ge\kappa$.}
By \eqref{eq euclidean},
$
    a_2q_1^{(\w)}(z)-a_1q_2^{(\w)}(z)=U(\w)z+V(\w),
$
where
$
    U=a_2b_1-a_1b_2, V=a_2c_1-a_1c_2.
$
In particular, $\|U\|_2\le2$ and $\|V\|_2\le2B$.

\paragraph{Case 1.1: $\|a_2q_1-a_1q_2\|_2\le\xi$.}
Then
$
    \|q_2-(a_2/a_1)q_1\|_2\le\xi/\kappa,
$
which gives the approximately proportional alternative.

\paragraph{Case 1.2: $\|a_2q_1-a_1q_2\|_2>\xi$.}
Suppose that $\|R(q_1^{(\w)},q_2^{(\w)})\|_2\le\eta$, where
$\eta\le\xi^2/10$. We will choose such an $\eta$ using
\Cref{lm close root to small resultant} at the end of the proof.
Define the auxiliary polynomial
\begin{equation}
    \mathcal R
    :=\frac{U^2c_1}{a_1}-\frac{UVb_1}{a_1}+V^2.
    \label{eq cal R}
\end{equation}
By \eqref{eq resultant representation},
$
    \mathcal R=a_1^{2-d_2}R(q_1^{(\w)},q_2^{(\w)}),
$
where $d_2\in\{0,1,2\}$ is the degree of $q_2^{(\w)}$ in $z$
for almost every $\w$. Hence $\|\mathcal R\|_2\le\eta$.

\begin{claim}\label{cl U lb}
Under the assumptions of Case~1.2,
$
    \|U\|_2\ge c\kappa\xi^2/B
$
for a universal constant $c>0$.
\end{claim}
\begin{proof}[Proof of \Cref{cl U lb}]
By \eqref{eq cal R} and \Cref{fct:gaussian-product-norm},
\[
    \|\mathcal R-V^2\|_2
    \le \frac{CB}{\kappa}\bigl(\|U\|_2^2+\|U\|_2\bigr).
\]
Since $\|V\|_2^2\le\|V^2\|_2$ and
$\|a_2q_1-a_1q_2\|_2^2=\|U\|_2^2+\|V\|_2^2$,
\[
    \xi^2
    \le\|U\|_2^2+\eta+
       \frac{CB}{\kappa}\bigl(\|U\|_2^2+\|U\|_2\bigr)
    \le\eta+\frac{C'B}{\kappa}\|U\|_2.
\]
The last inequality uses $\|U\|_2\le2$ and $B/\kappa\ge1$.
The assumption $\eta\le\xi^2/10$ proves the claim.
\end{proof}

We next show that $V$ is approximately divisible by $U$, with an
affine quotient.
\begin{claim}\label{cl V factorizable}
Under the assumptions of Case~1.2, there is an affine polynomial
$h(\w)$ such that
\[
    \|V-Uh\|_2
    \le C\frac{B^{4/3}\eta^{1/6}}
                    {\kappa^{2/3}\xi^{4/3}},
    \qquad
    \|h\|_2\le C\frac{B^2}{\kappa\xi^2}.
\]
\end{claim}
\begin{proof}[Proof of \Cref{cl V factorizable}]
Put
$
    \delta_*:=(B^{4/3}\eta^{1/6})/
                         (\kappa^{2/3}\xi^{4/3}).
$
If $\delta_*\ge c_0B$, where $c_0>0$ is a sufficiently small
universal constant, take $h=0$. Since $\|V\|_2\le2B$, the
claim follows after increasing $C$. We therefore assume
$\delta_*<c_0B$ and choose
\[
    \tau=\left(\frac{\kappa\xi^2\sqrt\eta}{B^2}\right)^{1/3}
         =\frac{\kappa\xi^2}{B^2}\delta_*.
\]
By \Cref{cl U lb} and the choice of $c_0$,
$\tau\le\|U\|_2/4$.

If $U$ is nonconstant, write $\w=t\u+\w_\perp$ and
$U(\w)=\beta_1t+\beta_0$, where
$\u\in\v^\perp$ is a unit vector, $\w_\perp\perp\u$, and
$\beta_1>0$. If $U$ is constant, set $\beta_1=0$,
$\beta_0=U$, and omit the terms involving $t$ in the second
subcase below.

Suppose first that $\beta_1\ge\tau$ and put
$s=-\beta_0/\beta_1$. Apply
\Cref{lm polynomial division}(1) in the Gaussian space
$\v^\perp$, with distinguished direction $\u$, to divide $V$ by
$U$. It gives an affine polynomial $h(\w)$ and a remainder
$E(\w_\perp)$ such that
\[
    V=Uh+E,\qquad E(\w_\perp)=V(s\u+\w_\perp).
\]
On the hyperplane $U=0$, \eqref{eq cal R} gives
$
    \mathcal R(s\u+\w_\perp)=E(\w_\perp)^2.
$
Since $\deg\mathcal R\le4$, the restriction bound in
\Cref{lm polynomial division}(2) implies
\[
    \|E\|_2^2\le\|E^2\|_2
    \le C(1+|s|)^4\|\mathcal R\|_2
    \le C\eta/\tau^4.
\]
Here $|\beta_0|\le\|U\|_2\le2$. Thus
$\|V-Uh\|_2\le C\sqrt\eta/\tau^2=C\delta_*$.

Suppose next that $\beta_1<\tau$. Since
$\tau\le\|U\|_2/4$, we have
    $|\beta_0|\ge\|U\|_2/2\ge c\kappa\xi^2/B$.
Take $h=V_{\le1}/\beta_0$. By the upper bound in \Cref{fct:gaussian-product-norm},
\begin{equation}
    \|V-Uh\|_2
    \le\|V_2\|_2+C\frac{B^2\tau}{\kappa\xi^2}.
    \label{eq factorization nearly constant U}
\end{equation}
To bound $\|V_2\|_2$, note that
$
    \|(V_2^2)_4\|_2\ge\sqrt2\|V_2\|_2^2.
$
Indeed, after diagonalization write
$V_2=\sum_i\gamma_i h_2(w_i)$. Gaussian orthogonality gives
\[
    \|(V_2^2)_4\|_2^2
    =6\sum_i\gamma_i^4+4\sum_{i<j}\gamma_i^2\gamma_j^2
    \ge2\left(\sum_i\gamma_i^2\right)^2.
\]
Taking degree-four Hermite components in \eqref{eq cal R},
\[
    \mathcal R_4
    =(V_2^2)_4+
      \frac{\beta_1^2}{a_1}\bigl(t^2(c_1)_2\bigr)_4
      -\frac{\beta_1}{a_1}\bigl(tV_2(b_1)_1\bigr)_4.
\]
Consequently, by \Cref{fct:gaussian-product-norm},
\[
    \sqrt2\|V_2\|_2^2
    \le\eta+C\frac{B\tau^2}{\kappa}
             +C\frac{\tau}{\kappa}\|V_2\|_2.
\]
Solving this quadratic inequality yields
\[
    \|V_2\|_2
    \le C\left(\sqrt\eta+
                  \tau\sqrt{B/\kappa}+\tau/\kappa\right).
\]
Combining this with \eqref{eq factorization nearly constant U},
using $B\ge1$ and $\kappa,\xi\le1$, gives
\[
    \|V-Uh\|_2
    \le C\left(\sqrt\eta+
                   \frac{B^2\tau}{\kappa\xi^2}\right)
    \le C\delta_*.
\]
Finally, in either nontrivial subcase,
$\|Uh\|_2\le\|V\|_2+\|V-Uh\|_2\le CB$.
The lower bound in \Cref{fct:gaussian-product-norm} and \Cref{cl U lb} give
\[
    \|h\|_2\le CB/\|U\|_2\le CB^2/(\kappa\xi^2).
\]
\end{proof}

Given $h$ from \Cref{cl V factorizable}, set $L(\x)=z+h(\w)$.
Apply \Cref{lm polynomial division}(1) to each $q_i$, now along
the original direction $\v$. It gives
\[
    q_i=L\ell_i+E_i(\w),\qquad
    \ell_i=a_i z+b_i-a_i h,\qquad
    E_i=c_i-b_i h+a_i h^2.
\]
Put $E=V-Uh$. Substitution into \eqref{eq cal R} gives
$
    \mathcal R
    =E^2+UE(2h-b_1/a_1)+U^2E_1/a_1,
$
and therefore
\begin{equation}
    U^2E_1=a_1\mathcal R-a_1E^2-UE(2a_1h-b_1).
    \label{eq factorization E1 identity}
\end{equation}
\Cref{fct:gaussian-product-norm} imply
\begin{equation}
    \|E_1\|_2
    \le\frac{C}{\|U\|_2^2}
       \left(\eta+\|E\|_2^2+
       \|U\|_2\|E\|_2(1+\|h\|_2)\right).
    \label{eq factorization E1 bound}
\end{equation}
Moreover, $a_2E_1-a_1E_2=V-Uh=E$, so
\begin{equation}
    \|E_2\|_2\le\kappa^{-1}(\|E\|_2+\|E_1\|_2).
    \label{eq factorization E2 bound}
\end{equation}
These inequalities give the required errors in the common-affine
alternative once the parameters are chosen below.

\paragraph{Case 2: $|a_1|<\kappa$ and $|a_2|\ge\kappa$.}
Interchange the two polynomials and apply Case~1.
\Cref{fact resultant order} preserves the absolute value, and hence
the norm, of the resultant. No other change is needed.

\paragraph{Case 3: $|a_1|<\kappa$ and $|a_2|<\kappa$.}
Put $\bar q_i=b_i(\w)z+c_i(\w)$. Then
    $\|q_i-\bar q_i\|_2=\sqrt3|a_i|\le\sqrt3\kappa$.
We first bound the elimination polynomial
$
    T(\w)=b_2(\w)c_1(\w)-b_1(\w)c_2(\w).
$
Let $F^{(\w)}$ be the jump points in the hypothesis, at which
$q_1^{(\w)}(s)=0$ and $0\le q_2^{(\w)}(s)\le\eta_c$.
Their weighted mass is at least $\alpha$. Choose
$R_\alpha=C\sqrt B$ with $C$ sufficiently large. Discard the
roots with $|s|>R_\alpha$ and the layers where
$|b_1(\w)|>R_\alpha$ or $|b_2(\w)|>R_\alpha$.
There are at most two roots of $q_1^{(\w)}$ on almost every
layer. Gaussian tails for $b_1,b_2$ and
$\phi(R_\alpha)\le\alpha/100$ therefore show that the remaining
weighted mass is at least $\alpha/2$.
In particular, the remaining layers have probability at least
$c\alpha$.

At a retained root $s$, the identity
$
    b_2q_1-b_1q_2=(a_1b_2-a_2b_1)z^2+T
$
gives
\[
    T(\w)=(a_2b_1(\w)-a_1b_2(\w))s^2
                         -b_1(\w)q_2^{(\w)}(s).
\]
Consequently, on a set of probability at least $c\alpha$,
$
    |T(\w)|\le C(\kappa B^{3/2}+\eta_c B^{1/2}).
$
Since $\deg T\le3$, Gaussian anti-concentration gives
\begin{equation}
    \|T\|_2\le
       C\alpha^{-3}(\kappa B^{3/2}+\eta_c B^{1/2}).
    \label{eq factorization T bound}
\end{equation}
The conclusion is immediate if $T=0$; otherwise apply
$\Pr_{\w}[|T(\w)|\le h]\le C(h/\|T\|_2)^{1/3}$
with the displayed small-value threshold.

Without loss of generality, assume $\|b_1\|_2\ge\|b_2\|_2$.
If $\|b_1\|_2\le\sqrt\rho$, then
$
    \|q_i-c_i(\w)\|_2\le\sqrt3\kappa+\sqrt\rho
$
for both $i$, giving the first alternative. Henceforth assume
$\|b_1\|_2>\sqrt\rho$. Define
\[
    \lambda=\frac{\E_{\w\sim\cN(0,\bI_{\v^\perp})}[b_1b_2]}
                       {\|b_1\|_2^2},
    \qquad b_2=\lambda b_1+r.
\]
Then $|\lambda|\le1$ and
$\E_{\w\sim\cN(0,\bI_{\v^\perp})}[b_1r]=0$.
If $\|r\|_2\le\rho$, the identity
$
    T=b_1(\lambda c_1-c_2)+rc_1
$
and \Cref{fct:gaussian-product-norm} gives
$
    \|\lambda c_1-c_2\|_2
       \le C(\|T\|_2+B\rho)/\sqrt\rho.
$
Thus
$
    \|q_2-\lambda q_1\|_2
       \le C\left(\kappa+B\sqrt\rho+
                         \|T\|_2/\sqrt\rho)\right),
$
which gives the second alternative.

It remains to consider $\|r\|_2>\rho$. We construct an affine
polynomial $h$ for which both $E_i:=c_i-b_i h$ are small.
For nonconstant $b_1$, write $\w=t\u+\w_\perp$ and
$b_1(\w)=\beta_1t+\beta_0$, where $\beta_1>0$.
If $b_1$ is constant, take $\beta_1=0$, $\beta_0=b_1$, and
omit the terms involving $t$ in the second subcase.

Suppose first that $\beta_1\ge\tau'$ and put
$s=-\beta_0/\beta_1$. Apply \Cref{lm polynomial division}(1)
in the Gaussian space $\v^\perp$, with distinguished direction
$\u$, to divide $c_1$ by $b_1$. We obtain an affine $h(\w)$ and
$E_1(\w_\perp)$ such that
$
    c_1=b_1h+E_1.
$
Restricting $T=b_2c_1-b_1c_2$ to $b_1=0$ gives
$
    T(s\u+\w_\perp)
       =r(s\u+\w_\perp)E_1(\w_\perp).
$
The degree-three restriction bound in
\Cref{lm polynomial division}(2) implies
\[
    \|r(s\u+\w_\perp)E_1(\w_\perp)\|_2
       \le C(1+|s|)^3\|T\|_2
       \le C(\tau')^{-3}\|T\|_2.
\]
Here $|\beta_0|\le\|b_1\|_2\le1$. To bound the restricted
$r$ from below, write
\[
    r(t\u+\w_\perp)=r_0+\gamma t+r_\perp(\w_\perp),
    \qquad \E_{\w_\perp}[r_\perp]=0.
\]
The orthogonality $\E_{\w}[b_1r]=0$ gives $\gamma=sr_0$.
Therefore,
\[
    \|r(s\u+\w_\perp)\|_2^2
       =(1+s^2)^2r_0^2+\|r_\perp\|_2^2
       \ge\|r\|_2^2>\rho^2.
\]
The lower bound in \Cref{fct:gaussian-product-norm} now yields
\begin{equation}
    \|E_1\|_2\le C\frac{\|T\|_2}{\rho(\tau')^3}.
    \label{eq factorization case3 large linear}
\end{equation}

Suppose next that $\beta_1<\tau'$. Since
$\|b_1\|_2>\sqrt\rho$ and $\tau'\le\sqrt\rho/4$,
$|\beta_0|\ge\sqrt\rho/2$. Take
$
    h=(c_1)_{\le1}/\beta_0,
     \|h\|_2\le2B/\sqrt\rho.
$
By the upper bound in \Cref{fct:gaussian-product-norm},
\begin{equation}
    \|E_1\|_2
       \le\|(c_1)_2\|_2+CB\tau'/\sqrt\rho.
    \label{eq factorization case3 small linear preliminary}
\end{equation}
Write $r=r_0+r_1$, where $r_1$ is its degree-one Hermite
component. Orthogonality gives
\[
    |r_0|\,|\beta_0|
       =|\beta_1\E_{\w}[t r_1(\w)]|
       \le\beta_1\|r_1\|_2.
\]
It follows that $|r_0|\le\|r_1\|_2/2$, and hence
$\|r_1\|_2\ge\|r\|_2/2>\rho/2$.
Taking degree-three Hermite components in
$T=b_1(\lambda c_1-c_2)+rc_1$ gives
$
    T_3=(r_1(c_1)_2)_3+
       \beta_1\bigl(t(\lambda(c_1)_2-(c_2)_2)\bigr)_3.
$
For a centered affine polynomial $P$ and a pure degree-two
Hermite polynomial $Q$, we have
$\|(PQ)_3\|_2\ge\|P\|_2\|Q\|_2$.
Indeed, rotate so that $P=\|P\|_2y$, and write
$Q=A h_2(y)+yD(\mathbf r)+C(\mathbf r)$ in orthogonal
second-chaos components. Then
\[
    \|(yQ)_3\|_2^2
       =3A^2+2\|D\|_2^2+\|C\|_2^2
       \ge\|Q\|_2^2.
\]
Applying this inequality and the upper bound in \Cref{fct:gaussian-product-norm} to
$T_3$, using $|\lambda|\le1$ and $\|c_i\|_2\le B$, gives
$
    \|(c_1)_2\|_2
       \le C(\|T\|_2+B\tau')/\rho.
$
Together with
\eqref{eq factorization case3 small linear preliminary}, this gives
\begin{equation}
    \|E_1\|_2\le C\frac{\|T\|_2+B\tau'}{\rho}.
    \label{eq factorization case3 small linear}
\end{equation}

For either choice of $h$, put $E_2=c_2-b_2h$.
Since $T=b_2E_1-b_1E_2$, \Cref{fct:gaussian-product-norm} gives
\[
    \|E_2\|_2\le
       \frac{C}{\|b_1\|_2}(\|T\|_2+\|b_2\|_2\|E_1\|_2)
       \le\frac{C}{\sqrt\rho}(\|T\|_2+\|E_1\|_2).
\]
Take $L(z\v+\w)=z+h(\w)$ and $\ell_i(\w)=b_i(\w)$.
Then
\[
    q_i-L\ell_i=a_i z^2+E_i(\w),
    \qquad
    \|q_i-L\ell_i\|_2\le\sqrt3\kappa+\|E_i\|_2.
\]
Equations~\eqref{eq factorization case3 large linear} and
\eqref{eq factorization case3 small linear} bound the errors in
the two subcases. Including the earlier independent and proportional
alternatives, one of the three conclusions holds with error at most
\begin{equation}
    C\left(\kappa+B\sqrt\rho+
       \frac{\|T\|_2}{\rho^{3/2}(\tau')^3}
       +\frac{B\tau'}{\rho^{3/2}}\right).
    \label{eq factorization case3 bound}
\end{equation}

\paragraph{Choice of parameters and conclusion.}
Let $D:=\eta_c/\alpha^4$. If $D\ge1$ or
$D^{1/2000}\ge\alpha$, the claimed error bound is at least an
absolute constant, up to its implicit constant. The first
alternative then follows by taking $\widetilde q_i(\w)=c_i(\w)$,
since $\|q_i-c_i\|_2\le\sqrt3|a_i|+\|b_i\|_2=O(1)$.
We henceforth assume $D<1$ and $D^{1/2000}<\alpha$. In particular,
\[
    \alpha^{-3}<D^{-3/2000},
    \qquad B\le C(1+\log(1/D)).
\]
By increasing the absolute constant in the asserted error bound,
we may also assume that $D$ is smaller than a sufficiently small
universal constant; the remaining values are covered by the same
constant-error approximation.

Choose
\[
    \kappa=D^{35/2000},\qquad
    \xi=D^{37/2000},\qquad
    \rho=D^{4/2000},\qquad
    \tau'=\tfrac18D^{8/2000}.
\]
These choices ensure $\tau'\le\sqrt\rho/4$.
Here and below, $\widetilde O$ and $\widetilde\Omega$ hide only
fixed powers of $B$ and absolute constants.
By \eqref{eq factorization T bound},
\[
    \|T\|_2
       \le C\alpha^{-3}(\kappa B^{3/2}+\eta_c B^{1/2})
       \le\widetilde O(D^{32/2000}).
\]
Substituting into \eqref{eq factorization case3 bound}, we have
\[
    B\sqrt\rho=\widetilde O(D^{2/2000}),\qquad
    \frac{\|T\|_2}{\rho^{3/2}(\tau')^3}
       =\widetilde O(D^{2/2000}),\qquad
    \frac{B\tau'}{\rho^{3/2}}
       =\widetilde O(D^{2/2000}).
\]
Thus Case~3 gives one of the required alternatives with error
$\widetilde O(D^{2/2000})$.

For Cases~1 and~2, take $\eta=C_0BD$, with $C_0$ sufficiently
large that \Cref{lm close root to small resultant} gives
$\|R(q_1^{(\w)},q_2^{(\w)})\|_2\le\eta$.
The smallness assumption on $D$ ensures $\eta\le\xi^2/10$.
The two claims give
\[
    \|U\|_2\ge\widetilde\Omega(D^{109/2000}),\qquad
    \|E\|_2\le
       C\frac{B^{4/3}\eta^{1/6}}{\kappa^{2/3}\xi^{4/3}}
       \le\widetilde O(D^{260/2000}),\qquad
    \|h\|_2\le\widetilde O(D^{-109/2000}).
\]
Using \eqref{eq factorization E1 bound},
\begin{align*}
    \|E_1\|_2
    &\le C\left(
        \frac{\eta}{\|U\|_2^2}
       +\frac{\|E\|_2^2}{\|U\|_2^2}
       +\frac{\|E\|_2(1+\|h\|_2)}{\|U\|_2}\right)\\
    &\le\widetilde O\left(
       D^{1782/2000}+D^{302/2000}
       +D^{151/2000}+D^{42/2000}\right)
     \le\widetilde O(D^{42/2000}).
\end{align*}
Equation~\eqref{eq factorization E2 bound} then gives
    $\|E_2\|_2\le\widetilde O(D^{7/2000})$.
The approximately proportional branch has error
$\xi/\kappa=D^{2/2000}$. Case~2 is symmetric.
Consequently, in every case one of the three alternatives holds
with error $\widetilde O(D^{2/2000})$. This is at most
\[
    \widetilde O\left(\alpha^{-1}D^{1/2000}\right)
       =\widetilde O\left(
          \alpha^{-1}(\eta_c/\alpha^4)^{1/2000}\right),
\]
which proves the claimed bound on $\epsilon_{\mathrm{fac}}$.
\end{proof}

\subsection{Completing the Construction of Certificate Polynomials}
\label{sec certificate completion}

We now combine the techniques developed so far to prove
\Cref{th certificate}. Our goal is to obtain correlation at least
$(\alpha/(Ck))^{2^{O(k)}}$ while keeping the degree of the certificate
at most $2k$. Recall from \Cref{lm error-mass} that large randomized
error gives a nontrivial amount of Gaussian-weighted decision boundary
along $\v$. At a selected boundary, \Cref{lm generic certificate} gives
a positive contribution when the derivative and the other defining
polynomials are quantitatively separated from zero. If these conditions
fail, the boundary corresponds to a nearly double root or a nearly shared
root. The analysis in \Cref{sec double roots} handles nearly double roots:
either their contribution to the weighted boundary is small, or a defining
inequality can be approximated by a Boolean function of $\w$ and absorbed
into $g(\w)$. For the remaining nearly shared roots,
\Cref{lm small resultant to factorization} either simplifies the instance
through approximate proportionality or supplies a common affine
polynomial; its approximately $\w$-dependent alternative is excluded on
the selected heavy boundary events in the proof below.

Once an affine polynomial $L$ with $\partial_{\v}L\ne0$ has been
found, \Cref{lm polynomial division} gives
\[
    q_i(\x)=L(\x)\ell_i(\x)+m_i(\w),
    \qquad m_i(\w)=q_i(r_L(\w)\v+\w),
\]
where $r_L(\w)$ is the root of $L$ along $\v$. Small remainders allow
us to approximate additional polynomials using $L$, whereas remainders
with nontrivial norm give separation at its roots by anti-concentration.
This suggests grouping the defining polynomials according to whether
they are approximately divisible by $L$. For an instance in which one
group has the exact representation $q_i=L\ell_i$, suppose that roots of
$L$ are active boundaries on a set of nontrivial weighted mass and that
the cofactors $\ell_i$ and the defining polynomials outside this group
are quantitatively separated from zero there. Then
\Cref{lm sec55 affine certificate} constructs a certificate with large
correlation. We use separation of these polynomial values, rather than
a condition stated only in terms of distances between roots.

Finding a factorization does not by itself guarantee these properties.
The newly obtained affine polynomial might not give an active boundary,
and replacing the defining polynomials by their approximations may remove
the boundary from which the factorization was obtained. We therefore use
an iterative construction. Denote by $F$ the current approximating
intersection. Each defining polynomial of $F$ is either an undecomposed
quadratic or has a recorded exact representation
$q_i=\ell_{i1}\ell_{i2}$, where the affine polynomials may be constant
and scalar signs are included in them. Once such a representation is
obtained, we keep it and do not factorize the same quadratic again.
At each round, the active boundaries of $F$ are associated with
undecomposed quadratics or affine polynomials already appearing in these
representations. We select one such polynomial whose associated
boundaries have nontrivial weighted mass, and handle its nearly double
roots using \Cref{sec double roots} when it is quadratic.

If the selected boundaries satisfy the required separation conditions,
the construction from \Cref{lm generic certificate} gives a certificate
when the selected polynomial is quadratic, and the construction from
\Cref{lm sec55 affine certificate} gives one when it is affine. Otherwise,
at the chosen separation scales, other undecomposed quadratics or
previously obtained affine polynomials have small values on substantial
collections of these selected roots. We use this information to simplify
$F$. If the selected polynomial is affine, say $L$, polynomial division
allows us to replace the relevant undecomposed quadratics by $L\ell_j$.
An affine polynomial that is small at these roots is instead replaced
by a scalar multiple of $L$, consistently at all its current occurrences.
Thus a round can make progress even when no quadratic remains to be
decomposed.

If the selected polynomial is an undecomposed quadratic $q_i$, we apply
\Cref{lm small resultant to factorization} to the relevant pairs. When
all selected partners are approximately proportional to $q_i$, we identify
them. Otherwise, one pair supplies a decomposition
$q_i\approx\ell_{i1}\ell_{i2}$. We then use these same two affine
polynomials for all the selected partners: the division lemma approximates
each partner using one of $\ell_{i1},\ell_{i2}$, with the choice allowed
to depend on the partner. Their small-value events need not occur on the
same layers. All these approximations are obtained using the selected
roots of the current instance, before making the replacements. We then
replace $F$ by the resulting intersection $F'$ and select an actual
boundary of $F'$ for the next round, without requiring the old boundary
to survive.

Each unsuccessful round therefore removes or decomposes an undecomposed
quadratic, or identifies affine polynomials already present. Each original
quadratic can be removed or decomposed at most once, and at most $2k$
affine polynomials are introduced in total. Hence there are only $O(k)$
such changes. Provided the accumulated approximation error is small
enough to preserve nontrivial randomized error, the construction must
reach an instance for which one of the two certificate constructions
succeeds.

It remains to choose the approximation accuracies so that the eventual
certificate also correlates with the original target. Write
$F_0=f,F_1,\ldots,F_t$ for the successive instances and
$e_s=\Pr_{\x\sim\cN(0,\bI)}[F_s(\x)\ne F_{s+1}(\x)]$.
If a normalized certificate $H$ satisfies
$\E_{\x\sim\cN(0,\bI)}[F_tH]\ge\beta$, then
\[
    \E_{\x\sim\cN(0,\bI)}[fH]
    \ge \beta-\sqrt{\sum_{s=0}^{t-1}e_s}.
\]
Thus the accumulated disagreement must be small compared with $\beta^2$,
not merely with $\alpha$. The factorization and division estimates give
replacement errors of the form $e\le b^{-A}u^c$, where $u$ is the
small-value cutoff, $b$ is a fixed power of $\alpha/(Ck)$, and $A,c>0$
are universal constants. Consequently, preserving a later correlation
requires only a fixed-power dependence of the current cutoff on that
correlation and $b$.

There is one additional issue: the certificate contains a product of
several polynomial values, so using the same cutoff for all of them could
introduce an extra factor of $k$ into the accuracy exponent at every
round. We avoid this by simplifying several polynomials together. If $j$
different polynomials are small, the same affine decomposition handles
them with one common error bound and makes at least $\lceil j/2\rceil$
changes. We therefore choose different cutoffs according to the number
of changes they allow. In the case where a certificate is obtained, the
logarithmic losses from the product are bounded by a geometric sum; in
the other case, the replacement error is small enough for the certificate
promised after those changes to transfer back. This keeps the accuracy
loss per unit of progress at a universal power. After $O(k)$ changes,
including the normalization and Hermite-block extraction, we obtain
correlation at least
\[
    \left(\frac{\alpha}{Ck}\right)^{D^{O(k)}}
    =\left(\frac{\alpha}{Ck}\right)^{2^{O(k)}},
\]
where $D>1$ is a universal constant. The detailed choice of cutoffs below
implements this calculation.

We first give the following fact on the stability of Boolean functions under perturbation,

\begin{fact}[Stability under polynomial perturbations]
\label{fact sec55 perturbation}
Suppose that $f$ is determined by $\w$ and the signs of $N$
polynomials of degree at most two and norm one. If each polynomial
is replaced, at all its occurrences, by a polynomial within $\delta$
in $L_2$ norm, then the resulting function $\widetilde f$ satisfies
\[
    \Pr_{\x\sim \cN(0,\bI)}[f\ne\widetilde f]\le CN\delta^{2/5}.
\]
For any Boolean functions $F,H$, with
$e=\Pr_{\x}[F\ne H]$, and every polynomial $p$,
\[
    |\operatorname{Rand}_F(\v)-\operatorname{Rand}_H(\v)|\le2e,
    \qquad |\E_{\x \sim \cN(0,\bI)}[(F-H)p]|\le\sqrt e\,\|p\|_2.
\]
\end{fact}

\paragraph{A certificate around a given affine polynomial.}
Suppose a collection of defining inequalities has already been
written using the same $L$. The following lemma gives the certificate
and relates its size directly to the remainders of the other polynomials.
Multiplication of an inequality by a positive constant does not change
the target, so the normalizations in the statement cause no restriction.

\begin{lemma}[Certificate using a common affine polynomial]
\label{lm sec55 affine certificate}
Let $\|L\|_2=1$ and $a=\partial_{\v}L\ne0$. Suppose
\[
 f(\x)=g(\w)\mathbf1\{L(\x)\ell_i(\x)\ge0\ (i\in S),\quad
                              q_j(\x)\ge0\ (j\notin S)\},
\]
where $\emptyset\ne S\subseteq[k]$, each $\ell_i$ is affine
or constant with norm one, and the remaining $q_j$ have degree at
most two and norm at most an absolute constant. Signs are included
in $\ell_i$ and $q_j$. Put $m=|S|$ and define
\begin{equation}
 Q_L=a\left(\prod_{i\in S}\ell_i\right)
          \left(\prod_{j\notin S}q_j\right)
 \begin{cases}
     1,&m\text{ odd},\\[1mm]
     \displaystyle\sum_{i\in S}\ell_i,&m\text{ even}.
 \end{cases}
 \label{eq sec55 affine certificate}
\end{equation}
Then $\deg Q_L\le2k-1$,
$\|M_{\v}Q_L\|_2\le |a|(Ck)^{Ck}$, and every boundary
contribution $\Delta f_{\w}(s)Q_L(s\v+\w)$ is nonnegative.

More quantitatively, let $r_L(\w)$ be the root of $L$ and suppose
it is a jump of $f_{\w}$ on a measurable set $E$ with
\[
    \E_{\w}[\varphi(r_L(\w))\mathbf1_E(\w)]\ge b>0.
\]
For $i\in S$ put $m_i(\w)=\ell_i(r_L(\w)\v+\w)$, and for
$j\notin S$ put $m_j(\w)=q_j(r_L(\w)\v+\w)$. Write
$d_i=\min\{1,\|m_i\|_2\}$. If every $d_i>0$, then
\begin{equation}
 \frac{\E_{\x}[fM_{\v}Q_L]}{\|M_{\v}Q_L\|_2}
 \ge \left(\frac b{Ck}\right)^{Ck}\prod_{i=1}^k d_i^2.
 \label{eq sec55 remainder certificate}
\end{equation}
\end{lemma}
\begin{proof}[Proof of \Cref{lm sec55 affine certificate}]
At an active root $s$ of $L$, \Cref{fact 1D boundary} implies that every
nonzero outside polynomial $q_j$ is positive. Moreover, for each
$i\in S$ with $\ell_i(s\v+\w)\ne0$,
\[
    \Delta f_{\w}(s)
    =\operatorname{sign}\bigl(\partial_{\v}(L\ell_i)(s\v+\w)\bigr)
    =\operatorname{sign}\bigl(a\ell_i(s\v+\w)\bigr).
\]
Thus all nonzero cofactors have the same sign. The product has the
correct sign when $m$ is odd, and the additional sum gives the
correct sign when $m$ is even. If a cofactor or an outside polynomial
vanishes, $Q_L$ is zero. At a jump not arising from $L$, one of these
polynomials must vanish, again by \Cref{fact 1D boundary}. This proves the
nonnegative contribution at every boundary.
The degree bound follows directly from
\eqref{eq sec55 affine certificate}; H\"older's inequality,
hypercontractivity, and \Cref{fact representation} give the norm bound.

By \Cref{lm polynomial division}(1), each $m_i$ is the
remainder obtained by dividing the corresponding $\ell_i$ or $q_i$
by $L$. In particular, $\deg m_i\le2$.
Take $h=(b/(Ck))^2$. Gaussian anti-concentration gives
   $\Pr_{\w}[|m_i(\w)|\le h d_i]\le C\sqrt h$.
A union bound discards weighted mass at most $b/2$ for sufficiently
large $C$. On the remaining roots all the absolute values are at
least $h d_i$. The sum in the even case has absolute value at least
any one of its summands. Since $d_i\le1$ and $h\le1$, in both cases
\[
   \Delta f_{\w}(r_L)Q_L(r_L\v+\w)
       \ge |a|h^{k+1}\prod_{i=1}^k d_i^2.
\]
Apply \Cref{lm M operator} and divide by the norm bound. The resulting factor
$b h^{k+1}/(2(Ck)^{Ck})$ is at least $(b/(Ck))^{C'k}$ after
increasing the absolute constant. Notice that $|a|$ cancels; no
separate lower bound on $|\partial_{\v}L|$ is needed for this
normalized correlation estimate.
\end{proof}

\begin{fact}[Evaluation at selected roots and normalization]
\label{fact evaluation}
Let $0<b\le b_*$, $R_b=\sqrt{20\log(1/b)}$, and let $s(\w)$
be measurable on $E\subseteq\v^\perp$ with $|s(\w)|\le R_b$.
The following hold.

\emph{(i)} If $\deg e\le2$ and $\|e\|_2\le\delta$, then,
outside a set of probability at most $b^3/100$,
\[
 \sup_{|z|\le R_b+1}
     (|e(z\v+\w)|+|\partial_{\v}e(z\v+\w)|)
       \le C\delta b^{-4}.
\]
\emph{(ii)} If $\deg R\le2$ and
$\Pr_{\w}[|R(\w)|\le h]\ge c_0b$, then
$\|R\|_2\le C_{c_0}b^{-4}h$.

\emph{(iii)} If $\deg P\le2$, $\|P\|_2=1$, and
\[
 \E_{\w}[\varphi(s(\w))\mathbf1_E(\w)
                     \mathbf1\{|P(s(\w)\v+\w)|\le u\}]\ge b,
 \qquad u\le b^{A_0},
\]
for sufficiently large absolute $A_0$, then $P$ is nonconstant.
Writing $P=\mu+\sigma H$, where $\E_{\x}[H]=0$ and
$\operatorname{Var}_{\x}(H)=1$, gives
\[
       \sigma\ge\frac c{1+\log(1/b)},\qquad
       |\mu|/\sigma\le C(1+\log(1/b)).
\]
\end{fact}
\begin{proof}[Proof of \Cref{fact evaluation}]
For (i), write $e=a(z^2-1)+B(\w)z+C(\w)$.
Orthogonality gives $2a^2+\|B\|_2^2+\|C\|_2^2\le\delta^2$.
Chebyshev bounds $|B|,|C|$ by $C\delta b^{-3}$ outside
probability $Cb^6$. Since $R_b+1\le b^{-1}$, the asserted
uniform bound follows; decrease $b_*$ to ensure $Cb^6\le b^3/100$.
For (ii), degree-two Gaussian anti-concentration gives
$c_0b\le C(h/\|R\|_2)^{1/2}$ when $R\ne0$.
The resulting bound $Cc_0^{-2}b^{-2}h$ implies the stated one;
the zero polynomial is immediate.

For (iii), a constant of norm one cannot satisfy the hypothesis,
so $\sigma>0$. Decompose the centered unit-variance polynomial
$H$ in the same way as in (i). Gaussian tails for its affine and
quadratic coefficient polynomials imply, outside probability $b/100$,
\[
           \sup_{|z|\le R_b}|H(z\v+\w)|\le C(1+\log(1/b)).
\]
The weighted event has ordinary probability at least $b/\varphi(0)$
and intersects this set. Thus
$|\mu|\le u+C\sigma(1+\log(1/b))$.
Using $\mu^2+\sigma^2=1$ and $u\le1/4$ proves the lower
bound on $\sigma$ and then the bound on $|\mu|/\sigma$.
\end{proof}

\paragraph{The iterative construction.}
\Cref{alg sec55 certificate} summarizes the construction used in
\Cref{th certificate}. It is a proof procedure with access to the defining
polynomials and exact Gaussian boundary quantities, not a sample-based
learning algorithm. We retain the normalization and replacement conventions
of the proof. In particular, $t$ counts changes, not rounds: a change is
the removal or decomposition of an old undecomposed quadratic, or the
disappearance of an old affine polynomial from the current expressions,
counting proportional affine polynomials only once. An affine polynomial
that remains as a new quotient is not counted as discarded.

Choose the universal constants as in the proof and set
\begin{equation}
\begin{aligned}
    \rho&=\alpha/(Ck), & b&=\rho^8,
    &\beta_t&=\rho^{KkD^{3k-t}},\\
    \theta_t&=\rho^{\Lambda kD^{3k-t-1}},
    &&&u_{t,j}&=\rho^{\Lambda kD^{3k-t-\lceil j/2\rceil}}.
\end{aligned}
\label{eq sec55 pseudocode scales}
\end{equation}
In a round with selected roots $s(\w)$ on $E$, put
$m_E(B)=\E_{\w}[\varphi(s(\w))\mathbf1_B(\w)]$ for $B\subseteq E$,
where $\w$ is standard Gaussian on $\v^\perp$; set
$R_\rho=\sqrt{20\log(1/\rho)}$.
If $P_1,\ldots,P_M$ are the other different current polynomials,
a batch $(j,\mathcal B)$ means
\begin{equation}
\begin{gathered}
    j\in[M],\qquad \mathcal B\subseteq\{P_1,\ldots,P_M\},
    \qquad |\mathcal B|=j,\\
    m_E\!\left(\left\{\w\in E:
       |P(s(\w)\v+\w)|\le u_{t,j}\right\}\right)\ge b
       \quad\text{for every }P\in\mathcal B.
\end{gathered}
\label{eq sec55 pseudocode batch}
\end{equation}
For $M=0$, no such batch exists. The expectations involving $\x$ below
are under the standard Gaussian distribution on $\mathbb R^d$.

\begin{algorithm}[htbp]
\caption{Iterative construction of a directional certificate}
\label{alg sec55 certificate}
\small
\begin{algorithmic}[1]
\State\textbf{Input:} The defining representation of the target function $f$, a unit direction $\v$,
  and $0<\alpha\le1/2$ with $\operatorname{Rand}_f(\v)\ge\alpha$.

\State \textbf{Output:} $H=M_{\v}Q/\|M_{\v}Q\|_2$ with $\deg Q\le2k-1$
  and $\E_{\x}[fH]\ge\beta_0/2$.

\smallskip

\State Set $F\gets f$ and $t\gets0$; normalize and simplify as in the proof.
\While{no certificate has been returned}
    \State Select a current polynomial $q_0$ and actual jumps $s(\w)$ of
      $F_{\w}$ on $E$, associated with $q_0$, with
      $m_E(E)\ge2\rho^4$ and $|s(\w)|\le R_\rho$.
    \If{$q_0=az^2+B(\w)z+C(\w)$ is undecomposed and
          $\|B^2-4aC\|_2\le\theta_t$}
        \State Approximate its inequality by a Boolean function of $\w$
          using \Cref{sec double roots}; absorb it into the gate and remove $q_0$, giving $F'$.
    \Else
        \If{$q_0$ is an undecomposed quadratic}
            \State Discard from $E$ the layers where
              $|\partial_{\v}q_0(s(\w)\v+\w)|\le\theta_t$.
        \EndIf
        \State List the other different current polynomials $P_1,\ldots,P_M$;
          search for a batch $(j,\mathcal B)$ satisfying
          \eqref{eq sec55 pseudocode batch}.
        \If{no such batch exists}
            \State If $q_0=q_i$ is quadratic, take
              $Q=(\partial_{\v}\prod_{j}q_j)$;
              otherwise use $Q_L$ from \Cref{lm sec55 affine certificate}.
            \State Reduce repeated affine powers by parity in the
              multiplicative part, as in the proof.
            \State \Return $H=M_{\v}Q/\|M_{\v}Q\|_2$.
        \EndIf
        \State Fix one batch $(j,\mathcal B)$ and put $u\gets u_{t,j}$.
        \If{$q_0=L$ is affine}
            \State Keep $L$ fixed. For each $P\in\mathcal B$, use
              \Cref{lm polynomial division} to write $P=L\ell_P+m_P(\w)$;
              prescribe $P\mapsto L\ell_P$.
        \Else
            \State Orient and normalize the pairs $(q_0,P)$ and apply
              \Cref{lm small resultant to factorization} as in the proof.
            \If{every selected partner is in the proportionality case}
                \State Prescribe scalar-multiple replacements using one
                  common polynomial from $\{q_0\}\cup\mathcal B$,
                  choosing an affine one when available.
            \Else
                \State Obtain $q_0\approx\kappa L_1L_2$ from one pair;
                  prescribe $q_0\mapsto\kappa L_1L_2$.
                \State For each $P\in\mathcal B$, choose one
                  $L(P)\in\{L_1,L_2\}$ using its small-value event at the
                  \emph{original} selected roots $s(\w)$.
                \State Apply \Cref{lm polynomial division} with divisor
                  $L(P)$; prescribe $P\mapsto L(P)\ell_P$.
            \EndIf
        \EndIf
        \State Apply the prescribed replacements simultaneously at all old
          occurrences, without substituting again into new quotients, giving $F'$.
    \EndIf
    \State Normalize $F'$, retain scalar signs and decomposed forms, identify
      proportional copies, and simplify repeated, constant, and square inequalities.
    \State Let $\Delta t$ be the actual number of changes from the old
      representation of $F$ to the resulting representation of $F'$.
    \State Set $F\gets F'$ and $t\gets t+\Delta t$;
      restart by selecting an actual boundary of this new $F$.
\EndWhile
\end{algorithmic}
\end{algorithm}

The proof below excludes the approximately $\w$-dependent alternative
in the quadratic-pair branch and justifies all choices in the construction.
After each update ending at $t'$, it establishes
\[
    \Pr_{\x}[F\ne F']\le\beta_{t'}^{20},
    \qquad \operatorname{Rand}_{F'}(\v)\ge\alpha/2.
\]
The small-discriminant branch makes at least one change; a batch of $j$
partners makes at least $\lceil j/2\rceil$ changes. A successful round
gives $\E_{\x}[FH]\ge\beta_t$, and stability transfers this to
$\E_{\x}[fH]\ge\beta_0/2$. The final paragraph of the proof then
extracts the required normalized block $h_r(\v\cdot\x)p(\w)$.
The error estimates are guarantees of the proof, not additional tests
performed by the procedure.

\begin{proof}[Proof of \Cref{th certificate}]
\Cref{alg sec55 certificate} records the construction.
We may assume $0<\alpha\le1/2$. Write $\x=z\v+\w$, where
$z\sim\cN(0,1)$ and $\w\sim\cN(0,\bI_{\v^\perp})$ are
independent.
We construct successive approximating intersections, starting with $F=f$.
At each round we select an actual boundary of $F$, and either obtain a
certificate or replace $F$ by a simpler intersection.

\paragraph{The representations and allowed changes.}
Normalize every undecomposed quadratic and every recorded affine
polynomial to norm one, keeping scalar signs in the defining inequalities.
An inequality that has been decomposed remains a product of affine
polynomials; we never treat this product as an undecomposed quadratic again.
Proportional affine polynomials are represented by the same normalized
polynomial, with their scalars retained. Delete repeated inequalities and
simplify constant and square inequalities. There are at most $k$
inequalities and at most $2k$ different nonconstant polynomials in these
expressions, counting each undecomposed quadratic and each recorded affine
polynomial once.

An affine replacement $M\mapsto cL$ is made at every current occurrence
of $M$, not by independently changing its coefficients in different
inequalities. Replacements in a round are simultaneous: we do not substitute
again into the newly obtained quotients. Count one change for each old
undecomposed quadratic no longer present as such, and one for each old
affine polynomial absent from the resulting expressions. In particular,
an affine polynomial that still occurs as a new quotient is not counted
as discarded.

Let $t$ be the total number of changes so far, not the number of rounds.
Each original quadratic is removed or decomposed at most once. New affine
polynomials arise only from such decompositions, with at most two
introductions per quadratic; a batch of $q+1$ quadratics sharing two
new affine polynomials introduces at most $q+2\le2(q+1)$.
Thus at most $2k$ affine polynomials are introduced in total, including
initial affine inequalities and later reintroductions. If $N$ different
polynomials currently remain, then
\begin{equation}
    t+N\le3k.
    \label{eq sec55 finite changes}
\end{equation}
Indeed, removed and remaining undecomposed quadratics total at most $k$,
while discarded and remaining affine polynomials total the number
introduced. This also rules out cycling: without a new quadratic
decomposition an affine merger strictly reduces the current collection.

\paragraph{Accuracies to be maintained.}
Let $A>1$ and $c\in(0,1)$ be universal constants large and small enough,
respectively, for the replacement estimates from
\Cref{lm small resultant to factorization,lm polynomial division},
\Cref{lm double roots}, and \Cref{fact sec55 perturbation}. The applications
below establish disagreement bounds $b^{-A}u^c$, for $u\le b^A$.
Choose universal constants $K,\Lambda,D>1$, in that order, such that
\begin{equation}
    K\ge4C_*,\qquad c\Lambda\ge20K+8A+20,\qquad
    \frac{\Lambda}{D}+\frac{6\Lambda}{D-1}\le\frac K2,
    \label{eq sec55 scale constants}
\end{equation}
where $C_*$ absorbs the certificate norm and boundary-mass constants below.
Set
\[
    \rho=\frac{\alpha}{Ck},\qquad b=\rho^8,\qquad
    \beta_t=\rho^{KkD^{3k-t}}\quad(0\le t\le3k).
\]
We take $C$ sufficiently large throughout. Every replacement ending at a
new count $t'$ will satisfy
\begin{equation}
    \Pr_{\x}[F\ne F']\le\beta_{t'}^{20}.
    \label{eq sec55 update invariant}
\end{equation}
Since each replacement increases $t$, the accumulated disagreement is at
most $\sum_{s=1}^{3k}\beta_s^{20}\le3k\rho^{20Kk}<\alpha/4$.
\Cref{fact sec55 perturbation} then ensures
$\operatorname{Rand}_F(\v)\ge\alpha/2$. Initially this holds without
any replacements; below we verify \eqref{eq sec55 update invariant}
for each possible update. In particular, positive randomized error ensures
$N\ge1$ and hence $t\le3k-1$ at the start of every round.

\paragraph{Selecting a boundary and handling nearly double roots.}
At the beginning of a round let $t$ be the current count. By
\Cref{lm error-mass}, whose proof applies to every finite-jump Boolean
layer,
\[
    \E_{\w}\left[\sum_{s\in S_{\w}}\varphi(s)\right]\ge c\alpha.
\]
Assign each jump to a sign-changing zero of an undecomposed quadratic or
a recorded affine polynomial. There are at most $2k$ choices and at most
two assigned roots per choice on each layer. Pigeonhole to one polynomial
$q_0$ and retain its assigned root of largest Gaussian density on each
layer. Discarding roots outside
$R_\rho=\sqrt{20\log(1/\rho)}$ loses at most
$2k\varphi(R_\rho)=O(k\rho^{10})$ in weighted mass. We obtain measurable selected roots $s(\w)$ on a set $E$ with
\[
    \E_{\w}[\varphi(s(\w))\mathbf1_E(\w)]\ge2\rho^4,
    \qquad |s(\w)|\le R_\rho.
\]
Use fixed ordering and tie-breaking for these selections.

If $q_0$ is quadratic, write $q_0=az^2+B(\w)z+C(\w)$ and put
$\Delta=B^2-4aC$ and $\theta=\rho^{\Lambda kD^{3k-t-1}}$.
If $\|\Delta\|_2\le\theta$, use \Cref{fact evaluation}(iii)
to write $q_0=\mu+\sigma p$, where $p$ has mean zero and variance
one, and $\sigma\ge c/(1+\log(1/b))$. The discriminant of
$q_0/\sigma$ is $\Delta/\sigma^2$. \Cref{lm double roots} therefore
approximates the inequality by a Boolean function of $\w$, with error
at most $b^{-A}\theta^c$, after increasing $A$ and decreasing $c$. Absorb that function into the gate. This
removes a quadratic, so $t'\ge t+1$, and
\[
    b^{-A}\theta^c
    =\rho^{c\Lambda kD^{3k-t-1}-8A}
    \le\beta_{t+1}^{20}\le\beta_{t'}^{20}.
\]
Thus the update satisfies \eqref{eq sec55 update invariant}; start the
next round. Otherwise, the root--discriminant identity and the argument
of \Cref{lm no double roots} give
\[
    \Pr_{\w}[\exists s:\ q_0(s\v+\w)=0,
                        \ |\partial_{\v}q_0(s\v+\w)|\le\theta]
    \le C\sqrt\theta.
\]
Discard these layers. The remaining selected mass is at least $\rho^4$,
and $|\partial_{\v}q_0|>\theta$ there. If $q_0$ is affine, no such
step is needed; define the same $\theta\le1$ for the bounds below.
Write $m(B)=\E_{\w}[\varphi(s(\w))\mathbf1_B(\w)]$ for $B\subseteq E$.

\paragraph{A finite test for success or simplification.}
Let $P_1,\ldots,P_M$ be the other different normalized polynomials
currently appearing. By \eqref{eq sec55 finite changes},
$M\le3k-t-1$. Put
\[
    d_j=\lceil j/2\rceil,\qquad
    u_j=\rho^{\Lambda kD^{3k-t-d_j}}\quad(j\in[M]).
\]
We ask whether, for some $j$, at least $j$ different polynomials satisfy
\begin{equation}
    m\bigl(\{\w\in E:|P_i(s(\w)\v+\w)|\le u_j\}\bigr)\ge b.
    \label{eq sec55 batch test}
\end{equation}
If this happens, choose any $j$ such polynomials and set $u=u_j$;
they are the partners to be simplified in this round.

Suppose first that it does not happen. Relabel the polynomials so that
\[
    m\bigl(\{|P_j(s(\w)\v+\w)|\le u_j\}\bigr)<b
    \quad\text{for every }j\in[M].
\]
Such a relabeling is obtained backwards: when assigning the $j$th position,
$j$ polynomials remain unassigned, and fewer than $j$ polynomials in the
whole collection satisfy \eqref{eq sec55 batch test} at $u_j$.
Choose one of the remaining polynomials that does not satisfy it.
Discard the union of these small-value events. Its weighted mass is at
most $Mb\le\rho^4/2$, so the remaining roots have mass at least
$\rho^4/2$ and $|P_j|>u_j$ for every $j$.

If $q_0=q_i$ is an undecomposed defining quadratic, take
$Q=(\partial_{\v}\prod_{j}q_j)$, as in
\Cref{lm generic certificate}. If $q_0=L$ is affine, use the polynomial
from \Cref{lm sec55 affine certificate}. \Cref{fact 1D boundary} gives the required
jump orientations. In the multiplicative part of $Q$, expand only the
recorded affine products and reduce each repeated power $L^a$ to $L$
for odd $a$ and to $L^2$ for even $a$. This preserves signs and zeros,
and hence nonnegative contributions at every boundary. Each different
polynomial then occurs at most twice. At active affine roots the
cofactors have a common sign, so the possible additional sum costs at
most one more clipped value.

The degree is at most $2k-1$. Hypercontractivity and \Cref{fact representation} give
$\|M_{\v}Q\|_2\le(Ck)^{Ck}$ in the quadratic case and
$\|M_{\v}Q\|_2\le|\partial_{\v}L|(Ck)^{Ck}$ in the affine case.
In the latter case the same constant derivative occurs in the boundary
contribution and cancels after normalization. \Cref{lm M operator} therefore gives
$H=M_{\v}Q/\|M_{\v}Q\|_2$ satisfying
\begin{align*}
    \E_{\x}[FH]
    \ge\rho^{C_*k}\theta\prod_{j=1}^M u_j^3
    =\rho^{C_*k+\Lambda kD^{3k-t-1}
                  +3\Lambda k\sum_{j=1}^M D^{3k-t-d_j}}
     \ge\rho^{KkD^{3k-t}}=\beta_t.
\end{align*}
Here
$\sum_{j=1}^M D^{3k-t-d_j}\le2D^{3k-t}/(D-1)$, and we used
\eqref{eq sec55 scale constants}. For $M=0$ the same argument uses an
empty product. We stop with this certificate.

\paragraph{An unsuccessful round at an affine boundary.}
Now suppose \eqref{eq sec55 batch test} holds for the selected $j$
partners, and $q_0=L$ is affine. Its derivative along $\v$ is nonzero,
and the selected points are its roots. For each selected partner $P$,
the small-value event has ordinary probability at least $cb$.
\Cref{lm polynomial division}(1),(3) gives
\begin{equation}
    P=L\ell+m_P(\w),\qquad
    \deg\ell\le\deg P-1,\qquad \|m_P\|_2\le Cb^{-2}u.
    \label{eq sec55 exact division}
\end{equation}
Replace every selected undecomposed quadratic by $L\ell$ and every
selected affine polynomial by the corresponding scalar multiple of $L$.
Thus the selected $L$ is kept fixed, and already decomposed inequalities
remain decomposed.

\paragraph{An unsuccessful round at a quadratic boundary.}
Suppose instead that $q_0$ is an undecomposed quadratic, including the
sign of its defining inequality. For each selected partner $P$, fix its
sign on the side where the current intersection is one, retaining weighted
mass at least $b/2$. By \Cref{fact 1D boundary} the selected root is a jump of this
oriented pair. \Cref{fact evaluation}(iii) allows us to divide their
centered parts by standard deviations at least $c/(1+\log(1/b))$.
The normalized pair has thresholds $O(1+\log(1/b))$ and partner value
at most $Cu(1+\log(1/b))$. Apply
\Cref{lm small resultant to factorization} and rescale back. Its error
is at most
    $\delta_{\rm pair}=Cb^{-C_1}u^{c_1}\le b^{110}$,
for universal $C_1,c_1>0$, by taking the cutoff exponent $A$ large enough.
The alternative of being nearly independent of $\v$ is excluded:
if a unit-norm polynomial is within $\delta_{\rm pair}$ of $R(\w)$,
then $\|R\|_2\ge1-\delta_{\rm pair}$, whereas
\Cref{fact evaluation}(i),(ii) on its selected small-value event gives
\[
    \|R\|_2\le Cb^{-4}(u+C\delta_{\rm pair}b^{-4})<1/2.
\]

If every selected partner is approximately proportional to $q_0$, retain
one common polynomial, choosing an affine one when available, and replace
the others by the appropriate scalar multiples. Unit norms ensure the
scalars are bounded above and below in absolute value. Otherwise one
pair supplies, after normalization,
\begin{equation}
    \|q_0-\kappa L_1L_2\|_2\le\delta_0\le C\delta_{\rm pair},
    \qquad \|L_1\|_2=\|L_2\|_2=1,\qquad c_0\le|\kappa|\le C_0.
    \label{eq sec55 seed decomposition}
\end{equation}
The last bounds follow from \Cref{fct:gaussian-product-norm}.
Keep these same $L_1,L_2$ for all selected partners. At the original
selected roots of $q_0$, \Cref{fact evaluation}(i) gives
$|L_1L_2|\le C\delta_0b^{-4}$ outside probability $b^3/100$.
For each partner, intersect with its small-value event and choose the
smaller affine value. One fixed $L\in\{L_1,L_2\}$ then satisfies
\[
    |P(s\v+\w)|\le u,\qquad
    |L(s\v+\w)|\le C\sqrt{\delta_0}\,b^{-2}
\]
on probability at least $cb$.

Write $L(z\v+\w)=az+B(\w)$. We have $|a|\ge b^4$:
otherwise $\|B\|_2\ge1/2$, but
$|B|\le b^4R_b+C\sqrt{\delta_0}b^{-2}\le b^3$ on probability
at least $cb$, contradicting affine anti-concentration.
\Cref{lm polynomial division}(1),(2) now gives $P=L\ell+m_P$
with $\|\ell\|_2\le Cb^{-8}$. Part (3) of that lemma gives
\begin{equation}
    \|m_P\|_2\le Cb^{-2}
       \bigl(u+\sqrt{\delta_0}\,b^{-10}\sqrt{\log(2/b)}\bigr)
       \le Cb^{-C_2}(u+\sqrt{\delta_0}).
    \label{eq sec55 approximate division}
\end{equation}
Replace $q_0$ by $\kappa L_1L_2$ and each partner by its approximation
$L\ell$. Different partners may use different choices of $L$, and their
events need not overlap. All estimates use the old selected roots;
no boundary of the new intersection has yet been assumed to survive.

\paragraph{Progress and error of an unsuccessful round.}
Make the replacements simultaneously as specified at the start, retain
scalar signs, normalize the new affine polynomials, and simplify repeated,
constant, and square inequalities. Let $F'$ be the resulting intersection.
The bounds above approximate every changed normalized polynomial within
$Cb^{-C_3}u^{c_3}$, for universal $C_3,c_3>0$. Since their number is at
most $2k\le b^{-1}$, \Cref{fact sec55 perturbation} gives
\begin{equation}
    e:=\Pr_{\x}[F\ne F']\le b^{-A}u^c.
    \label{eq sec55 joint error}
\end{equation}
This justifies the universal constants used earlier; the small-discriminant
replacement is covered by the same constants after adjustment.

To count the changes, suppose $q$ selected partners are quadratic and
$a$ are affine, so $j=q+a$. At an affine boundary all $q$ quadratics
are decomposed. Of the selected affine polynomials, at most $q$ can
remain among the new quotients. Thus there are at least
$q+\max\{0,a-q\}\ge\lceil j/2\rceil$ changes. In the proportional
case there are at least $j$ changes. In the common-affine quadratic case,
$q_0$ and all $q$ quadratic partners are decomposed, and at most $q+2$
selected affine polynomials can remain among $L_1,L_2$ and the new
quotients. There are therefore at least
$q+1+\max\{0,a-q-2\}\ge\lceil j/2\rceil$ changes.
These counts allow a new quotient to coincide with an old polynomial.
Hence the new count satisfies $t'\ge t+d_j$. Since $u=u_j$,
\[
    e\le\rho^{c\Lambda kD^{3k-t-d_j}-8A}
       \le\beta_{t+d_j}^{20}\le\beta_{t'}^{20}.
\]
This proves \eqref{eq sec55 update invariant}. All cutoffs used above
are valid: $u_j,\theta\le b^A$, $R_\rho\le R_b$, and $2k\le b^{-1}$.
Replace $F$ by $F'$ and begin the next round by selecting an actual
boundary of $F'$. We do not continue at an assumed boundary of $L_1$
or $L_2$.

\paragraph{Termination and transfer to the original target.}
Every unsuccessful round increases $t$. By
\eqref{eq sec55 finite changes}, at most $3k$ changes are possible;
if no polynomial remained, the intersection would depend only on $\w$
and have zero randomized error. Since every update preserves randomized
error at least $\alpha/2$, a successful round must occur.
Let $t_*$ be its count, and let $H$ be its normalized certificate.
The sequence $\beta_t$ is increasing. Thus the sum of the disagreement
errors from all earlier rounds is at most $3k\beta_{t_*}^{20}$, and
\Cref{fact sec55 perturbation} gives
\[
    \E_{\x}[fH]\ge\beta_{t_*}-\sqrt{3k}\,\beta_{t_*}^{10}
       \ge\tfrac12\beta_0.
\]
Finally $\deg Q\le2k-1$, so \Cref{fact representation} gives
\[
    H(z\v+\w)=\sum_{r=1}^{2k}h_r(z)p_r(\w),\qquad
    \deg p_r\le2k-r,\qquad \sum_r\|p_r\|_2^2=1.
\]
One summand has correlation at least $\beta_0/(4k)$. Normalize its
nonzero $p_r$, which can only increase that positive correlation.
Since $KkD^{3k}=2^{O(k)}$, the resulting $r,p$ satisfy
\[
    \|p\|_2=1,\qquad r+\deg p\le2k,\qquad
    \E_{\x}[f(\x)h_r(\v\cdot\x)p(\w)]
       \ge\left(\frac{\alpha}{Ck}\right)^{2^{O(k)}}.
    \qedhere
\]
\end{proof}

\section{Efficient Correlation Finding through Influence Matrix}\label{sec direction finding}

By \Cref{th certificate}, we know that if there is a direction $\v \in \s^{d-1}$, such that $\rand_f(\v) \ge \alpha$, then there exists some polynomial $p(\w)$ such that $\E_{\x \sim \cN(0,\bI)}[yh_k(\v\cdot\x)p(\w)]>\poly(\alpha)$.
In this section, we give an algorithm to approximately find such a direction $\v$.
Notice that in general $k+\deg(p) \ge 2$, so
if $\x$ is sampled from a general distribution, then finding such a direction would require us to solve a higher-order tensor problem, which is generally hard \cite{hillar2013most}. However, the structure of the Gaussian distribution allows us to do this much more efficiently.

Suppose that $k+\deg(p) \le \ell$, then we know that for every $\v$ and every $p$, $\E_{(\x,y)\sim D}[yh_k(\v\cdot\x)p(\w)]$ only depends on the projection of $y$ onto the subspace spanned by Hermite polynomials with degree at most $\ell$.
Thus, we instead consider $F(\x) = \proj_{P_{\le \ell}}(y)$, the low degree projection of $y$.
The central part of our analysis relies on \Cref{lm influential matrix}, which consider the influential matrix defined by $G_\ell:= \E_{\x \sim \cN(0,\bI)}[\nabla F\nabla F ^\top]$.
In fact, denote by $i \in [\ell]$, $B_i(\v) = \sup_{p(\w) \in \mathcal{P}_{\le \ell-i},\norm{p}\le 1} \E_{(\x,y)\sim D}[yh_i(\v\cdot\x)p(\w)]$, then for every $\v$, we have $\v^\top G_\ell \v = \sum_{i=1}^\ell i B_i^2(\v)$.
This immediately gives two consequences. First, the top eigenvector must have large correlation with respect $y$. Second, any eigenvector of $G_\ell$ with a small eigenvalue must have tiny correlation with respect to $y$. In fact, since $\tr(G) \le \ell$, there cannot be too many large eigenvectors of $G_\ell$. Combine this with \Cref{th certificate}, we can find a subspace $V$ of dimension $\poly_k(1/\alpha)$ that contains all directions in $\R^d$ with a randomized error at least $\alpha$ with respect to any intersections of $k$ degree 2 PTFs $f(\x)$ in one shot.

However, we still need to estimate $G_\ell$ well enough. Fortunately, the under Hermite basis, the coefficient of $F$ with respect to a Hermite polynomial $H_\gamma$ is just $\E_{(\x,y)\sim D}[yH_\gamma(\x)]$. Using basic properties of Hermite polynomials, this gives us a simple way to estimate $G_\ell$ to a desired accuracy.

\begin{algorithm}[htbp]
		\caption{\textsc{CorrelationFinding}}\label{alg direction finding}
		\begin{algorithmic} [1]
\State \textbf{Input:}  Sample access from $D$, degree budget $\ell$, correlation parameter $\alpha$
\State \textbf{Output:} Subspace $V \subseteq \R^d$ and when $V \neq \{0\}$, a direction $\hat \v \in \s^{d-1}$, along which $y$ has a non-trivial correlation

\smallskip

\State Draw $N = d^{O(\ell)}\poly(1/\alpha)$ examples $\{(\x^{(i)},y^{(i)})\}_{i=1}^N$ from $D$
\State
For each multi-index $\gamma$ over $[d]$ with $\card{\gamma} \le \ell$, let $\hat{c}_\gamma = N^{-1} \sum_{i=1}^N (y^{(i)}H_\gamma(\x^{(i)}))$, where $H_\gamma$ is the normalized Hermite polynomial with multi-index $\gamma$.

\State Construct matrix $\hat{G} \in \R^{d \times d}$ such that for $i,j \in [d]$, $\hat{G}_{ij} = \sum_{\gamma:\card{\gamma}\le \ell-1}\sqrt{(\gamma_i+1)(\gamma_j+1)}\hat c_{\gamma+\e_i}\hat c_{\gamma+\e_j}$

\State Return $V$ as the subspace spanned by eigenvectors of $\hat{G}$ with large eigenvalue. And if $V \neq \{0\}$, also return $\v$ the top eigenvector of $\hat{G}_\ell$.
\end{algorithmic}
\end{algorithm}

\begin{theorem}\label{th direction finding efficient}
    Let $D$ be a distribution over $\R^d \times \{0,1\}$ such that $D_\x = \cN(0,\bI)$. There is an algorithm such that for $\alpha \in (0,1)$, suppose there exists a direction $\v \in \s^{d-1}$, $k \in \N$ and a degree $r$ polynomial $p(\w)$, with $\norm{p(\w)} \le 1$ such that $\abs{\E_{(\x,y)\sim D}[yh_k(\v\cdot\x)p(\w)]}>\alpha$, then it draws $d^{(r+k)}\poly(1/\alpha)$ examples from $D$, runs in $\poly(d^{r+k}/\alpha)$ time and outputs a vector $\hat{\v}$ such that there exists some polynomial $q(\hat{\w})$ $\norm{q(\hat{\w})}\le 1$, that satisfies $\abs{\E_{(\x,y)\sim D}[yh_i(\hat \v\cdot\x)q(\hat \w)]}>\Omega(\alpha/(k+r))$, where $i+\deg(q) \le r+k$.
\end{theorem}
Furthermore, we give an even stronger algorithm that finds all such directions in one shot.

\begin{theorem}\label{th direction finding one shot}
 Let $D$ be a distribution over $\R^d \times \{0,1\}$ such that $D_\x = \cN(0,\bI)$ and $\ell \in N_+$.  There is an algorithm such that for $\beta \in (0,1)$, it draws
    $d^{O(\ell)}\poly(1/\beta)$ examples from $D$, runs in $\poly(d^{\ell}/\beta)$ time and outputs a subspace $V \subseteq \R^d$ with dimension at most $O(\ell/\beta^2)$ such that for every $\v \perp V$, $\sup_{p,k: \norm{p}=1, k+\deg(p) \le \ell, k\ge 1}\E_{(\x,y)\sim D}[yh_k(\v\cdot\x)p(\w)] \le \beta$.
\end{theorem}

To prove the above two theorems, we present the following technical lemmas.

\begin{lemma}\label{lm influential matrix}
    Let $D$ be a distribution over $\R^d \times \{0,1\}$ such that $D_\x = \cN(0,\bI)$. For $\ell \in \N_+$, denote by $F_\ell = \proj_{\le \ell}(y)$, the projection of $y$ onto the subspace of polynomials of degree no larger than $\ell$ under Gaussian distribution and denote by $G_\ell = \E_{\x \sim \cN(0,\bI)}[\nabla F(\x) \nabla F(\x)^\top]$. Then, for every $\v \in \s^{d-1}$, we have
    \begin{align*}
        \v^\top G_\ell \v = \sum_{i=1}^\ell i B_i^2(\v) \;,
    \end{align*}
    where for $i \in [\ell]$, $B_i(\v) = \sup_{p(\w) \in \mathcal{P}_{\le \ell-i},\norm{p}\le 1} \E_{(\x,y)\sim D}[yh_i(\v\cdot\x)p(\w)]$.
    Furthermore, there is an algorithm such that for every $\eps \in (0,1)$, it draws $d^{O(\ell)}\poly(\ell/\eps)$, runs in $d^{O(\ell)}\poly(\ell/\eps)$ time and outputs a matrix $\hat{G}$ such that with probability at least $99\%$, $\norm{G_\ell-\hat{G}}_{op} \le \eps$.
\end{lemma}

\begin{proof}[Proof of \Cref{lm influential matrix}]
    We expand $F_\ell$ using normalized Hermite polynomials along direction $\v$ and its orthogonal subspace.
    \begin{align*}
        F_\ell(\x) = \sum_{i, \alpha: \card{\alpha}+i \le \ell} c_{i,\alpha} h_i(\v\cdot\x)H_{\alpha}(\w).
    \end{align*}
    Let $p(\w)$ be any polynomial over $\v^\perp$ with degree at most $\ell -i$, with $i \le \ell$. We expand $p(\w)$ using normalized Hermite polynomial.
    \begin{align*}
        p(\w) = \sum_{\alpha:\card{\alpha}\le \deg(p)} a_\alpha H_\alpha(\w).
    \end{align*}
    In particular, we have $\norm{p}^2 = \sum_{\alpha} a_\alpha^2$. Using orthogonality of Hermite polynomials, we have
    \begin{align*}
        \E_{(\x,y)\sim D}[yh_i(\v\cdot\x)p(\w)] = \E_{(\x,y)\sim D}[F_\ell h_i(\v\cdot\x)p(\w)] = \sum_{\alpha:\card{\alpha}+i \le \ell} c_{i,\alpha} a_\alpha.
    \end{align*}
Under the constraint $\sum_\alpha a_\alpha^2 \le 1$,
Cauchy--Schwarz gives $    B_i(\v)^2
    = \sum_{\alpha:\card{\alpha}+i \le \ell} c_{i,\alpha}^2.$
When this sum is positive, the maximum is attained at
\begin{align*}
    p(\w)
    =
    \frac{\sum_{\alpha:\card{\alpha}+i \le \ell}
          c_{i,\alpha}H_\alpha(\w)}
         {\left(\sum_{\alpha:\card{\alpha}+i \le \ell}
          c_{i,\alpha}^2\right)^{1/2}}.
\end{align*}
Otherwise, take $p=0$.
On the other hand, for each $\v \in \s^{d-1}$, we have
\begin{align*}
    \v^\top G_\ell \v
    = \norm{\partial_\v F_\ell}^2
    = \sum_{i \ge 1,\alpha: i+\card{\alpha} \le \ell}
      i c_{i,\alpha}^2
    = \sum_{i=1}^\ell i B_i(\v)^2.
\end{align*}
To finalize the proof of \Cref{lm influential matrix}, we show that we can estimate $G_\ell$ to a good accuracy using samples.
Write $F_\ell = \sum_{\alpha:\card{\alpha} \le \ell}c_\alpha H_\alpha(\x)$. We notice that for each $\alpha$ such that $\alpha_i>0$, there exists a multi-index $\gamma$ such that $\alpha = \gamma + \e_i$, where $i \in [d]$. This gives
\begin{align*}
    \partial_i F_\ell = \sum_{\gamma:\card{\gamma}\le \ell-1}\sqrt{\gamma_i+1}c_{\gamma+\e_i}H_{\gamma}(\x)
\end{align*}
This implies that
for each $i,j \in [d]$,
we have
\begin{align}\label{eq G matrix}
G_{ij} = \E_{\x \sim \cN(0,\bI)}[\partial_i F_\ell \partial_j F_{\ell}] = \sum_{\gamma}\sqrt{(\gamma_i+1)(\gamma_j+1)}c_{\gamma+\e_i}c_{\gamma+\e_j}.
\end{align}
This implies that if we can estimate each $c_\alpha = \E_{(\x,y)\sim D}[yH_\alpha(\x)]$ to error $\eps d^{-O(\gamma)}$ using some $\hat{c}_\alpha$, then we have $ \norm{\hat{G}-G} \le \eps$.
By Chebyshev inequality and union bound, we only need to use $d^{O(r+k)}\poly(1/\alpha)$ to run empirical estimation for all $c_\alpha = \E_{(\x,y)\sim D}[yH_\alpha(\x)]$.

\end{proof}

With \Cref{lm influential matrix}, the proofs of \Cref{th direction finding one shot} and \Cref{th direction finding efficient} are immediate.

\begin{proof}[Proof of \Cref{th direction finding efficient}]
By \Cref{lm influential matrix} with $\ell = r+k$, we know that using $d^{O(r+k)}\poly(1/\alpha)$ examples, we can get a matrix $\hat{G}$ such that $\norm{\hat{G}-G} \le \alpha^2/8$. Since $\abs{\E_{(\x,y)\sim D}[yh_k(\v\cdot\x)p(\w)]}>\alpha$, we know that $G_\ell$ has an eigenvalue at least $\alpha^2$ and thus $\hat{G}$ has an eigenvalue at least $7\alpha^2/8$.
Let $\u$ be the top eigenvector of $\hat{G}$ and we know that $\u^\top \hat{G} \u \ge 7\alpha^2/8$.
As a result $\u^\top G \u \ge 3\alpha^2/4$. By \Cref{lm influential matrix}, we know that there exists some $i \in [\ell]$ such that $B_i(\u) \ge \Omega(\alpha/\ell)$.
\end{proof}

\begin{proof}[Proof of \Cref{th direction finding one shot}]
    By \Cref{lm influential matrix} with $\ell$, we know that using $d^{O(\ell)}\poly(1/\beta)$ examples, we can get a matrix $\hat{G}$ such that $\norm{\hat{G}-G} \le \beta^2/d^{O(\ell)}$.
    By \eqref{eq G matrix}, since $\tr(G) \le \ell$, we know that $\tr(\hat{G}) \le 2\ell$. We take $V$ as the subspace spanned by all eigenvectors of $\hat{G}$ with an eigenvalue at least $\beta^2/100$. Then $\dim(V) \le O(\ell/\beta^2)$.
    Notice that for every $\u \perp V$, we have $\abs{\u^\top \hat{G} \u} \le \beta^2/100$, which implies $\u^\top G \u \le \beta^2$. By \Cref{lm influential matrix}, we finish the proof.
\end{proof}

\section{A Tree-Based Algorithm for Learning Intersections of PTFs}\label{sec tree building}

So far, we have already obtained all technical ingredients to obtain a learning algorithm that runs in $d^{O(k)}2^{\poly_k(1/\eps)}$ time and outputs a hypothesis with error $\eps$.
\begin{theorem}\label{th exponential time}
    Consider the problem of learning an intersection of $k$ degree-$2$ PTFs under a Gaussian distribution. There is an algorithm such that given any $\eps \in (0,1)$, it draws $d^{O(k)}2^{\poly_k(1/\eps)}$ samples from $D$, runs in $d^{O(k)}2^{\poly_k(1/\eps)}$ time and outputs a hypothesis $\hat{h}: \R^d \to \{0,1\}$ such that with probability at least $99\%$, $\err_D(\hat{h}) \le \eps$.
\end{theorem}

Recall that \Cref{th certificate} says for any instance of intersections of $k$ degree-two PTFs $f$, if there is a direction $\u$ such that $\rand_f(\u)$ is not very small, $f$ must have correlation with respect to some low degree Hermite polynomial that involves $\u$ direction.
Furthermore, \Cref{th direction finding efficient} shows that once such a direction exists, we can efficiently find one direction $\v$, along which $f$ actually has correlation.
Once we have such a direction $\v$, we consider fixing $\x \sim \cN(0,\bI)$ along $\v$ direction using tiny grids.
This can also be viewed as sampling $z\sim\cN(0,1)$ and
considering the restricted function
$f_z(\w):=f(z\v+\w)$ on $\v^\perp$.
In \Cref{lm hermite energy increment}, we show that for any fixed $j \in \N$ in expectation, $\norm{\proj_{\le j}(f^*_z(\w))}^2_2$ is at least $\norm{\proj_{\le j}(f^*(\w))}^2_2$. In fact, we show in \Cref{lm certificate to progress} that the property of $\v$ can in fact increase the low-degree Hermite energy by a non-trivial amount.

However, no matter how we fix these values, the low-degree Hermite energy cannot exceed $1$.
This implies that if we create a tree rooted at $f^*$, the ground truth hypothesis, and grow it by creating children that represent the functions, which fix $\x$ along some directions, in expectation the tree cannot be too deep, which implies that there will be at most $2^{\poly_k(1/\eps)}$ many leaves.
In other words, for all but $o(\eps)$ fraction of the leaves, there is no direction in $\R^d$ that can have a large randomized error for the function represented at the leaf. Importantly, each function at a leaf is also an instance of intersections of $k$ degree-$2$ PTFs under a Gaussian distribution.

Recall in \Cref{th error to regularity}, we proved that if for an instance of intersections of $k$ degree-$2$ PTFs, there is no direction in $\R^d$ that has a large randomized error, then the instance is close to another instance that is regular. As a result, by \Cref{th regular instance}, we can efficiently learn the function at the leaf with a high accuracy. This means by calling the direction finding algorithm, \Cref{alg direction finding} and \Cref{th regular instance} $2^{\poly_k(1/\eps)}$ times, we return a decision tree that is $\eps$ close to $f^*$.

\begin{algorithm}[htbp]
    \caption{\textsc{FiniteGridPTFLearning}}
    \label{alg finite grid learning}
    \begin{algorithmic}[1]
        \State \textbf{Input:} Sample access to $D$ with $D_{\x}=\cN(0,\bI)$ and
        $y=f^*(\x)$ for an intersection of $k$ degree-two PTFs; accuracy $\eps$ and failure probability $\delta$
        \State \textbf{Output:} A decision-tree hypothesis $\hat f:\R^d\to\{0,1\}$

\smallskip

        \State Set $\ell=2k+1$ and $\rho=\eps/10$; choose
        $\eps_{\rm reg},\eta,\eps_{\rm learn},\tau$ so that
        \Cref{th regular instance} achieves leaf error at most $\eps/2$
        whenever a $\tau$-regular intersection has error at most
        $\eps_{\rm reg}+\eta$
        \State Set $\alpha=\alpha_k(\eps_{\rm reg},\tau)$ from
        \Cref{th error to regularity}, and $\beta=\beta_k(\alpha)$ from
        \Cref{th certificate}
        \State Set $\Delta_0=\min\{\beta^2/16,\eta\}$,
        $B=\left\lceil 256k\ell^4/(\rho\beta^2\Delta_0)\right\rceil$
        and $D_{\max}=\left\lceil 128\ell^4/(\rho\beta^2)\right\rceil$
        \State Let $\theta_b=\Phi_{\rm G}^{-1}(b/B)$ for $b=1,\ldots,B-1$,
        with $\theta_0=-\infty$ and $\theta_B=+\infty$
        \State Set $L_{\max}=B^{D_{\max}}$,
        $m_{\rm mom}=(d+1)^{O(\ell)}\poly(1/\beta,\log(8L_{\max}/\delta))$
        for estimating each $G_\ell$ to operator-norm error $\beta^2/16$
        \State Let $m_{\rm reg}$ be the conditional sample budget of
        \textsc{RegularLearn} at failure probability $\delta/(8L_{\max})$,
        enlarged if necessary to be at least
        $C\eps^{-2}\log(8L_{\max}/\delta)$
        \State Initialize a tree $\mathcal T$ whose root is the active cell
        $C_\varnothing=\R^d$
        \For{$t=0,1,\ldots,D_{\max}-1$}
            \State Draw a fresh global batch $S_t$ of
            $\left\lceil C B^t(m_{\rm mom}+\log(8L_{\max}/\delta))\right\rceil$
            examples from $D$; route them to the active depth-$t$ cells
            \For{each active cell $C$ at depth $t$}
                \State Let $U_C$ be the span of the directions selected on
                the path to $C$ and let $W_C=U_C^\perp$
                \If{$W_C=\{0\}$}
                    \State Set $V_C=\{0\}$
                \Else
                    \State From
                    $\{(\operatorname{proj}_{W_C}\x,y):(\x,y)\in S_t,\,\x\in C\}$,
                    run \Cref{alg direction finding} with parameters
                    $(\ell,\beta)$, obtaining $V_C$ and, if $V_C\neq\{0\}$,
                    a direction $\v_C\in W_C$
                \EndIf
                \If{$V_C=\{0\}$}
                    \State Mark $C$ as a \emph{spectral terminal leaf}
                \Else
                    \For{$b=1,\ldots,B$}
                        \State Create an active child cell
                        $C_b=C\cap\{\x:\theta_{b-1}\le \v_C\cdot\x<\theta_b\}$
                    \EndFor
                \EndIf
            \EndFor
        \EndFor
        \State Mark all active cells at depth $D_{\max}$ as \emph{capped leaves}
        \State Draw a fresh global batch $S_{\rm leaf}$ of
        $\left\lceil C B^{D_{\max}}(m_{\rm reg}+\log(8L_{\max}/\delta))\right\rceil$
        examples and route them to the spectral terminal leaves
        \For{each spectral terminal leaf $C$}
            \If{$W_C=\{0\}$}
                \State Let $\hat h_C$ be the empirical majority label
                among the conditional samples in $C$
            \Else
                \State Apply $\textsc{RegularLearn}$ from
                \Cref{th regular instance} to the conditional samples
                $\{(\operatorname{proj}_{W_C}\x,y):
                (\x,y)\in S_{\rm leaf},\,\x\in C\}$,
                with parameters $(\tau,\eps_{\rm learn})$,
                obtaining $\hat h_C$
            \EndIf
        \EndFor
        \State Set $\hat h_C\equiv0$ on each capped leaf
        \State \textbf{Return}
        $\hat f(\x)=\hat h_{C(\x)}(\operatorname{proj}_{W_{C(\x)}}\x)$,
        where $C(\x)$ is the leaf of $\mathcal T$ containing $\x$
    \end{algorithmic}
\end{algorithm}

\begin{lemma}\label{lm hermite energy increment}
    Let $f^*(\x): \R^d \to [0,1]$ be a function. Let $\v \in \s^{d-1}$ be a unit vector and write $f^*(\x) = f^*(z\v+\w)$, where $\w = \proj_{\v^\perp}(\x)$.
    Write the $f^*(\x)$ into Hermite decomposition,
    \begin{align*}
        f^*(\x) = \sum_{k \ge 0}\sum_{\alpha}\widehat{f^*}(k,\alpha) h_k(\v\cdot\x)h_\alpha(\w).
    \end{align*}
    For each $z \in \R$, we write $f^*_z(\w) = f^*(z\v+\w)$ as a function of $\w$, when $\v\cdot\x =z$, then for every $j \in \N$, it holds
    \begin{align*}
        \E_{z \sim \cN(0,1)}[\norm{\proj_{\le j}(f^*_z(\w))}^2_2] - \norm{\proj_{\le j}(f^*(\x))}^2_2 = \sum_{\alpha: \card{\alpha} \le j}\sum_{k: k + \card{\alpha}>j} \widehat{f^*}(k,\alpha)^2,
    \end{align*}
    where for a function $f$, $\proj_{\le j}(f)$ is the projection of $f$ onto the subspace of polynomials with degree no larger than $j$ and $\norm{f}_2$ is the $L_2$ norm of $f$ under $\cN(0,\bI)$.
\end{lemma}

\begin{proof}[Proof of \Cref{lm hermite energy increment}]
    Notice that for each $z \in \R$, we notice that $f^*_z(\w) = \sum_{k \ge 0}\sum_{\alpha}\widehat{f^*}(k,\alpha) h_k(z)h_\alpha(\w)$. This implies that $\proj_{\le j}(f^*_z(\w)) = \sum_{\alpha: \card{\alpha} \le j} \left(\sum_{k \ge 0}\widehat{f^*}(k,\alpha) h_k(z)\right)h_\alpha(\w)$ and thus,
    \begin{align*}
        \norm{\proj_{\le j}(f^*_z(\w))}_2^2 = \sum_{\alpha: \card{\alpha} \le j} \left(\sum_{k \ge 0}\widehat{f^*}(k,\alpha) h_k(z)\right)^2.
    \end{align*}
By the orthogonality of the Hermite polynomials, after taking expectations over $z\sim \cN(0,1)$, we obtain that
\begin{align*}
    \E_{z \sim \cN(0,1)}[\norm{\proj_{\le j}(f^*_z(\w))}^2_2] = \sum_{\alpha: \card{\alpha} \le j} \sum_{k \ge 0}\widehat{f^*}(k,\alpha)^2.
\end{align*}
Since $\norm{\proj_{\le j}(f^*(\x)}^2_2 = \sum_{k,\alpha: k+ \card{\alpha} \le j} \widehat{f^*}(k,\alpha)^2$, we get
\begin{align*}
        \E_{z \sim \cN(0,1)}[\norm{\proj_{\le j}(f^*_z(\w))}^2_2] - \norm{\proj_{\le j}(f^*(\x))}^2_2 = \sum_{\alpha: \card{\alpha} \le j}\sum_{k: k>j-\card{\alpha}} \widehat{f^*}(k,\alpha)^2.
    \end{align*}
\end{proof}

Next, we will show that if $\E_{\x \sim \cN(0,\bI)}[f^*(\x)h_k(\v\cdot\x)p(\w)]$ is lower bounded, then we can also lower bound the increment in \Cref{lm hermite energy increment}.

\begin{lemma}\label{lm certificate to progress}
        Let $f^*(\x): \R^d \to [0,1]$ be a function. Let $\v \in \s^{d-1}$ be a unit vector and write $f^*(\x) = f^*(z\v+\w)$, where $\w = \proj_{\v^\perp}(\x)$.
    For each $z \in \R$, we write $f^*_z(\w) = f^*(z\v+\w)$ as a function of $\w$, when $\v\cdot\x =z$. Suppose there exist $k \ge 1$ and a degree-$r$ polynomial $p(\w)$ with $\norm{p(\w)}_2 \le 1$ such that $\E_{\x \sim \cN(0,\bI)}[f^*(\x)h_k(\v\cdot \x)p(\w)] \ge \beta$, for some $\beta>0$. Then, there exists some $j^* \le r$ such that
    \begin{align*}
        \E_{z \sim \cN(0,1)}[\norm{\proj_{\le j^*}(f^*_z(\w))}^2_2] - \norm{\proj_{\le j^*}(f^*(\x))}^2_2 \ge \beta^2/(r+1),
    \end{align*}
    where for a function $f$, $\proj_{\le j^*}(f)$ is the projection of $f$ onto the subspace of polynomials with degree no larger than $j^*$ and $\norm{f}_2$ is the $L_2$ norm of $f$ under $\cN(0,\bI)$.
\end{lemma}

\begin{proof}[Proof of \Cref{lm certificate to progress}]
    We write $p(\w)$ into Hermite decomposition, $p(\w) = \sum_{\alpha: \card{\alpha}\le r}\widehat{p}(\alpha)h_\alpha(\w)$. Since $\norm{p(\w)}_2 \le 1$, we have $\sum_{\alpha} \widehat{p}(\alpha)^2 \le 1$.
    Similarly, we write $f^*(\x) = \sum_{k \ge 0}\sum_{\alpha}\widehat{f^*}(k,\alpha) h_k(\v\cdot\x)h_\alpha(\w)$.
    Then, we have
    \begin{align*}
        \E_{\x \sim \cN(0,\bI)}[f^*(\x)h_k(\v\cdot \x)p(\w)] = \sum_{\alpha: \card{\alpha}\le r} \widehat{f^*}(k,\alpha) \widehat{p}(\alpha) \ge \beta.
    \end{align*}
    By Cauchy--Schwarz inequality, we have $\sum_{\alpha:\card{\alpha}\le r} \widehat{f^*}(k,\alpha)^2 \ge \beta^2$, which implies that there exists some $j^* \le r$ such that $\sum_{\alpha: \card{\alpha} = j^*} \widehat{f^*}(k,\alpha)^2 \ge \beta^2/(r+1)$.
    Apply \Cref{lm hermite energy increment} with $j = j^*$, we have
    \begin{align*}
        \E_{z \sim \cN(0,1)}[\norm{\proj_{\le j^*}(f^*_z(\w))}^2_2] - \norm{\proj_{\le j^*}(f^*(\x))}^2_2 & = \sum_{\alpha: \card{\alpha} \le j^*}\sum_{k: k>j^*-\card{\alpha}} \widehat{f^*}(k,\alpha)^2 \\
        &\ge \sum_{\alpha: \card{\alpha} = j^*} \widehat{f^*}(k,\alpha)^2 \ge \beta^2/(r+1) \;.
    \end{align*}
\end{proof}

We now analyze the finite-grid implementation of the preceding freezing
argument.
Throughout the rest of this section, set $\ell=2k+1$ and write
\[
 \Phi_\ell(f)=\frac1\ell\sum_{j=0}^{\ell-1}
             \norm{\proj_{\le j}f}_2^2.
\]
We use the influence matrix $G_\ell$ and its empirical estimate
$\hat G_\ell$ from \Cref{sec direction finding}.

Fix the $B$ intervals $I_1,\ldots,I_B$ of \Cref{alg finite grid learning},
each of Gaussian probability $1/B$, and let
$z_b=\Phi_{\rm G}^{-1}((b-1/2)/B)\in I_b$ be a representative of $I_b$.
Consider a cell $C$ at depth $t$, defined by mutually orthogonal unit
vectors $\v_1,\ldots,\v_t$ and constraints $\v_i\cdot\x\in I_{b_i}$.
Set
\[
 U_C=\operatorname{span}\{\v_1,\ldots,\v_t\},\quad
 W_C=U_C^\perp,\quad
 a_C=\sum_{i=1}^t z_{b_i}\v_i.
\]
Let $\nu_C$ be the distribution of $\proj_{U_C}\x$ conditional on
$\x\in C$, where $\x\sim\cN(0,\bI_d)$. Conditional on $\x\in C$,
the random variables $\u\sim\nu_C$ and
$\w\sim\cN(0,\bI_{W_C})$ are independent. Define
\begin{equation}
\begin{aligned}
 g_C(\w)&=\E_{\u\sim\nu_C}[f^*(\u+\w)],\\
 h_C(\w)&=f^*(a_C+\w),\\
 \Delta_C&=\Pr_{\substack{\u\sim\nu_C\\
        \w\sim\cN(0,\bI_{W_C})}}
        [f^*(\u+\w)\ne h_C(\w)].
\end{aligned}
\label{eq:sec7-grid-node}
\end{equation}
In particular, $h_C$ is an intersection of at most $k$ degree-two PTFs,
whereas $g_C$ is the conditional mean of the observed labels. At the root,
$g_{C_\varnothing}=h_{C_\varnothing}=f^*$ and $\Delta_{C_\varnothing}=0$.

\begin{lemma}[Finite-grid approximation]
\label{lm finite grid progress}
Fix a cell $C$ and a unit vector $\v\in W_C$. Put
$E=W_C\cap\v^\perp$ and $C_b=C\cap\{\x:\v\cdot\x\in I_b\}$.
For $z\in\R$ and $\w'\in E$, write
$g_{C,z}(\w')=g_C(z\v+\w')$. Then, for each $j\ge0$,
\begin{equation}
 \frac1B\sum_{b=1}^B\norm{\proj_{\le j}g_{C_b}}_2^2
 \ge \E_{Z\sim\cN(0,1)}
     \left[\norm{\proj_{\le j}g_{C,Z}}_2^2\right]
       -\frac{k}{2B}.
\label{eq:sec7-grid-energy}
\end{equation}
Furthermore,
\begin{equation}
 \frac1B\sum_{b=1}^B\Delta_{C_b}
      \le \Delta_C+\frac{2k}{B},
\label{eq:sec7-grid-error}
\end{equation}
and
\begin{equation}
 \E_{\w\sim\cN(0,\bI_{W_C})}
 [|g_C(\w)-h_C(\w)|]=\Delta_C,
 \qquad \norm{g_C-h_C}_2^2\le\Delta_C.
\label{eq:sec7-grid-L2}
\end{equation}
\end{lemma}

\begin{proof}
For fixed $\u\in U_C$ and $\w'\in E$, define
$F_{\u,\w'}(z)=f^*(\u+z\v+\w')$. Each defining quadratic has
at most two zeros as a function of $z$, unless it is identically zero.
Thus $F_{\u,\w'}$ is constant on all but at most $2k$ of the intervals
$I_1,\ldots,I_B$, up to Gaussian-null sets. Since the variance of a
Boolean random variable is at most $1/4$, we have
\begin{equation}
 \frac1B\sum_{b=1}^B
 \operatorname{Var}_{Z\sim\cN(0,1)\mid Z\in I_b}
 [F_{\u,\w'}(Z)]\le\frac{k}{2B}.
\label{eq:sec7-root-grid}
\end{equation}
By the definition of $g_C$ and Jensen's inequality,
\[
 \operatorname{Var}_{Z\sim\cN(0,1)\mid Z\in I_b}
 [g_C(Z\v+\w')]
 \le \E_{\u\sim\nu_C}
 \left[\operatorname{Var}_{Z\sim\cN(0,1)\mid Z\in I_b}
 [F_{\u,\w'}(Z)]\right].
\]
Averaging over $\w'\sim\cN(0,\bI_E)$ and $b\in[B]$ gives
\begin{equation}
 q_C(\v):=\frac1B\sum_{b=1}^B
 \E_{\w'\sim\cN(0,\bI_E)}
 \left[\operatorname{Var}_{Z\sim\cN(0,1)\mid Z\in I_b}
       [g_C(Z\v+\w')]\right]\le\frac{k}{2B}.
\label{eq:sec7-grid-variance}
\end{equation}
Let $R_j$ denote the orthogonal projection onto Gaussian Hermite
polynomials of degree at most $j$ in $E$. Since
\[
 g_{C_b}(\w')=
 \E_{Z\sim\cN(0,1)\mid Z\in I_b}[g_{C,Z}(\w')],
\]
$R_jg_{C_b}$ is the conditional mean of $R_jg_{C,Z}$ over the same
interval. The variance decomposition in the Hilbert space $L^2(E)$
therefore gives
\begin{align*}
 &\E_{Z\sim\cN(0,1)}\left[\norm{R_jg_{C,Z}}_2^2\right]
      -\frac1B\sum_{b=1}^B\norm{R_jg_{C_b}}_2^2\\
 &=\frac1B\sum_{b=1}^B
 \E_{Z\sim\cN(0,1)\mid Z\in I_b}
 \left[\norm{R_j(g_{C,Z}-g_{C_b})}_2^2\right]\\
 &\le\frac1B\sum_{b=1}^B
 \E_{\substack{Z\sim\cN(0,1)\mid Z\in I_b\\
                       \w'\sim\cN(0,\bI_E)}}
 [(g_{C,Z}(\w')-g_{C_b}(\w'))^2]
 =q_C(\v)\le\frac{k}{2B}.
\end{align*}
This proves \eqref{eq:sec7-grid-energy}.

For the approximation error, let $b(Z)$ be the index of the interval
containing $Z$. Since $\u\sim\nu_C$, $Z\sim\cN(0,1)$, and
$\w'\sim\cN(0,\bI_E)$ are mutually independent,
\begin{align*}
 \frac1B\sum_{b=1}^B\Delta_{C_b}
 &=\Pr_{\substack{\u\sim\nu_C,\ Z\sim\cN(0,1)\\\w'\sim\cN(0,\bI_E)}}[f^*(\u+Z\v+\w')
                  \ne f^*(a_C+z_{b(Z)}\v+\w')]\\
 &\le\Pr_{\substack{\u\sim\nu_C,\ Z\sim\cN(0,1)\\\w'\sim\cN(0,\bI_E)}}[f^*(\u+Z\v+\w')
                  \ne f^*(a_C+Z\v+\w')]\\
 &\quad+\Pr_{\substack{Z\sim\cN(0,1)\\\w'\sim\cN(0,\bI_E)}}[f^*(a_C+Z\v+\w')
                  \ne f^*(a_C+z_{b(Z)}\v+\w')].
\end{align*}
The first term equals $\Delta_C$. For every fixed $\w'$, the
defining polynomials that are not identically zero on the line
$\x=a_C+z\v+\w'$ have at most $2k$ roots in total.
Thus $z\mapsto f^*(a_C+z\v+\w')$ agrees with its interval
representative except possibly on at most $2k$ intervals
containing such roots, whose total Gaussian mass is at most $2k/B$.
This proves \eqref{eq:sec7-grid-error}.
Finally, since $h_C(\w)$ is Boolean,
\[
 \Pr_{\u\sim\nu_C}[f^*(\u+\w)\ne h_C(\w)]
 =|g_C(\w)-h_C(\w)|.
\]
Averaging over $\w\sim\cN(0,\bI_{W_C})$ gives the first identity
in \eqref{eq:sec7-grid-L2}, and the second follows from
$|g_C-h_C|^2\le|g_C-h_C|$.
\end{proof}

\begin{proof}[Proof of \Cref{th exponential time}]
Let $\eps_{\rm reg},\tau$ and $\alpha$ be the parameters of the
randomized-error form of \Cref{th error to regularity}, and write
$\beta=\beta_k(\alpha)$ for the correlation bound of
\Cref{th certificate}. Let $\eta>0$ be chosen so that the regular-case
learner of \Cref{th regular instance} learns conditional labels to
error at most $\eps/2$ whenever the labels disagree with some
$\tau$-regular intersection with probability at most
$\eta+\eps_{\rm reg}$. This is the label-robustness guarantee
required for the use of that learner. Put
\[
 \rho=\eps/10,\qquad
 \Delta_0=\min\{\beta^2/16,\eta\},\qquad
 B=\left\lceil\frac{256k\ell^4}{\rho\beta^2\Delta_0}\right\rceil,
 \qquad
 D_{\max}=\left\lceil\frac{128\ell^4}{\rho\beta^2}\right\rceil.
\]
Algorithm~\ref{alg finite grid learning} makes the following
spectral decision at an active cell $C$.
If $W_C=\{0\}$ or the subspace $V_C$ returned by
\Cref{alg direction finding} is $\{0\}$, the algorithm
stops at $C$. Otherwise, it splits $C$ along the
returned direction $\v_C$ using the intervals
$I_1,\ldots,I_B$.

\emph{Potential progress and tree depth.}
At every internal node $C$, we have $V_C\neq\{0\}$.
By the spectral threshold in \Cref{alg direction finding}
and the estimation guarantee
$\|\widehat G_\ell-G_\ell\|_{\op}\le\beta^2/16$,
the returned direction $\v_C$ satisfies
$
\v_C^\top G_\ell\v_C\ge 3\beta^2/16.
$
By \Cref{lm influential matrix}, there exist an integer $s\in[\ell]$
and a polynomial $p$ on $W_C\cap\v_C^\perp$
such that
\begin{equation}
 s+\deg(p)\le\ell,\quad \norm{p}_2=1,\quad
 \E_{\w\sim\cN(0,\bI_{W_C})}
 [g_C(\w)h_s(\v_C\cdot\w)p(\proj_{\v_C^\perp}\w)]
 \ge\frac\beta{4\ell}.
\label{eq:sec7-certificate-from-matrix}
\end{equation}

Apply \Cref{lm hermite energy increment} and \Cref{lm certificate to progress}
to $g_C$ and the certificate
\eqref{eq:sec7-certificate-from-matrix}. It supplies
$j^*\le\deg(p)\le\ell-1$ such that
\begin{equation}
 \E_{Z\sim\cN(0,1)}
 [\norm{\proj_{\le j^*}g_{C,Z}}_2^2]
 -\norm{\proj_{\le j^*}g_C}_2^2
 \ge\frac{\beta^2}{16\ell^3}.
\label{eq:sec7-progress-one-degree}
\end{equation}
For every other $j\in\{0,\ldots,\ell-1\}$, \Cref{lm hermite energy increment} gives a
nonnegative exact-freezing increment. Applying
\eqref{eq:sec7-grid-energy} of \Cref{lm finite grid progress} for every such $j$
and averaging over $j$ yields
\begin{align}
 \frac1B\sum_{b=1}^B\Phi_\ell(g_{C_b})-\Phi_\ell(g_C)
 &\ge\frac{\beta^2}{16\ell^4}-\frac{k}{2B}
 \ge\frac{\beta^2}{32\ell^4}=: \mu.
\label{eq:sec7-split-progress}
\end{align}
The last inequality holds by the chosen value of $B$.

For every node, put
$\pi_C=\Pr_{\x\sim\cN(0,\bI_d)}[\x\in C]$.
Each child has mass $\pi_{C_b}=\pi_C/B$, because the splitting
direction is orthogonal to all earlier splitting directions on its
path. Let $\mathcal I$ and $\mathcal L$ be the internal nodes and
leaves, respectively. For any real-valued function $A$ on the nodes,
finite-tree telescoping gives
\begin{equation}
 \sum_{C\in\mathcal L}\pi_C A(C)-A(C_\varnothing)
 =\sum_{C\in\mathcal I}\pi_C
 \left(\frac1B\sum_{b=1}^BA(C_b)-A(C)\right).
\label{eq:sec7-tree-telescope}
\end{equation}
Since $0\le\Phi_\ell(g_C)\le1$, applying
\eqref{eq:sec7-tree-telescope} to $A(C)=\Phi_\ell(g_C)$ and using
\eqref{eq:sec7-split-progress} gives
\begin{equation}
 \sum_{C\in\mathcal I}\pi_C\le\frac1\mu
  =\frac{32\ell^4}{\beta^2}.
\label{eq:sec7-total-internal-mass}
\end{equation}
A fresh Gaussian point reaching depth $D_{\max}$ passes through at
least $D_{\max}$ internal nodes. Thus, by Markov's inequality,
\begin{equation}
 \Pr_{\x\sim\cN(0,\bI_d)}
 [\x\text{ reaches a capped leaf}]
 \le\frac{32\ell^4}{D_{\max}\beta^2}\le\rho/4.
\label{eq:sec7-capped-mass}
\end{equation}

\emph{Accumulated gridding error.}
Using \eqref{eq:sec7-tree-telescope} with $A(C)=\Delta_C$,
noting $\Delta_{C_\varnothing}=0$, and applying
\eqref{eq:sec7-grid-error} and
\eqref{eq:sec7-total-internal-mass}, we obtain
\[
 \sum_{C\in\mathcal L}\pi_C\Delta_C
 \le\frac{2k}{B}\sum_{C\in\mathcal I}\pi_C
 \le\frac{64k\ell^4}{B\beta^2}.
\]
In particular,
\begin{equation}
 \Pr_{\x\sim\cN(0,\bI_d)}[\Delta_{C(\x)}>\Delta_0]
 \le\frac{64k\ell^4}{B\beta^2\Delta_0}
 \le\rho/4,
\label{eq:sec7-bad-grid-mass}
\end{equation}
where $C(\x)$ is the leaf containing $\x$. By
\eqref{eq:sec7-capped-mass} and \eqref{eq:sec7-bad-grid-mass},
a fresh Gaussian point reaches a noncapped leaf with
$\Delta_C\le\Delta_0$ except with probability at most $\rho/2$.

\emph{Regularity at a good terminal cell.}
Fix such a leaf $C$ with $W_C\ne\{0\}$. By
\eqref{eq:sec7-grid-L2},
\[
 \norm{g_C-h_C}_2\le\sqrt{\Delta_C}\le\beta/4,
 \qquad\Delta_C\le\eta.
\]
Suppose some unit vector $\u\in W_C$ satisfies
$\rand_{h_C}(\u)\ge\alpha$. By
\Cref{th certificate}, there are $s\in[\ell]$ and a polynomial
$p$ with $\norm{p}_2=1$ and $s+\deg(p)\le\ell$ such that
\[
 \E_{\w\sim\cN(0,\bI_{W_C})}
 [h_C(\w)h_s(\u\cdot\w)p(\proj_{\u^\perp}\w)]\ge\beta.
\]
Cauchy--Schwarz implies that the same correlation with $g_C$ in
place of $h_C$ is at least $3\beta/4$.
On the other hand, since $C$ is a spectral terminal
leaf with $W_C\neq\{0\}$, Algorithm 3 returned
$V_C=\{0\}$. By the spectral stopping condition
and the accuracy of $\widehat G_\ell$, for every
unit vector $\u\in W_C$,
$
\u^\top G_\ell\u\le 5\beta^2/16.
$
By \Cref{lm influential matrix}, for every $s\ge 1$ and polynomial
$p$ with $\|p\|_2=1$ and $s+\deg(p)\le\ell$,
\[
\left|
\E_{\w\sim\cN(0,\bI_{W_C})}
[g_C(\w)h_s(\u\cdot\w)
p(\operatorname{proj}_{\u^\perp}\w)]
\right|
\le \frac{\sqrt{5}\beta}{4}
<\frac{3\beta}{4}.
\]
This gives a contradiction. Hence,
\begin{equation}
 \rand_{h_C}(\u)<\alpha\qquad
 \text{for every unit }\u\in W_C.
\label{eq:sec7-small-rand-leaf}
\end{equation}
If $W_C=\{0\}$, this assertion is vacuous and $h_C$ is constant.

By the randomized-error form of \Cref{th error to regularity},
$h_C$ is $\eps_{\rm reg}$-close to a $\tau$-regular intersection
$\widetilde h_C$ under $\cN(0,\bI_{W_C})$. Moreover,
\begin{align*}
 \Pr_{\substack{\u\sim\nu_C\w\sim\cN(0,\bI_{W_C})}}
 [f^*(\u+\w)\ne\widetilde h_C(\w)]
 \le\Delta_C+\eps_{\rm reg}
 \le\eta+\eps_{\rm reg}.
\end{align*}
Thus the stipulated robust regular-case learner returns a hypothesis
of conditional error at most $\eps/2$ at every good terminal cell.
The exceptional cells have combined Gaussian probability at most
$\rho/2$, so the resulting decision tree has error at most
$\eps/2+\rho/2<\eps$.

\emph{Sample complexity.}
At a fixed cell $C$, conditioned on its earlier splitting directions,
$\proj_{W_C}\x\sim\cN(0,\bI_{W_C})$ under
$D\mid\{\x\in C\}$. The moment-estimation guarantee of
\Cref{lm influential matrix} therefore applies to these conditional samples.
With fresh independent sample batches for each depth and a union bound
over at most $B^{D_{\max}}$ nodes, all calls of \Cref{alg direction finding} succeed with probability at least
$1-\delta/2$. Every depth-$t$ cell has Gaussian probability $B^{-t}$,
so a global batch of size $B^t$ times the number of samples required
per node (up to logarithmic factors) supplies the needed conditional
samples. A final fresh batch gives the samples for the regular-case
learner at every terminal cell. Allocating failure probability
$\delta/(2B^{D_{\max}})$ per learner call and taking a union bound
gives success probability at least $1-\delta$. The total running time
is bounded by
\[
 B^{D_{\max}}
 \left((d+1)^{O(\ell)}\operatorname{poly}
     \left(\beta^{-1},\log\frac{B^{D_{\max}}}{\delta}\right)
     +T_{\rm reg}\right),
\]
where $T_{\rm reg}$ is the cost of the regular-case learner per leaf.
With $\ell=2k+1$ and the parameter bounds required by
\Cref{th error to regularity,th regular instance}, this is
$d^{O(k)}2^{\operatorname{poly}_k(1/\eps)}$.
\end{proof}

\section{Polynomial Approximation with Hinge Loss}\label{sec hinge loss}
In this section, we give an algorithm for learning intersections
of degree-$2$ PTFs supported on a known lifted subspace by minimizing
hinge loss over bounded-norm polynomials. The main technical lemma
establishes the existence of a low-degree polynomial with small
expected hinge loss against the target intersection and controlled
Gaussian $L_2$ norm, while preserving the given lifted subspace.
This polynomial provides a good comparator for convex optimization,
and its norm bound controls the additional loss caused by label
disagreement. The resulting learning guarantee is stated in the
following theorem.
For fixed $k$, the running time is polynomial in the ambient dimension and quasipolynomial in the subspace dimension and inverse accuracy.
\begin{theorem} [Learning in a known lifted subspace] \label{thm:hinge-main}
Let $d,k\in \Z_+$, $\eps\in (0,1/2)$ and $L=k^2\log^3(k)\log^2(1/\eps)$.
Let \(\Phi_2:\R^d\to\R^N\), where \(N=d(d+3)/2\), consist of the normalized Gaussian Hermite polynomials of degrees one and two, and let \(S\subseteq\R^N\) be a subspace of dimension \(n\).
Suppose \(D\) is a joint distribution of $(\x,y)$ over \(\R^d\times\{\pm1\}\) with marginal \(D_\x=\cN(0,\bI_d)\), and there exists an intersection
$
f(\x)=\bigwedge_{i=1}^k
\sign\!\bigl(c_i+\u_i\cdot\Phi_2(\x)\bigr)$, where $c_i\in\R$ and $\u_i\in S$ such that
$\Pr_{(\x,y)\sim D}[f(\x)\ne y]\le 2^{-CL}$.

There is an algorithm based on hinge-loss minimization over
bounded-norm polynomials with the following guarantee.
Given $d,k,\eps$, a basis for $S$, and i.i.d.\ sample access to $D$,
the algorithm has sample complexity and running time
at most
    $d^{O(1)}
    n^{O(k^2\log^2(2k)\log(1/\eps))}
    2^{O(k^2\log^3(2k)\log^2(1/\eps))}$
and, with probability at least $2/3$, outputs a hypothesis
$h:\R^d\to\{\pm1\}$ satisfying
$\Pr_{(\x,y)\sim D}[h(\x)\neq y]\le\eps$.

\end{theorem}

The high-level intuition is as follows. For convenience, first consider
the case $S=\R^N$. We minimize the expected hinge loss
$\Hinge(y,y')=\max\{0,1-yy'\}$ over polynomials of degree at most
$m$ and Gaussian $L_2$ norm at most $W$, where $m$ and $W$ are chosen
according to the technical lemma below. This objective is convex in the
polynomial coefficients. The lemma guarantees the existence of a
polynomial with small hinge loss against the target intersection, and
its norm bound controls the additional loss caused by the label
disagreement allowed in \Cref{thm:hinge-main}. Stochastic convex
optimization then finds a polynomial $p$ with small expected hinge loss.
Since hinge loss upper-bounds the classification error of
$h(\x)=\sign(p(\x))$, thresholding at zero gives the desired hypothesis.

Using the normalized Hermite basis, this optimization has runtime and
sample complexity polynomial in $M,W,1/\eps$, where
$M=\binom{d+m}{m}$ is the dimension of the space of polynomials of degree
at most $m$. When a subspace $S\subseteq\R^N$ is given, the technical
lemma further guarantees that the approximating polynomial can be
written as
\[
    p(\x)=P\bigl(\proj_S\Phi_2(\x)\bigr),
    \qquad \deg(P)\le m/2.
\]
Writing $n=\dim(S)$, we can therefore restrict the search to a space
spanned by at most $\binom{n+m/2}{m/2}$ monomial features. Together with
bounds on the coefficients and feature moments, this yields a runtime
that is polynomial in $d$ and, for fixed $k$, quasipolynomial in $n$
and $1/\eps$, as stated in \Cref{thm:hinge-main}.

We now state the main technical lemma, which establishes the existence
of the required polynomial, bounds its degree and Gaussian $L_2$ norm,
and shows that it preserves the given lifted subspace.

\begin{lemma} \label{lem:polynomial-low-hinge-loss}
Let $f:\R^d\to\{\pm1\}$ be an intersection of $k$
degree-$2$ PTFs of the form $f(\x)=\bigwedge_{i=1}^k \sign(p_i(\x))$.
Then, for any $\eps\in(0,1/2)$, there exist an even integer
$m$ and a polynomial $g$ of degree at most $m$ with
$
\E_{\x\sim\cN(0,\bI_d)}
\big[\Hinge(f(\x),g(\x))\big]
\le \eps$.
Moreover, $g$ satisfies the following properties:
\begin{enumerate}
    \item \textbf{Degree and norm bounds.}
    We have
    \[
    m=O\!\left(k^2\log^2(2k)\log(1/\eps)\right),
    \qquad
    \norm{g}_{L_2(\cN(0,\bI_d))}
    \le
    2^{O(k^2\log^3(2k)\log^2(1/\eps))}.
    \]
    Furthermore, for every $\delta\in[0,1]$ and every measurable
    weight $a:\R^d\to[0,1]$ with
    $\E_{\x\sim\cN(0,\bI_d)}[a(\x)]\le\delta$,
    \[
    \E_{\x\sim\cN(0,\bI_d)}
    \big[|a(\x)g(\x)|\big]
    \le
    \sqrt{\delta}\,\operatorname{quasipoly}_k(1/\eps).
    \]

    \item \textbf{Preservation of the lifted subspace.}
    If all $p_i$ are of the form
$p_i(\x)=c_i+\u_i\cdot \Phi_2(\x)$
for $\u_i\in S$, where $S\subseteq\R^N$ is a subspace.
    Then $g$ can be chosen in the form $g(\x)=p(\proj_S\Phi_2(\x))$, $p$ is a polynomial of degree at most $m/2$.
\end{enumerate}
\end{lemma}

Our construction builds on the rational-function approach to constructing a sign-matching polynomial for intersection of halfspaces with margins in \cite{KOS:04,KS04}.
We adapt this approach to intersections of degree-$2$ PTFs under Gaussian marginals, without assuming a uniform positive margin.
We retain the strategy of approximating the sign function by rational functions and rescaling by the denominators to obtain a polynomial.
We use additional analysis to bound the expected hinge loss near the threshold boundaries and in the Gaussian tails, and Gaussian ($L_2$) norm of the rescaled polynomial.

The motivation is that directly approximating the value of sign function and then build the intersection incurs an $\Omega(1/\eta)$ degree dependence on the boundary width $\eta$. Such construction does not leverage the fact that we only need sign matching. To avoid this, we approximate the individual sign values by rational functions (which has $O(\log(1/\eta))$ dependence), combine them to obtain a rational function whose sign agrees with the intersection away from the threshold boundaries, and then clear the denominators by rescaling.
If this denominator is always positive, this scaling yields a polynomial with the same sign.
We subsequently rescale it to obtain margin on a high-probability region and control its hinge loss elsewhere.

We first give the following fact on rational approximation of sign functions.
\begin{fact} [Rational approximation of the sign function] \label{fct:rational-approximate-sign}
    For any boundary width $\eta\in (0,1/2)$ and approximation error $\eps\in (0,1/2)$, let
    $m = \left\lceil c\log(1/\eta) \right\rceil$ and
    $s = \left\lceil c\log(1/\eps) \right\rceil$,
    where $c$ is a sufficiently large universal constant
    and define
    \(P_m(t) = \prod_{j=0}^{m} (t + 2^{-j})\),
    \[
    N(t) = P_m(t)^{2s} - P_m(-t)^{2s},
    \qquad
    D(t) = P_m(t)^{2s} + P_m(-t)^{2s}.
    \]
    Each has degree at most $O(\log(1/\eta)\log(1/\eps))$
    and the rational function $r(t)=N(t)/D(t)$ satisfies:
    \begin{enumerate}
        \item {\bf Positive denominator:} For all $t\in \R$, $D(t)>0$.
        \item {\bf Boundedness and sign preservation:} For all $t\in \R$, $|r(t)|\leq 1$ and $\sign(r(t))= \sign(t)$.
        \item {\bf Approximation away from 0:} For $|t|\in [\eta,1]$, $|\sign(t)-r(t)|\leq \eps$.
    \end{enumerate}
\end{fact}
\begin{proof}[Proof of \Cref{fct:rational-approximate-sign}]
Both $N$ and $D$ have degree at most $2s(m+1)$, which is
$O(\log(1/\eta)\log(1/\eps))$.

For the first property, notice that roots of $P_m(t)$ are all negative, while roots of $P_m(-t)$ are all positive. Therefore, $P_m(t)$ and $P_m(-t)$ have no common roots and $D(t)$ must be always positive.

For the second property, $P_m(t)^{2s}$ and $P_m(-t)^{2s}$
are nonnegative, so $|N(t)|\le D(t)$ and hence
$|r(t)|\le 1$ for all $t$.
For $t>0$, every factor in $P_m$ satisfies
$|t-2^{-j}|<t+2^{-j}$. Thus $|P_m(-t)|<P_m(t)$ and $N(t),r(t)\neq 0$ for any $t\neq 0$.
Furthermore, since $N$ is odd and $D$ is even,
we must have $\sign(r(t))= \sign(t)$.

It only remains to prove the approximation property.
For $t>0$, the goal of the construction is to have $|P_m(-t)^{2s}|$ much smaller
relative to $P_m(t)^{2s}$, so that $r(t)$ is close to $1$. Notice that this also suffices for the purpose of the negative side.
At a fixed scale $a>0$, we have
\[
    \frac{|a-t|}{a+t}\le\frac13
    \qquad\text{when }a\le t\le2a,
\]
and this ratio is at most $1$ for every $t>0$.
Thus one factor gives a constant separation over $[a,2a]$.
By multiplying over dyadic scales, we cover the entire desired interval
without worsening the separation.
Using this to construct $P_m$, we get for any $t\in [2^{-m},1]$,
\[
    \frac{|P_m(-t)|}{P_m(t)}
    =\prod_{j=0}^{m}\frac{|2^{-j}-t|}{2^{-j}+t}
    \le\frac13.
\]
We then take it to the power of $2s$ to amplify this separation, this gives $|P_m(-t)^{2s}|/|P_m(t)^{2s}|\leq 1/3^{2s}$. Therefore,
\[
|1-r(t)|=\left |\frac{2P_m(-t)^{2s}}{P_m(-t)^{2s}+P_m(t)^{2s}}\right |\leq \frac{2}{3^{2s}+1}\; .\]
By choosing $c$ be a sufficiently large universal constant in $m = \left\lceil c\log(1/\eta) \right\rceil$ and
$s = \left\lceil c\log(1/\eps) \right\rceil$ gives $|1-r(t)|\leq\eps$ for $t\in [\eta,1]$.
Then the negative side follows the exact same argument.
\end{proof}

Given the rational function from the preceding fact, one can easily scale to construct a rational function \(r(t)=N(t)/D(t)\)
with approximation error \(\eps\le 1/(2k)\) over the interval $[-B,-\eta B]\cup[\eta B,B]$.
Then it is immediate that for the intersection of degree-$2$ PTFs $f(\x)=\bigwedge_{i=1}^k\sign(p_i(\x))$.
We can construct a sign matching polynomial for $f$ by first having the following rational
\[
g(\x)=\sum_{i=1}^k r(p_i(\x))-(k-1)=\sum_{i=1}^k \frac{N(p_i(\x))\Pi_{j\in [k]\backslash {i}}D(p_j(\x))}{\Pi_{j\in [k]}D(p_j(\x))}-(k-1) \; .
\]
Since the preceding fact has shown that $D(t)$ is always positive, we can simply have the sign matching polynomial as $p(\x)=g(\x)\Pi_{j\in [k]}D(p_j(\x))$. It remains to check that 1. its hinge loss is bounded near the decision boundaries; 2. it is sufficiently large for any point sufficiently far from the decision boundary, so the overall Hinge can be upper bounded; and 3. the Hinge loss incurred outside the decision box is small due to Gaussian decaying tail. We first give the following fact on the sign matching polynomial.
\begin{fact} \label{fct:polynomial-sign-matching}
Let $f:\R^k\to \{\pm 1\}$ be an intersection of $k$ sign functions of the form $f(\x)=\bigwedge_{i=1}^k\sign(x_i)$.
For any $\eta\in (0,1/2)$, there exists a polynomial $q:\R^k\to \R$ of degree at most $O(k^2\log(k)\log(1/\eta))$ of the form $q(\x)=\sum_\alpha c_\alpha\x^{\alpha}$
where $\x^{\alpha}$ are monomials and $\sum_\alpha |c_\alpha|=2^{O(k^2\log(k)\log(1/\eta)^2)}$ such that
\begin{enumerate}
    \item \textbf{Unit margin sign matching away from 0:}
    If $\eta\le |x_i|\le 1$ for every $i$, then $f(\x)q(\x)\ge 1$.

    \item \textbf{Bounded pointwise hinge loss:}
    If $|x_i|\le 1$ for every $i$, then
    $\max\{0,1-f(\x)q(\x)\}\le 2$.
\end{enumerate}
\end{fact}
\begin{proof}[Proof of \Cref{fct:polynomial-sign-matching}]
Set
\[
    L=\left\lceil\log_2(1/\eta)\right\rceil,\qquad
    \gamma=2^{-L},\qquad
    \ell=4kL,\qquad
    \delta=2^{-\ell},
\]
and let
    $\rho=\frac1{4k},
    \qquad
    s=\left\lceil\log_9(8k)\right\rceil$.
We use the explicit rational construction $r=N/D$ from
\Cref{fct:rational-approximate-sign} with product-depth parameter
$\ell$ and power parameter $s$.
Thus the boundary width and approximation error used there are
$\delta$ and $\rho$, respectively.
The estimate proved in \Cref{fct:rational-approximate-sign} gives
\[
    |r(t)-\sign(t)|
    \le\frac{2}{9^s+1}\le\rho
    \qquad\text{for }\delta\le|t|\le1.
\]

Define the rational separator
\[
    R(\x)=\sum_{i=1}^k r(x_i)-(k-1).
\]
For Boolean inputs, the corresponding sum is $1$ when all inputs
are positive and at most $-1$ otherwise.
The approximation changes this sum by at most $k\rho=1/4$, so
\[
    f(\x)R(\x)\ge\frac12
    \qquad\text{whenever }\delta\le|x_i|\le1
    \text{ for every }i.
\]
Also, $|r|\le1$ gives $|R|\le2k-1$ everywhere.

Clear the positive denominator and normalize:
\[
    S(\x)=R(\x)\prod_{i=1}^kD(x_i),
    \qquad
    \lambda=\frac12P_\ell(\gamma)^{2sk},
    \qquad
    q(\x)=\frac{S(\x)}{\lambda}.
\]
Then $q$ is a polynomial of degree at most
    $2ks(\ell+1)
    =O\!\left(k^2\log(2k)\log(1/\eta)\right)$.

We now reduce the two required properties to denominator estimates.
Since $D$ has only even powers with nonnegative coefficients,
it is even and nondecreasing on $[0,\infty)$.
We will show
\begin{equation}\label{eq:sign-matching-denominator-bounds}
    D(\gamma)^k\ge2\lambda,
    \qquad
    D(\delta)D(1)^{k-1}\le\frac{\lambda}{2k-1}.
\end{equation}
The first bound ensures a sufficiently large denominator product
when every coordinate is away from zero.
The second ensures a sufficiently small product when one coordinate
is within width $\delta$, even if all other coordinates are as large
as $1$.

For the first bound, directly from the construction,
    $D(\gamma)\ge P_\ell(\gamma)^{2s}$,
and hence $D(\gamma)^k\ge2\lambda$.
For the second bound, let
    $C_0=\prod_{j=0}^{\infty}(1+2^{-j})<e^2<8$.
For $|t|\le1$, we have $|P_\ell(\pm t)|\le C_0$.
For $|t|\le\delta$, the smaller scale gives
\[
    |P_\ell(\pm t)|
    \le P_\ell(0)\prod_{j=0}^{\ell}(1+2^{j-\ell})
    \le C_0P_\ell(0).
\]
Therefore,
\[
    D(1)\le2C_0^{2s},
    \qquad
    D(\delta)\le2C_0^{2s}P_\ell(0)^{2s}.
\]
Moreover,
\[
    P_\ell(0)=2^{-\ell(\ell+1)/2},
    \qquad
    P_\ell(\gamma)
    \ge
    \left(\prod_{j=0}^{L-1}2^{-j}\right)\gamma^{\ell-L+1}
    =2^{-L\ell+L(L-1)/2}.
\]
Using $\ell=4kL$, these estimates imply
\[
    \frac{P_\ell(0)^{2s}}{P_\ell(\gamma)^{2sk}}
    \le
    2^{-s((8k^2+k)L^2+3kL)}.
\]
Consequently,
\begin{align*}
    \frac{(2k-1)D(\delta)D(1)^{k-1}}{\lambda}
    &\le
    2(2k-1)(2C_0^{2s})^k
    \frac{P_\ell(0)^{2s}}{P_\ell(\gamma)^{2sk}}\\
    &\le
    2^{9sk-s((8k^2+k)L^2+3kL)}
    \le1.
\end{align*}
Here we used $2(2k-1)(2C_0^{2s})^k\le2^{9sk}$, and the last
inequality holds for $k,L,s\ge1$.
This proves \eqref{eq:sign-matching-denominator-bounds}.

We can now deduce the two properties.
For Property~1, suppose $\eta\le|x_i|\le1$ for every $i$.
Since $\delta<\gamma\le\eta$, monotonicity and the first bound
in \eqref{eq:sign-matching-denominator-bounds} give
\[
    f(\x)q(\x)
    =\frac{f(\x)R(\x)}{\lambda}\prod_iD(x_i)
    \ge\frac{D(\gamma)^k}{2\lambda}
    \ge1.
\]

For Property~2, take any $\x\in[-1,1]^k$.
If $f(\x)q(\x)\ge0$, its hinge loss is at most $1$.
Otherwise, positivity of the denominator product implies
$f(\x)R(\x)<0$, so some $|x_i|<\delta$.
The second bound in \eqref{eq:sign-matching-denominator-bounds}
then gives
\[
    |q(\x)|
    \le
    \frac{(2k-1)D(\delta)D(1)^{k-1}}{\lambda}
    \le1.
\]
Thus its hinge loss is at most $1+|q(\x)|\le2$.

Finally, we bound the monomial coefficients of $q$.
Both $N$ and $D$ have nonnegative coefficients: their definitions
retain the odd and even coefficients of $P_\ell^{2s}$,
respectively, each multiplied by $2$.
Their coefficient sums are therefore $N(1)$ and $D(1)$.
Expanding
\[
    q(\x)=\frac1\lambda\left(
        \sum_iN(x_i)\prod_{j\ne i}D(x_j)
        -(k-1)\prod_iD(x_i)\right)
\]
and using $N(1)\le D(1)$, we obtain
\[
    \sum_\alpha|c_\alpha|
    \le
    \frac{kN(1)D(1)^{k-1}+(k-1)D(1)^k}{\lambda}
    \le
    \frac{(2k-1)D(1)^k}{\lambda}.
\]
The preceding lower bound on $P_\ell(\gamma)$ gives
    $\lambda^{-1}\le2^{1+8sk^2L^2}$,
while $D(1)^k\le(2C_0^{2s})^k=2^{O(sk)}$.
Hence
\[
    \sum_\alpha|c_\alpha|
    \le2^{O(sk^2L^2)}
    =
    2^{O(k^2\log(2k)\log^2(1/\eta))},
\]
as required.
\end{proof}
We now use \Cref{fct:polynomial-sign-matching} to prove
\Cref{lem:polynomial-low-hinge-loss}. Applying the fact to the rescaled
defining polynomials $\{p_i/B\}_{i\in [k]}$ for the intersection of PTFs, we obtain a polynomial $q$ whose hinge loss
is zero whenever $\eta B\le |p_i(\x)|\le B$ for every $i$, by
Property~1. It remains to control the contribution from points (i.e. $\{p_i/B\}_{i\in [k]}$) near
the threshold boundaries and points in the tails.

Within the region where all $|p_i(\x)|\le B$, Property~2 bounds the
hinge loss by $2$. Thus, the near-boundary contribution is controlled
by its probability, which we bound using Gaussian anticoncentration.
For the remaining tail region, we combine upper bound on the
size of the constructed polynomial with Gaussian tail decay.
With appropriate parameter choice of both $B$ and $\eta$, the contribution from both regions will be sufficiently small.
\begin{proof}[Proof of \Cref{lem:polynomial-low-hinge-loss}]
The proof follows the following structure: We first choose the parameter $\eta$ and $B$ for rescaling the polynomial from \Cref{fct:polynomial-sign-matching}.
Then as we discussed,
Property 1 of \Cref{fct:polynomial-sign-matching} already ensures the Hinge loss is 0 if $\eta B\le |p_i(\x)|\le B$ for all $i$. We then show that for the near decision boundary case, i.e., there exists $i$ such $p_i(\x)\in [-\eta B,\eta B]$.
Since we have Property 2 of \Cref{fct:polynomial-sign-matching}, it suffices for us to bound $\Pr_{\x\sim \cN(0,\bI_d)} [\exists i,p_i(\x)\in [-\eta B,\eta B]]$ from the anticoncentration.
The other case that $p_i(\x)\not\in [-B,B]$ can be analyzed from the Gaussian tail bound and the polynomial $L_2$ norm. After that, we can simply put these together to bound the Hinge loss.

\paragraph{Step 1: Choose $\eta,B$ and apply \Cref{fct:polynomial-sign-matching} to construct the polynomial.}
We WLOG normalize the defining polynomials to satisfy
$\norm{p_i}_{L_2(\cN(0,\bI_d))}=1$.
Let $c$ be a sufficiently large universal constant and choose
\[
    \eta=\left(\frac{\eps}{k}\right)^C,
    \qquad
    B=Ck^2\log(k)^3\log^2(1/\eta).
\]
Let $q$ be the polynomial from
\Cref{fct:polynomial-sign-matching} and define
\[
    g(\x)=q\left(\frac{p_1(\x)}{B},\ldots,
                       \frac{p_k(\x)}{B}\right).
\]
The choice of $\eta$ makes the boundary region small, while $B$ is
chosen large enough that Gaussian tail decay dominates the possible
growth of $g$.

\paragraph{Step 2: Control the contribution near the decision
boundaries.}
Let ${\cal B}={\x:\exists i, p_i(\x)\in [-\eta B,\eta B]}$.
From Carbery--Wright anticoncentration and union bound, we must have that
\[
\Pr_{\x\sim \cN(0,\bI_d)}[\x\in {\cal B}]\leq \sum_i\Pr_{\x\sim \cN(0,\bI_d)}[p_i(\x)\in [-\eta B,\eta B]]=O(k\sqrt{B\eta})\; .
\]
Using the second property of \Cref{fct:polynomial-sign-matching} directly implies that
\[
\E_{\x\sim \cN(0,\bI_d)}[\Hinge(f(\x),g(\x))\mathbf{1}_{\mathcal B}(\x)]=O(k\sqrt{B\eta})\; .
\]
\paragraph{Step 3: Control the contribution from the tails.}
Let
    $\mathcal{T}=\{\x:\max_i|p_i(\x)|>B\}$
denote the tail region. We first bound the $L_2$ norms of $g$ and
$\mathbf{1}_{\mathcal{T}}$, then combine them to control the hinge
contribution from this region.

Since each $p_i$ has degree at most $2$ and $\norm{p_i}_2=1$,
hypercontractivity gives
\[
    \norm{\frac{p_i}{B}}_u
    \le\frac{u-1}{B},
    \qquad u\ge2.
\]
Taking a power requires a higher moment: for every integer $a\ge1$,
\[
    \norm{\left(\frac{p_i}{B}\right)^a}_2
    =
    \norm{\frac{p_i}{B}}_{2a}^{a}
    \le
    \left(\frac{2a-1}{B}\right)^a.
\]
More generally, for a monomial of total degree $r=|\alpha|\ge1$,
H\"older's inequality gives
\[
    \norm{\prod_i(p_i/B)^{\alpha_i}}_2
    \le
    \prod_i\norm{\frac{p_i}{B}}_{2r}^{\alpha_i}
    \le
    \left(\frac{2r-1}{B}\right)^r
    \le1,
\]
where the last inequality uses $r\le\deg q$ and $B\ge2\deg q$.
The constant monomial also has norm $1$.

Write
    $q(\z)=\sum_\alpha c_\alpha\z^\alpha$.
By the triangle inequality, the preceding bound on each
substituted monomial, and the coefficient bound in
\Cref{fct:polynomial-sign-matching}, we have
\begin{align*}
    \norm{g}_2
    \le
    \sum_\alpha |c_\alpha|
        \norm{\prod_i(p_i/B)^{\alpha_i}}_2
    \le
    \sum_\alpha |c_\alpha|
    \le
    2^{O(k^2\log(2k)\log^2(2k/\eta))}.
\end{align*}

Next, the squared $L_2$ norm of the indicator is exactly the tail
probability. Hypercontractivity and Markov's inequality give
$\Pr[|p_i(\x)|>B]\le e^{-cB}$ for a universal constant $c>0$.
Thus a union bound yields
\[
    \norm{\mathbf{1}_{\mathcal{T}}}_2^2
    =\Pr[\x\in\mathcal{T}]
    \le\sum_{i=1}^k\Pr[|p_i(\x)|>B]
    \le ke^{-cB}.
\]

We can now bound the $L_1$ norm of $g$ restricted to the tail
region. By Cauchy--Schwarz,
\[
    \norm{g\mathbf{1}_{\mathcal{T}}}_1
    \le\norm{g}_2\norm{\mathbf{1}_{\mathcal{T}}}_2
    \le\sqrt{k}\,\norm{g}_2e^{-cB/2}.
\]
Since $\Hinge(f,g)\le1+|g|$, the hinge contribution
from this region satisfies
\begin{align*}
\E\left[
        \Hinge(f(\x),g(\x))
        \mathbf{1}_{\mathcal{T}}(\x)
    \right]
    \le
        \Pr[\x\in\mathcal{T}]
        +\norm{g\mathbf{1}_{\mathcal{T}}}_1
    \le
        ke^{-cB}+\sqrt{k}\,\norm{g}_2e^{-cB/2}.
\end{align*}
\paragraph{Step 4: Combine the bounds.}
Taking $C$ sufficiently large makes
\[
    O\!\left(k\sqrt{B\eta}\right)\le\eps/2,
    \qquad
    ke^{-cB}+\sqrt{k}\,\norm{g}_2e^{-cB/2}\le\eps/2.
\]
For the first inequality, the power
$(\eps/k)^{C/2}$ dominates the polynomial and logarithmic
factors in $k\sqrt{B}$.
For the second, $B$ is a sufficiently large multiple of the exponent
in the norm bound.
Therefore,
\[
    \E[\Hinge(f(\x),g(\x))]\le\eps.
\]

Finally, set $m=2\deg q$. Then $m$ is even, and quadratic
substitution gives $\deg g\le m$. Since
\[
    \log(2k/\eta)
    =O(\log(2k/\eps))
    =O(\log(2k)\log(1/\eps)),
\]
the degree and norm bounds become
\[
    m
    =O\!\left(k^2\log^2(2k)\log(1/\eps)\right),
    \qquad
    \norm{g}_2
    \le
    2^{O(k^2\log^3(2k)\log^2(1/\eps))}.
\]
The weighted-error estimate follows from Cauchy--Schwarz
and $a^2\le a$.
For the preservation claim, recall that
$g=q(p_1/B,\ldots,p_k/B)$ with $\deg q=m/2$.
Under the lifted-subspace assumption, each $p_i$ is affine
in $\proj_S\Phi_2(\x)$, so $g$ has the required form.
\end{proof}

Given \Cref{lem:polynomial-low-hinge-loss},
we can now prove \Cref{thm:hinge-main} by applying standard stochastic convex optimization over
the corresponding polynomial ball to find a polynomial $g$ with
small expected hinge loss.
Then obtain the hypothesis $h(\x)=\sign(g(\x))$ with small
$0$--$1$ error, since any point that $h$ makes mistake for $0$--$1$ error would also incur constant error for Hinge loss. We will use the following fact about stochastic convex optimization and the whole algorithm procedure is summarized in
\Cref{alg:hinge-learning}.

\begin{fact} [Projected stochastic subgradient descent; \cite{Duc18}, Corollary 3.4.9] \label{fct:stochastic-convex-optimization}
Let $\mathcal{C}\subseteq\R^N$ be a nonempty compact
convex set of diameter at most $R>0$, with efficiently computable
Euclidean projections, and let $F:\mathcal{C}\to\R$ be
continuous and convex.
Suppose a stochastic oracle, using fresh independent randomness
on each call, satisfies
\[
    \E_{\xi}[G(\w,\xi)]\in\partial F(\w),
    \qquad
    \E_{\xi}\!\left[\norm{G(\w,\xi)}_2^2\right]\le M^2
\]
for every $\w\in\mathcal{C}$, where $M>0$.
Then projected stochastic subgradient descent with step sizes
$\alpha_t=R/(M\sqrt{t})$ and iterate averaging returns
$\widehat{\w}\in\mathcal{C}$ after $T\ge1$ oracle calls such that
\[
    \E[F(\widehat{\w})]
    \le
    \min_{\w\in\mathcal{C}}F(\w)
    +\frac{3RM}{2\sqrt{T}}\; .
\]
\end{fact}

\begin{algorithm}[H]
\caption{HingeLearning}
\label{alg:hinge-learning}
\begin{algorithmic}[1]
\State \textbf{Input:} $d,k,\eps$, a basis for $S\subseteq\R^N$,
and i.i.d.\ sample access to $D$.
\State \textbf{Output:} A hypothesis $h:\R^d\to\{\pm1\}$.

\State Let $m$ be the even degree bound from
\Cref{lem:polynomial-low-hinge-loss} with accuracy $\eps/12$,
and set $t=m/2$. Choose the coefficient radius $R$ as in
the proof of \Cref{thm:hinge-main}.

\State Compute an orthonormal basis $\v_1,\ldots,\v_n$ of $S$,
where $n=\dim S$, and define
\[
    \Psi_t(\x)
    =
    \left(
        \prod_{j=1}^{n}
        \langle \v_j,\Phi_2(\x)\rangle^{\alpha_j}
    \right)_{\alpha\in\Z_{\ge0}^{n},\,|\alpha|\le t}.
\]

\State Using fresh samples from $D$, apply standard projected
stochastic subgradient descent to
\[
    \min_{\norm{\w}_2\le R}
    \E_{(\x,y)\sim D}
    \left[
        \max\{0,1-y\langle \w,\Psi_t(\x)\rangle\}
    \right]
\]
to expected additive optimization error at most $\eps/12$.
Let $\widehat \w$ denote its output.

\State \textbf{Return:}
$h(\x)=\sign(\langle\widehat \w,\Psi_t(\x)\rangle)$.
\end{algorithmic}
\end{algorithm}

We now prove \Cref{thm:hinge-main}.

\begin{proof}[Proof of \Cref{thm:hinge-main}]
We analyze \Cref{alg:hinge-learning}. Apply the construction in the
proof of \Cref{lem:polynomial-low-hinge-loss} with accuracy $\eps/12$.
Let
\[
    m=O\!\left(k^2\log^2(2k)\log(1/\eps)\right),
    \qquad
    t=m/2,
    \qquad
    W=2^{O(k^2\log^3(2k)\log^2(1/\eps))},
\]
where $m$ is even and the constant in the definition of $W$ is
sufficiently large. This gives a polynomial $g^\star$ satisfying
\[
    \E_{\x\sim\cN(0,\bI_d)}
    \left[\Hinge(f(\x),g^\star(\x))\right]
    \le \eps/12,
    \qquad
    \norm{g^\star}_2\le W.
\]

We first bound the coefficients of $g^\star$ in the feature
representation used by the algorithm. Normalize the defining
polynomials
\[
    p_i(\x)=c_i+\u_i\cdot\Phi_2(\x)
\]
to have Gaussian $L_2$ norm one. The construction gives
\[
    g^\star(\x)
    =
    q\!\left(p_1(\x)/B,\ldots,p_k(\x)/B\right),
\]
where $B\ge1$, $\deg q\le t$, and, by
\Cref{fct:polynomial-sign-matching}, the sum of the absolute
coefficients of $q$ is at most $W$, after increasing the constant
defining $W$.

Let $\v_1,\ldots,\v_n$ be the orthonormal basis of $S$ computed by
the algorithm, and write
\[
    z_j(\x)=\langle \v_j,\Phi_2(\x)\rangle,
    \qquad
    p_i(\x)=c_i+\sum_{j=1}^n b_{ij}z_j(\x).
\]
Orthonormality of the Hermite features and the normalization
of $p_i$ imply
\[
    c_i^2+\sum_{j=1}^n b_{ij}^2=1,
    \qquad
    |c_i|+\sum_{j=1}^n|b_{ij}|\le\sqrt{n+1}.
\]
Expanding the composition defining $g^\star$, we therefore obtain
\[
    g^\star(\x)=\langle \w^\star,\Psi_t(\x)\rangle,
    \qquad
    \norm{\w^\star}_2
    \le \norm{\w^\star}_1
    \le W(n+1)^{t/2}.
\]
Accordingly, set
    $R=W(n+1)^{t/2}$.
This radius is determined by the known parameters and does not
require knowledge of the defining polynomials.

Next, we bound the comparator's hinge loss under the observed
labels. Let
    $\delta=\Pr_{(\x,y)\sim D}[y\ne f(\x)]$.
The hinge loss satisfies
\[
    \Hinge(y,a)
    \le
    \Hinge(f(\x),a)
    +2|a|\mathbf{1}_{\{y\ne f(\x)\}}.
\]
Since the input marginal of $D$ is Gaussian, Cauchy--Schwarz gives
\[
\begin{aligned}
    \E_{(\x,y)\sim D}
    [\Hinge(y,g^\star(\x))]
    &\le
    \eps/12
    +2\E_{(\x,y)\sim D}
    \left[
        |g^\star(\x)|\mathbf{1}_{\{y\ne f(\x)\}}
    \right] \\
    &\le \eps/12+2W\sqrt{\delta}
    \le \eps/4,
\end{aligned}
\]
where the last inequality follows from the theorem's noise
assumption for a sufficiently large universal constant $C$.

We now verify the moment bound needed to apply
\Cref{fct:stochastic-convex-optimization}. Let
    $M=\binom{n+t}{t}$
be the number of coordinates of $\Psi_t$. Each $z_j$ is a
polynomial of degree at most two with Gaussian $L_2$ norm one.
Thus, for every multi-index $\alpha$ with $1\le|\alpha|=r\le t$,
H\"older's inequality and Gaussian hypercontractivity give
\[
    \norm{\prod_{j=1}^n z_j^{\alpha_j}}_2
    \le
    \prod_{j=1}^n\norm{z_j}_{2r}^{\alpha_j}
    \le (2r-1)^r
    \le (2t)^t.
\]
The constant feature has norm one, so
\[
    \E_{\x\sim\cN(0,\bI_d)}
    \norm{\Psi_t(\x)}_2^2
    \le M(2t)^{2t}
    =:\sigma^2.
\]

Consider the convex objective
\[
    F(\w)
    =
    \E_{(\x,y)\sim D}
    \left[
        \Hinge
        \bigl(y,\langle \w,\Psi_t(\x)\rangle\bigr)
    \right]
\]
over $\mathcal{C}=\{\w\in\R^M:\norm{\w}_2\le R\}$. It is continuous,
since
\[
    |F(\w)-F(\v)|\le \sigma\norm{\w-\v}_2.
\]
A fresh labeled example provides the stochastic subgradient
\[
    G(\w;\x,y)
    =
    -y\Psi_t(\x)\,
    \mathbf{1}_{\{y\langle \w,\Psi_t(\x)\rangle<1\}},
\]
whose expectation belongs to $\partial F(\w)$ and whose second
moment is at most $\sigma^2$. Moreover, $\mathcal{C}$ has diameter
$2R$ and admits an $O(M)$-time Euclidean projection.

Applying \Cref{fct:stochastic-convex-optimization} with
\[
    T=
    \left\lceil
        \frac{1296R^2\sigma^2}{\eps^2}
    \right\rceil
\]
samples therefore returns $\widehat \w\in\mathcal{C}$ satisfying
\[
    \E[F(\widehat \w)]
    \le
    F(\w^\star)+\frac{3R\sigma}{\sqrt{T}}
    \le \eps/4+\eps/12
    =\eps/3.
\]
For the returned hypothesis
\[
    h(\x)=\sign\langle\widehat \w,\Psi_t(\x)\rangle,
\]
misclassification loss is at most hinge loss. Hence
\[
    \E\left[
        \Pr_{(\x,y)\sim D}[h(\x)\ne y]
    \right]
    \le \eps/3.
\]
Markov's inequality implies that $h$ has classification error at
most $\eps$ with probability at least $2/3$.

Finally, the algorithm uses $T$ samples. Computing the lifted
coordinates and their monomials, followed by a subgradient step
and projection, takes $(d+1)^{O(1)}M$ arithmetic operations per
iteration. The basis and feature preprocessing also fit within
the claimed bound. Using $M\le(n+1)^t$,
$R^2=W^2(n+1)^t$, and $\sigma^2=M(2t)^{2t}$, we obtain
\[
    MT
    =O\!\left(
        \frac{W^2(n+1)^{3t}(2t)^{2t}}{\eps^2}
    \right).
\]
Since $t=O(k^2\log^2(2k)\log(1/\eps))$ and
\[
    t\log(2t)+\log(W/\eps)
    =O\!\left(k^2\log^3(2k)\log^2(1/\eps)\right),
\]
both the sample complexity and arithmetic running time are at most
\[
    d^{O(1)}
    n^{O(k^2\log^2(k)\log(1/\eps))}
    2^{O(k^2\log^3(k)\log^2(1/\eps))}.
\]
This proves the theorem.
\end{proof}

\section{Improved Complexity through Dimension Reduction and Hinge Loss Optimization}\label{sec quasi}
In this section, we combine two stages of dimension reduction with hinge-loss minimization to improve the dependence on the accuracy parameter from
$2^{\poly_k(1/\eps)}$ to $2^{\poly_k(\log(1/\eps))}$.
The main idea is to identify a low-dimensional space of polynomial features that supports a sufficiently accurate approximation to the target, allowing us to apply the hinge-loss learner of \Cref{thm:hinge-main}.
Our first dimension-reduction step identifies a subspace $V$ such that the randomized error is small along every direction orthogonal to $V$.
While this provides the regularity needed to control the residual polynomial contributions, the resulting feature space may still be too large for efficient hinge-loss minimization.
To overcome this, we develop a second dimension-reduction procedure that uses labeled moments to recover a low-dimensional subspace $U$ capturing the relevant residual polynomial features, without explicitly identifying the defining polynomials.
We show that the combined space $\mathcal{P}_2(V)+U$ supports a sufficiently accurate intersection, enabling hinge-loss minimization in this reduced feature space.

\begin{theorem}[Learning by dimension reduction and hinge loss]
\label{thm:sec9-conclusion}
    Consider the problem of learning an intersection of $k$ degree-$2$ PTFs under a Gaussian distribution. There is an algorithm such that given any $\eps,\delta \in (0,1)$, it draws $d^{O(k)}2^{O_k(\log^3(1/\eps))}\log(1/\delta)$ samples from $D$, runs in $d^{O(k)}2^{O_k(\log^3(1/\eps))}\log(1/\delta)$ time and outputs a hypothesis $\hat{h}: \R^d \to \{0,1\}$ such that with probability at least $1-\delta$, $\err_D(\hat{h}) \le \eps$.
\end{theorem}

We give the intuition below.
Notice that \Cref{alg finite grid learning} makes use of \Cref{th certificate} to adaptively find directions that are important to the current intersections of PTFs, and when we cannot do so, the resulting instance is close to a jointly regular instance, where we can make use of the subspace recovery algorithm over the subspace spanned by the degree-$2$ Hermite polynomials.
However, such a method will essentially produces $2^{\poly(1/\eps)}$ many subproblems to solve, which introduces a $2^{\poly(1/\eps)}$ complexity.
The intuition is that the regularity condition required for implementing the algorithm from \cite{DKS18a} turns out to be much stronger than what the learning task really needs.
To obtain an improved complexity, our motivation is that combined with \Cref{th certificate}, we want to construct a single subspace of the space of degree-$2$ Hermite polynomials, such that some intersection of $k$ degree-$2$ PTFs from that subspace is very close to the target instance.

Our approach works as follows. By \Cref{th certificate} and \Cref{th direction finding one shot}, we can find a single subspace $V \subseteq \R^d$ of dimension $2^{O_k(\log^2(1/\eps))}$ such that for every unit vector $\v \perp V$, $\rand_{f^*}(\v)$ is tiny.
We write $\x = \z + \w$, where $\z \in V$ and $\w \perp V$. To simplify the notation, we denote by $n = \dim(V)$ and $m = d-n$.
Notice that for each quadratic function $q_i(\x)$ that defines the target function, we can write
\begin{align*}
    q_i(\z,\w) = a_i(\z) + (\bb_i + \bL_i^\top\z)^\top \w + \w^\top \mathbf{A}_i \w - \tr(\mathbf{A}_i),
\end{align*}
where $a_i(\z)$ is a quadratic function that purely depends on $\z$, $\bb_i \in \R^m$, $\bL_i: \R^m \to \R^n$ and $\mathbf{A}_i \in \R^{m \times m}$.
We will use the subspace $V$ that we found to construct another subspace $\bar{V} \subseteq P_{\le 2}(\R^d)$ such that $P_{\le 2}(V)$ together with $\bar V$ contains some intersections of $k$ degree-$2$ PTFs that is close to $f^*$.
Similar to \Cref{sec:gaussian-comparison}, where we use algorithm from \cite{DKS18a}, and look at the second moment matrix over the space $P_{\le 2}(\R^d)$ and find a subspace that almost spans the linear combination coefficients.
Our method will focus on the subspace of the polynomials of the form $(\bb+\bL^\top\z)^\top \w +\w^\top \mathbf{A} \w - \tr(\mathbf{A})$. In particular, we will use $\u = (\bb,\bL,\mathbf{A})$ to vectorize the coefficient of the polynomial. Denote by $\mathcal{S}$ this space of coefficients. We will build a bilinear map $\mathcal{M}: \mathcal{S} \times P_{\le 2k}(\R^d) \to \R$, such that the required subspace $\bar V$ would be the space of large singular vectors of $\mathcal{M}$.

\subsection{Subspace Recovery through Gaussian Comparison}
To begin with, we first work on an idealized Gaussian model, to get a stronger property \Cref{lm structre idealized gaussian} than the one we used from \cite{DKS18a}.

\begin{lemma}\label{lm structre idealized gaussian}
    Let $V \subseteq \R^d$. Write $\x = \z + \w \in \R^d$, where $\z \in V$ and $\w \in V^\perp$. Let $f(\x) = \Ind\{q_1(\x) \ge 0, \dots, q_k(\x) \ge 0\}$, where for $i \in [k]$, $q_i(\x) = a_i(\z) + \u_i \cdot \w$, $\norm{q_i(\x)}=1$ ,where $a_i(\z)$ is a polynomial of degree at most $2$ that only depends on $\z$. Let $\v \in V^\perp$ be any unit vector and let $\alpha \in (0,1)$.
    Suppose $\rand_{f}(\v) \ge \alpha$, there exists a polynomial $p: \R^d \to \R$, with $\deg(p) \le 2k$ and $\norm{p} \le 1$ such that $\E_{\x\sim \cN(0,\bI)}[f(\x) M_\v p(\x)] \ge \Omega((\alpha/k)^{3k})$. In particular, there exists a polynomial $P:\R^k\to\R$ such that
    $p(\x)=P(q_1(\x),\dots,q_k(\x))$.
\end{lemma}

Roughly speaking, \Cref{lm structre idealized gaussian} considers a family of intersections of $k$ degree-$2$ polynomials, whose defining polynomials have a special form of $q_i(\x) = a_i(\z) + \u_i \cdot \w$, where $a_i$ is a quadratic function that only depends on a given subspace $V$, while in $V^\perp$, $q_i$ only has a linear dependence.
Notice that if $V = \emptyset$, then the instance degenerates to an intersections of $k$ halfspaces. The property given by \cite{DKS18a} says that if a direction $\v \in V^\perp$ has large randomized error, then there is a degree-$2$ polynomial associated with $\v$ to certify this fact. Thus, by using the first 2 Chow-moments, we are able to find all these important directions.
Here, for a general subspace $V$, we still want to identify all directions in $V^\perp$ that have large randomized error through moment methods. However, the existence of the quadratic functions $a_i(\z)$ can hide the signals from the first two moments and thus we need to use higher moments for this purpose.

The idea is similar to the certificate polynomial construction in \Cref{sec certificate polynomial}, where fix a pair of $\z \in V$, $\u \in V^\perp, \u \perp \v$, $f_{(\z,\u)}$ gives a one dimensional Boolean function. We would like to construct a polynomial $h$ such that $\Delta_{f_{(\z,\u)}}(s)h(s)\phi(s)$ is non-trivially large at any boundary of $f_{(\z,\u)}$.
The special form of $f$ implies that once $(\z,\u)$ is fixed, the function is just an indicator function of an interval. This fact makes the construction much easier than \Cref{sec certificate polynomial}.
In fact, we can simply choose $h(\x) = \prod_{i:\u_i\cdot \v<0} q_i(\x) - \prod_{i:\u_i\cdot \v \ge 0} q_i(\x)$. Once $(\z,\u)$ is fixed, each $q_i$ is a monotone linear function along $\v$ direction, which naturally gives the certificate polynomial that we want.

\begin{proof}[Proof of \Cref{lm structre idealized gaussian}]
We consider polynomial $h(\x) = \prod_{i:\u_i\cdot \v<0} q_i(\x) - \prod_{i:\u_i\cdot \v \ge 0} q_i(\x)$.
We write $\x = \z + s\v + \u$, where $\u \perp V, \u \perp \v$, $s \in \R$.
By \Cref{lm M operator}, we know that
\begin{align*}
   & \E_{\x\sim \cN(0,\bI)}[f(\x) M_\v h(\x)]  \\
    = & \E_{\z,\u \sim \cN(0,\bI)} [\sum_{s \in S_{(\z,\u)}} \Delta_{f_{(\z,\u)}}(s)h(s\v+\u+\z)\phi(s)].
\end{align*}
Since each $q_i(\x)$ depends linearly on $\w$, for each fixed
$(\z,\u)$ the function $f_{(\z,\u)}(s)$ agrees almost everywhere
with the indicator of an interval, possibly empty or all of $\R$.
Layers on which $f_{(\z,\u)}$ is constant almost everywhere
contribute zero to both the correlation and the randomized error,
and are omitted from the following layerwise analysis.
For each remaining layer, denote the interval by
$(l_{(\z,\u)},r_{(\z,\u)})$, allowing infinite endpoints.
All endpoint identities below are asserted only at finite endpoints.
Boundary terms at infinite endpoints are interpreted as zero by
taking limits, and any condition in the definition of $B_t$
involving an infinite endpoint is omitted.
This implies that
\begin{align*}
    \E_{\x\sim \cN(0,\bI)}[f(\x) M_\v h(\x)] = \E_{\z,\u \sim \cN(0,\bI)} [h(l_{(\z,\u)}\v+\u+\z)\phi(l_{(\z,\u)}) -  h(r_{(\z,\u)}\v+\u+\z)\phi(r_{(\z,\u)})].
\end{align*}
Notice that for some $i \in [k]$. If $\v \cdot \u_i < 0$, then $q_i(s\v+\u+\z)$ is a decreasing function and $q_i(l_{(\z,\u)}\v + \z + \u) \ge 0$, while if $\v \cdot \u_i \ge  0$, then $q_i(s\v+\u+\z)$ is a nondecreasing function and $q_i(r_{(\z,\u)}\v+\z+\u) \ge 0$. In particular, this implies $\prod_{i:\u_i\cdot \v \ge 0} q_i(l_{(\z,\u)}\v + \z + \u) = 0$ and $\prod_{i:\u_i\cdot \v < 0} q_i(r_{(\z,\u)}\v + \z + \u) = 0$. Otherwise, it contradicts the definition of the $r_{(\z,\u)},l_{(\z,\u)}$.
So, we have
\begin{align*}
    \E_{\x\sim \cN(0,\bI)}[f(\x) M_\v h(\x)] = \E_{\z,\u \sim \cN(0,\bI)} [\prod_{i:\u_i\cdot \v < 0} q_i(l_{(\z,\u)}\v + \z + \u)\phi(l_{(\z,\u)}) +  \prod_{i:\u_i\cdot \v \ge 0} q_i(r_{(\z,\u)}\v + \z + \u)\phi(r_{(\z,\u)})],
\end{align*}
which is nonnegative.
We will show that $\rand_{f}(\v)$ can be upper bounded in terms of $\E_{\x\sim \cN(0,\bI)}[f(\x) M_\v h(\x)]$.
Recall that
\begin{align*}
\rand_f(\v) & = \E_{\u,\z \sim \cN(0,\bI)}\left[ \Pr_{s,s' \sim \cN(0,1)}[f(\u+\z+s \v) \neq f(\u+\z+s' \v)] \right]    \\
& = \E_{\u,\z \sim \cN(0,\bI)}\left[ 2\Pr_{s \sim \cN(0,1)}[f(\u+\z+s \v) =1](1-\Pr_{s \sim \cN(0,1)}[f(\u+\z+s \v) =1]) \right].
\end{align*}
To simplify the notations, we denote by $I_-=\{i \in [k] \mid \u_i \cdot \v < 0\}$ and $I_+=\{i \in [k] \mid \u_i \cdot \v \ge 0\}$.
Let $B_t$ be the event over $(\z,\u)$ such that there exists some $q_i(l_{(\z,\u)}\v + \z + \u) \le t$, $ i \in I_-$ or $q_i(r_{(\z,\u)}\v + \z + \u) \le t$, $ i \in I_+$.
We will separate the analysis for $(\u,\z) \in B_t$ and $(\u,\z) \not\in B_t$.

We start by analyzing the case where $(\u,\z) \in B_t$.
By the monotonicity of $q_i$ as a function of $s$ along direction $\v$, we know that if $f(\x) = 1$ and $B_t$ happen together, then there must be some $ i \in [k]$ such that $\abs{q_i(\x)} \le t$.
This implies that
\begin{align*}
    \Pr_{\x \sim \cN(0,\bI)}[f(\x) = 1, B_t] \le \sum_{i \in [k]}[\Pr_{\x \sim \cN(0,\bI)}\abs{q_i(\x)} \le t] \le O(k\sqrt{t}).
\end{align*}
This implies that
\begin{align*}
    \E_{\u,\z \sim \cN(0,\bI)}\left[ 2\Pr_{s \sim \cN(0,1)}[f(\u+\z+s \v) =1](1-\Pr_{s \sim \cN(0,1)}[f(\u+\z+s \v) =1]) \Ind\{(\u,\z) \in B_t\} \right] \le O(k\sqrt{t}).
\end{align*}

Next, we analyze the case where $(\u,\z) \notin B_t$. In this case, we have
\begin{align*}
   & \left( \prod_{i:\u_i\cdot \v < 0} q_i(l_{(\z,\u)}\v + \z + \u)\phi(l_{(\z,\u)}) +  \prod_{i:\u_i\cdot \v \ge 0} q_i(r_{(\z,\u)}\v + \z + \u)\phi(r_{(\z,\u)})  \right)\Ind\{(\u,\z) \notin B_t\} \\
    &\ge t^k(\phi(l_{(\z,\u)})+\phi(r_{(\z,\u)})) \Ind\{ (\u,\z) \notin B_t \}
\end{align*}
By \Cref{fact gaussian isopermetry}, we know that
\begin{align*}
    \phi(l_{(\z,\u)})+\phi(r_{(\z,\u)}) \ge \Omega(\Pr_{s \sim \cN(0,1)}[f(\u+\z+s \v) =1](1-\Pr_{s \sim \cN(0,1)}[f(\u+\z+s \v) =1])),
\end{align*}
which implies that
\begin{align*}
   &t^{-k}\E_{\x\sim \cN(0,\bI)} [f(\x) M_\v h(\x)] \\
   &\ge \Omega(\E_{\u,\z \sim \cN(0,\bI)}\left[ 2\Pr_{s \sim \cN(0,1)}[f(\u+\z+s \v) =1](1-\Pr_{s \sim \cN(0,1)}[f(\u+\z+s \v) =1]) \Ind\{(\u,\z) \not \in B_t\} \right] ).
\end{align*}
Together, this gives
\begin{align*}
    \rand_{f}(\v) \le O(t^{-k} \E_{\x\sim \cN(0,\bI)} [f(\x) M_\v h(\x)] + k\sqrt{t}  ).
\end{align*}
Now suppose that $\rand_f(\v)\ge\alpha$.
Choose $t=(c\alpha/k)^2$, where $c>0$ is a
sufficiently small universal constant.
By the preceding inequality,
$
\E_{\x\sim\cN(0,\bI)}[f(\x)M_\v h(\x)]
\ge \Omega(\alpha t^k)$.
By H\"older's inequality and Gaussian
hypercontractivity, for every nonempty
$I\subseteq[k]$,
\[
\left\|\prod_{i\in I}q_i\right\|_2
\le (2|I|)^{|I|}
\le (2k)^k.
\]
Consequently, $\norm{h}_2\le H_k=2(2k)^k$.
Taking $p(\x)=\frac{h(\x)}{H_k}$, we finish the proof.

\end{proof}

That is to say once we can write each $q_i$ that defines $f(\x)$ in a special form of $a_i(\z) + \u_i \cdot \w$, then we can run a singular value decomposition of the bilinear form $\E_{\x \sim \cN(0,\bI)}[f(\x)M_\v p]$ to find the directions in the $\w$ space that is important to $f$. To use this fact to design a learning algorithm for learning a general intersection of $k$ PTFs, the intuition follows from \Cref{sec:gaussian-comparison} and \Cref{sec regular case reduction}.

By \Cref{th direction finding one shot}, the subspace $V$ can be chosen so that for every $\v \perp V$, $\rand_f(\v)$ is tiny. By \Cref{th error to regularity}, we know that $f$ must be close to a $(\tau,V)$-regular instance for some sufficiently small $\tau$. This implies
some form of Gaussian comparison can be done, the random polynomial vector, $(\z,N_{\u_1}(\z,\w),\dots,N_{\u_k}(\z,\w))$, where $\z,\w \sim \cN(0,\bI)$ and $N_{\u_i}(\z,\w) = (\bb_i + \bL_i^\top\z)^\top \w + \w^\top \mathbf{A}_i \w - \tr(\mathbf{A}_i)$ is close to $(\z, \u_1\cdot G,\dots,\u_k \cdot G)$, where $G$ is an independent Gaussian.
Then, since we are able to use the SVD type method in the idealized Gaussian model, we translate this into the original model over $\R^d$ that we care about.

Recall that the bilinear form we consider in \Cref{lm structre idealized gaussian} satisfies
\begin{align*}
\E_{\x \sim \cN(0,\bI)}[f(\x)M_\v p] = \E_{\x \sim \cN(0,\bI)}[p \partial_\v f],
\end{align*}
which is the key property used in the proof of \Cref{lm structre idealized gaussian}. However, in the set up we care about a vector $\v$ represents a polynomial $N_\v$ instead of a point in $\R^d$, which makes the notion $\partial_v f$ not well defined in our set up.
So, the next step will be to properly design a bilinear form in the space we are working on to preserve the property of $\E_{\x \sim \cN(0,\bI)}[f(\x)M_\v p] = \E_{\x \sim \cN(0,\bI)}[p \partial_\v f]$ in some way so that we can use Gaussian comparison to show that an SVD over it can recover important $N_{\v}$.

Specifically, under the idealized Gaussian model, since $f(\x) = \Ind\{q_1(\z,G)\ge 0,\dots, q_k(\z,G) \ge 0\}$, which can be seen as a function of $q_1,\dots, q_k$,
for a unit vector $\v$ in $V^\perp$, we have
\begin{align}\label{eq boundary operator}
    \E_{\x \sim \cN(0,\bI)}[f(\x)M_\v p] = \E_{\x \sim \cN(0,\bI)}[p \partial_\v f] = \sum_{i=1}^k (\v \cdot \u_i) \E_{\z,\w \sim \cN(0,\bI)} [p\frac{\partial f}{\partial q_i}].
\end{align}
We would like to build another bilinear form $B_f: \mathcal{S} \times P_{\le 2k}(\R^d)$ such that for every $\v \in \mathcal{S}$ and for every polynomial $p$, $\abs{B_f(\v,p) - \sum_{i=1}^k (\v \cdot \u_i) \E_{\x \sim \cN(0,\bI)} [p\frac{\partial f}{\partial q_i}]}$ is small.
If this approximation holds, we can recover the subspace $\bar V$ by using the large singular vector spaces of $B_f$.
The proof follows Gaussian comparison results that we have established earlier. We give an overview about this.
On the one hand, for every $\v \perp \mathrm{span}(\u_1,\dots,\u_k)$, $\abs{B_f(\v,p)}$ is small, which implies that $B_f$ is approximately low-rank. On the other hand, by the special certificate polynomial we construct in \Cref{lm structre idealized gaussian} and gaussian comparison, every important direction $\v$ must be captured in the large singular vector spaces of $B_L$.
This implies that if we take the top $k$ singular vectors of this bilinear form $B_f$ and span a subspace $\bar V$, then any vector $\u \perp \bar V$ cannot introduce large error in the idealized Gaussian model. Thus, \Cref{lm subspace preserve} immediately shows that we can search an intersection of $k$ PTFs in the found subspace together with $P_{\le 2}(V)$ that is sufficiently close to $f$. We next formalize this argument.

We first describe the approximation property required of the bilinear
form, without specifying how it is constructed. Fix a smooth function
$\chi:\R\to[0,1]$ equal to zero on $(-\infty,-1]$ and one on
$[1,\infty)$. For $h\in(0,1)$, define
\[
\psi_h(t_1,\ldots,t_k)=\prod_{i=1}^k\chi(t_i/h),
\qquad f_h=\psi_h(q_1,\ldots,q_k),
\qquad \widetilde f_h=\psi_h(\widetilde q_1,\ldots,\widetilde q_k).
\]
We use the notation
\[
\frac{\partial f_h}{\partial q_i}
:=\partial_i\psi_h(q_1,\ldots,q_k).
\]
These regularized partial derivatives are used only in the analysis.
In particular, the algorithm will not need the defining polynomials or
values of $f_h$.

For a bilinear form on
$\mathcal S\times\mathcal P_{\le2k}(\R^d)$, set
\[
\|B\|_{\mathrm{op}}
=\sup_{\|\v\|\le1,\,\|p\|_2\le1}|B(\v,p)|.
\]
Representing both spaces in orthonormal bases identifies this with the
usual matrix operator norm; the left singular vectors belong to
$\mathcal S$.

\begin{lemma}[Subspace recovery from derivative approximation]
\label{lem:sec9-recovery}
Let $q_1,\ldots,q_k$ satisfy the normalization above and suppose each
$q_i$ is $(\tau,V)$-regular, with $\tau\in(0,1)$.
Let $h\in(0,1)$ and $\alpha\in(0,1/2)$.
Suppose a bilinear form
$B_f:\mathcal S\times\mathcal P_{\le2k}(\R^d)\to\R$ satisfies,
for every $\v\in\mathcal S$ and $p\in\mathcal P_{\le2k}(\R^d)$,
\begin{equation}
\left|
B_f(\v,p)-\sum_{i=1}^k\langle\v,\u_i\rangle
\E_{\z,\w}\left[p\frac{\partial f_h}{\partial q_i}\right]
\right|
\le\xi\|\v\|\|p\|_2.
\label{eq:sec9ac-abstract-hypothesis}
\end{equation}
Let $\widehat B$ satisfy
$\|\widehat B-B_f\|_{\mathrm{op}}\le\nu$, and let
$\bar V\subseteq\mathcal S$ be the span of its top
$\min\{k,\dim\mathcal S\}$ left singular vectors.
There is $C_k=2^{O(k)}$ such that, if
\begin{equation}
2\xi+2\nu+C_k\bigl(h^{1/4}+\tau h^{-3}\bigr)
\le\frac{\beta_k(\alpha)}2,
\label{eq:sec9ac-abstract-condition}
\end{equation}
then there exists an intersection
$g=\mathbf1\{g_i\ge0\text{ for all }i\in[k]\}$ with
\[
g_i\in\mathcal P_{\le2}(V)\oplus\{N_{\u}:\u\in\bar V\},
\qquad
\Pr_{\z,\w}[f\ne g]
\le C_{\mathrm{tr},k}(k\alpha+\tau)^{1/20}.
\]
Here $C_{\mathrm{tr},k}$ is the constant in the quantitative conclusion
of \Cref{lm subspace preserve}.
\end{lemma}

\begin{proof}[Proof of \Cref{lem:sec9-recovery}]
If $\bar V=\mathcal S$, take $g=f$. We henceforth assume
$k<\dim\mathcal S$.

Define the bilinear form
\[
A_h(\v,p)=\sum_{i=1}^k\langle\v,\u_i\rangle
\E_{\z,\w}\left[p\frac{\partial f_h}{\partial q_i}\right].
\]
It has rank at most $k$, and
\eqref{eq:sec9ac-abstract-hypothesis} gives
$\|B_f-A_h\|_{\mathrm{op}}\le\xi$.
Thus
$\sigma_{k+1}(\widehat B)\le\xi+\nu$.
By the definition of $\bar V$, for every unit $\v\perp\bar V$,
\begin{equation}
\sup_{\deg(p)\le2k,\,\|p\|_2\le1}|B_f(\v,p)|
\le\xi+2\nu.
\label{eq:sec9ac-discarded-bound}
\end{equation}

We show that every such direction has small randomized error in the
ideal Gaussian model. Suppose instead that
$\operatorname{Rand}_{\widetilde f}(\v)\ge\alpha$.
Let $P$ be the certificate from \Cref{lm structre idealized gaussian}. Then
\begin{equation}
\E_{\z,G}[P(\widetilde q)\partial_{\v}\widetilde f]
\ge\beta_k(\alpha).
\label{eq:sec9ac-ideal-signal}
\end{equation}
We will compare this quantity with $A_h(\v,P(q))$.

First, if $Q_1,\ldots,Q_k$ are degree-at-most-two polynomials of any
standard Gaussian input with $\|Q_i\|_2\le1$, H\"older's inequality and
Gaussian hypercontractivity give
\[
\left\|\prod_{i\in J}Q_i\right\|_2
\le\prod_{i\in J}\|Q_i\|_{2|J|}
\le(2|J|-1)^{|J|}\le(2k)^k
\quad (J\ne\varnothing).
\]
The empty product has norm one. Every partial derivative of $P$ of order
at most two is a signed sum of at most two such products divided by
$H_k$. Consequently,
\begin{equation}
\|P(Q)\|_2\le1,\qquad
\|(\partial_aP)(Q)\|_2\le1,\qquad
\|(\partial_{ab}P)(Q)\|_2\le1.
\label{eq:sec9ac-product-bounds}
\end{equation}
In particular, $P(q)$ and $P(\widetilde q)$ have degree at most $2k$
and norm at most one.

For $s\in[0,1]$, introduce the interpolation
\[
q_{s,i}=a_i(\z)+\sqrt{s}\,N_{\u_i}(\z,\w)
+\sqrt{1-s}\,\langle\u_i,G\rangle,
\qquad Q_s=(q_{s,1},\ldots,q_{s,k}).
\]
The residuals are centered conditional on $\z$, and $G$ is independent
of $(\z,\w)$. Therefore every $q_{s,i}$ has degree at most two and
$L_2$ norm one under the joint input $(\z,\w,G)$.

Fix $i\in[k]$ and put $J_i(t)=P(t)\partial_i\psi_h(t)$.
The interpolation identity in \Cref{fact:gaussian-interpolation-identity}, applied to $J_i$, gives
\begin{equation}
\frac{d}{ds}\E_{\z,\w,G}[J_i(Q_s)]
=\frac12\sum_{a,b=1}^k
\E_{\z,\w,G}\bigl[
(\Gamma_{ab}-\Sigma_{ab})\partial_{ab}J_i(Q_s)
\bigr],
\label{eq:sec9ac-interpolation}
\end{equation}
where, writing $c_i(\z)=b_i+L_i^\top\z$,
\[
\Gamma_{ab}=(c_a+\mathbf A_a\w)^\top
(c_b+2\mathbf A_b\w),
\qquad
\Sigma_{ab}=\langle\u_a,\u_b\rangle.
\]
The regularity estimate of \Cref{fact:gaussian-interpolation-error} gives
$\|\Gamma_{ab}-\Sigma_{ab}\|_2\le C\tau$.
Although $J_i$ need not have bounded second derivatives, it and its
derivatives have polynomial growth. Thus the differentiation and
Gaussian integrations by parts in \Cref{fact:gaussian-interpolation-identity} remain valid.
We use their identity directly, before taking a supremum norm of the
second derivatives.

For $j\le3$, the order-$j$ partial derivatives of $\psi_h$ are bounded
by $C_jh^{-j}$. Moreover,
\begin{align*}
\partial_{ab}J_i
=(\partial_{ab}P)\partial_i\psi_h
+(\partial_aP)\partial_{bi}\psi_h+(\partial_bP)\partial_{ai}\psi_h
+P\partial_{abi}\psi_h.
\end{align*}
By \eqref{eq:sec9ac-product-bounds},
\[
\sup_{s\in[0,1]}\|\partial_{ab}J_i(Q_s)\|_2\le Ch^{-3}.
\]
Cauchy--Schwarz in \eqref{eq:sec9ac-interpolation}, followed by integration
over $s$, therefore gives
\begin{equation}
\left|
\E_{\z,\w}[P(q)\partial_i\psi_h(q)]
-\E_{\z,G}[P(\widetilde q)\partial_i\psi_h(\widetilde q)]
\right|
\le Ck^2\tau h^{-3}.
\label{eq:sec9ac-weighted-comparison}
\end{equation}
All endpoint expectations are continuous by Gaussian moment bounds and
polynomial growth.

Since $\|\u_i\|\le1$ and $\|\v\|=1$, summing
\eqref{eq:sec9ac-weighted-comparison} with weights
$\langle\v,\u_i\rangle$ and applying the ideal chain rule gives
\begin{equation}
\left|
A_h(\v,P(q))
-\E_{\z,G}[P(\widetilde q)\partial_{\v}\widetilde f_h]
\right|
\le Ck^3\tau h^{-3}.
\label{eq:sec9ac-smooth-ideal-comparison}
\end{equation}
Here the exact identity used on the ideal side is
\[
\partial_{\v}\widetilde f_h
=\sum_{i=1}^k\langle\v,\u_i\rangle
\partial_i\psi_h(\widetilde q).
\]

It remains to remove smoothing in this ideal derivative correlation.
Degree-two Gaussian
anti-concentration and a union bound give
\[
\|\widetilde f-\widetilde f_h\|_{2,\z,G}^2
\le\sum_{i=1}^k
\Pr_{\z,G}[|\widetilde q_i|\le h]
\le Ck\sqrt h.
\]
For a degree-at-most-$2k$ polynomial $R(\z,G)$, Gaussian integration
by parts and the normalized Hermite recurrence give
\[
\E_{\z,G}[R\partial_{\v}H]
=\E_{\z,G}[H((\v\cdot G)R-\partial_{\v}R)],
\qquad
\|(\v\cdot G)R-\partial_{\v}R\|_2
\le\sqrt{2k+1}\|R\|_2.
\]
These identities also hold for bounded $H$ in the weak sense.
Taking $R=P(\widetilde q)$ and
$H=\widetilde f-\widetilde f_h$ yields
\[
\left|
\E_{\z,G}[P(\widetilde q)
\partial_{\v}(\widetilde f-\widetilde f_h)]
\right|
\le C\sqrt{k(2k+1)}\,h^{1/4}.
\]
Together with \eqref{eq:sec9ac-smooth-ideal-comparison}, this proves
\begin{equation}
\left|
A_h(\v,P(q))
-\E_{\z,G}[P(\widetilde q)\partial_{\v}\widetilde f]
\right|
\le C_k(h^{1/4}+\tau h^{-3}).
\label{eq:sec9ac-abstract-bridge}
\end{equation}

Using \eqref{eq:sec9ac-abstract-hypothesis},
\eqref{eq:sec9ac-ideal-signal}, and
\eqref{eq:sec9ac-abstract-bridge}, we obtain
\[
B_f(\v,P(q))
\ge\beta_k(\alpha)-\xi-C_k(h^{1/4}+\tau h^{-3}).
\]
Since $\|P(q)\|_2\le1$, comparison with
\eqref{eq:sec9ac-discarded-bound} would imply
\[
\beta_k(\alpha)
\le2\xi+2\nu+C_k(h^{1/4}+\tau h^{-3})
\le\beta_k(\alpha)/2,
\]
a contradiction. Hence
\[
\operatorname{Rand}_{\widetilde f}(\v)<\alpha
\qquad\text{for every unit }\v\perp\bar V.
\]

Set
\[
E=\operatorname{span}\{
\operatorname{proj}_{\bar V^\perp}\u_i:i\in[k]\}.
\]
Then $E\subseteq\bar V^\perp$, $\dim(E)\le k$, and $\widetilde f$
depends on the $\bar V^\perp$ coordinates only through their projection
onto $E$. Choose an orthonormal basis of $E$ and independently resample
its coordinates one at a time. At each step the current Gaussian vector
is standard and independent of $\z$, so the probability that the label
changes is less than $\alpha$. A union bound therefore gives
\[
\Pr_{\z,G,G'}\bigl[
\widetilde f(\z,G)\ne
\widetilde f(\z,\operatorname{proj}_{\bar V}G
+\operatorname{proj}_{\bar V^\perp}G')
\bigr]\le k\alpha,
\]
where $G'$ is an independent standard Gaussian on $\mathcal S$.
Applying \Cref{lm subspace preserve} with $\delta=\min\{k\alpha,1\}$ gives an intersection
with defining polynomials in
$\mathcal P_{\le2}(V)\oplus\{N_{\u}:\u\in\bar V\}$ and error at most
$C_{\mathrm{tr},k}(k\alpha+\tau)^{1/20}$, as required.
\end{proof}

Finally it suffices to design a bilinear form that satisfies the requirement in the statement of \Cref{lem:sec9-recovery}. The intuition of designing this bilinear form is simple.
Recall that in the Gaussian model, by integral by part, for a direction $\v$ and a (smooth) function $f$ and polynomial $p$, we have $\E_{\x \sim \cN(0,\bI)}[(\v \cdot \x)f(\x)] = \E_{\x \sim \cN(0,\bI)}[\partial_\v f]$, as well as $\E_{\x \sim \cN(0,\bI)}[f(\x)M_\v p] = \E_{\x \sim \cN(0,\bI)}[\partial_\v f p]$.
We would like to map these two properties to the setup, where we replace $\v$ by the corresponding polynomial $N_\v$.
Using integral by part, we have
\begin{align*}
    \E_{(\z,\w)\sim \cN(0,\bI)}[N_\u(\z,\w)f] & = \E_{(\z,\w)\sim \cN(0,\bI)}[((\bb + \bL^\top\z)^\top \w + \w^\top \mathbf{A} \w - \tr(\mathbf{A}))f] \\
    & =  \E_{(\z,\w)\sim \cN(0,\bI)}[(\bb + \bL^\top\z)^\top \nabla_\w f + \tr(\mathbf{A}\nabla^2_\w f)]
\end{align*}
We simplify the notation by defining $d_\u(f) = (\bb + \bL^\top\z)^\top \nabla_\w f + \tr(\mathbf{A}\nabla^2_\w f)$, which is used as a suitable replacement of partial derivative.
We give the following guarantee under the regularity assumption.
\begin{lemma}\label{lm derivative operator}
Let $V\subseteq\R^d$, and write $\x=\z+\w$, where
$\z\in V$ and $\w\in V^\perp$ are independent
standard Gaussian vectors. For $i\in[k]$, let $q_i(\z,\w)=a_i(\z)+N_{\u_i}(\z,\w),$
where $\u_i=(\bb_i,\bL_i,\mathbf A_i)$ and $N_{\u_i}(\z,\w)
=(\bb_i+\bL_i^\top\z)^\top\w
+\w^\top\mathbf A_i\w-\tr(\mathbf A_i).$
Suppose that each $q_i$ is $(\tau,V)$-regular
and $\Var(q_i)\le 1$.

Let $\psi:\R^k\to\R$ be a $C^2$ function with
bounded first and second derivatives, and set
\[
F(\z,\w)=\psi(q_1(\z,\w),\ldots,q_k(\z,\w)).
\]
For $j\in\{1,2\}$, define
\[
D_j(\psi)=
\max_{i_1,\ldots,i_j\in[k]}
\|\partial_{i_1}\cdots\partial_{i_j}\psi\|_\infty.
\]
Then for every unit coefficient direction
$\v=(b,L,\mathbf A)$ and every polynomial $p$,
\begin{align*}
&\abs{
\E_{\z,\w \sim \cN(0,\bI)}[p\,d_\v F]
-\sum_{i=1}^k
\langle\v,\u_i\rangle
\E_{\z,\w \sim \cN(0,\bI)}[p\,\partial_i\psi(q_1,\ldots,q_k)]
}\le
C\tau\bigl(kD_1(\psi)+k^2D_2(\psi)\bigr)
\|p\|_2,
\end{align*}
where $C>0$ is a universal constant.
\end{lemma}

\begin{proof}[Proof of \Cref{lm derivative operator}]
To simplify the notation, write $a(\z)=b+L^\top\z,
a_i(\z)=b_i+L_i^\top\z,$
and let $g_i(\z,\w)
=\nabla_\w q_i(\z,\w)
=a_i(\z)+2\mathbf A_i\w.$
Since $\nabla_\w^2q_i=2\mathbf A_i$, the chain
rule gives $\nabla_\w F
=\sum_{i=1}^k\partial_i\psi(q)\,g_i$
and $\nabla_\w^2 F
=\sum_{i=1}^k
2\partial_i\psi(q)\mathbf A_i+
\sum_{i,j=1}^k
\partial_{ij}\psi(q)\,g_i g_j^\top,$
where $q=(q_1,\ldots,q_k)$.

Substituting these expressions into the
definition of $d_\v F$, we obtain
\begin{align*}
d_\v F
=a(\z)^\top\nabla_\w F
+\tr(\mathbf A\nabla_\w^2 F)
=\sum_{i=1}^k
\partial_i\psi(q)
\left(a(\z)^\top g_i
+2\langle\mathbf A,\mathbf A_i\rangle_F\right)
+
\sum_{i,j=1}^k
\partial_{ij}\psi(q)\,
g_i^\top\mathbf A g_j.
\end{align*}

For $i\in[k]$, define $K_i(\z,\w)
=a(\z)^\top g_i(\z,\w)
+2\langle\mathbf A,\mathbf A_i\rangle_F.$
By independence and Gaussian orthogonality,
\begin{align*}
\E_{\z,\w \sim \cN(0,\bI)}[K_i]
=\bb^\top \bb_i+\langle \bL,\bL_i\rangle_F
+2\langle\mathbf A,\mathbf A_i\rangle_F
=\langle\v,\u_i\rangle.
\end{align*}

We first show that $\|K_i-\langle\v,\u_i\rangle\|_2
\le C\tau.$
Expanding the centered polynomial gives
\begin{align*}
K_i-\langle\v,\u_i\rangle
=\bb^\top \bL_i^\top\z+(\bL^\top\z)^\top \bb_i
+\z^\top \bL \bL_i^\top\z
-\langle \bL,\bL_i\rangle_F+2(\bb+\bL^\top\z)^\top\mathbf A_i\w.
\end{align*}

By $(\tau,V)$-regularity and \Cref{fct:regular-to-coefficient}, $\|\bb_i\|_2,\ \|\bL_i\|_{\rm op},\
\|\mathbf A_i\|_{\rm op}\le\tau.$
Moreover, since $\v$ is a unit coefficient
direction, $\|\bb\|_2^2+\|\bL\|_F^2+
2\|\mathbf A\|_F^2=1.$
Consequently,
\begin{align*}
\|\bb^\top \bL_i^\top\z\|_2
&=\|\bL_i \bb\|_2\le\tau,\\
\|(\bL^\top\z)^\top \bb_i\|_2
&=\|\bL \bb_i\|_2\le\tau,\\
\|\z^\top \bL \bL_i^\top\z
-\langle \bL,\bL_i\rangle_F\|_2
&\le\sqrt2\|\bL \bL_i^\top\|_F
\le\sqrt2\tau.
\end{align*}
For the remaining term, conditioning on $\z$
and taking the expectation over $\w$ gives
\begin{align*}
\|2(\bb+\bL^\top\z)^\top\mathbf A_i\w\|_2^2=4\E_\z
\left[\|\mathbf A_i(\bb+\bL^\top\z)\|_2^2\right]\le
4\|\mathbf A_i\|_{\rm op}^2
\bigl(\|\bb\|_2^2+\|\bL\|_F^2\bigr)
\le4\tau^2.
\end{align*}
Combining the four bounds proves $\|K_i-\langle\v,\u_i\rangle\|_2
\le C\tau.$

Next, we show that for every $i,j\in[k]$, $\|g_i^\top\mathbf A g_j\|_2\le C\tau.$
By the variance bound on $q_i$ and Gaussian
orthogonality, $\|\bb_i\|_2^2+\|\bL_i\|_F^2
+2\|\mathbf A_i\|_F^2\le1.$
It follows that
\begin{align*}
\E_{\z,\w \sim \cN(0,\bI)}\|g_i\|_2^2
=\|\bb_i\|_2^2+\|\bL_i\|_F^2
+4\|\mathbf A_i\|_F^2\le2.
\end{align*}
Since $g_i$ is an affine Gaussian vector,
Gaussian fourth-moment bounds imply $\left(\E_{\z,\w \sim \cN(0,\bI)}\|g_i\|_2^4\right)^{1/4}
\le C.$
On the other hand,
\begin{align*}
\E_{\z,\w \sim \cN(0,\bI)}\|\mathbf A g_j\|_2^2
&=\|\mathbf A \bb_j\|_2^2
+\|\mathbf A \bL_j^\top\|_F^2
+4\|\mathbf A\mathbf A_j\|_F^2\\
&\le
\|\mathbf A\|_F^2
\left(
\|\bb_j\|_2^2
+\|\bL_j\|_{\rm op}^2
+4\|\mathbf A_j\|_{\rm op}^2
\right)\\
&\le C\tau^2.
\end{align*}
Applying the same fourth-moment bound gives $\left(
\E_{\z,\w \sim \cN(0,\bI)}\|\mathbf A g_j\|_2^4
\right)^{1/4}
\le C\tau.$
Thus, by Cauchy--Schwarz,
\begin{align*}
\|g_i^\top\mathbf A g_j\|_2
\le
\left(\E\|g_i\|_2^4\right)^{1/4}
\left(\E\|\mathbf A g_j\|_2^4\right)^{1/4}\le C\tau.
\end{align*}

Finally, using the expansion of $d_\v F$,
\begin{align*}
&\abs{
\E_{\z,\w \sim \cN(0,\bI)}[p\,d_\v F]
-\sum_{i=1}^k
\langle\v,\u_i\rangle
\E_{\z,\w \sim \cN(0,\bI)}[p\,\partial_i\psi(q)]
} \\
&\le
\sum_{i=1}^k
\abs{
\E_{\z,\w \sim \cN(0,\bI)}\left[
p\,\partial_i\psi(q)
(K_i-\langle\v,\u_i\rangle)
\right]
}+
\sum_{i,j=1}^k
\abs{
\E_{\z,\w \sim \cN(0,\bI)}\left[
p\,\partial_{ij}\psi(q)
g_i^\top\mathbf A g_j
\right]
}\\
&\le
\|p\|_2
\left(
D_1(\psi)\sum_{i=1}^k
\|K_i-\langle\v,\u_i\rangle\|_2
+
D_2(\psi)\sum_{i,j=1}^k
\|g_i^\top\mathbf A g_j\|_2
\right)\\
&\le
C\tau\bigl(kD_1(\psi)+k^2D_2(\psi)\bigr)
\|p\|_2.
\end{align*}
This completes the proof.
\end{proof}

To complete the proof, we only need to find an operator $T_\u$ such that $\E_{(\z,\w)\sim \cN(0,\bI)}[p d_\u f] = \E_{(\z,\u)\sim \cN(0,\bI)}[f T_\u p]$, for every polynomial $p$.
Such an operator is obtained through integration by parts.

\begin{lemma}\label{lm derivative operator construction}
Let $\v=(b,L,\mathbf A)$ be a residual
coefficient direction and let $p$ be a polynomial.
Define
\begin{align*}
T_\v p
=pN_\v(\z,\w)
-(\bb+\bL^\top\z+2\mathbf A\w)^\top\nabla_\w p +\tr(\mathbf A\nabla_\w^2p),
\end{align*}
where
\[
N_\v(\z,\w)
=(\bb+\bL^\top\z)^\top\w
+\w^\top\mathbf A\w-\tr(\mathbf A).
\]
Then, for every sufficiently smooth function $F$,
\[
\E_{\z,\w \sim \cN(0,\bI)}[p\,d_\v F]
=\E_{\z,\w \sim \cN(0,\bI)}[F\,T_\v p].
\]
The identity also holds for bounded measurable
$F$ when $d_\v F$ is interpreted weakly.
\end{lemma}

\begin{proof}[Proof of
\Cref{lm derivative operator construction}]
Recall that
\[
d_\v F
=(\bb+\bL^\top\z)^\top\nabla_\w F
+\tr(\mathbf A\nabla_\w^2F).
\]
We apply Gaussian integration by parts
with respect to $\w$, keeping $\z$ fixed.
For the linear part, since $b+L^\top\z$ does not
depend on $\w$, we have
\begin{align*}
\E_\w[
p(\bb+\bL^\top\z)^\top\nabla_\w F]=
\E_\w\left[
F\left(
p(\bb+\bL^\top\z)^\top\w
-(\bb+\bL^\top\z)^\top\nabla_\w p
\right)\right].
\end{align*}

For the quadratic part, using the symmetry
of $\mathbf A$ and applying Gaussian
integration by parts once gives
\begin{align*}
\E_\w[p\tr(\mathbf A\nabla_\w^2F)]=
\E_\w[
(p\mathbf A\w-\mathbf A\nabla_\w p)^\top
\nabla_\w F].
\end{align*}
Applying Gaussian integration by parts
a second time, we obtain
\begin{align*}
\E_\w[
(p\mathbf A\w-\mathbf A\nabla_\w p)^\top
\nabla_\w F]=
\E_\w\Big[
F\big(
\w^\top(p\mathbf A\w-\mathbf A\nabla_\w p)
-\mathrm{div}_\w(
p\mathbf A\w-\mathbf A\nabla_\w p)
\big)\Big].
\end{align*}
Notice that $\mathrm{div}_\w(p\mathbf A\w)
=(\mathbf A\w)^\top\nabla_\w p
+p\tr(\mathbf A), \mathrm{div}_\w(\mathbf A\nabla_\w p)
=\tr(\mathbf A\nabla_\w^2p).$
Therefore,
\begin{align*}
\w^\top(p\mathbf A\w-\mathbf A\nabla_\w p)
-\mathrm{div}_\w(
p\mathbf A\w-\mathbf A\nabla_\w p)=
p(\w^\top\mathbf A\w-\tr(\mathbf A))
-2(\mathbf A\w)^\top\nabla_\w p
+\tr(\mathbf A\nabla_\w^2p).
\end{align*}
This gives
\begin{align*}
\E_\w[p\tr(\mathbf A\nabla_\w^2F)]=
\E_\w\Big[
F\big(
p(\w^\top\mathbf A\w-\tr(\mathbf A))
-2(\mathbf A\w)^\top\nabla_\w p
+\tr(\mathbf A\nabla_\w^2p)
\big)\Big].
\end{align*}

Combining the linear and quadratic parts,
and taking expectation over $\z$, we obtain
\begin{align*}
\E_{\z,\w \sim \cN(0,\bI)}[p\,d_\v F]
=\E_{\z,\w \sim \cN(0,\bI)}\Big[
F\big(
pN_\v(\z,\w) -(\bb+\bL^\top\z+2\mathbf A\w)^\top
\nabla_\w p
+\tr(\mathbf A\nabla_\w^2p)
\big)\Big]
=\E_{\z,\w \sim \cN(0,\bI)}[F\,T_\v p],
\end{align*}
as required.

For bounded measurable $F$, the same
identity defines the weak action of
$d_\v F$ on polynomial test functions,
and the conclusion follows by smooth
approximation.
\end{proof}

\subsection{Proof of \Cref{thm:sec9-conclusion}}

We now combine the dimension-reduction guarantees with the
known-lifted-subspace learner of \Cref{thm:hinge-main}. The regular
intersection supplied by \Cref{th error to regularity} and the approximating
intersection supplied by \Cref{lem:sec9-recovery} to prove \Cref{thm:sec9-conclusion}.

\begin{proof}[Proof of \Cref{thm:sec9-conclusion}]
Let $C_8$ be the constant in the approximation requirement of
\Cref{thm:hinge-main}, and put
\[
 L=k^2\log^3(2k)\log^2(1/\varepsilon),\qquad
 \rho=2^{-C_8L}.
\]
Increase $C_{\mathrm{tr},k}$ to be at least one. Choose
\begin{align*}
 a&=\left(\frac{\rho}{2C_{\mathrm{tr},k}}\right)^{20},
 &\alpha&=\frac{a}{2k},
 &\beta&=\beta_k(\alpha),\\
 \tau&=\min\left\{\frac a2,
             \left(\frac{\beta}{8A_k}\right)^{13}\right\},
 &\nu_{\mathrm{est}}&=\frac{\beta}{16},
 &\eta_0&=\min\left\{\frac\rho2,
             \left(\frac{\beta}{16K_k}\right)^2\right\}.
\end{align*}
These choices imply
\begin{equation}
 \eta_0+C_{\mathrm{tr},k}(k\alpha+\tau)^{1/20}\le\rho
 \quad\text{and}\quad
 A_k\tau^{1/13}+2\nu_{\mathrm{est}}+2K_k\sqrt{\eta_0}
 \le\frac{3\beta}{8}.
 \label{eq:sec9-final-parameter-check}
\end{equation}
All logarithms of the inverse parameters are $O_k(L)$.

\paragraph{Finding the physical subspace.}
Set
\[
 \theta=\alpha_k(\eta_0,\tau)
 =\frac{\tau^2\eta_0^2}{C_4k^{12}},
 \qquad \ell=2k+1,
\]
where $C_4$ is the constant in \Cref{th error to regularity}. Let $2\gamma$ be a
valid certificate lower bound from \Cref{th certificate} at randomized-error
level $\theta$; we may take
$\gamma=\tfrac12(\theta/(C_5k))^{2^{C_5k}}$
for a sufficiently large universal $C_5$.
Apply \Cref{th direction finding one shot} to the original labels with degree budget $\ell$
and correlation threshold $\gamma$. With failure probability at most
$\delta/3$, it returns a subspace $V$ such that
\[
 n:=\dim V\le O(\ell/\gamma^2),\qquad
 \operatorname{Rand}_{f^*}(\v)<\theta
 \quad\text{for every unit }\v\perp V.
\]
The randomized-error conclusion follows by contradiction from
\Cref{th certificate}. \Cref{th error to regularity} then supplies an intersection $f$ of at most
$k$ normalized $(\tau,V)$-regular quadratics with
$\Pr_{\x\sim\cN(0,\bI_d)}[f^*(\x)\ne f(\x)]\le\eta_0$.
The required confidence in the moment-estimation step is obtained by
applying entrywise median-of-means amplification to the moment
estimates used in the proof of \Cref{lm influential matrix}.

\paragraph{Recovering the residual polynomial subspace.}
Condition on the first-stage samples and use fresh examples to
estimate $B_y$. Let $R=\dim\mathcal S\le d(d+3)/2$ and
$J=\binom{d+2k}{2k}$. For orthonormal bases $(e_a)_{a=1}^R$
and $(H_\gamma)_{|\gamma|\le2k}$, estimate
\[
 B_y(e_a,H_\gamma)
 =\E_{(\x,y)\sim D}[yT_{e_a}H_\gamma(\x)].
\]
Each sample contribution has second moment at most $K_k^2$. Entrywise median-of-means estimates
with accuracy $\nu_{\mathrm{est}}/\sqrt{RJ}$ therefore give
$\|\widehat B-B_y\|_{\mathrm{op}}\le\nu_{\mathrm{est}}$ with
failure probability at most $\delta/3$, using
\[
 O\!\left(K_k^2RJ\nu_{\mathrm{est}}^{-2}
              \log\frac{2RJ}{\delta}\right)
\]
examples. The operator-norm conclusion follows from the
Frobenius-norm bound. If $R=0$, this estimation step is omitted.
Constructing the matrix and taking its singular value decomposition
uses $(d+1)^{O(k)}\operatorname{poly}_k
(1/\nu_{\mathrm{est}},\log(2/\delta))$ arithmetic operations.

By
\eqref{eq:sec9-final-parameter-check}, \Cref{lem:sec9-recovery} applies to its
leading left singular subspace $\bar V$, of dimension at most $k$.
It supplies an intersection $g$ with defining polynomials in
$
 \mathcal H=\mathcal P_{\le2}(V)
             \oplus\{N_\u:\u\in\bar V\}
$
and
\[
 \Pr_{(\x,y)\sim D}[g(\x)\ne y]
 \le\eta_0+C_{\mathrm{tr},k}(k\alpha+\tau)^{1/20}
 \le\rho.
\]
Let $S\subseteq\R^N$ be the coefficient subspace satisfying
\[
 \mathcal H=\R\oplus\{u\cdot\Phi_2(\cdot):u\in S\}.
\]
Constants are excluded from $S$ and are treated separately in
\Cref{thm:hinge-main}. Since the residual polynomials are orthogonal to
$\mathcal P_{\le2}(V)$,
\begin{equation}
 r:=\dim S
 \le\binom{n+2}{2}-1+k
 =\frac{n(n+3)}2+k.
 \label{eq:sec9-final-lifted-dimension}
\end{equation}

\paragraph{Learning in the recovered subspace.}
Use a fresh batch and apply \Cref{thm:hinge-main} to $S$ and the relabeled
examples $(\x,2y-1)$, with accuracy $\varepsilon$ and failure
probability $\delta/3$. The function $2g-1$ satisfies its
approximation hypothesis because $\rho=2^{-C_8L}$. If its output
is $h:\R^d\to\{-1,1\}$, return $\widehat f=(h+1)/2$.
Conditionally on the successful preceding stages, its error against
the original labels is at most $\varepsilon$. A union bound gives
overall success probability at least $1-\delta$.
The degree statement follows from \Cref{thm:hinge-main}, since substituting
quadratic features at most doubles the degree.

\paragraph{Complexity.}
The parameter choices and \Cref{th certificate} give
\[
 \log(1/\gamma)=O_k(L),\qquad
 \log(n+2)=O_k(L),\qquad
 \log(r+2)=O_k(L).
\]
The two dimension-reduction stages have sample and arithmetic cost
at most
\[
 (d+1)^{O(k)}2^{O_k(L)}\log^2(2/\delta).
\]
Here polynomial factors in $\log(d+2)$ are absorbed into
$(d+1)^{O(k)}$.

By \Cref{thm:hinge-main}, the hinge-loss stage has sample and
arithmetic cost at most
\[
 (d+1)^{O(1)}
 (r+2)^{O(k^2\log^2(2k)\log(1/\varepsilon))}
 2^{O(k^2\log^3(2k)\log^2(1/\varepsilon))}
 \log^2(2/\delta).
\]
Since $\log(r+2)=O_k(L)$ and
$L=O_k(\log^2(1/\varepsilon))$, this is at most
\[
 (d+1)^{O(1)}
 2^{O_k(L\log(1/\varepsilon))}
 \log^2(2/\delta)
 \le
 (d+1)^{O(1)}
 2^{O_k(\log^3(1/\varepsilon))}
 \log^2(2/\delta).
\]
Combining the bounds for the two dimension-reduction stages
and hinge-loss minimization proves the claimed dependence
on $d$ and $\varepsilon$.
\end{proof}

\begin{algorithm}[htbp]
\caption{\textsc{LiftedSubspaceLearning}}
\label{alg lifted subspace learning}
\begin{algorithmic}[1]
\State \textbf{Input:} Sample access from $D$, a subspace
$V\subseteq\R^d$ satisfying the condition of \Cref{sec quasi},
accuracy $\eps$ and failure probability $\delta$
\State \textbf{Output:} A hypothesis
$\hat f:\R^d\to\{0,1\}$

\smallskip

\State Let $m,W$ be the degree and norm bounds from
\Cref{lem:polynomial-low-hinge-loss} with accuracy $\eps/8$.
Set $t=m/2$ and $\eta=(\eps/(20W))^2$.
Choose $\xi$ as required by the subspace recovery
guarantee for approximation error $\eta$.

\State Write $\x=\z+\w$, where
$\z=\proj_V\x$ and $\w=\proj_{V^\perp}\x$.
Let $\mathcal S$ be the space of residual degree-$2$
polynomials $N_\v(\z,\w)$ defined in \Cref{sec quasi}.

\State Let $R_1,\ldots,R_M$ be an orthonormal basis
of $\mathcal S$ under the Gaussian $L_2$ inner product.

\State Draw
$N=d^{O(k)}\poly(1/\xi,\log(1/\delta))$
independent examples
$\{(\x^{(i)},y^{(i)})\}_{i=1}^N$ from $D$.

\State Construct a matrix
$\hat B\in\R^{M\times \dim(\mathcal P_{\le 2k}(\R^d))}$
such that, for every $a\in[M]$ and
multi-index $\gamma$ with $|\gamma|\le 2k$,
\[
\hat B_{a,\gamma}
=\frac1N\sum_{i=1}^N
y^{(i)}T_{R_a}H_\gamma(\x^{(i)}),
\]
where $T_{R_a}$ is the operator in \Cref{lm derivative operator construction}.

\State Let $\u_1,\ldots,\u_r$ be the top $r$
left singular vectors of $\hat B$, where
$r=\min\{k,M\}$.
Set
\[
U=\operatorname{span}\left\{
\sum_{a=1}^M (\u_j)_aR_a:
j\in[r]\right\}.
\]

\State Set
$\mathcal H=\mathcal P_{\le 2}(V)\oplus U$,
and let $1,\phi_1,\ldots,\phi_{s-1}$ be an
orthonormal basis of $\mathcal H$.

\State Construct the lifted polynomial feature map
\[
\Psi_t(\x)=
\left(
\prod_{j=1}^{s-1}\phi_j(\x)^{\gamma_j}
\right)_{|\gamma|\le t}.
\]

\State Using fresh samples from $D$, apply the
bounded-norm hinge-loss minimization method of
\Cref{sec hinge loss} to the features $\Psi_t(\x)$ and
labels $2y-1$, with degree budget $t$ and
the corresponding coefficient-norm bound
$R_{\eps,s}$.

\State \textbf{Return}
$\hat f(\x)=
\mathbf{1}\{\langle\hat\theta,\Psi_t(\x)\rangle
\ge 0\}$,
where $\hat\theta$ is the coefficient vector
returned by hinge-loss minimization.
\end{algorithmic}
\end{algorithm}

\section{Statistical Query Lower Bound}\label{sec sq}
In this section, we prove a statistical query (SQ) lower bound
for learning intersections of degree-2 PTFs. We first recall
the SQ model. Throughout this section, we use the equivalent
$\{\pm1\}$-label convention.

The SQ model was introduced by Kearns~\cite{Kearns:98}
for learning Boolean functions in the PAC
setting~\cite{Valiant:84}. It provides access to approximate
expectations of bounded functions through an oracle.
The model has also been studied in unsupervised
learning~\cite{Feldman16b}, and many familiar techniques admit
SQ implementations, including spectral algorithms, methods
based on moments and tensors, and local search such as
EM~\cite{FeldmanGRVX17,FeldmanGV17}.

\begin{definition}[Statistical Query Model]
\label{def:sq-model}
Let $D$ be a distribution over $\R^d\times\{\pm1\}$.
For a tolerance parameter $\tau>0$, the oracle
$\mathrm{STAT}_D(\tau)$ accepts a query function
$\phi:\R^d\times\{\pm1\}\to[-1,1]$ and returns a value $a$
satisfying
$|a-\E_{(\x,y)\sim D}[\phi(\x,y)]|\le\tau$.
An \emph{SQ algorithm} accesses $D$ through these oracle
calls and may choose each query and its tolerance adaptively
based on previous responses. Its guarantee must hold for
every sequence of responses satisfying the tolerance bounds.
\end{definition}

The main theorem of this section is the following.
\begin{theorem} \label{thm:SQ-LB-main}
    Let \(k\in \Z_+\) and let \(d\in \Z_+\) be sufficiently large depending on \(k\). Any SQ algorithm that learns intersections of $k$ degree-$2$ PTFs with Gaussian marginals (as defined in \Cref{def:learning-degree-2-ptf-intersections}) to any constant error smaller than $1/2$ must either (i) use a query with tolerance at most \(d^{-\Omega(k)}\), or (ii) make at least \(2^{d^{\Omega(1)}}\) queries.
\end{theorem}

The main technical fact of the section is the following.
It states that there exists a function $g:\R\to\{\pm1\}$
that is discontinuous at at most $2k$ points and satisfies
$\E_{t\sim\cN(0,1)}[g(t)p(t)]=0$
for every polynomial $p$ of degree at most $2k-1$.

\begin{fact} \label{fct:sign-matching-function}
    For every \(k\in\Z_+\), there exists a function \(g:\R\to\{\pm1\}\) that is discontinuous at at most \(2k\) points and satisfies $\E_{t\sim\cN(0,1)}[g(t)t^m]=0$ for all $m\in\{0,\ldots,2k-1\}$.
\end{fact}
\begin{proof}[Proof of \Cref{fct:sign-matching-function}]
We construct \(g\) by taking a best \(L_1\) approximation \(p^*\) to \(t^{2k}\) among polynomials of degree at most \(2k-1\), and setting
\[
g(t)=\sign\bigl(p^*(t)-t^{2k}\bigr),
\]
where \(\sign(0)=1\). The idea is that optimality of \(p^*\) forces \(g\) to be orthogonal to every polynomial of degree at most \(2k-1\), while the degree of \(p^*(t)-t^{2k}\) bounds the number of discontinuities. We justify this construction by checking existence of the minimizer, the moment identities, and the discontinuity bound.

First, let \(\mathcal P_{2k-1}\) be the space of real polynomials of degree at most \(2k-1\), and define
\[
F(p):=\E_{t\sim\cN(0,1)}
\bigl[|p(t)-t^{2k}|\bigr].
\]
The functional \(F\) is continuous and satisfies
\[
F(p)\ge \norm{p}_1-\E_{t\sim\cN(0,1)}[|t|^{2k}],
\]
where \(\norm{\cdot}_1\) denotes the Gaussian \(L_1\)-norm. Since this is a norm on the finite-dimensional space \(\mathcal P_{2k-1}\), the sublevel sets of \(F\) are compact. Hence, \(F\) attains its minimum at some \(p^*\in\mathcal P_{2k-1}\).

Next, fix \(m\in\{0,\ldots,2k-1\}\) and consider the perturbation \(p^*(t)+st^m\). The residual \(p^*(t)-t^{2k}\) is a polynomial of degree exactly \(2k\), so it is nonzero almost surely under the Gaussian distribution. Moreover, for \(s\ne0\),
\[
\left|
\frac{|p^*(t)+st^m-t^{2k}|-|p^*(t)-t^{2k}|}{s}
\right|
\le |t|^m.
\]
Since \(\E_{t\sim\cN(0,1)}[|t|^m]<\infty\), dominated convergence justifies differentiating under the expectation. Optimality of \(p^*\) therefore gives
\[
0=
\left.\frac{d}{ds}
\E_{t\sim\cN(0,1)}
\bigl[|p^*(t)+st^m-t^{2k}|\bigr]
\right|_{s=0}
=
\E_{t\sim\cN(0,1)}[g(t)t^m].
\]

Finally, \(g\) can be discontinuous only at a real root of \(p^*(t)-t^{2k}\). This polynomial has degree \(2k\), so there are at most \(2k\) such points. Thus, \(g\) satisfies both required properties, proving the fact.
\end{proof}
Given \Cref{fct:sign-matching-function}, we are ready to prove \Cref{thm:SQ-LB-main}. Let \(g\) be the function constructed in the proof of \Cref{fct:sign-matching-function}. For each unit vector \(\v\in\s^{d-1}\), define \(D_\v\) to be the joint distribution of \((\x,y)\) obtained by sampling \(\x\sim\cN(0,\bI_d)\) and setting \(y=g(\v\cdot \x)\). As we verify below, the labels under \(D_\v\), after conversion to \(\{0,1\}\), are consistent with an intersection of \(k\) degree-$2$ PTFs. Let \(D_0\) be the joint distribution of \((\x,y)\), where \(\x\sim\cN(0,\bI_d)\) and \(y\sim\operatorname{Unif}(\{\pm1\})\) independently.

Using the framework introduced in $\cite{DKS17-sq}$, we can show that any SQ algorithm that distinguish between $D_\v$ with unknown $\v$ and $D_0$ must either (i) use a query with tolerance at most \(d^{-\Omega(k)}\), or (ii) make at least \(2^{d^{\Omega(1)}}\) queries. Notice that an algorithm for learning intersections of $k$ degree-$2$ PTFs can be used to distinguish between $D_\v$ and $D_0$, as when input distribution is $D_\v$, the algorithm must with high probability return a hypothesis with constant error better than $1/2$, while if the input distribution is $D_0$, no such hypothesis could exists.

For convenience, we invoke the SQ lower bound of \cite{DKPZ21}, which is based on the argument outlined above. We recall the relevant definition and proposition below.

\begin{definition} [\cite{DKPZ21}, Definition 18] \label{def:SQ-testing}
    Let \(g:\R\to[-1,1]\) be a measurable function. For each \(\v\in\s^{d-1}\), let \(D_\v\) be the distribution of \((\x,y)\) over \(\R^d\times\{\pm1\}\) satisfying
    $\x\sim\cN(0,\bI_d)$ and
    $\E[y\mid \x]=g(\v\cdot \x)$.
    Define \(\D_g:=\{D_\v:\v\in\s^{d-1}\}\) and
    $D_0:=\cN(0,\bI_d)\times\operatorname{Unif}(\{\pm1\})$.
    We denote by \(\mathcal B(\D_g,D_0)\) the problem of distinguishing \(D\in\D_g\) from \(D=D_0\), given SQ access to an unknown distribution \(D\) promised to satisfy one of these conditions.
\end{definition}

\begin{fact} [\cite{DKPZ21}, Proposition 19] \label{fct:SQ-function-to-LB}
Let \(\ell\in\Z_+\), and let \(d\) be sufficiently large depending on \(\ell\). Suppose that \(g:\R\to[-1,1]\) is measurable and satisfies
$\E_{t\sim\cN(0,1)}[g(t)t^j]=0$
for all $j\in\{0,\ldots,\ell-1\}$.
Any SQ algorithm that solves \(\mathcal B(\D_g,D_0)\) with probability at least \(2/3\) must either (i) use a query with tolerance at most \(d^{-\Omega(\ell)}\), or (ii) make at least \(2^{d^{\Omega(1)}}\) queries.
\end{fact}

We now prove \Cref{thm:SQ-LB-main}.
\begin{proof} [Proof of \Cref{thm:SQ-LB-main}]
The proof has two parts. First, we use
\Cref{fct:sign-matching-function,fct:SQ-function-to-LB} to
obtain a hard distinguishing problem of the form in
\Cref{def:SQ-testing}. Second, we show that an SQ learner would
solve this problem by learning from the input distribution and
checking the error of its output hypothesis.

For the first part, let $g$ be the function constructed in the
proof of \Cref{fct:sign-matching-function}. Applying
\Cref{fct:SQ-function-to-LB} with $\ell=2k$, distinguishing an
unknown $D_{\v}\in\D_g$ from $D_0$ requires either a
query of tolerance at most $d^{-ck}$ or at least
$N:=2^{d^{c_0}}$ queries, for universal constants $c,c_0>0$.

Moreover, each $D_{\v}$ is realizable by the target class.
Indeed, $g$ is negative on both tails, and its positive set is
a union of at most $k$ bounded intervals. A union of $r$ bounded
intervals can be expressed using $r$ quadratic inequalities:
one restricting to their convex hull and $r-1$ excluding the
intervening gaps. Substituting $t=\v\cdot\x$ gives the required
intersection of quadratic PTFs, up to Gaussian-null endpoints
and conversion of the labels. We use the equivalent
$\{\pm1\}$-label convention below.

For the second part, fix a constant $\gamma\in(0,1/2)$.
Suppose for contradiction that an SQ learner achieves error
at most $1/2-\gamma$ with probability at least $2/3$,
using fewer than $N/2$ queries, all of tolerance greater
than $d^{-ck}$.
Since $\gamma$ is a positive constant, for every $k\ge1$
and sufficiently large $d$ depending on $k$ and $\gamma$,
we have $\gamma/4>d^{-ck}$.

Given an input distribution $D$, run this learner, enforcing
its claimed complexity bounds, and use one additional SQ query
of tolerance $\gamma/4$ to estimate the error of its output
hypothesis $h$.
If $D=D_{\v}$, then with probability at least $2/3$, the error
of $h$ is at most $1/2-\gamma$, so the estimate is at most
$1/2-3\gamma/4$.
If $D=D_0$, every hypothesis has error exactly $1/2$, so the
estimate is at least $1/2-\gamma/4$.
Comparing the estimate with $1/2-\gamma/2$ therefore
distinguishes the two cases with probability at least $2/3$.

For sufficiently large $d$, this distinguishing algorithm
uses fewer than $N$ queries, all of tolerance greater than
$d^{-ck}$, contradicting \Cref{fct:SQ-function-to-LB}.
\end{proof}

\bibliography{mydb}

\newcommand{\etalchar}[1]{$^{#1}$}
\begin{thebibliography}{DMRT25b}
\expandafter\ifx\csname url\endcsname\relax
  \def\url#1{\texttt{#1}}\fi
\expandafter\ifx\csname doi\endcsname\relax
  \def\doi#1{\burlalt{doi:#1}{http://dx.doi.org/#1}}\fi
\expandafter\ifx\csname urlprefix\endcsname\relax\def\urlprefix{URL }\fi
\expandafter\ifx\csname href\endcsname\relax
  \def\href#1#2{#2}\fi
\expandafter\ifx\csname burlalt\endcsname\relax
  \def\burlalt#1#2{\href{#2}{#1}}\fi

\bibitem[ABAM23]{ABM23}
E.~Abbe, E.~Boix-Adser{\`a}, and T.~Misiakiewicz.
\newblock {SGD} learning on neural networks: Leap complexity and saddle-to-saddle dynamics.
\newblock In {\em Proceedings of the Thirty-Sixth Conference on Learning Theory}, volume 195 of {\em Proceedings of Machine Learning Research}, pages 2552--2623, 2023.
\newblock \urlprefix\url{https://proceedings.mlr.press/v195/abbe23a.html}.

\bibitem[APS26]{APS26}
J.~Alman, S.~Patel, and R.~A. Servedio.
\newblock Learning functions of halfspaces.
\newblock In {\em Proceedings of the 58th Annual ACM Symposium on Theory of Computing}, STOC '26. Association for Computing Machinery, 2026.
\newblock \doi{10.1145/3798129.3800866}.

\bibitem[Bau90]{Baum:91}
E.~B. Baum.
\newblock A polynomial time algorithm that learns two hidden unit nets.
\newblock {\em Neural Computation}, 2(4):510--522, 1990.
\newblock \doi{10.1162/neco.1990.2.4.510}.

\bibitem[BH25]{BH25}
J.~Bruna and D.~Hsu.
\newblock Survey on algorithms for multi-index models.
\newblock {\em Statistical Science}, 40(3):378--391, 2025.
\newblock \doi{10.1214/25-STS993}.

\bibitem[BK97]{blum1997learning}
A.~L. Blum and R.~Kannan.
\newblock Learning an intersection of a constant number of halfspaces over a uniform distribution.
\newblock {\em Journal of Computer and System Sciences}, 54(2):371--380, 1997.

\bibitem[BR88]{blum1988training}
A.~Blum and R.~L. Rivest.
\newblock Training a 3-node neural network is {NP}-complete.
\newblock In D.~S. Touretzky, editor, {\em Advances in Neural Information Processing Systems 1}, pages 494--501, 1988.

\bibitem[CDG{\etalchar{+}}23]{chen2023learning}
S.~Chen, Z.~Dou, S.~Goel, A.~Klivans, and R.~Meka.
\newblock Learning narrow one-hidden-layer {ReLU} networks.
\newblock In {\em The Thirty Sixth Annual Conference on Learning Theory}, pages 5580--5614. PMLR, 2023.

\bibitem[CKM21]{ChenKM21}
S.~Chen, A.~R. Klivans, and R.~Meka.
\newblock Learning deep {ReLU} networks is fixed-parameter tractable.
\newblock In {\em 62nd {IEEE} Annual Symposium on Foundations of Computer Science, {FOCS} 2021, Denver, CO, USA, February 7-10, 2022}, pages 696--707. {IEEE}, 2021.
\newblock \doi{10.1109/FOCS52979.2021.00073}.

\bibitem[CN24]{ChenN24}
S.~Chen and S.~Narayanan.
\newblock A faster and simpler algorithm for learning shallow networks.
\newblock In S.~Agrawal and A.~Roth, editors, {\em The Thirty Seventh Annual Conference on Learning Theory}, volume 247 of {\em Proceedings of Machine Learning Research}, pages 981--994. {PMLR}, 2024.
\newblock \urlprefix\url{https://proceedings.mlr.press/v247/chen24b.html}.

\bibitem[CW01]{CW:01}
A.~Carbery and J.~Wright.
\newblock {Distributional and $L^q$ norm inequalities for polynomials over convex bodies in $R^n$}.
\newblock {\em Mathematical Research Letters}, 8(3):233--248, 2001.

\bibitem[Dan16]{daniely2016complexity}
A.~Daniely.
\newblock Complexity theoretic limitations on learning halfspaces.
\newblock In {\em Proceedings of the forty-eighth annual ACM symposium on Theory of Computing}, pages 105--117, 2016.

\bibitem[DDS14]{de2014deterministic}
A.~De, I.~Diakonikolas, and R.~A. Servedio.
\newblock Deterministic approximate counting for juntas of degree-2 polynomial threshold functions.
\newblock In {\em 2014 IEEE 29th Conference on Computational Complexity (CCC)}, pages 229--240. IEEE, 2014.

\bibitem[DHK{\etalchar{+}}10]{diakonikolas2010bounding}
I.~Diakonikolas, P.~Harsha, A.~Klivans, R.~Meka, P.~Raghavendra, R.~A. Servedio, and L.-Y. Tan.
\newblock Bounding the average sensitivity and noise sensitivity of polynomial threshold functions.
\newblock In {\em Proceedings of the forty-second ACM symposium on Theory of computing}, pages 533--542, 2010.

\bibitem[DIKR25]{DIKR25}
I.~Diakonikolas, G.~Iakovidis, D.~M. Kane, and L.~Ren.
\newblock Algorithms and {SQ} lower bounds for robustly learning real-valued multi-index models.
\newblock In {\em Advances in Neural Information Processing Systems}, volume~38, 2025.
\newblock \urlprefix\url{https://proceedings.neurips.cc/paper_files/paper/2025/hash/5debeab774db7ca61ea15f1117f7fa33-Abstract-Conference.html}.

\bibitem[DIKZ25]{DiakonikolasIKZ25}
I.~Diakonikolas, G.~Iakovidis, D.~M. Kane, and N.~Zarifis.
\newblock Robust learning of multi-index models via iterative subspace approximation.
\newblock In {\em 66th {IEEE} Annual Symposium on Foundations of Computer Science, {FOCS} 2025, Sydney, Australia, December 14-17, 2025}, pages 2231--2239. {IEEE}, 2025.

\bibitem[DIM26]{diakonikolas2026polynomial}
I.~Diakonikolas, G.~Iakovidis, and M.~Ma.
\newblock Polynomial-time robust multiclass linear classification under {Gaussian} marginals.
\newblock In {\em Advances in Neural Information Processing Systems}, 2026.
\newblock \urlprefix\url{https://arxiv.org/abs/2605.21428}.
\newblock To appear.

\bibitem[DK24]{diakonikolas2024efficiently}
I.~Diakonikolas and D.~M. Kane.
\newblock Efficiently learning one-hidden-layer {ReLU} networks via {Schur} polynomials.
\newblock In {\em The Thirty Seventh Annual Conference on Learning Theory}, pages 1364--1378. PMLR, 2024.

\bibitem[DKKZ20]{diakonikolas2020algorithms}
I.~Diakonikolas, D.~M. Kane, V.~Kontonis, and N.~Zarifis.
\newblock Algorithms and {SQ} lower bounds for {PAC} learning one-hidden-layer {ReLU} networks.
\newblock In {\em Conference on Learning Theory}, pages 1514--1539. PMLR, 2020.

\bibitem[DKM26]{diakonikolas2026robust}
I.~Diakonikolas, D.~M. Kane, and M.~Ma.
\newblock Robust learning of multi-index models under logconcave distributions via weak partial learners.
\newblock In {\em Proceedings of the 67th Annual IEEE Symposium on Foundations of Computer Science (FOCS)}, 2026.

\bibitem[DKN10]{diakonikolas2010bounded}
I.~Diakonikolas, D.~M. Kane, and J.~Nelson.
\newblock Bounded independence fools degree-2 threshold functions.
\newblock In {\em 2010 IEEE 51st Annual Symposium on Foundations of Computer Science}, pages 11--20. IEEE, 2010.

\bibitem[DKPZ21]{DKPZ21}
I.~Diakonikolas, D.~M. Kane, T.~Pittas, and N.~Zarifis.
\newblock The optimality of polynomial regression for agnostic learning under {Gaussian} marginals in the {SQ} model.
\newblock In M.~Belkin and S.~Kpotufe, editors, {\em Proceedings of the 34th Conference on Learning Theory}, volume 134 of {\em Proceedings of Machine Learning Research}, pages 1552--1584, 2021.
\newblock \urlprefix\url{https://proceedings.mlr.press/v134/diakonikolas21c.html}.

\bibitem[DKS17]{DKS17-sq}
I.~Diakonikolas, D.~M. Kane, and A.~Stewart.
\newblock Statistical query lower bounds for robust estimation of high-dimensional {Gaussians} and {Gaussian} mixtures.
\newblock In {\em 58th {IEEE} Annual Symposium on Foundations of Computer Science ({FOCS})}, pages 73--84, 2017.
\newblock \doi{10.1109/FOCS.2017.16}.

\bibitem[DKS18]{DKS18a}
I.~Diakonikolas, D.~Kane, and A.~Stewart.
\newblock Learning geometric concepts with nasty noise.
\newblock In {\em Proceedings of the 50\textsuperscript{th} Annual {ACM} {SIGACT} Symposium on Theory of Computing, {STOC} 2018}, pages 1061--1073, 2018.

\bibitem[DLB25]{DLB25}
A.~Damian, J.~D. Lee, and J.~Bruna.
\newblock The generative leap: Tight sample complexity for efficiently learning {Gaussian} multi-index models.
\newblock In {\em Advances in Neural Information Processing Systems}, volume~38, 2025.
\newblock \doi{10.52202/085713-0953}.

\bibitem[DMRT25a]{DMRT25}
I.~Diakonikolas, M.~Ma, L.~Ren, and C.~Tzamos.
\newblock Learning intersections of two margin halfspaces under factorizable distributions.
\newblock In {\em Proceedings of the Thirty Eighth Conference on Learning Theory}, volume 291 of {\em Proceedings of Machine Learning Research}, pages 1472--1530, 2025.
\newblock \urlprefix\url{https://proceedings.mlr.press/v291/diakonikolas25a.html}.

\bibitem[DMRT25b]{diakonikolas2025statistical}
I.~Diakonikolas, M.~Ma, L.~Ren, and C.~Tzamos.
\newblock Statistical query hardness of multiclass linear classification with random classification noise.
\newblock In {\em International Conference on Machine Learning}, pages 13693--13712. PMLR, 2025.

\bibitem[DS14]{DeS14}
A.~De and R.~A. Servedio.
\newblock Efficient deterministic approximate counting for low-degree polynomial threshold functions.
\newblock In D.~B. Shmoys, editor, {\em Symposium on Theory of Computing, {STOC} 2014, New York, NY, USA, May 31 - June 03, 2014}, pages 832--841. {ACM}, 2014.
\newblock \doi{10.1145/2591796.2591800}.

\bibitem[DSTW10]{diakonikolas2010regularity}
I.~Diakonikolas, R.~A. Servedio, L.-Y. Tan, and A.~Wan.
\newblock A regularity lemma, and low-weight approximators, for low-degree polynomial threshold functions.
\newblock In {\em 2010 IEEE 25th Annual Conference on Computational Complexity}, pages 211--222. IEEE, 2010.

\bibitem[Duc18]{Duc18}
J.~C. Duchi.
\newblock Introductory lectures on stochastic optimization.
\newblock In M.~W. Mahoney, J.~C. Duchi, and A.~C. Gilbert, editors, {\em The Mathematics of Data}, volume~25 of {\em IAS/Park City Mathematics Series}, pages 99--185. 2018.

\bibitem[Fel16]{Feldman16b}
V.~Feldman.
\newblock Statistical query learning.
\newblock In {\em Encyclopedia of Algorithms}, pages 2090--2095. 2016.

\bibitem[FGR{\etalchar{+}}17]{FeldmanGRVX17}
V.~Feldman, E.~Grigorescu, L.~Reyzin, S.~Vempala, and Y.~Xiao.
\newblock Statistical algorithms and a lower bound for detecting planted cliques.
\newblock {\em J. {ACM}}, 64(2):8:1--8:37, 2017.

\bibitem[FGV17]{FeldmanGV17}
V.~Feldman, C.~Guzman, and S.~S. Vempala.
\newblock Statistical query algorithms for mean vector estimation and stochastic convex optimization.
\newblock In P.~N. Klein, editor, {\em Proceedings of the Twenty-Eighth Annual {ACM-SIAM} Symposium on Discrete Algorithms, {SODA} 2017}, pages 1265--1277. {SIAM}, 2017.

\bibitem[FWNL25]{FWNL25}
H.~Fu, Z.~Wang, E.~Nichani, and J.~D. Lee.
\newblock Learning hierarchical polynomials of multiple nonlinear features.
\newblock In {\em International Conference on Learning Representations}, 2025.
\newblock \urlprefix\url{https://proceedings.iclr.cc/paper_files/paper/2025/hash/599221d7ebf6b3403190f38a3f282a1c-Abstract-Conference.html}.

\bibitem[HL13]{hillar2013most}
C.~J. Hillar and L.-H. Lim.
\newblock Most tensor problems are {NP}-hard.
\newblock {\em Journal of the ACM}, 60(6):45:1--45:39, 2013.
\newblock \doi{10.1145/2512329}.

\bibitem[Kan11]{Kane11}
D.~M. Kane.
\newblock The {Gaussian} surface area and noise sensitivity of degree-\emph{d} polynomial threshold functions.
\newblock {\em Computational Complexity}, 20(2):389--412, 2011.

\bibitem[Kea98]{Kearns:98}
M.~J. Kearns.
\newblock Efficient noise-tolerant learning from statistical queries.
\newblock {\em Journal of the ACM}, 45(6):983--1006, 1998.

\bibitem[KKSK11]{kakade2011efficient}
S.~Kakade, V.~Kanade, O.~Shamir, and A.~Kalai.
\newblock Efficient learning of generalized linear and single index models with isotonic regression.
\newblock {\em Advances in Neural Information Processing Systems}, 24, 2011.

\bibitem[KOS04]{KOS:04}
A.~Klivans, R.~O'Donnell, and R.~Servedio.
\newblock Learning intersections and thresholds of halfspaces.
\newblock {\em Journal of Computer \& System Sciences}, 68(4):808--840, 2004.

\bibitem[KOS08]{KOS:08}
A.~Klivans, R.~O'Donnell, and R.~Servedio.
\newblock Learning geometric concepts via {G}aussian surface area.
\newblock In {\em Proc.\ 49th IEEE Symposium on Foundations of Computer Science (FOCS)}, pages 541--550, Philadelphia, Pennsylvania, 2008.

\bibitem[KP96]{kwek1996pac}
S.~Kwek and L.~Pitt.
\newblock {PAC} learning intersections of halfspaces with membership queries.
\newblock In {\em Proceedings of the ninth annual conference on Computational learning theory}, pages 244--254, 1996.

\bibitem[KS04]{KS04}
A.~Klivans and R.~Servedio.
\newblock Learning intersections of halfspaces with a margin.
\newblock In {\em Proceedings of the 17\textsuperscript{th} Annual Conference on Learning Theory}, pages 348--362, 2004.

\bibitem[KS07]{klivans2007unconditional}
A.~R. Klivans and A.~A. Sherstov.
\newblock Unconditional lower bounds for learning intersections of halfspaces.
\newblock {\em Machine Learning}, 69(2--3):97--114, 2007.

\bibitem[KS08]{khot2008hardness}
S.~Khot and R.~Saket.
\newblock On hardness of learning intersection of two halfspaces.
\newblock In {\em Proceedings of the fortieth annual ACM symposium on Theory of computing}, pages 345--354, 2008.

\bibitem[KS09]{klivans2009cryptographic}
A.~R. Klivans and A.~A. Sherstov.
\newblock Cryptographic hardness for learning intersections of halfspaces.
\newblock {\em Journal of Computer and System Sciences}, 75(1):2--12, 2009.

\bibitem[MT94]{MT:94}
W.~Maass and G.~Turan.
\newblock How fast can a threshold gate learn?
\newblock In S.~Hanson, G.~Drastal, and R.~Rivest, editors, {\em Computational Learning Theory and Natural Learning Systems}, pages 381--414. MIT Press, 1994.

\bibitem[MZ10]{MZstoc10}
R.~Meka and D.~Zuckerman.
\newblock Pseudorandom generators for polynomial threshold functions.
\newblock In {\em STOC}, pages 427--436, 2010.

\bibitem[NDL23]{NDL23}
E.~Nichani, A.~Damian, and J.~D. Lee.
\newblock Provable guarantees for nonlinear feature learning in three-layer neural networks.
\newblock In {\em Advances in Neural Information Processing Systems ({NeurIPS})}, volume~36, 2023.

\bibitem[Nic26]{Nichani26}
E.~Nichani.
\newblock A tutorial on feature learning in shallow neural networks.
\newblock \url{https://eshaannichani.com/blog/multi-index-models/index.html}, September 27 2026.
\newblock Section 5: Open Directions.

\bibitem[O'D14]{Don14}
R.~O'Donnell.
\newblock {\em Analysis of Boolean Functions}.
\newblock Cambridge University Press, 2014.

\bibitem[PV26]{rev-PV26}
S.~Patel and S.~Vempala.
\newblock Tight bounds for learning polyhedra with a margin.
\newblock In {\em Proceedings of the IEEE Symposium on Foundations of Computer Science (FOCS)}. IEEE, 2026.
\newblock \urlprefix\url{https://arxiv.org/abs/2604.14614}.

\bibitem[Ros58]{rosenblatt1958perceptron}
F.~Rosenblatt.
\newblock The perceptron: A probabilistic model for information storage and organization in the brain.
\newblock {\em Psychological Review}, 65(6):386--408, 1958.
\newblock \doi{10.1037/h0042519}.

\bibitem[She10]{sherstov2010optimal}
A.~A. Sherstov.
\newblock Optimal bounds for sign-representing the intersection of two halfspaces by polynomials.
\newblock In {\em Proceedings of the Forty-second ACM Symposium on Theory of Computing}, pages 523--532, 2010.

\bibitem[She13]{sherstov2013intersection}
A.~A. Sherstov.
\newblock The intersection of two halfspaces has high threshold degree.
\newblock {\em SIAM Journal on Computing}, 42(6):2329--2374, 2013.

\bibitem[Tie24]{tiegel2024improved}
S.~Tiegel.
\newblock Improved hardness results for learning intersections of halfspaces.
\newblock In {\em The Thirty Seventh Annual Conference on Learning Theory}, pages 4764--4786. PMLR, 2024.

\bibitem[Val84]{Valiant:84}
L.~Valiant.
\newblock A theory of the learnable.
\newblock {\em Communications of the ACM}, 27(11):1134--1142, 1984.

\bibitem[Vem10a]{Vempala10a}
S.~Vempala.
\newblock Learning convex concepts from {Gaussian} distributions with {PCA}.
\newblock In {\em 51\textsuperscript{st} Annual {IEEE} Symposium on Foundations of Computer Science, {FOCS}}, pages 124--130, 2010.

\bibitem[Vem10b]{Vempala10}
S.~Vempala.
\newblock A random-sampling-based algorithm for learning intersections of halfspaces.
\newblock {\em J. {ACM}}, 57(6):32:1--32:14, 2010.

\bibitem[WNL24]{WNL24}
Z.~Wang, E.~Nichani, and J.~D. Lee.
\newblock Learning hierarchical polynomials with three-layer neural networks.
\newblock In {\em The Twelfth International Conference on Learning Representations, {ICLR}}. OpenReview.net, 2024.

\end{thebibliography}
\bibliographystyle{halpha-abbrv}

\end{document}